%% file: evaluation.tex
\documentclass[10pt,twoside]{article}
\usepackage{fullpage, amsthm, subfig}

\usepackage[utf8]{inputenc} 
\usepackage[T1]{fontenc}    
\usepackage{url}            
\usepackage{booktabs}       
\usepackage{amsfonts,amsmath}       
\usepackage{nicefrac}       
\usepackage{microtype}      
\usepackage[dvipsnames]{xcolor}

\usepackage[unicode,psdextra]{hyperref}

\input{macros_evaluation}
\title{\bf \Large Debiasing Evaluations That are Biased by Evaluations}

\date{}

\author{\\%
    Jingyan Wang$^\dagger$, Ivan Stelmakh$^\dagger$, Yuting Wei$^\ast$ and Nihar B. Shah$^\dagger$\\~\\
    School of Computer Science$^\dagger$\\
    Department of Statistics \& Data Science$^\ast$\\
    Carnegie Mellon University\\
    \texttt{\{jingyanw, stiv\}@cs.cmu.edu, ytwei@cmu.edu, nihars@cs.cmu.edu}
}

\usepackage[switch]{lineno}
\usepackage{footnote}

\usepackage{amsfonts, amsmath, enumitem, mathtools, dsfont, xspace, xcolor}
\usepackage[linesnumbered, algoruled, boxed, lined]{algorithm2e}

\DeclareMathOperator*{\argmax}{arg\,max}
\DeclareMathOperator*{\argmin}{arg\,min}
\DeclarePairedDelimiter{\abs}{\lvert}{\rvert}
\DeclarePairedDelimiter{\norm}{\lVert}{\rVert}

\newcommand{\normfro}[1]{\norm{#1}_F}
\newcommand{\normtwo}[1]{\norm{#1}_2}
\newcommand{\norminf}[1]{\norm{#1}_\infty}
\DeclareMathOperator*{\median}{median}
\DeclareMathOperator*{\sign}{sign}

\newtheorem{theorem}{Theorem}
\newtheorem{lemma}[theorem]{Lemma}
\newtheorem{proposition}[theorem]{Proposition}
\newtheorem{definition}[theorem]{Definition}
\newtheorem{corollary}[theorem]{Corollary}

\newcommand{\figurewidth}{5.2cm}
\newcommand{\figurewidthlong}{5.5cm} 

\let\citep\cite
\let\citet\cite
\let\citealp\cite

\begin{document}

\maketitle

\begin{abstract}
It is common to evaluate a set of items by soliciting people to rate them. For example, universities ask students  to rate the teaching quality of their instructors, and conference organizers ask authors of submissions to evaluate the quality of the reviews. However, in these applications, students often give a higher rating to a course if they receive higher grades in a course, and authors often give a higher rating to the reviews if their papers are accepted to the conference. In this work, we call these external factors the ``outcome'' experienced by people, and consider the problem of mitigating these outcome-induced biases in the given ratings when some information about the outcome is available. We formulate the information about the outcome as a known partial ordering on the bias. We  propose a debiasing method by solving a regularized optimization problem under this ordering constraint, and also provide a carefully designed cross-validation method that adaptively chooses the appropriate amount of regularization. We provide theoretical guarantees on the performance of our algorithm, as well as experimental evaluations.
\end{abstract}

\date{}


\input{text_main}

\section*{Acknowledgments}
The work of J.W., I.S., and N.S. was supported in part by NSF CAREER award 1942124 and in part by NSF CIF 1763734. 
Y.W. was partially supported by the NSF grants CCF-2007911 and
DMS-2015447.

{
\bibliographystyle{plain}
\bibliography{isotonic}
}

\appendix
\input{text_appendix}
\input{text_proof}

\end{document}

%% file: macros_evaluation.tex
\newcommand{\indicator}{\mathds{1}}
\newcommand{\defn}{\vcentcolon=}

\newcommand{\reals}{\mathbb{R}}
\newcommand{\integers}{\mathbb{Z}}

\newcommand{\Expect}{{\mathbb E}}
\newcommand{\Prob}{\mathbb{P}}
\newcommand{\prob}{\Prob}
\newcommand{\Probbig}[1]{\Prob\Big(#1\Big)}
\newcommand{\dd}{\,\textrm{d}}

\newcommand{\normal}{\mathcal{N}}
\newcommand{\identitymtx}{I}

\newcommand{\given}{\mid}
\newcommand{\intersect}{\cap}
\newcommand{\union}{\cup}
\newcommand{\sample}{\sim}
\newcommand{\gaussian}{\ensuremath{\mathcal{N}}}

\newcommand{\uniform}{\text{Unif}}

\newcommand{\vecone}{\mathbf{1}}

\newcommand{\stepone}{\text{(i)}\xspace}
\newcommand{\steptwo}{\text{(ii)}\xspace}
\newcommand{\stepthree}{\text{(iii)}\xspace}

\newcommand{\thcount}{\text{th}}
\newcommand{\limn}{\lim_{\numstudent\rightarrow \infty}}
\newenvironment{quotex}%
  {\list{}{\leftmargin=0.1in\rightmargin=0.1in}\item[]}%
  {\endlist}

\newcommand{\const}{c}
\newcommand{\course}{course\xspace}
\newcommand{\courses}{{\course}s\xspace}
\newcommand{\student}{student\xspace}
\newcommand{\students}{{\student}s\xspace}
\newcommand{\grade}{grade\xspace}

\newcommand{\cv}{cross-validation\xspace}
\newcommand{\nn}{nearest-neighbor\xspace}

\newcommand{\subsampling}{reweighted mean\xspace}
\newcommand{\Subsampling}{Reweighted mean\xspace}

\newcommand{\element}{element\xspace}
\newcommand{\elements}{elements\xspace}

\newcommand{\designmatrix}{A}

\newcommand{\numcourse}{d}
\newcommand{\numstudent}{n}

\newcommand{\idxcourse}{i}
\newcommand{\idxstudent}{j}
\newcommand{\idxcoursealt}{{\idxcourse'}}
\newcommand{\idxstudentalt}{{\idxstudent'}}
\newcommand{\idxcoursenn}{\idxcourse^\totalorder}
\newcommand{\idxstudentnn}{\idxstudent^\totalorder}

\newcommand{\idxcoursescope}{\idxcourse\in [\numcourse]}
\newcommand{\idxstudentscope}{\idxstudent\in [\numstudent]}
\newcommand{\setscope}{\set\subseteq [\numcourse]\times [\numstudent]}
\newcommand{\idxgroupscope}{\idxgroup\in [\numgroup]}

\newcommand{\idxpair}{{\idxcourse\idxstudent}}
\newcommand{\idxpairparen}{(\idxcourse, \idxstudent)}

\newcommand{\numgroup}{r}
\newcommand{\idxgroup}{k}
\newcommand{\idxgroupalt}{{\idxgroup'}}

\newcommand{\bias}{b}
\newcommand{\noise}{z}

\newcommand{\obssingle}{y}
\newcommand{\biassingle}{b}
\newcommand{\noisesingle}{z}

\newcommand{\obsmtx}{Y}
\newcommand{\biasmtx}{B}
\newcommand{\noisemtx}{Z}

\newcommand{\meancourse}{x}
\newcommand{\truemean}{\meancourse^*}
\newcommand{\estmean}{\widehat{\meancourse}}

\newcommand{\textmean}{\mathrm{mean}}
\newcommand{\estmeanmean}{\estmean_{\textmean}}
\newcommand{\textsubsample}{\mathrm{rw}}
\newcommand{\estmeansubsample}{\estmean_{\textsubsample}}
\newcommand{\textmed}{\mathrm{med}}
\newcommand{\estmeanmed}{\estmean_{\textmed}}

\newcommand{\estbias}{\widehat{\bias}}
\newcommand{\estbiasmtx}{\widehat{\biasmtx}}
\newcommand{\estbiasvalat}[1]{\widetilde{\bias}^{(#1)}}
\newcommand{\estbiasmtxval}{\widetilde{\biasmtx}}
\newcommand{\estbiasmtxvalat}[1]{\widetilde{\biasmtx}^{(#1)}}
\newcommand{\estbiasmtxvalatoforder}[2]{\widetilde{\biasmtx}^{(#1)}_{#2}}

\newcommand{\estmeanat}[1]{\widehat{\meancourse}^{(#1)}}
\newcommand{\estbiasat}[1]{\widehat{\bias}^{(#1)}}
\newcommand{\estbiasmtxat}[1]{\widehat{\biasmtx}^{(#1)}}

\newcommand{\setpartialorder}{\mathcal{O}}
\newcommand{\settotalorder}{\mathcal{T}}

\newcommand{\totalorder}{\pi}
\newcommand{\totalordersplit}{\pi_0}

\newcommand{\obs}{y}

\newcommand{\set}{\Omega}
\newcommand{\setnn}{\mathrm{NN}}
\newcommand{\texttrain}{\mathrm{t}}
\newcommand{\textval}{\mathrm{v}}

\newcommand{\settrain}{{\set^\texttrain}}
\newcommand{\setval}{{\set^\textval}}

\newcommand{\setcoursegroup}[2]{{\setgroup_{#1#2}}}
\newcommand{\settraincoursegroup}[2]{{\setgroup^\texttrain_{#1#2}}}
\newcommand{\setvalcoursegroup}[2]{{\setgroup^\textval_{#1#2}}}
\newcommand{\setvalcourse}[1]{{\set^\textval_{#1}}}

\newcommand{\position}{t}
\newcommand{\setreg}{\Lambda}

\newcommand{\blocklengthgroupmin}[1]{\blocklength_{#1, \textmin}}

\newcommand{\treelevel}{\ell}
\newcommand{\numelementspernode}{k}

\newcommand{\reg}{\lambda}

\newcommand{\textcv}{\mathrm{cv}}
\newcommand{\regcv}{\reg_\textcv}

\newcommand{\fraction}{r}
\newcommand{\setgroup}{G}

\newcommand{\textlow}{\mathrm{L}}
\newcommand{\texthigh}{\mathrm{H}}

\newcommand{\gaussianwidthbias}{\sigma}
\newcommand{\gaussianwidthnoise}{\eta}

\newcommand{\errbound}{\epsilon}
\newcommand{\numstudentlb}{\numstudent_0}
\newcommand{\probbound}{\delta}
\newcommand{\setgroupidx}{R}

\newcommand{\blocklength}{\ell}
\newcommand{\blocklengthgroup}[1]{\blocklength_{#1}}
\newcommand{\blocklengthgroupval}[1]{\blocklength^\textval_{#1}}
\newcommand{\blocklengthgrouptrain}[1]{\blocklength^\texttrain_{#1}}
\newcommand{\err}{e}
\newcommand{\errat}[1]{\err^{(#1)}}
\newcommand{\erratcoursegroup}[3]{\err^{(#1)}_{#2#3}}

\newcommand{\meanplusbiasmtx}{W}
\newcommand{\meanplusbiasmtxopt}{W_0}

\newcommand{\constcoercive}{M}
\newcommand{\normcoercive}{R_M}

\newcommand{\diffmeancourse}{\Delta \meancourse}

\newcommand{\contradictionsumbias}{\gamma}

\newcommand{\minimalvalfirstterm}{V^*}
\newcommand{\valfirsttermatinfty}{V}
\newcommand{\margin}{\gamma}
\newcommand{\randsign}{U}

\newcommand{\convdist}{\xrightarrow{d}}
\newcommand{\convprob}{\xrightarrow{P}}
\newcommand{\maxspacing}{M}

\newcommand{\setpairconstraint}{S}
\newcommand{\setpairconstrainttrain}{\setpairconstraint^\texttrain}

\newcommand{\rank}{t}

\newcommand{\rankalt}{\rank'}
\newcommand{\textmax}{\mathrm{max}}
\newcommand{\textmin}{\mathrm{min}}
\newcommand{\rankrand}{T}

\newcommand{\numsupernodes}{s}
\newcommand{\setsupernode}{V}
\newcommand{\setsupernodeleft}{\setsupernode_\textlow}
\newcommand{\setsupernoderight}{\setsupernode_\texthigh}
\newcommand{\estmeanbiasleft}{\estbias_\textlow}
\newcommand{\estmeanbiasright}{\estbias_\texthigh}

\newcommand{\supernode}{hypernode\xspace}
\newcommand{\supernodes}{hypernodes\xspace}
\newcommand{\lengthcycle}{L}
\newcommand{\idxnode}{m}
\newcommand{\idxnodealt}{{\idxnode'}}

\newcommand{\truebiasrank}[1]{\bias^{*(#1)}}
\newcommand{\meanbiasgroup}[1]{\overline{\bias}_{\setgroupof{#1}}}
\newcommand{\truemeanbiasgroup}[1]{\overline{\bias}^*_{\setgroupof{#1}}}
\newcommand{\biascourse}[1]{\overline{\bias}_{#1}}
\newcommand{\idxrank}{k}

\newcommand{\biasmaxminusmin}{\Delta_\biasmtx}
\newcommand{\truebiasmaxminusmin}{\Delta_{\truebiasmtx}}

\newcommand{\solutionincrease}{\Delta}
\newcommand{\obscoursegroupmin}[2]{\obs_{#1#2, \textmin}}
\newcommand{\obscoursegroupmax}[2]{\obs_{#1#2, \textmax}}

\newcommand{\tildeu}{\widetilde{U}}
\newcommand{\tildev}{\widetilde{V}}

\newcommand{\biascoursegroupmax}[2]{\bias_{#1#2, \textmax}}
\newcommand{\biascoursegroupmaxtrain}[2]{\bias_{#1#2, \textmax}^\texttrain}
\newcommand{\biascoursegroupmin}[2]{\bias_{#1#2,\textmin}}
\newcommand{\biascoursegroupmintrain}[2]{\bias_{#1#2,\textmin}^\texttrain}
\newcommand{\biasgroupmax}[1]{\bias_{#1, \textmax}}
\newcommand{\biasgroupmaxtrain}[1]{\bias_{#1, \textmax}^\texttrain}
\newcommand{\biasgroupmin}[1]{\bias_{#1,\textmin}}
\newcommand{\biasgroup}[1]{\bias_#1}
\newcommand{\term}{T}

\newcommand{\idxgeneral}{i}
\newcommand{\idxgeneralalt}{j}

\newcommand{\meanobs}{\overline{\obs}}
\newcommand{\meanobscourse}[1]{\overline{\obs}_#1}
\newcommand{\eventbad}{\event_\mathrm{bad}}
\newcommand{\idxnodesplit}{\idxnode^*}

\newcommand{\truebias}{\bias^*}

\newcommand{\truebiasmtx}{\biasmtx^*}

\newcommand{\estbiasinterp}{\widetilde{\bias}}
\newcommand{\estbiasinterpat}[1]{\widetilde{\bias}^{(#1)}}
\newcommand{\estbiasinterpgroup}[1]{\widetilde{\bias}_#1}
\newcommand{\estbiasinterpatgroup}[2]{\widetilde{\bias}^{(#1)}_#2}

\newcommand{\biasmeangrouptrain}[1]{\bias^{\texttrain}_#1}
\newcommand{\biasmeangroupval}[1]{\bias^{\textval}_#1}
\newcommand{\biasmeancoursegroup}[2]{\bias_{#1#2}}
\newcommand{\biasmeancoursegroupval}[2]{\bias^{\textval}_{#1#2}}

\newcommand{\estbiasmeangrouptrain}[1]{\estbias^{\texttrain}_#1}

\newcommand{\setgroupof}[1]{{\setgroup_{#1}}}
\newcommand{\settraingroup}[1]{{G^\texttrain_{#1}}}
\newcommand{\setcourse}[1]{\set_{#1}}
\newcommand{\settraincourse}[1]{\set^\texttrain_{#1}}
\newcommand{\setvalgroup}[1]{G^\textval_{#1}}

\newcommand{\sizesettrain}{\abs{\settrain}}
\newcommand{\sizesetval}{\abs{\setval}}

\newcommand{\event}{E}

\newcommand{\numstudenttrain}{\numstudent^\texttrain}
\newcommand{\numstudentval}{\numstudent^\textval}

\newcommand{\blocklengthcoursegroup}[2]{\blocklength_{#1#2}}
\newcommand{\blocklengthcoursegroupval}[2]{\blocklength_{#1#2}^\textval}
\newcommand{\blocklengthcoursegrouptrain}[2]{\blocklength_{#1#2}^\texttrain}
\newcommand{\sumconstruct}{S}
\newcommand{\meanconstruct}{\mu}
\newcommand{\valuecondition}{a}
\newcommand{\setcondition}{A}

\newcommand{\vargaussian}{X}
\newcommand{\vargaussianorder}[1]{X^{(#1)}}

\newcommand{\valueconditionmax}{\valuecondition_7}
\newcommand{\valueconditionmin}{\valuecondition_1}
\newcommand{\setregoutside}{\setreg_{\errbound}}

\newcommand{\settotalorderin}{\settotalorder_\mathrm{in}}
\newcommand{\settotalorderout}{\settotalorder_\mathrm{out}}
\newcommand{\sizesettotalorder}{\abs{\settotalorder}}

\newcommand{\biascoursetrain}[1]{\bias_{#1}^\texttrain}
\newcommand{\biastrain}{\bias^\texttrain}
\newcommand{\biasorder}[1]{\bias^{(#1)}}

\newcommand{\objgivenx}{L}

\newcommand{\proj}{\Pi}
\newcommand{\conemonotone}{\mathcal{M}}
\newcommand{\eventnonoverlaplinearval}[2]{\event^\textval_{#1,#2}}
\newcommand{\eventnonoverlaplinear}[2]{\event_{#1,#2}}
\newcommand{\eventnonoverlaplinearrank}[3]{\event_{#1,#2,#3}}

\newcommand{\constfrac}{\const_\mathrm{f}}

\newcommand{\setinterleaving}{S}
\newcommand{\setinterleavingplus}{S^+}
\newcommand{\setinterleavingminus}{S^-}
\newcommand{\setinterleavingval}{\setinterleaving^\textval}
\newcommand{\setinterleavingvalplus}{\setinterleaving^{\textval+}}
\newcommand{\setinterleavingvalminus}{\setinterleaving^{\textval-}}

\newcommand{\mtxdiff}{U}
\newcommand{\valdiff}{u}
\newcommand{\diffmax}{\valdiff_\textmax}
\newcommand{\diffmin}{\valdiff_\textmin}
\newcommand{\biasat}[1]{\bias^{(#1)}}
\newcommand{\biasmtxat}[1]{\biasmtx^{(#1)}}

\newcommand{\idxgroupconstfrac}{{\idxgroup_\mathrm{f}}}

\newcommand{\setpairindices}{S}
\newcommand{\setpairindicesnonoverlap}{\setpairindices'}
\newcommand{\setpairrank}{R}
\newcommand{\pdf}{g}
\newcommand{\cdf}{G}
\newcommand{\invcdf}{\cdf^{-1}}
\newcommand{\cdfinv}{\invcdf}

\newcommand{\setinterleavingpairs}{S_\mathrm{pairs}}

\newcommand{\marginpair}{\gamma}
\newcommand{\numnonoverlappairs}{t}
\newcommand{\numremaining}{T}

\newcommand{\meanblock}{\mu}

\newcommand{\unifpos}{V}
\newcommand{\unifneg}{U}
\newcommand{\meanunifneg}{\overline{\unifneg}}
\newcommand{\meanunifpos}{\overline{\unifpos}}
\newcommand{\rvunif}{X}

\newcommand{\rvunifmin}{\rvunif_\textmin}
\newcommand{\unifposmin}{\unifpos_\textmin}
\newcommand{\unifnegmax}{\unifneg_\textmax}
\newcommand{\setcomplement}[1]{\overline{#1}}
\newcommand{\minstat}{\wedge}
\newcommand{\maxstat}{\vee}

\newcommand{\soliso}{u}
\newcommand{\solopt}{\soliso^*}
\newcommand{\yseq}[1]{\soliso^{(#1)}}
\newcommand{\numpava}{T}
\newcommand{\idxpava}{t}

\newcommand{\setpartition}{P}

%% file: text_main.tex
\section{Introduction}

It is common to aggregate information and evaluate items by collecting ratings on these items from people. In this work, we focus on the bias introduced by people's observable outcome or experience from the entity under evaluation, and we call it the ``outcome-induced bias''. Let describe this notion of bias with the help of two common applications -- teaching evaluation and peer review.

Many universities use student ratings for teaching evaluation. However, numerous studies have shown that student ratings are affected by the grading policy of the instructor~\citep{greenwald1997leniency,johnson2003inflation,boring2016effectiveness}. For instance, as noted in~\citet[Chapter 4]{johnson2003inflation}:%
\begin{quotex}
\emph{``...the effects of grades on teacher-course evaluations are both substantively and statistically important, and suggest that instructors can often double their odds of receiving high evaluations from students simply by awarding A's rather than B's or C's.''}
\end{quotex}
As a consequence, the association between student ratings and teaching effectiveness can become  negative~\citep{boring2016effectiveness}, and student ratings  serve as a poor predictor on the follow-on course achievement of the students~\citep{carrell2008assignment,braga2014evaluation}:
\begin{quotex}
\emph{``...teachers who are associated with better subsequent
performance receive worst evaluations from their students.''}~\citep{braga2014evaluation}
\end{quotex}
The outcome we consider in teaching evaluation is the grades that the students receive in the course under evaluation\footnote{We use the term ``grades'' broadly to include letter grades, numerical scores, and rankings. We do not distinguish the difference between evaluation of a course and evaluation of the instructor teaching the course, and use them interchangeably.} and the goal is to correct for the bias in student evaluations induced by the grades given by the instructor.

An analogous issue arises in conference peer review, where conference organizers survey authors to rate their received reviews in order to understand the quality of the review process. It is well understood that authors are more likely to give higher ratings to a positive review than a to negative review~\citep{weber2002author,apagiannaki2007pam,khosla2012cvpr}:
\begin{quotex}
\emph{``Satisfaction had a strong, positive association with acceptance of the manuscript for publication... Quality of the review of the manuscript was not associated with author satisfaction.''}~\citep{weber2002author}
\end{quotex}
Due to this problem, an author feedback experiment~\citep{apagiannaki2007pam} conducted at the PAM 2007 conference concluded that:
\begin{quotex}
\emph{``...some of the TPC members from academia paralleled the collected feedback to faculty evaluations within universities... while author feedback may be useful in pinpointing extreme cases, such as exceptional or problematic reviewers, it is not quite clear how such feedback could become an integral part of the process behind the organization of a conference.''}
\end{quotex}
With this motivation, for the application of peer review, the outcome we consider is the review rating or paper decision received by the author, and the goal is to correct for the bias induced by it in the feedback provided by the author.

Although the existence of such bias is widely acknowledged, student and author ratings are still widely used~\citep{becker1999economics}, and such usage poses a number of issues. First, these biased ratings can be uninformative and unfair for instructors and reviewers who are not lenient. Second, instructors, under the possible consideration of improving their student-provided evaluation, may be incentivized to ``teach to the test'', raising concerns such as inflating grades and reducing content~\citep{carrell2008assignment}. Furthermore, author-provided ratings can be a factor for selecting reviewer awards~\citep{khosla2012cvpr}, and student-provided ratings can be a heavily-weighted component for salary or promotion and tenure decision of the faculty members~\citep{becker1999economics,carrell2008assignment,boring2016effectiveness}. If the ratings are highly unreliable and sometimes even follow a trend that reverses the true underlying ordering, then na\"ively using these ratings or simply taking their mean or median will not be sufficient. Therefore, interpreting and correcting these ratings properly is an important and practical problem.

The goal of this work is to mitigate such outcome-induced bias in ratings. Incidentally, in teaching evaluation and peer review, the ``outcome'' that people (students or authors) encounter in the process is the evaluation they receive (grades from instructors or reviews from reviewers), and hence we call this bias ``evaluations that are biased by evaluations''. That said, we note that the general problem we consider here is applicable to other settings with outcomes that are not necessarily evaluations. For example, in evaluating whether a two-player card game is fair or not, the outcome can be whether the player wins or loses the game~\citep{molina2019game}.

The key insight we use in this work is that the outcome (e.g., grades and paper decisions) is naturally available to those conduct the evaluation (e.g., universities and conference organizers). These observed outcomes provide directional information about the manner that evaluators are likely to be biased. For example, it is known~\cite{greenwald1997leniency,johnson2003inflation,boring2016effectiveness} that students receiving higher grades are biased towards being more likely to give  higher ratings to the course instructor than students receiving lower grades. To use this structural information, we model it as a known partial ordering constraint on the biases given people's different outcomes. This partial ordering, for instance, is simply a relation on the students based on their grades or ranking, or on the authors in terms of acceptance decisions of their papers. 

\subsection{Our contributions}

We identify and formulate a problem of mitigating biases in evaluations that are biased by evaluations (Section~\ref{sec:formulation}). Specifically, this bias is induced by observable outcomes, and the outcomes are formulated as a known partial ordering constraint. We then propose an estimator that solves an optimization jointly in the true qualities and the bias, under the given ordering constraint (Section~\ref{sec:estimator}). The estimator includes a regularization term that balances the emphasis placed on bias versus noise. To determine the appropriate amount of regularization, we further propose a cross-validation algorithm that chooses the amount of regularization in a data-dependent manner by minimizing a carefully-designed validation error (Section~\ref{sec:cv}).

We then provide a theoretical analysis of the performance of our proposed algorithm (Section~\ref{sec:theory}). First, we show that our estimator, under the two extremal choices of the regularization hyperparameter ($0$ and $\infty$), converges to the true value in probability under only-bias (Section~\ref{sec:theory_bias_only_consistent}) and only-noise (Section~\ref{sec:theory_noise_only_minimax}) settings respectively. Moreover, our estimator reduces to the popular sample-mean estimator when the regularization hyperparameter is set to $\infty$, which is known to be minimax-optimal in the only-noise case.  We then show (Section~\ref{sec:theory_cv}) that the cross-validation algorithm correctly converges to the solutions corresponding to hyperparameter values of $0$ and $\infty$ in probability in the two aforementioned settings, under various conditions captured by our general formulation. We finally conduct synthetic and semi-synthetic experiments that establish the effectiveness of our proposed approach via numerical experiments in more general settings not covered by the theoretical results (Section~\ref{sec:experiment}). 

\subsection{Related work}

In terms of correcting rating biases, past work has studied the problem of adjusting student GPAs due to different grading policies across courses and disciplines. Proposed models include introducing a single parameter for each course and each student solved by linear regression~\citep{caulkins1995gpa}, and more complicated parametric generative models~\citep{johnson1997gpa}. Though grade adjustment seems to be a perfect counterpart of teaching evaluation adjustment, the non-parametric ordering constraint we consider is unique to teaching evaluation, and do not have obvious counterpart in grade adjustment. For the application of peer review, there are many works~\cite{ge13bias,lee2015commensuration,tomkins2017reviewer,noothigattu2018choosing,stelmakh2018forall,wang2018your,stelmakh2019testing,fiez2020super,jecmen2020manipulation,manzoor2020uncovering} addressing various biases and other issues in the review process, but to the best of our knowledge none of them addresses biases in author-provided feedback. It is of interest in the future to design schemes that combine our present work with these past works in order to jointly address multiple problems such as simultaneous existence of outcome-dependent bias and miscalibration.

In terms of the models considered, one statistical problem related to our work is the isotonic regression, where the goal is to estimate a set of parameters under a total ordering constraint (see, e.g.~\citealp{barlow1972isotonic,zhang2002risk,mammen2007additive,groeneboom2014nonparametric}). 
Specifically, our problem becomes isotonic regression, if in our exact formulation~\eqref{eq:optimization} to be presented, we set $\reg =0, \meancourse = 0$ and the partial ordering to a total ordering. 

Another type of related models in statistics literature concerns the semiparametric additive models  (e.g.~\citealp{hastie1990generalized,cuzick1992semiparametric,wood2004stable,yu2011semi}) with shape constraints~\citep{chen2016generalized}. In particular, one class of semiparametric additive models involves linear components and components with ordering (isotonic) constraints~\citep{huang2002monotonic,cheng2009semiparametric,meyer2013semi,rueda2013degrees}. Our optimization~\eqref{eq:optimization} falls within this class of semiparametric models, if we set the second term of $\ell_2$-regularization to $0$. To see the connection, we write the first term of~\eqref{eq:optimization} in a linearized form as $\norm{\obs - \designmatrix \meancourse-\bias}_2^2$, where $\obs, \bias\in \reals^{\numcourse\numstudent}, \meancourse\in \reals^\numcourse$ and $\designmatrix\in \reals^{\numcourse\numstudent\times \numcourse}$ is a $0/1$ matrix that specifies the course membership of each rating: if a rating is from course $\idxcourse$, then in corresponding of row of $\designmatrix$, the $\idxcourse^{th}$ entry is $1$ and all other entries are $0$. Past work has studied the least-squares estimator for this problem, but the results such as consistency and asymptotic normality rely on assumptions such as $\designmatrix$ being random design or each coordinate of $\meancourse$ being i.i.d., which are not applicable to our setting. The special $0/1$ structure of $\designmatrix$ makes our problem unique and differ from past work in terms of the theoretical analysis.

In terms of the technical approach, our estimator (Equation~\ref{eq:optimization}) is partly inspired by permutation-based models~\cite{shah2017stochastically,shah2017learning} which focuses only on shape constraints rather than parameters, but with the key difference that here we can exploit the crucial information pertaining to the ordering of the bias. 

The idea of adopting cross-validation to select the right amount of penalization is classical
in statistics literature (see, e.g.~\cite{stone1974cross,kohavi1995study,hastie2009elements}). 
Yet, this generic scheme cannot be directly applied to models where training samples are not exchangeable---in which case, both the sub-sampling step and the test-error estimation are highly non-trivial. Therefore caution needs to be exercised when order restrictions, therefore non-exchangeability, are involved.
The cross-validation algorithm proposed in this work is partly inspired by the cross-validation used in nearly-isotonic regression~\citep{tibshirani2011nearly}. 
In nearly-isotonic regression, the hard ordering constraint is replaced by a soft regularization term, and the extent of regularization is determined by cross-validation. However, introducing the linear term of $\meancourse$ as the quantity of interest significantly changes the problem. Thus, our cross-validation algorithm and its analysis are quite different.

\section{Problem formulation}\label{sec:formulation}
For ease of exposition, throughout the paper we describe our problem formulation using the running example of course evaluation, but we note that our problem formulation is general and applies to other problems under outcome-induced bias as well. Consider a set of $\numcourse$ \courses. Each \course $\idxcourse \in [\numcourse]$ has an unknown true quality value $\truemean_\idxcourse\in \reals$ to be estimated. Each \course is evaluated by $\numstudent$ \students.\footnote{
    For ease of exposition, we assume that each course is evaluated by $\numstudent$ students, but the algorithms and the results extend to regimes where the number of students is different across courses. 
}
Denote $\obssingle_{\idxcourse\idxstudent}\in \reals$ as the rating given by the $\idxstudent^{th}$ student in course $\idxcourse$, for each $\idxcourse\in [\numcourse]$ and $\idxstudent\in [\numstudent]$. Note that we do not require the same set of $\numstudent$ students to take all $\numcourse$ courses; students in different courses are considered different individuals. We assume that each rating $\obssingle_{\idxcourse\idxstudent}$ is given by:
\begin{align}\label{eq:model}
    \obssingle_{\idxcourse\idxstudent} = \truemean_\idxcourse + \biassingle_{\idxcourse\idxstudent} + \noisesingle_{\idxcourse\idxstudent},
\end{align}
where $\biassingle_{\idxcourse\idxstudent}$ represents a bias term, and $\noisesingle_{\idxcourse\idxstudent}$ represents a noise term. 
 We now describe these terms in more detail.

The term $\noise_{\idxcourse\idxstudent}$ captures the noise involved in the ratings, assumed to be i.i.d. across $\idxcourse\in[\numcourse]$ and $\idxstudent\in [\numstudent]$.
The term $\bias_{\idxcourse\idxstudent}$ captures the bias that is induced by the observed ``outcome'' of \student $\idxstudent$ experienced in \course $\idxcourse$. In the example of teaching evaluation, the outcome can be the grades of the students that are known to the university, and the bias captures the extent that student ratings are affected by their received grades. Given these observed outcomes (grades), we characterize the information provided by these outcomes as a known partial ordering, represented by a collection of ordering constraints $\setpartialorder\subseteq ([\numcourse]\times [\numstudent])^2$. Each ordering constraint is represented by two pairs of $(\idxcourse, \idxstudent)$ indices.
An ordering constraint $((\idxcourse, \idxstudent), (\idxcourse', \idxstudent')) \in \setpartialorder$ indicates that the bias terms obey the relation $\bias_{\idxcourse\idxstudent} \le \bias_{\idxcoursealt\idxstudentalt}$. We say that this ordering constraint is on the \elements $\{\idxpairparen\}_{\idxcoursescope, \idxstudentscope}$ and on the bias $\{\bias_\idxpair\}_{\idxcoursescope,\idxstudentscope}$ interchangeably. We assume the terms $\{\bias_{\idxcourse\idxstudent}\}_{\idxcourse\in[\numcourse], \idxstudent \in [\numstudent]}$ satisfy the partial ordering $\setpartialorder$. In teaching evaluations, the partial ordering $\setpartialorder$ can be constructed by, for example, taking $((\idxcourse, \idxstudent), (\idxcoursealt, \idxstudentalt))\in \setpartialorder$ if and only if  student $\idxstudentalt$ in course $\idxcoursealt
$ receives a strictly higher grade than student $\idxstudent$ in course $\idxcourse$.

For ease of notation, we denote $\obsmtx \in \reals^{\numcourse\times \numstudent}$ as the matrix of observations whose $(\idxcourse,\idxstudent)^\thcount$ entry equals $\obssingle_\idxpair$ for every $\idxcourse\in [\numcourse]$ and $\idxstudent\in [\numstudent]$. We define matrices $\biasmtx\in \reals^{\numcourse\times\numstudent}$ and $\noisemtx\in \reals^{\numcourse\times\numstudent}$ likewise. We denote $\truemean\in \reals^\numcourse$ as the vector of $\{\truemean_\idxcourse\}_{\idxcourse\in [\numcourse]}$.

\paragraph{Goal.}
Our goal is to estimate the true quality values $\truemean \in \reals^\numcourse$. For model identifiability, we assume $\Expect[\noise_{\idxcourse\idxstudent}] = 0$ and $\sum_{\idxcourse\in [\numcourse], \idxstudent\in [\numstudent]}\Expect[\bias_{\idxcourse\idxstudent}] = 0$. An estimator takes as input the observations $\obsmtx$ and the partial ordering $\setpartialorder$, and outputs an estimate $\estmean\in \reals^\numcourse$. We measure the performance of any estimator in terms of its (normalized) squared $\ell_2$ error $\frac{1}{\numcourse}\norm{\estmean-\truemean}_2^2$.

\section{Proposed estimator}\label{sec:estimator}
Our estimator takes as input the observations $\obsmtx$ and the given partial ordering $\setpartialorder$. The estimator is associated with a tuning parameter $\reg \ge 0$, and is given by:
\begin{align}\label{eq:optimization}
    \estmeanat{\reg} \in
    \argmin_{ \meancourse\in \reals^\numcourse}
    \min_{\substack{
        \biasmtx\in \reals^{\numcourse\times \numstudent}\\
        \biasmtx \text{ satisfies } \setpartialorder}} 
    \norm{\obsmtx - \meancourse\vecone^T - \biasmtx}_F^2 + \reg \norm{\biasmtx}_F^2,
\end{align}
where $\vecone
$ denotes the all-one vector of dimension $\numstudent$. We let $\estbiasmtxat{\reg}$ denote the value of $\biasmtx$ that attains the minimum of the objective~\eqref{eq:optimization}, so that the objective~\eqref{eq:optimization} is minimized at $(\estmeanat{\reg}, \estbiasmtxat{\reg})$. Ties are broken by choosing the solution $(\meancourse, \biasmtx)$ such that $\biasmtx$ has the minimal Frobenius norm $\norm{\biasmtx}_F^2$. We show that the estimator under this tie-breaking rule defines a unique solution in Proposition~\ref{prop:uniqueness} in Appendix~\ref{app:property_of_estimator}. Furthermore, as explained in Appendix~\ref{app:simulation_implementation}, the optimization~\eqref{eq:optimization} is a convex quadratic programming (QP) in $(\meancourse, \biasmtx)$, and therefore can be solved in polynomial time in terms of $(\numcourse, \numstudent)$.

While the first term $ \norm{\obsmtx - \meancourse\vecone^T - \biasmtx}_F^2$ of~\eqref{eq:optimization} captures the squared difference between the bias-corrected observations $(\obsmtx - \biasmtx)$ and the true
qualities $\meancourse\vecone^T$, the second term $\norm{\biasmtx}_F^2$ captures the magnitude of the bias.
Since the observations in~\eqref{eq:model} include both the bias $\biasmtx$ and the noise $\noisemtx$, there is fundamental ambiguity pertaining to the relative contributions of the bias and noise to the observations.
The penalization parameter $\reg$ is introduced to balance the bias and the variance, and at the same time preventing overfitting to the noise. More specifically, consider the case when the noise level is relatively large and the partial ordering $\setpartialorder$ is not sufficiently restrictive --- in which case, it is sensible to select a larger $\reg$ to prevent $\biasmtx$ overly fitting the observations $\obsmtx$. 

For the rest of this section, we first describe intuition about the tuning parameter $\reg$ by considering two extreme choices of $\reg$ which are by themselves of independent interest. We then propose a carefully-designed cross-validation algorithm to choose the value of $\reg$ in a data-dependent manner. 

\subsection{Behavior of our estimator under some fixed choices of \texorpdfstring{$\reg$}{\unichar{"03BB}}}\label{sec:estimator_extremal}

To facilitate understandings of the estimator~\eqref{eq:optimization}, we discuss its behavior for two important choices of $\reg$ --- $0$ and $\infty$ --- that may be of independent interest. 

\paragraph{\boldmath{$\reg=0$}:} When $\reg=0$, intuitively the estimator~\eqref{eq:optimization} allows the bias term $\biasmtx$ to be arbitrary in order to best fit the data, as long as it satisfies the ordering constraint $\setpartialorder$.  Consequently with this choice, the estimator attempts to explain the observations $\obsmtx$ as much as possible in terms of the bias. One may use this choice if domain knowledge suggests that bias considerably dominates the noise. Indeed, as we show subsequently in Section~\ref{sec:theory_bias_only_consistent}, our estimator with $\reg=0$ is consistent in a noiseless setting (when only bias is present), whereas common baselines are not.

\paragraph{\boldmath{$\reg=\infty$}:} We now discuss the other extremity, namely when $\reg$ approaches infinity. Intuitively, this case sets the bias term to zero in~\eqref{eq:optimization} (note that $\estbiasmtx=0$ trivially satisfies any partial ordering $\setpartialorder$). Therefore, it aims to explain the observations in terms of the noise.  Formally we define $(\estmeanat{\infty}, \estbiasmtxat{\infty}) = \lim_{\reg\rightarrow \infty} (\estmeanat{\reg}, \estbiasmtxat{\reg})$. In the subsequent result of Proposition~\ref{prop:property_existence_at_infty}, we show that this limit exists, where we indeed have $\estbiasmtxat{\infty}=0$ and our estimator simply reduces to the sample mean as $[\estmeanat{\infty}]_\idxcourse= \frac{1}{\numstudent}\sum_{\idxstudent=1}^\numstudent \obs_{\idxcourse\idxstudent}$ for every $i \in [\numcourse]$.
We thus see that perhaps the most commonly used estimator for such applications --- the sample mean --- also lies in our family of estimators specified in~\eqref{eq:optimization}. Given the well-known guarantees of the sample mean in the absence of bias (under reasonable conditions of the noise), one may use this choice if domain knowledge suggests that noise is highly dominant as compared to the bias.

\paragraph{\boldmath{$\reg\in (0, \infty)$}:} More generally, the estimator interpolates between the behaviors at the two extremal values $\lambda = 0$ and $\infty$ when both bias and noise is present. As we increase $\reg$ from $0$, the magnitude of the estimated bias $\estbiasmtxat{\reg}$ gradually decreases and eventually goes to $0$ at $\reg=\infty$. The estimator hence gradually explains the observations less in terms bias, and more in terms of noise. Our goal is to choose an appropriate value for $\reg$, such that the contribution of bias versus noise determined by the estimator approximately matches the true relative contribution that generates the observations. The next subsection presents a principled method to choose the value for $\reg$.

\subsection{A cross-validation algorithm for selecting \texorpdfstring{$\reg$}{\unichar{"03BB}}}\label{sec:cv}

We now present a carefully designed cross-validation algorithm to select the tuning parameter $\reg$ in a data-driven manner. Our cross-validation algorithm determines an appropriate value of $\reg$ from a finite-sized set of candidate values $\setreg\subseteq [0, \infty]$ that is provided to the algorithm.
For any matrix $A\in \reals^{\numcourse\times \numstudent}$, we define its squared norm restricted to a subset of \elements $\set\subseteq [\numcourse]\times [\numstudent]$ as $\norm{A}_\set^2 = \sum_{(\idxcourse, \idxstudent)\in \set} A_{\idxcourse\idxstudent}^2$. Let $\settotalorder$ denote the set of all total orderings (of the $\numcourse\numstudent$ \elements) that are consistent with the partial ordering $\setpartialorder$.  
The cross-validation algorithm is presented in Algorithm~\ref{alg:cv}. It consists of two steps: a data-splitting step
 (Lines~\ref{line:split_start}-\ref{line:split_end}) and a validation step (Lines~\ref{line:validation_step_start}-\ref{line:validation_step_end}). 
 
\paragraph{Data-splitting step} In the data-splitting step, our algorithm splits the observations $\{\obs_{\idxpair}\}_{\idxcourse\in [\numcourse], \idxstudent\in [\numstudent]}$ into a training set $\settrain\subseteq [\numcourse]\times [\numstudent]$ and a validation set $\setval \subseteq [\numcourse] \times [\numstudent]$. 
To obtain the split, our algorithm first samples uniformly at random a total ordering $\totalordersplit$ from $\settotalorder$ (Line~\ref{line:sample_total_order}). 
For every course $\idxcourse\in [\numcourse]$, we find the sub-ordering of the $\numstudent$ \elements within this course  (that is, the ordering of the \elements $\{(\idxcourse, \idxstudent)\}_{\idxstudent\in [\numstudent]}$) according to $\totalordersplit$ (Line~\ref{line:find_sub_order}). For each consecutive pair of \elements in this sub-ordering, we assign one \element in this pair to the training set and the other \element to the validation set uniformly at random (Lines~\ref{line:assign_start}-\ref{line:assign_end}). 
We note that in comparison to classical cross-validation methods, our algorithm uses the total ordering $\totalordersplit$ to guide the split, instead of independently assigning each individual \element to either the training set or the validation set uniformly at random. 
This splitting procedure ensures that for each \element in the validation set there is an \element that is ``close'' in the training set with respect to the partial ordering $\setpartialorder$. This property is useful for interpolation in the subsequent validation step.

\input{text_algo}

\paragraph{Validation step} Given the training set and the validation set, our algorithm iterates over the choices of $\reg\in \setreg$ as follows. For each value of $\reg$, 
the algorithm first computes our estimator with penalization parameter $\reg$ on the training set $\settrain$ to obtain $(\estmeanat{\reg},\estbiasmtxat{\reg})$. 
The optimization (Line~\ref{line:compute_err_start}) is done by replacing the Frobenius norm on the two terms in the original objective~\eqref{eq:optimization} by the Frobenius norm restricted to $\settrain$. Note that this modified objective is independent from the parameters $\{\bias_\idxpair\}_{\idxpairparen\in \setval}$. Therefore, by the tie-breaking rule of minimizing $\normfro{\estbiasmtxat{\reg}}$, we have $[\estbiasmtxat{\reg}]_\idxpair = 0$ for each $\idxpairparen\in \setval$.

Next, our algorithm evaluates these choices of $\reg$ by their corresponding cross-validation (CV) errors. 
The high-level idea is to evaluate the fitness of $(\estmeanat{\reg}, \estbiasmtxat{\reg})$ to the validation set $\setval$, by computing $\frac{1}{\sizesetval} \norm{\obsmtx- \estmeanat{\reg}\vecone^T - \estbiasmtxat{\reg}}_\setval^2$. 
However, recall that the estimate $\estbiasmtxat{\reg}$ only estimates the bias on the training set meaningfully, and we have $\estbiasmtxat{\reg}_\idxpair = 0$ for each \element $\idxpairparen$ in the validation set $\setval$. Therefore, we ``synthesize'' the estimated bias $\estbiasmtxvalat{\reg}$ on the validation from the estimated bias $\estbiasmtxat{\reg}$ on the training set via an interpolation procedure (Lines~\ref{line:iterate_all_val_entry}-\ref{line:interpolation_end}), as explained below.

\paragraph{Interpolation} We now discuss how the algorithm interpolates the bias $\estbiasvalat{\reg}_\idxpair$ at each \element $\idxpairparen\in \setval$ from $\estbiasmtxat{\reg}$. We first explain how to perform interpolation with respect to some given total ordering $\totalorder$ (Line~\ref{line:interpolation_per_total_order}), and then compute a mean of these interpolations by iterating over $\totalorder\in \settotalorder$ (Line~\ref{line:interpolation}).

\begin{itemize}
    \item \textbf{Interpolating with respect to a total ordering (Line~\ref{line:interpolation_per_total_order}):}
    Given some total ordering $\totalorder$, we find the \element in the training set that is the closest to $\idxpairparen$ in the total ordering $\totalorder$. We denote this closest \element from the training set as $(\idxcoursenn, \idxstudentnn)$, and simply interpolate the bias at $\idxpairparen$ with respect to $\totalorder$ (denoted $[\estbiasvalat{\reg}_{\totalorder}]_\idxpair$) using the value of $\estbias_{\idxcoursenn\idxstudentnn}$. That is, we set $[\estbiasvalat{\reg}_{\totalorder}]_\idxpair = \estbiasat{\reg}_{\idxcoursenn\idxstudentnn}$. If there are two closest \elements of equal distance to $\idxpairparen$ (one ranked higher than $\idxpairparen$ and one lower than $\idxpairparen$ in $\totalorder$), we use the mean of the estimated bias $\estbiasmtxat{\reg}$ of these two \elements. This step is similar to the CV error computation in~\cite{tibshirani2011nearly}.
    
    \item  \textbf{Taking the mean over all total orderings in $\settotalorder$ (Line~\ref{line:interpolation}):}
    After we find the interpolated bias $\estbiasmtxvalat{\reg}_\totalorder$ on the validation set with respect to each $\totalorder$, the final interpolated bias $\estbiasvalat{\reg}$ is computed as the mean of the interpolated bias over all total orderings $\totalorder\in \settotalorder$. The reason for taking the mean over $\totalorder\in \settotalorder$ is as follows.
    When we interpolate by sampling a single ordering $\totalorder\in \settotalorder$, this sampling of the ordering introduces randomness in terms of which training \elements are chosen for which validation \elements, and hence increasing the variance of the CV error.\footnote{
        In more detail, this variance on the CV error due to sampling causes the algorithm to choose an excessively large $\reg$ to underestimate the bias. A large $\reg$ shrinks the the magnitude of the estimated bias towards $0$, and therefore the estimated bias becomes closer to each other, reducing this variance -- in the extreme case, if the estimated bias is $0$ on all \elements from the training set, then the interpolated bias is $0$ in the validation set regardless of the ordering $\totalorder$, giving no variance due to sampling $\totalorder$.
    }
    Taking the mean over all total orderings eliminates this source of the variance of the CV error due to sampling, and therefore leads to a better choice of $\reg$.

\end{itemize}

After interpolating the bias $\estbiasmtxvalat{\reg}$ on the validation set, the CV error is computed as $\frac{1}{\sizesetval}\norm{\obsmtx -\estmeanat{\reg}\vecone^T- \estbiasmtxvalat{\reg})}_\setval$ (Line~\ref{line:compute_err_end}). Finally, the value of $\regcv\in \setreg$ is chosen by minimizing the CV error (with ties broken arbitrarily). This completes the description of the cross-validation algorithm. 

\paragraph{Implementation} Now we comment on two important operations in Algorithm~\ref{alg:cv}: sampling a total ordering from the set $\settotalorder$ of total orderings consistent with the partial ordering $\setpartialorder$ (Line~\ref{line:sample_total_order}), and iterating over the set $\settotalorder$ (Line~\ref{line:iterate_all_total_orders}).
For sampling a total ordering from $\settotalorder$ uniformly at random, many algorithms have been proposed that are approximate~\citep{matthews1991ordering,bubley1999ordering} or exact~\citep{huber2006perfect}. For iterating over $\settotalorder$ which can be computationally intractable, we approximate the true mean over $\settotalorder$ by sampling from $\settotalorder$ multiple times, and take their empirical mean. 
In many practical settings, the partial ordering contains a structure on which these two operations are simple to implement and run in polynomial time -- we discuss a subclass of such partial orderings termed ``group orderings'' in the theoretical results (Section~\ref{sec:theory_preliminaries}); this subclass of partial orderings is also evaluated in the experiments (Section~\ref{sec:experiment}).

\section{Theoretical guarantees}\label{sec:theory}

We now present theoretical guarantees for our proposed estimator (cf. \eqref{eq:optimization}) along with our cross-validation algorithm (Algorithm~\ref{alg:cv}). In Section~\ref{sec:theory_bias_only_consistent} and~\ref{sec:theory_noise_only_minimax}, we establish properties of our estimator at the two extremal choices of $\reg$ ($\reg=0$ and $\reg=\infty$) for no noise  and no bias settings respectively. Then in Section~\ref{sec:theory_cv}, we analyze the cross-validation algorithm. The proofs of all results are in Appendix~\ref{app:proof}.

\subsection{Preliminaries}\label{sec:theory_preliminaries}

\paragraph{Model assumptions:} To introduce our theoretical guarantees, we start with several model assumptions that are used throughout the theoretical result of this paper. 
Specifically, we make the following assumptions on the model~\eqref{eq:model}:
\begin{enumerate}[label={(A\arabic*)}]
    \item \label{assumption:noise}
    \textbf{Noise:} The noise terms $\{\noise_{\idxcourse\idxstudent}\}_{\idxcourse\in [\numcourse], \idxstudent\in [\numstudent]}$ are i.i.d.  $\normal(0, \gaussianwidthnoise^2)$ for some constant $\gaussianwidthnoise \ge 0$. 
    
    \item \label{assumption:bias}
    \textbf{Bias:} The bias terms $\{\bias_{\idxpair}\}_{\idxcourse\in [\numcourse], \idxstudent\in [\numstudent]}$ are marginally distributed as $\normal(0, \gaussianwidthbias^2)$ for some constant $\gaussianwidthbias \ge 0$ unless specified otherwise, and obey one of the total orderings (selected uniformly at random from the set of total orderings) consistent with the partial ordering $\setpartialorder$. That is, we first sample $\numcourse\numstudent$ values i.i.d. from $\normal(0, \gaussianwidthbias^2)$, and then sample one total ordering uniformly at random from all total orderings consistent with the partial ordering $\setpartialorder$. Then we assign these $\numcourse\numstudent$ values to $\{\bias_{\idxpair}\}$ according to the sampled total ordering.
    
    \item \label{assumption:d}
    \textbf{Number of courses:} The number of courses $\numcourse$ is assumed to be a fixed constant.
\end{enumerate}
All theoretical results hold for any arbitrary $\truemean\in \reals^{\numcourse}$.
It is important to note that the estimator~\eqref{eq:optimization} and the cross-validation algorithm (Algorithm~\ref{alg:cv}) requires no knowledge of these distributions or standard deviation parameters $\gaussianwidthbias$ and $\gaussianwidthnoise$. 

Throughout the theoretical results, we consider the solution $\estmeanat{\regcv}$ as solution at $\reg=\regcv$ on the training set.

Our theoretical analysis focuses on a general subclass of partial orderings, termed ``group orderings'', where each rating belongs to a group, and the groups are totally ordered.
\begin{definition}[Group ordering]\label{def:group_ordering}
A partial ordering $\setpartialorder$ is called a group ordering with $\numgroup$ groups if there is a partition $\setgroupof{1}, \ldots, \setgroupof{\numgroup} \subseteq[\numcourse]\times [\numstudent]$ of the $\numcourse\numstudent$ ratings such that $((\idxcourse, \idxstudent), (\idxcoursealt, \idxstudentalt))\in \setpartialorder$ if and only if $(\idxcourse, \idxstudent) \in \setgroupof{\idxgroup}$ and $(\idxcoursealt, \idxstudentalt)\in \setgroupof{\idxgroupalt}$ for some $1 \leq \idxgroup < \idxgroupalt \leq \numgroup$. 
\end{definition}
Note that in Definition~\ref{def:group_ordering}, if two samples are in the same group, we do not impose any relation restriction between these two samples.

Group orderings arise in many practical settings. For example, in course evaluation, the groups can be letter grades (e.g., $\{\text{A}, \text{B}, \text{C}, \text{D}, \text{F}\}$ or $\{\text{Pass}, \text{Fail}\}$), or numeric scores (e.g., in the range of $[0, 100]$) of the students. The group ordering intuitively says that a student receiving a strictly higher \grade is more positively biased in rating than a student receiving a lower \grade. A total ordering is also group ordering, with the number of groups equal to the number of samples. We assume that the number of groups is $\numgroup \geq 2$ since otherwise groups are vacuous.

Denote $\blocklength_{\idxcourse\idxgroup}$ as the number of students of group $\idxgroup \in [\numgroup]$ in course $\idxcourse \in [\numcourse]$. We further introduce some regularity conditions used in the theoretical results.
The first set of regularity conditions is motivated from the case where students receive a discrete set of letter grades. 
\begin{definition}[Group orderings with the single constant-fraction assumption]
A group ordering is said to satisfy the single $\const$-fraction assumption for some constants $\const\in (0, 1)$ if there exists some group $\idxgroup\in [\numgroup]$ such that $\blocklength_{\idxcourse\idxgroup} >  \const\numstudent~\forall~\idxcourse \in [\numgroup]$.
\end{definition}

\begin{definition}[Group orderings with the all constant-fraction assumption]\label{def:all_constant_fraction}
A group ordering of $\numgroup$ groups is said to satisfy the all $\const$-fraction assumption for some constant $\const\in (0, \frac{1}{\numgroup})$, if $ \blocklength_{\idxcourse\idxgroup} \ge \const\numstudent~\forall~\idxcourse\in [\numcourse],\ \idxgroup\in [\numgroup]$.
\end{definition}
Note that group orderings with all $\const$-fractions is a subset of group orderings with single $\const$-fraction.
The final regularity condition below is motivated from the scenario where student performances are totally ranked in the course.
\begin{definition}[Total orderings with the constant-fraction interleaving assumption]\label{def:interleaving}
    Let $\setpartialorder$ be a total ordering (of the $\numcourse\numstudent$ \elements $\{\idxpairparen\}_{\idxcoursescope, \idxstudentscope}$). We define an interleaving point as any number $\rank\in [\numcourse\numstudent-1]$, such that the $\rank^{\thcount}$ and the $(\rank + 1)^{\thcount}$ highest-ranked \elements according to the total ordering $\setpartialorder$ belong to different courses. 
    A total ordering $\setpartialorder$ is said to satisfy the $\const$-fraction interleaving assumption for some constant $\const \in (0, 1)$, if there are at least $\const\numstudent$ interleaving points in $\setpartialorder$.
\end{definition}

With these preliminaries in place, we now present our main theoretical results. 

\subsection{\texorpdfstring{$\reg=0$}{\unichar{"03BB}=0} is consistent when there is no noise}\label{sec:theory_bias_only_consistent}

We first consider the extremal case where there is only bias but no noise involved. 
The following theorem states that our estimator with $\reg=0$ is consistent in estimating the underlying quantity $\truemean$, that is $\estmeanat{0}\rightarrow \truemean$ in probability.

\begin{theorem}\label{thm:consistency}[Consistency in estimating $\truemean$]
    Suppose the assumptions~\ref{assumption:noise},~\ref{assumption:bias} and~\ref{assumption:d} hold. Suppose there is no noise, or equivalently suppose $\gaussianwidthnoise=0$ in~\ref{assumption:noise}. Consider any $\truemean\in\reals^\numcourse$.
    Suppose the partial ordering is one of:
    \begin{enumerate}[label={(\alph*)}]
        \item \label{part:thm_consistency_constant_fraction}
        any group ordering of $\numgroup$ groups satisfying the all $\const$-fraction assumption, where $\const\in (0, \frac{1}{\numgroup}]$ is a constant, or

        \item \label{part:thm_consistency_binary}
        any group ordering with $\numcourse=2$ courses and $2$ groups, or
        
        \item \label{part:thm_consistency_total_order}
        any total ordering. 
    \end{enumerate}
    Then for any $\errbound>0$ and $\probbound>0$, there exists an integer $\numstudentlb$ (dependent on $\errbound, \probbound, \const, \numcourse, \gaussianwidthnoise$), such that for every $\numstudent\ge \numstudentlb$ and every partial ordering satisfying at least one of the conditions \ref{part:thm_consistency_constant_fraction}, \ref{part:thm_consistency_binary} or \ref{part:thm_consistency_total_order}:
    \begin{align*}
        \Prob\Big(\norm{\estmeanat{0} - \truemean}_2 < \errbound\Big) \ge 1- \probbound.
    \end{align*}
\end{theorem}
The proof of this result is provided in Appendix~\ref{app:proof_thm_consistency}. The convergence of the estimator to the true qualities $\truemean$ implies the following corollary on ranking the true qualities $\truemean$. In words, our estimator $\estmeanat{0}$ is consistent in comparing the true qualities $\truemean_\idxcourse$ and $\truemean_\idxcoursealt$ of any pair of courses $\idxcourse, \idxcoursealt\in [\numcourse]$ with $\idxcourse\ne \idxcoursealt$, as long as their values are distinct.
\begin{corollary}[Consistency on the ranking of $\truemean$]\label{cor:consistency_comparison}
Suppose the assumptions~\ref{assumption:noise},~\ref{assumption:bias} and~\ref{assumption:d} hold. Consider any $\truemean\in\reals^\numcourse$. Assume there is no noise, or equivalently assume $\gaussianwidthnoise=0$ in~\ref{assumption:noise}.
Then for any $\probbound > 0$, there exists an integer $\numstudentlb$ (dependent on $\truemean, \probbound, \const, \numcourse, \gaussianwidthnoise$),  such that for all $\numstudent\ge \numstudentlb$ and every partial ordering satisfying at least one of the conditions (a), (b) or (c) in Theorem~\ref{thm:consistency}:
\begin{align*}
   \Prob\Big(\sign(\estmean_\idxcourse - \estmean_\idxcoursealt) = \sign(\truemean_\idxcourse - \truemean_\idxcoursealt)\Big) \ge 1-\probbound \qquad \textrm{for all $\idxcourse, \idxcoursealt \in [\numcourse]$ such that $\idxcourse\ne \idxcoursealt$ and $\truemean_\idxcourse \ne \truemean_{\idxcoursealt}$}.
\end{align*}
\end{corollary}

In Appendix~\ref{app:mean}, we also evaluate the mean estimator. We show that under the conditions of Theorem~\ref{thm:consistency}, the mean estimator is provably not consistent. This is because the mean estimator does not account for the biases and only tries to correct for the noise. In order to obtain a baseline that accommodates the outcome-dependent bias (since to the best of our knowledge there is no prior literature on it), in Appendix~\ref{app:subsample} we then propose a \subsampling estimator. It turns out that our estimator at $\reg=0$ also theoretically outperforms this \subsampling estimator (see Proposition~\ref{prop:uniform_example} in Appendix~\ref{app:subsample}). 

\subsection{\texorpdfstring{$\reg=\infty$}{\unichar{"03BB}=\unichar{"221E}} is minimax-optimal when there is no bias}\label{sec:theory_noise_only_minimax}

We now move to the other extremity of $\reg = \infty$, and consider the other extremal case when there is only noise but no bias. Recall that we define the estimator at $\reg=\infty$ as $\estmeanat{\infty} = \lim_{\reg\rightarrow \infty} \estmeanat{\reg}$. The following proposition states that this limit is well-defined, and our estimator reduces to taking the sample mean at this limit.

\begin{proposition}[Estimator at $\reg= \infty$]\label{prop:property_existence_at_infty}
    The limit of $(\estmeanat{\infty}, \estbiasmtxat{\infty})\defn \lim_{\reg\rightarrow \infty} (\estmeanat{\reg}, \estbiasmtxat{\reg})$ exists and is given by
    \begin{align}
        \begin{split}\label{eq:properties_expression_at_infty}
        [\estmeanat{\infty}]_\idxcourse & = \frac{1}{\numstudent} \sum_{\idxstudent=1}^\numstudent \obs_{\idxcourse\idxstudent}, \qquad \text{for each $\idxcourse\in [\numcourse]$, and}\\
        \estbiasmtxat{\infty} & = 0.
        \end{split}
    \end{align}
\end{proposition}
The proof of this result is provided in Appendix~\ref{app:proof_prop_property_existence_at_infty}. With no bias, estimating the true quality $\truemean$ reduces to estimating the mean of a multivariate normal distribution with the covariance matrix $\gaussianwidthnoise^2 \identitymtx_\numcourse$, where $\identitymtx_\numcourse$ denotes the identity matrix of size $\numcourse\times \numcourse$. Standard results in the statistics literature imply that taking the sample mean is minimax-optimal in this setting if $\numcourse$ is a fixed dimension, formalized in the following proposition for completeness.
    \begin{proposition}[Implication of Example 15.8 in~\citealp{wainwright2019nonasymptotic}]\label{prop:sample_mean_minimax_optimal}
    Let $\numcourse\ge 1$ be a fixed constant.
    Let $\obsmtx = \truemean\vecone^T + \noisemtx$, where $\truemean\in \reals^{\numcourse}$ is an unknown vector and each entry of $\noisemtx$ is i.i.d. $\gaussian(0, \gaussianwidthnoise^2)$ with unknown $\gaussianwidthnoise$. Then the sample mean estimator $\estmean = \frac{1}{\numstudent}\obsmtx \vecone$ is minimax-optimal for the squared $\ell_2$-risk $\frac{1}{\numcourse}\Expect\norm{\estmean - \truemean}_2^2$, up to a constant factor that is independent of $\numcourse$.
    \end{proposition}

This concludes the properties of our estimator at the two extremal cases.

\subsection{Cross-validation effectively selects  \texorpdfstring{$\reg$}{\unichar{"03BB}}}\label{sec:theory_cv}

This section provides the theoretical guarantees for our proposed cross-validation algorithm. Specifically, we show that in the two extremal cases, cross-validation outputs a solution that converges in probability to the solutions at $\reg = 0$ and $\reg=\infty$, respectively. Note that the cross-validation algorithm is agnostic to the values of $\gaussianwidthbias$ and $\gaussianwidthnoise$, or any specific shape of the bias or the noise. 

The first result considers the case when there is only bias and no noise, and we show that \cv obtains a solution that is close to the solution using a fixed choice of $\reg=0$. The intuition for this result is as follows. The CV error $\norm{\obsmtx - \estmeanat{\reg} \vecone^T - \estbiasmtxvalat{\reg}}_{\setval}^2$ measures the difference between the bias-corrected observations $\obsmtx - \estbiasmtxvalat{\reg}$ and the estimated qualities $\estmeanat{\reg}\vecone^T$. By construction, the values in $\estmeanat{\reg}\vecone^T$ are identical within each row. Hence, to minimize the CV error we want $\estbiasmtxvalat{\reg}$ to capture as much variance as possible within each row of $\obsmtx$. Now consider $\reg= 0$. In this case $\estbiasmtxat{\reg}$ correctly captures the intra-course variance of the bias on the training set due to the noiseless assumption. Due to the nearest-neighbor interpolation, we expect that the interpolated $\estbiasmtxvalat{\reg}$ captures most of the intra-course variance of the bias on the validation set, giving a small CV error. However, for larger $\reg > 0$, the bias estimated from the training set shrinks in magnitude due to the regularization term. The bias $\estbiasmtxat{\reg}$ and hence $\estbiasmtxvalat{\reg}$ only capture a partial extent of the actual bias in the observations. The rest of the uncaptured bias within each course contributes to the residue $\norm{\obsmtx - \estmeanat{\reg} \vecone^T - \estbiasmtxvalat{\reg}}_{\setval}^2$, giving a larger CV error. Hence, \cv is likely to choose $\reg=0$ (or some sufficiently small value of $\reg$). The following theorem shows that cross-validation is consistent in estimating $\truemean$ under the only-bias setting.

\begin{theorem}\label{thm:cv_bias_only}

Suppose the assumptions~\ref{assumption:noise},~\ref{assumption:bias} and~\ref{assumption:d} hold. Consider any $\truemean\in\reals^\numcourse$. Suppose there is no noise, or equivalently suppose $\gaussianwidthnoise=0$ in~\ref{assumption:noise}. Suppose $\const\in (0, 1)$ is a constant. Suppose the partial ordering is either:
\begin{enumerate}[label={(\alph*)}]
    \item \label{part:cv_bias_only_constant_fraction}
    any group ordering satisfying the all $\const$-fraction assumption, or
    \item \label{part:cv_bias_only_total}
    any total ordering with $\numcourse=2$.
\end{enumerate}
Let $0\in \setreg$.
Then for any $\probbound>0$ and $\errbound > 0$, there exists some integer $\numstudentlb$  (dependent on $\errbound, \probbound, \const, \numcourse, \gaussianwidthbias$), such that
for every $\numstudent \ge \numstudentlb$ and every partial ordering satisfying \ref{part:cv_bias_only_constant_fraction} or \ref{part:cv_bias_only_total}:
\begin{align*}
\prob \Big(\norm{\estmeanat{\regcv} - \truemean}_2 < \epsilon\Big) \geq 1-\delta.
\end{align*}
\end{theorem}
The proof of this result is provided in Appendix~\ref{app:proof_thm_cv_bias_only}. From Theorem~\ref{thm:consistency} we have that the estimator $\estmeanat{0}$ (at $\reg=0$) is also consistent under the only-bias setting. Combining Theorem~\ref{thm:consistency} with Theorem~\ref{thm:cv_bias_only}, we have $\estmeanat{\regcv}$ approaches $\estmeanat{0}$. Formally, under the conditions of Theorem~\ref{thm:cv_bias_only}, we have
\begin{align*}
    \prob \Big(\norm{\estmeanat{\regcv} - \estmeanat{0}}_2 < \epsilon\Big) \geq 1-\delta.
\end{align*}

The next result considers the case when there is only noise and no bias, and we show that \cv obtains a solution that is close to the solution using a fixed choice of $\reg=\infty$ (sample mean). Intuitively, at small values of $\reg$ the estimator still tries to estimate a non-trivial amount of the interpolated bias $\estbiasmtxvalat{\reg}$. However, any such non-trivial interpolated bias is erroneous since there is no bias in the observations to start with, increasing the CV error $\norm{\obsmtx - \estmeanat{\reg} \vecone^T - \estbiasmtxvalat{\reg}}_{\setval}^2$ by doing a wrong bias ``correction''. On the other hand, at $\reg=\infty$ (or some $\reg$ that is sufficiently large), the interpolated bias $\estbiasmtxvalat{\reg}$ is zero (or close to zero), which is the right thing to do and hence gives a smaller CV error. The following theorem shows that cross-validation is consistent in estimating $\truemean$ under the only-noise setting.

\begin{theorem}\label{thm:cv_noise_only}
Suppose the assumptions~\ref{assumption:noise},~\ref{assumption:bias} and~\ref{assumption:d} hold. Consider any $\truemean\in\reals^\numcourse$. Suppose there is no bias, or equivalently assume $\gaussianwidthbias=0$ in~\ref{assumption:bias}. Suppose $\const_1, \const_2\in (0, 1)$ are constants. Suppose the partial ordering is either:
\begin{enumerate}[label={(\alph*)}]
    \item \label{part:cv_noise_only_constant_fraction}
    any group ordering satisfying the single $\const_1$-fraction assumption, or
    
    \item \label{part:cv_noise_only_total}
    any total ordering satisfying the $\const_2$-fraction interleaving assumption with $\numcourse=2$.
\end{enumerate}
Let $\infty\in \setreg$.
Then for any $\delta>0$ and $\epsilon > 0$, there exists some integer $\numstudentlb$  (dependent on $\errbound, \probbound, \const_1, \const_2, \numcourse, \gaussianwidthnoise$), such that for every $\numstudent \ge \numstudentlb$ and every partial ordering satisfying \ref{part:cv_noise_only_constant_fraction} or \ref{part:cv_noise_only_total}:
\begin{align*}
    \Prob\Big(\norm{\estmeanat{\regcv} - \truemean}_2 < \epsilon \Big) \ge 1-\probbound.
\end{align*}
\end{theorem}
The proof of this result is provided in Appendix~\ref{app:proof_thm_cv_noise_only}.
By the consistency of $\estmeanat{\infty}$ implied from Proposition~\ref{prop:sample_mean_minimax_optimal} under the only-noise setting, this result implies that the estimator $\estmeanat{\regcv}$ approaches $\estmeanat{\infty}$. Formally, under the conditions of Theorem~\ref{thm:cv_noise_only}, we have
\begin{align*}
        \Prob\Big(\norm{\estmeanat{\regcv} - \estmeanat{\infty}}_2 < \epsilon \Big) \ge 1-\probbound.
\end{align*}

Recall that the sample mean estimator is commonly used and minimax-optimal in the absence of bias. This theorem suggests that our \cv algorithm, by adapting the amount of regularization in a data-dependent manner, recovers the sample mean estimator under the setting when sample mean is suitable (under only noise and no bias).

These two theorems, in conjunction to the properties of the estimator at $\reg=0$ and $\reg=\infty$ given in Sections~\ref{sec:theory_bias_only_consistent} and~\ref{sec:theory_noise_only_minimax} respectively, indicate that our proposed cross-validation algorithm achieves our desired goal in the two extremal cases.  The main intuition underlying these two results is that if the magnitude of the estimated bias from the training set aligns with the true amount of bias, the interpolated bias from the validation set also aligns with the true amount of bias and hence gives a small CV error. Extending this intuition to the general case where there is both bias and noise, one may expect \cv to still able to identify an appropriate value of $\reg$.

\section{Experiments}\label{sec:experiment}

We now conduct experiments to evaluate our estimator and our cross-validation algorithm under various settings. We consider the metric of the squared $\ell_2$ error. To estimate the qualities using our \cv algorithm, we first use Algorithm~\ref{alg:cv} to obtain a value of the hyperparameter $\regcv$; we then compute the estimate $\estmeanat{\regcv}$ as the solution to~\eqref{eq:optimization} at $\reg=\regcv$ (that is, we solve~\eqref{eq:optimization} on the entire data combining the training set and the validation set).\footnote{
    Note that this is different from the theoretical results in Section~\ref{sec:theory_cv}, where we solve~\eqref{eq:optimization} at $\reg=\regcv$ only on the training set.
}
Implementation details for the cross-validation algorithm (Algorithm~\ref{alg:cv}) are provided in Appendix~\ref{app:simulation_implementation}. Throughout the experiments, we use $\setreg = \{2^i: -9\le i\le 5, i\in \integers \}\union \{0, \infty\}$. We also plot the error incurred by the best fixed choice of $\reg\in \setreg$, where for each point in the plots, we pick the value of $\reg\in \setreg$ which minimizes the empirical $\ell_2$ error over all fixed choices in $\setreg$. Note that this best fixed choice is not realizable in practice since we cannot know the actual value of the $\ell_2$ error.

We compare our cross-validation algorithm with the mean, median, and also the \subsampling estimator introduced in Appendix~\ref{app:subsample}. The mean estimator is the sample mean for each course (same as our estimator at $\reg=\infty$) defined as $[\estmeanmean]_\idxcourse = \frac{1}{\numstudent}\sum_{\idxstudentscope} \obs_{\idxpair}$ for each $\idxcoursescope$, and the median estimator is defined as $[\estmeanmed]_{\idxcourse} = \median(\obs_{\idxcourse1}, \ldots, \obs_{\idxcourse\numstudent})$ for each $\idxcoursescope$. The \subsampling estimator is not applicable to total orderings. 

In the model~\eqref{eq:model}, we assume that the noise terms $\{\noise_\idxpair\}_{\idxcoursescope, \idxstudentscope}$ and the bias terms $\{\bias_\idxpair\}_{\idxcoursescope, \idxstudentscope}$ follow the assumptions~\ref{assumption:noise} and~\ref{assumption:bias} respectively for our theoretical results in Section~\ref{sec:theory_preliminaries}.
In our simulations, we consider three cases for the amounts of bias and noise: only bias ($\gaussianwidthbias=1, \gaussianwidthnoise=0$), only noise ($\gaussianwidthbias=0, \gaussianwidthnoise=1$), and both bias and noise  ($\gaussianwidthbias=0.5, \gaussianwidthnoise=0.5$). Throughout the experiments we use $\truemean=0$, and as explained in Proposition~\ref{prop:property_shift_invariance} in Appendix~\ref{app:property_of_estimator}, the results remain the same for any value of $\truemean$.

Each point in all the plots is computed as the empirical mean over $250$ runs. Error bars in all the plots represent the standard error of the mean.

\subsection{Dependence on \texorpdfstring{$\numstudent$}{n}}\label{sec:exp_vary_n}

We first focus on group orderings. We evaluate the performance of our estimator under different values of $\numstudent$, under the following types of group orderings.
    
\begin{itemize}
    \item \textbf{Non-interleaving total ordering: } We call a total ordering a ``non-interleaving'' total ordering, if the total ordering is $\bias_{11} \le \ldots \le \bias_{1\numstudent} \le \bias_{21}\le \ldots \le \bias_{2\numstudent} \le \ldots \le \bias_{\numcourse1} \le \ldots \bias_{\numcourse\numstudent}$. In the non-interleaving total ordering, the values of the bias terms vary quite significantly across courses. Our goal is to evaluate whether our estimator provides good estimates under such imbalanced bias.

    \item \textbf{Interleaving total ordering:} We call a total ordering an ``interleaving'' total ordering, if the total ordering is $\bias_{11} \le \bias_{21} \le \ldots \le \bias_{\numcourse1} \le \bias_{12} \le \ldots \le \bias_{\numcourse2} \le \bias_{1\numstudent} \le \ldots \le \bias_{\numcourse\numstudent}$. In contrast to the non-interleaving total ordering, in the interleaving total ordering the bias terms are more balanced across different courses, and we expect the mean and the median baselines to work well in this setting. Our goal is to evaluate whether the \cv algorithm deviates much from the baselines when the baselines work well.

    \item \textbf{Binary ordering:}  We call a group ordering a ``binary'' ordering, if there are $\numgroup=2$ groups. Specifically, we consider a group distribution where $(\blocklength_{\idxcourse1}, \blocklength_{\idxcourse2}) = (0.9\numstudent, 0.1\numstudent)$ for half of the courses $\idxcourse$, and $(\blocklength_{\idxcourse1}, \blocklength_{\idxcourse2}) = (0.1\numstudent, 0.9\numstudent)$ for the other half of the courses $\idxcourse$.
\end{itemize}

\input{text_fig_vary_d}
We consider $\numcourse=3$ courses for the non-interleaving and interleaving total orderings, and consider $\numcourse=4$ for the binary ordering. 
The results are shown in Fig.~\ref{fig:vary_d_group_order_full}. In the non-interleaving case (Fig.~\ref{fig:vary_d_group_order_full}\subref*{float:vary_d_noninterleaving}) and the binary case (Fig.~\ref{fig:vary_d_group_order_full}\subref*{float:vary_d_binary}) where the distribution of the bias is quite imbalanced, our estimator performs better than the mean and median baselines when there is bias (with or without noise). The improvement is the most significant in the case when there is only bias and no noise. In the case where there is only noise, our estimator still performs reasonably as compared to the the baselines -- the performance of our estimator is worse, but this is not unexpected, because while our algorithm tries to compensate for possible bias, the mean and median baselines do not. Indeed, as the theory (Proposition~\ref{prop:sample_mean_minimax_optimal}) suggests, the mean estimator is ideal for the only-noise setting, but in practice we do not know whether we operate in this only-noise setting a priori. In the interleaving case where the bias is more balanced (Fig.~\ref{fig:vary_d_group_order_full}\subref*{float:vary_d_interleaving}), our estimator performs on par with the baselines, and is still able to correct the small amount of bias in the only-bias case.

We also compare our estimator with the \subsampling estimator in the binary case. Recall that the \subsampling estimator is more specialized and not applicable to total orderings or more general partial orderings. Our estimator performs slightly better than the \subsampling estimator in the two extremal (only-bias and only-noise) cases. In the noisy case, the best fixed $\reg$ is better than the \subsampling estimator but the cross-validation algorithm is worse. In general, we observe that there remains a non-trivial gap between the best fixed $\reg$ and cross-validation in the noisy case (also see the non-interleaving total ordering in the noisy case). If prior knowledge about the relative amounts of bias and noise is given, we may be able to achieve better performance with our estimator by setting the value of $\reg$ manually.

\subsection{Choices of \texorpdfstring{$\reg$}{\unichar{"03BB}} by cross-validation}
    \input{text_fig_lambda}
    
    We inspect the choices of the hyperparameter $\reg$ made by our cross-validation algorithm. We use the binary setting from Section~\ref{sec:exp_vary_n}, with $\numstudent=50$. The histograms in Fig.~\ref{fig:lambda} plot the fraction of times that each value of $\reg\in \setreg$ is chosen by cross-validation.
    When there is only bias, the chosen value of $\reg$ is small  (with $\reg=0$ as the most chosen); when there is only noise, the chosen value of $\reg$ is large  (with $\reg=\infty$ as the most chosen). When there is both bias and noise, the value of $\reg$ lies in the middle of the two extremal cases. These trends align with our intuition and theoretical results about cross-validation in Section~\ref{sec:theory_cv}, and show that cross-validation is indeed able to adapt to different amounts of bias and noise present in the data.

    \subsection{The regime of \texorpdfstring{$\numcourse > \numstudent$}{d>n}}

    In our theoretical results from Section~\ref{sec:theory}, we restricted our attention to the case where the number of courses $\numcourse$ is a fixed constant. We now evaluate the regime where the number of courses $\numcourse$ becomes large compared to the number of students $\numstudent$, in order to test the general applicability of our estimator. We again consider the three types of group orderings from Section~\ref{sec:exp_vary_n}. We set $\numstudent=10$ for the non-interleaving and interleaving total orderings, and $\numstudent=20$ for the binary ordering.
    
    The results with different choices of $\numcourse$ are shown in Fig.~\ref{fig:vary_n}. The mean baseline has a flat curve (except for the small sample-size regime of small values of $\numcourse$) and converges to some non-zero constant in all of the settings. The flat curves come from the fact that the number of parameters (i.e., the number of courses $\numcourse$) grows linearly in the number of observations. The median baseline also has a relatively flat curve, with the exception that in the only-bias case for the interleaving ordering, the error decreases rapidly for small values of $\numcourse$, and eventually converges to a very small constant (not shown), because the median observations across courses have very close bias due to the interleaving ordering). Again, our estimator performs better than the mean and median baselines when there is bias. In the binary case, our estimator also performs better than the  \subsampling estimator for large values of $\numcourse$. One notable setting where our estimator does not perform as well is the only-noise case for the non-interleaving ordering. Note that this is a case not covered by the theory in Theorem~\ref{thm:cv_noise_only}\ref{part:cv_noise_only_total} because the non-interleaving ordering does not satisfy the constant-fraction interleaving assumption. 
    In this case, our estimator at $\reg=0$ (or small values of $\reg$) incurs a large error. Therefore, despite the fact that we empirically observe that cross-validation still chooses large values of $\reg$ for a large fraction of times, due to the very large error when small values of $\reg$ are chosen, the overall error is still large. The reason that our estimator at $\reg=0$ (or small values of $\reg$) gives a large error is that our estimator attempts to explain the data (that has no bias and only noise) as much as possible by the bias. Since in the non-interleaving ordering, course $\idxcourse$ has smaller bias than course $(\idxcourse+1)$, our estimator at $\reg=0$ mistakenly estimates that $\estmean_{\idxcourse}$ is about a constant larger than $\estmean_{\idxcourse+1}$ for each $\idxcourse\in [\numcourse-1]$, incurring a large error. 
    
    \input{text_fig_vary_n}

\subsection{General partial orderings}\label{sec:exp_tree}

In our theoretical results from Section~\ref{sec:theory}, we restricted our attention to group orderings. While group orderings cover a large range of common cases in practice, there may exist other types of partial orderings. 
We now consider the following two types of general partial orderings that are not group orderings to test the general applicability of our estimator. 

\begin{itemize}
    \item \textbf{Total binary tree:} We consider a binary tree, and denote the number of levels (depth) of the tree as $\treelevel$. Each node in the tree represents a single \element from the observations. Each node has a direct edge to both of its children, and the partial ordering is the set of all directed edges. Specifically, we consider $\numcourse=2$ courses. In this case, the total number of observations $\numcourse\numstudent$ is even. Therefore, we construct a binary tree with one (arbitrary) leaf node removed. We assign all the $2^{\treelevel-1} - 1$ nodes from levels $1$ to $(\treelevel-1)$ to the first course, and assign all the $2^{\treelevel-1} - 1$ nodes from level $\treelevel$ (leaf nodes) to the second course. This construction is conceptually similar to total orderings in group orderings, where each \element takes a distinct role in the partial ordering. In this construction we have the relation $\numcourse\numstudent = 2^\treelevel-2$.
    \item \textbf{Binary tree of $3$ levels:} We consider a binary tree of $3$ levels and therefore $7$ nodes in total. Each node contains $\numelementspernode$ \elements. There is an ordering constraint between two \elements if and only if there is an edge between the corresponding nodes they belong to. We have the relation $\numcourse\numstudent = 7\numelementspernode$. We consider $\numcourse=3$, and therefore we have $\numstudent = \frac{7}{3}\numelementspernode$. The three courses have the following assignment, where the \elements in each level are sampled uniformly at random from all \elements in this level:
    \begin{itemize}
        \item Course 1: all $\numelementspernode$ \elements from level $1$; $\numelementspernode$ \elements from level $2$; $\frac{\numelementspernode}{3}$ \elements from level $3$,
        \item Course 2: $\numelementspernode$ \elements from level $2$; $\frac{4}{3}\numelementspernode$ \elements from level $3$,
        \item Course 3: $\frac{7}{3}\numelementspernode$ \elements from level $3$.
    \end{itemize}
    This construction is conceptually similar to a group ordering with a constant number of groups.
\end{itemize}

We evaluate our estimator under these two types of tree partial orderings for various values of $\numstudent$ (setting the values of $\treelevel$ and $\numelementspernode$ accordingly). Given that the \subsampling estimator is defined only for group orderings, we also consider its two extensions that are tailored to tree orderings, termed ``\subsampling (node)'' and ``\subsampling (level)'' as explained in Appendix~\ref{app:subsampling_tree}. Similar to the case of group orderings, these two \subsampling estimators are applicable to the binary tree of $3$ levels but not the total binary tree.

The results are shown in Fig.~\ref{fig:tree}. Again, when there is noise, we observe that our estimator performs better than the mean and median baselines in both of these two tree orderings. In the binary tree of $3$ levels, the construction procedure specifies the number of \elements in each course from each level, but there is randomness in which nodes in the level these \elements from belong to. Due to this randomness, the \subsampling (node) estimator is not always applicable, and we use hollow squares to indicate these settings and only compute the error across the runs where the estimator is applicable. We observe that our \cv algorithm performs better than the two \subsampling estimators in the only-bias case. When there is noise (with or without bias), our \cv algorithm performs on par while the best fixed $\reg$ performs better than the \subsampling estimators.

\input{text_fig_tree}

\subsection{Semi-synthetic grading data}\label{sec:experiment_indiana}

In this section we conduct a semi-synthetic experiment using real grading statistics. We use the grading data from Indiana University Bloomington~\cite{indiana}, where the possible grades that students receive are A+ through D-, and F. We consider three ways to construct the group orderings:
\begin{itemize}
    \item \textbf{Fine grades:} The $13$ groups correspond to the grades of A+ through D-, and F.
    \item \textbf{Coarse grades:} The fine grades are merged to $5$ groups of A, B, C, D and F, where grades in $\{\text{A+}, \text{A}, \text{A-}\}$ are all considered A, etc.
    \item \textbf{Binary grades: } The grades are further merged to $2$ groups of P and F (meaning pass and fail), where all grades except F are considered P. According to the university's policies, D- is the lowest passing grade.
\end{itemize}

We use the grading data from the course ``Business Statistics'' from Spring 2020. This course consists of $10$ sessions taught by multiple instructors. The average number of students per session is $50$. We choose this course because this course has multiple sessions, so that the grading distributions across different sessions are more balanced. Therefore, many common grades (A+ through B) appear in all sessions, allowing the \subsampling estimator to use more observations and perform well. Instead, if we consider all $31$ statistics courses taught in the semester, then the only grade appearing in all courses is A, and the \subsampling estimator has to discard the data from all other grades.

We use the number of students and the grade distribution from this course, and synthesize the observations using our model~\eqref{eq:model} under the Gaussian assumptions~\ref{assumption:bias} and~\ref{assumption:noise}. The true quality is set as $\truemean=0$ (again the results are independent from the value of $\truemean$); the bias is generated according to the group ordering induced by the fine grades, with a marginal distribution of $\normal(0, \gaussianwidthbias^2)$, and the noise is generated i.i.d. from $\normal(0, \gaussianwidthnoise^2)$. We set $\gaussianwidthnoise = 1- \gaussianwidthbias$, and consider different choices of  $\gaussianwidthbias$. The estimators are given one of the three group orderings listed above.

Note that the number of students is unequal in different sessions of the course. The mean and median baselines are still defined as taking the mean and median of each course respectively. The precise definitions of the \subsampling estimator and our estimator are in Appendix~\ref{app:experiment_unequal}. We estimate the quality of the $10$ sessions of the course individually, even if some sessions are taught by the same instructor.

\input{text_fig_indiana}

The results are shown in Fig~\ref{fig:indiana}. As in previous simulations, the mean and median baselines do not perform well when there is considerable bias (corresponding to a large value of $\gaussianwidthbias$). As the number of groups increases from the binary grades to coarse grades and then to the fine grades, the performance of both our estimator and the \subsampling estimator improves, because the finer orderings provide more information about the bias. Our estimator performs slightly better than the \subsampling estimator for the fine grades (Fig.~\ref{fig:indiana}\subref*{float:indiana_fine_grades}), and slightly better on a subset of values of $\gaussianwidthbias$ for the coarse grades (Fig.~\ref{fig:indiana}\subref*{float:indiana_coarse_grades}). For the binary grades, the error of both our estimator and the \subsampling estimator increases as the relative amount of bias increases (Fig.~\ref{fig:indiana}\subref*{float:indiana_binary_grades}). This increase is likely due to the model mismatch as the data is generated from fine grades. In this case our estimator performs better than the \subsampling estimator for large values of $\gaussianwidthbias$.

\section{Discussion}
Evaluations given by participants in various applications are often spuriously biased by the evaluations received by the participant. We formulate the problem of correcting such outcome-induced bias, and propose an estimator and a cross-validation algorithm to address it. The cross-validation algorithm adapts to data without prior knowledge of the relative extents of bias and noise. Access to any such prior knowledge can be challenging in practice, and hence not requiring such prior knowledge provides our approach more flexibility. 

\paragraph{Open problems} There are a number of open questions of interest resulting out of this work. An interesting and important set of open questions pertains to extending our theoretical analysis of our estimator and cross-validation algorithm to more general settings: in the regime where there is both bias and noise, under other types of partial orderings, in a non-asymptotic regime, and in a high-dimensional regime with $\numcourse \gg \numstudent$. 
In addition, while our work aims to correct biases that already exist in the data, it is also helpful to mitigate such biases during data elicitation itself. This may be done from a mechanism design perspective where we align the users with proper incentives to report unbiased data, or from a user-experience perspective where we design multitude of questions that jointly reveal the nature of any bias.

\paragraph{Limitations} There are several caveats that need to be kept in mind when interpreting or using our work. 
First, our work only claims to address biases obeying the user-provided information such as biases associated with the grading practice of the instructor (which follow the ordering constraints), and does {\em not} address biases associated with aspects such as the demographics of the instructor (which may not align with the ordering constraints). Second, the user should be careful in supplying the appropriate ordering constraints to the algorithm, ensuring these constraints have been validated separately. 
Third, our theoretical guarantees hold under specific shape assumptions of the bias and the noise. Our algorithm is designed distribution-free, and we speculate similar guarantees to hold under other reasonable, well-behaved shape assumptions; however, formal guarantees under more general models remain open.  
Our algorithm consequently may be appropriate for use as an assistive tool along with other existing practices (e.g., sample mean) when making decisions, particularly in any high-stakes scenario. Aligned results between our algorithm and other practices give us more confidence that the result is correct; different results between our algorithm and other practices suggests need for additional information or deliberation before drawing a conclusion.

%% file: text_algo.tex
\newcommand\mycommfont[1]{\footnotesize\ttfamily {#1}}
\SetCommentSty{mycommfont}
\SetNoFillComment

\begin{algorithm*}
    \DontPrintSemicolon
    \tcc{Step 1: Split the data}
    Initialize the training and validation sets as $\settrain \leftarrow \{\}$, $\setval \leftarrow \{\}$.\label{line:split_start}\;
    Sample a total ordering of $\totalordersplit$ uniformly at random from the set $\settotalorder$ of all total orderings (of the $\numcourse\numstudent$ \elements) consistent with the partial ordering $\setpartialorder$.\label{line:sample_total_order}\;
    \ForEach{\normalfont $\idxcourse\in [\numcourse]$}{
        Find the sub-ordering of the $\numstudent$ \elements in course $\idxcourse$ according to $\totalordersplit$, denoted in increasing order as $(\idxcourse, \idxstudent^{(1)}), \ldots, (\idxcourse, \idxstudent^{(\numstudent)})$.\label{line:find_sub_order}\;
        \For{\normalfont $\position= 1, \ldots, \frac{\numstudent}{2}$\label{line:assign_start}}{
            Assign $(\idxcourse, \idxstudent^{(2\position-1)}), (\idxcourse, \idxstudent^{(2\position)})$ to $\settrain$ and $\setval$, one each uniformly at random. If $\numstudent$ is odd, assign the last \element $(\idxcourse, \idxstudent^{(\numstudent)})$ to the validation set. \label{line:assign}\; 
        }\label{line:assign_end}
    }\label{line:split_end}
    \tcc{Step 2: Compute validation error}
    \ForEach{\normalfont $\reg\in \setreg$}{\label{line:validation_step_start}
        Obtain $(\estmeanat{\reg}, \estbiasmtxat{\reg})$ as a solution to the following optimization problem:
        \begin{align*} 
        (\estmean_{\reg}, \estbiasmtxat{\reg}) \in \argmin_{ \substack{\meancourse\in \reals^\numcourse,\ 
        \biasmtx\in \reals^{\numcourse\times \numstudent,}\\
        \biasmtx \text{ satisfies } \setpartialorder}} \ 
    \norm{\obsmtx - \meancourse\vecone^T - \biasmtx}_\settrain^2 + \reg \norm{\biasmtx}_\settrain^2,
    \end{align*}
    where ties are broken by minimizing $\normfro{\estbiasmtxat{\reg}}$.
     \label{line:compute_err_start}\; 
        \ForEach{\normalfont $(\idxcourse, \idxstudent)\in \setval$}{\label{line:iterate_all_val_entry}
            \ForEach{\normalfont $\totalorder\in \settotalorder$\label{line:iterate_all_total_orders}}{
                Find the \element $(\idxcoursenn, \idxstudentnn)\in \settrain$ that is closest to $(\idxcourse, \idxstudent)$ with respect to $\totalorder$, and set $[\estbiasvalat{\reg}_{\totalorder}]_\idxpair = \estbiasat{\reg}_{\idxcoursenn\idxstudentnn}$. There may be two closest \elements at equal  distance to $\idxpairparen$, in which case call them $(\idxcoursenn_1, \idxstudentnn_1)$ and $(\idxcoursenn_2, \idxstudentnn_2)$ and set
                $[\estbiasvalat{\reg}_{\totalorder}]_\idxpair = \frac{
                    \estbiasat{\reg}_{\idxcoursenn_1\idxstudentnn_1} + \estbiasat{\reg}_{\idxcoursenn_2\idxstudentnn_2}
                }{2}
                $.\label{line:interpolation_per_total_order}\;
            }
            Interpolate the bias as $\estbiasmtxvalat{\reg} = \frac{1}{\abs*{\settotalorder}} \sum_{\totalorder\in \settotalorder} \estbiasmtxvalatoforder{\reg}{\totalorder}$. \label{line:interpolation}\;
        }\label{line:interpolation_end}
        Compute the CV error $\errat{\reg} \defn \frac{1}{\abs*{\setval}}\norm{\obsmtx - \estmean_\reg\vecone^T - \estbiasmtxvalat{\reg}}_{\setval}^2$.\label{line:compute_err_end}\;
    }
    Output $\regcv \in \argmin_{\reg\in \setreg} \errat{\reg}$. \qquad (Ties are broken arbitrarily)\label{line:validation_step_end}
    \caption{Cross-validation. Inputs: observations $\obsmtx$, partial ordering $\setpartialorder$, and set $\setreg$.}\label{alg:cv}

\end{algorithm*}

%% file: text_fig_vary_d.tex
\begin{figure*}
\centering

\includegraphics[width=0.6\linewidth]{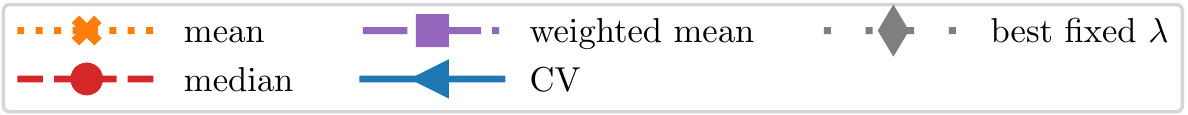}

\vspace{1mm}

\input{text_fig_mode_bias_noise}
    
\subfloat[Non-interleaving total ordering]{\label{float:vary_d_noninterleaving}
    \centering
    \includegraphics[width=\figurewidthlong]{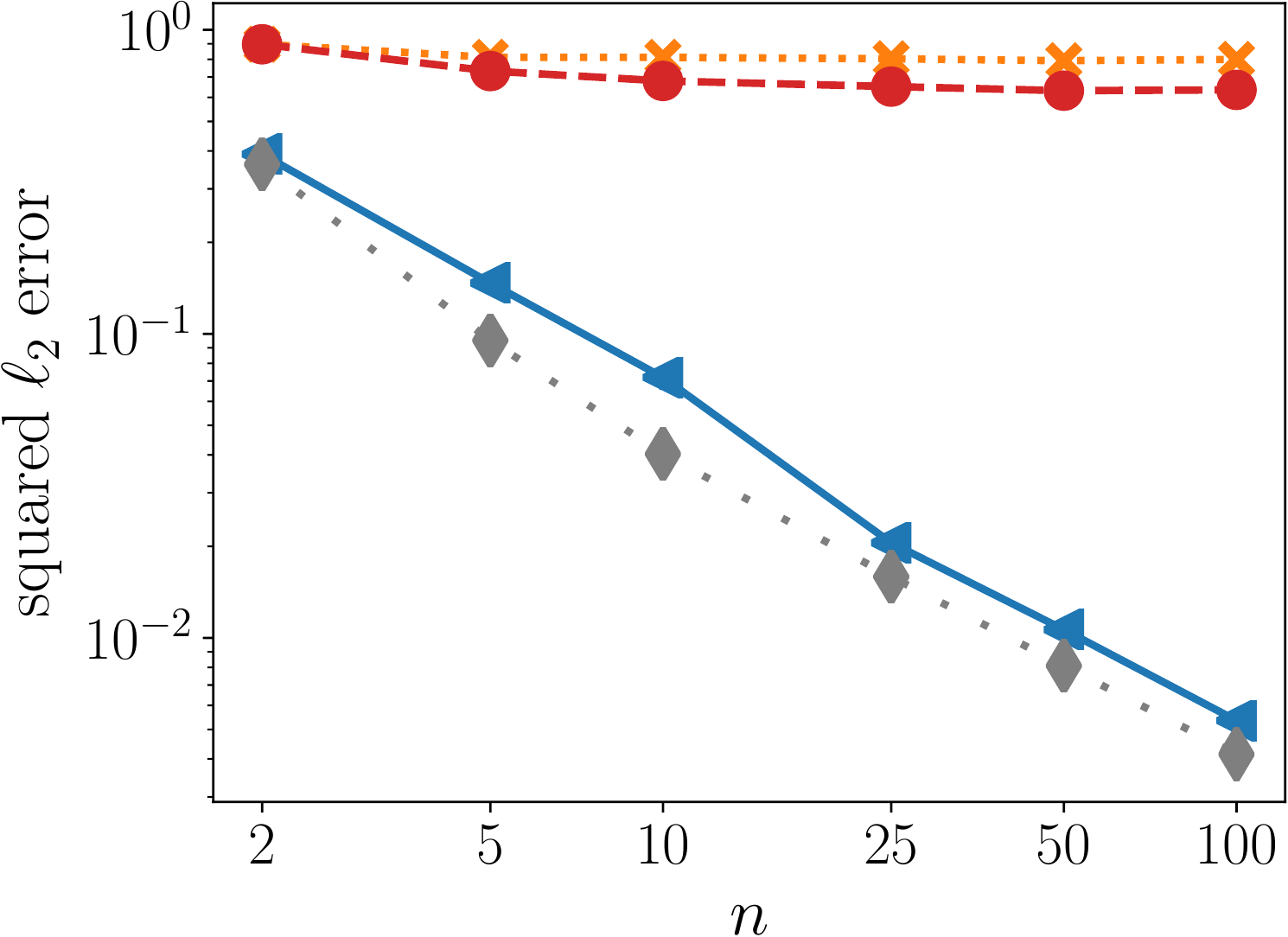}
    \includegraphics[width=\figurewidth]{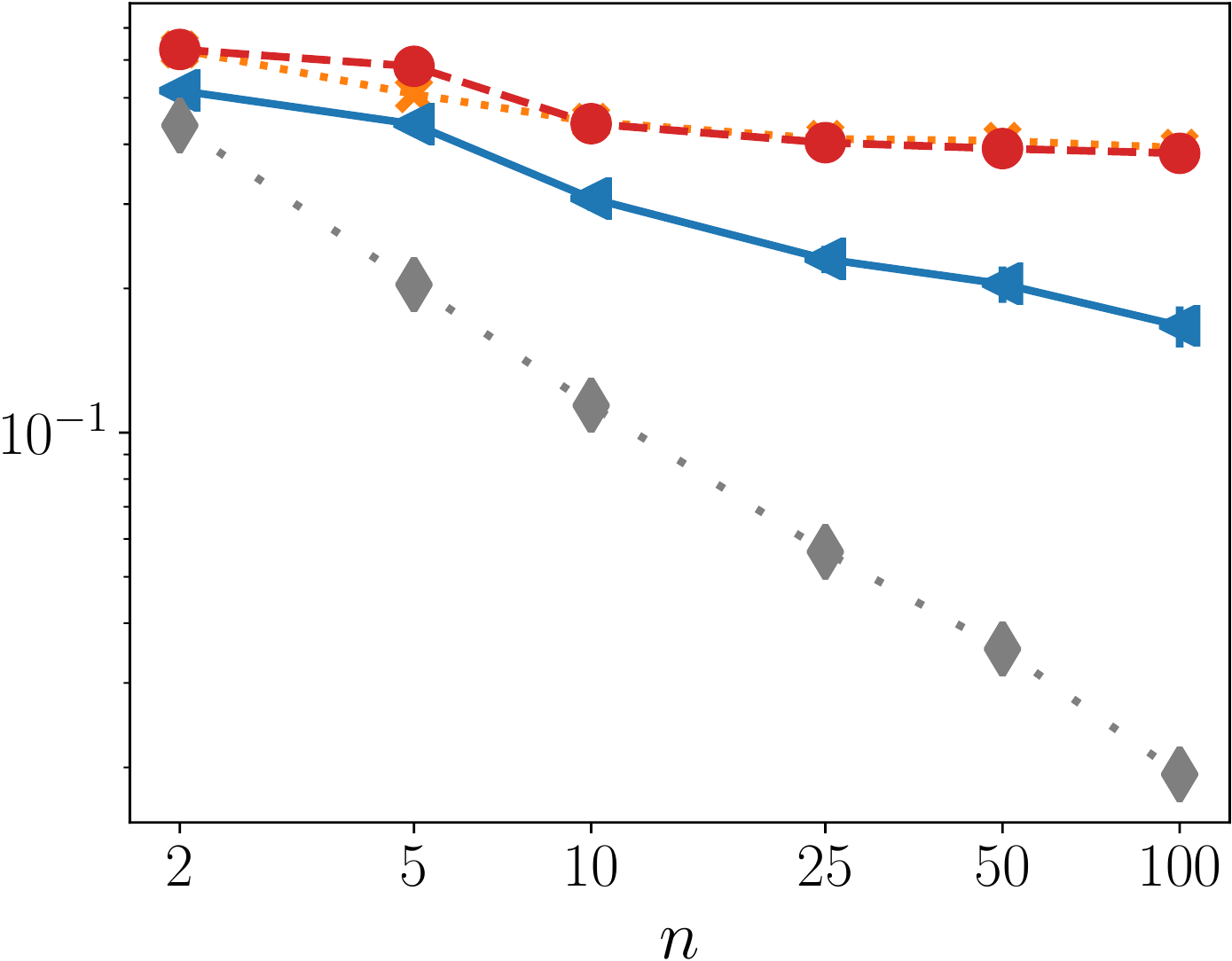}
    \includegraphics[width=\figurewidth]{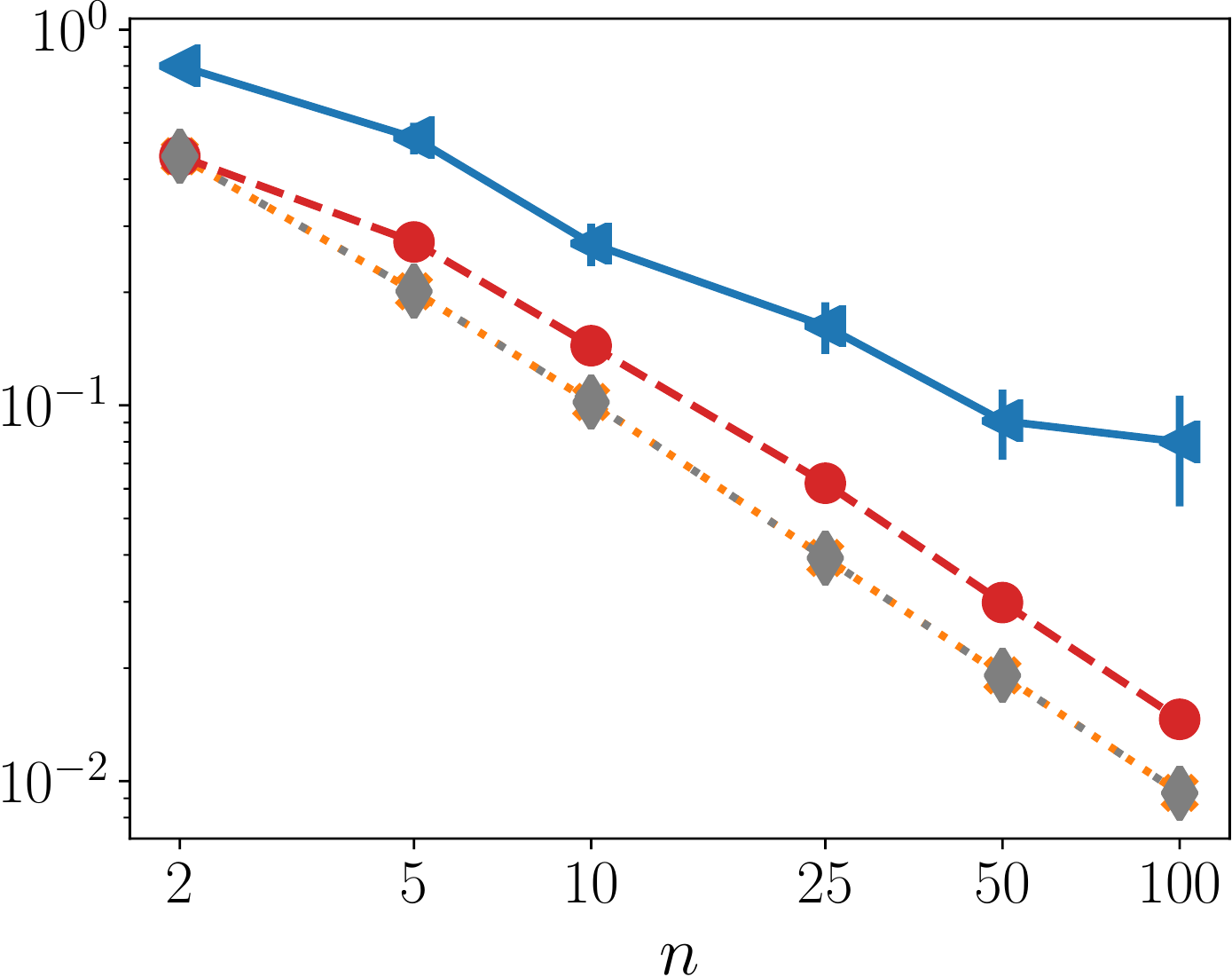}
}

\subfloat[Interleaving total ordering]{\label{float:vary_d_interleaving}
    \centering
    \includegraphics[width=\figurewidthlong]{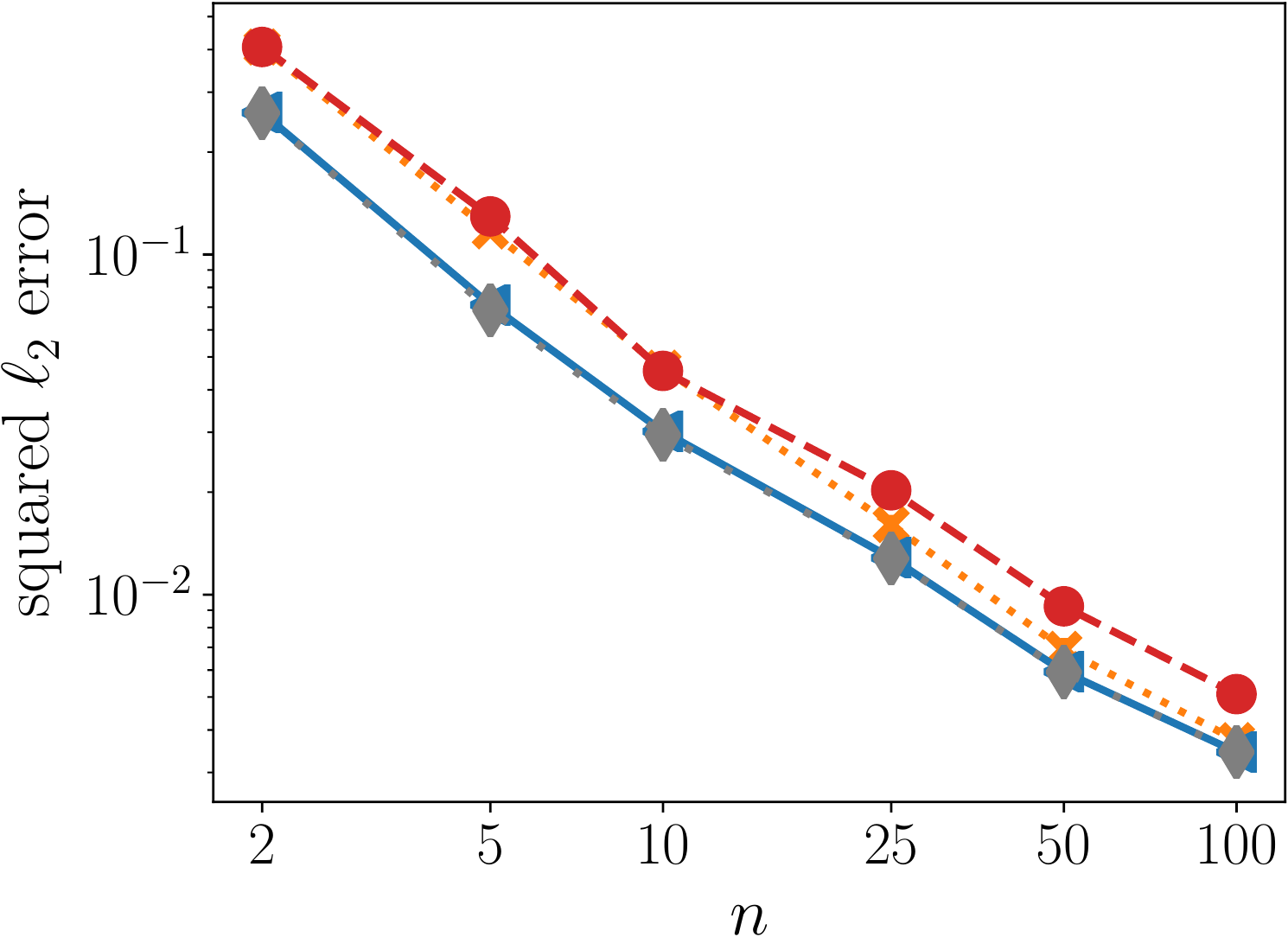}
    \includegraphics[width=\figurewidth]{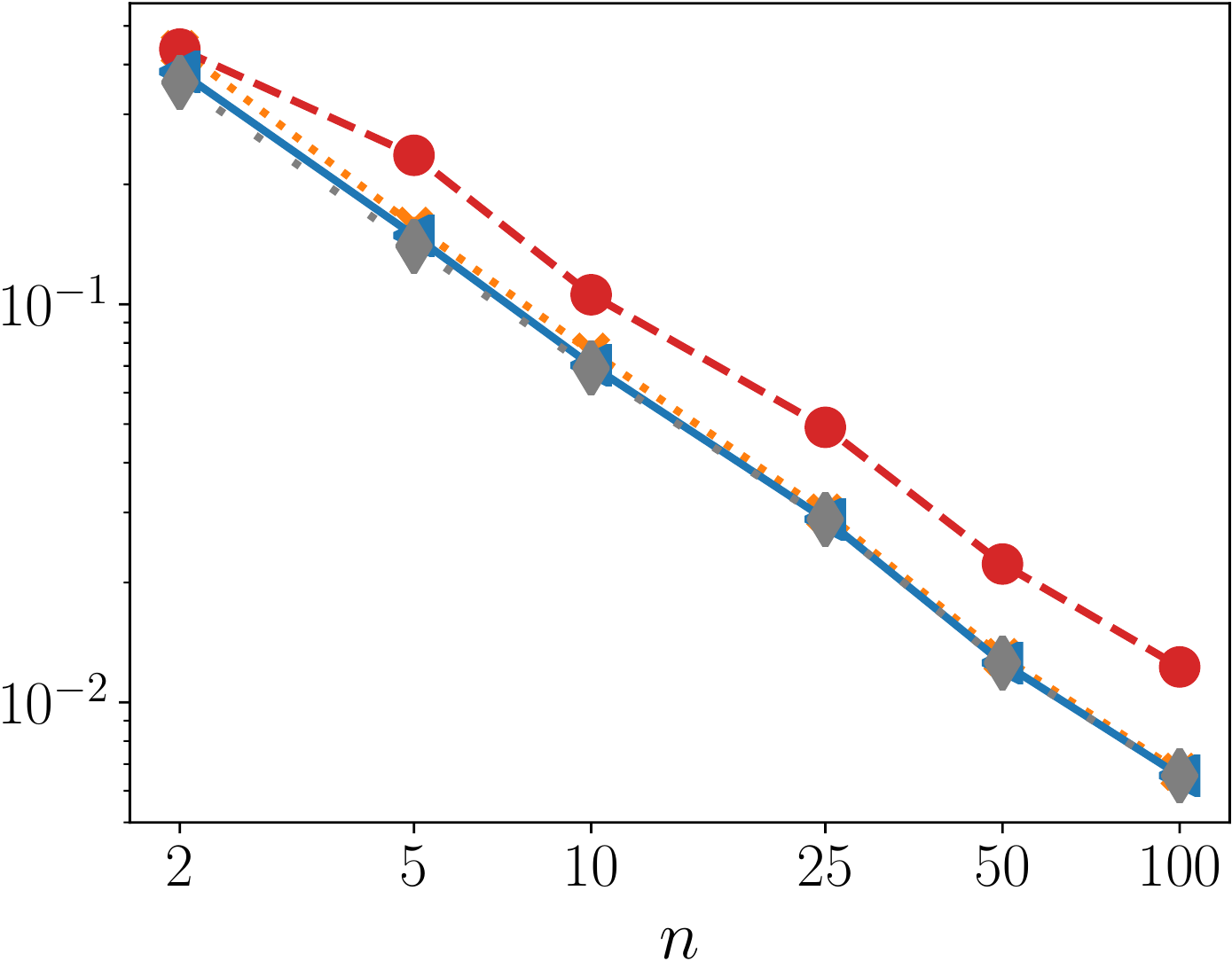}
    \includegraphics[width=\figurewidth]{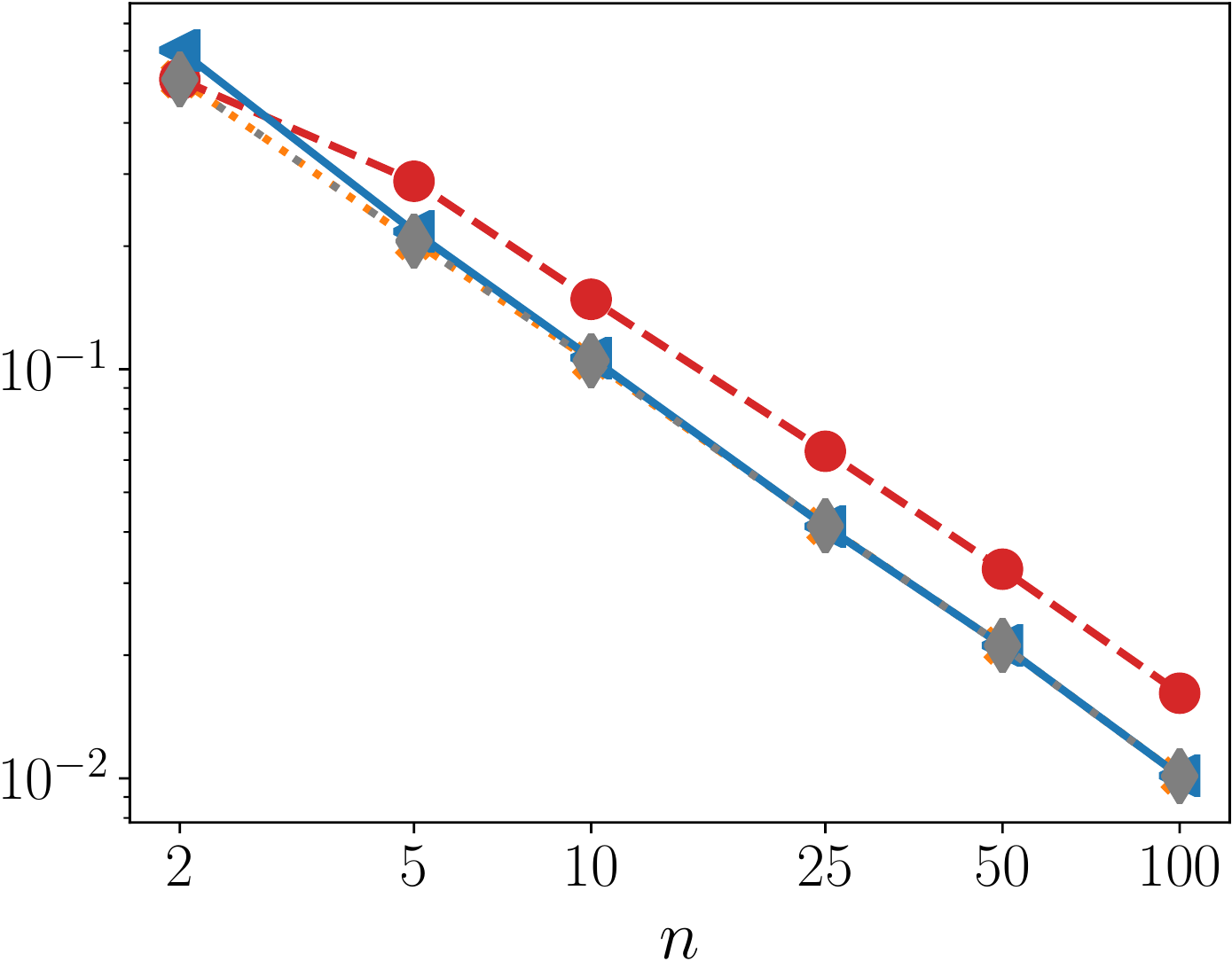}
}
    
\subfloat[Binary ordering]{\label{float:vary_d_binary}
    \centering
    \includegraphics[width=\figurewidthlong]{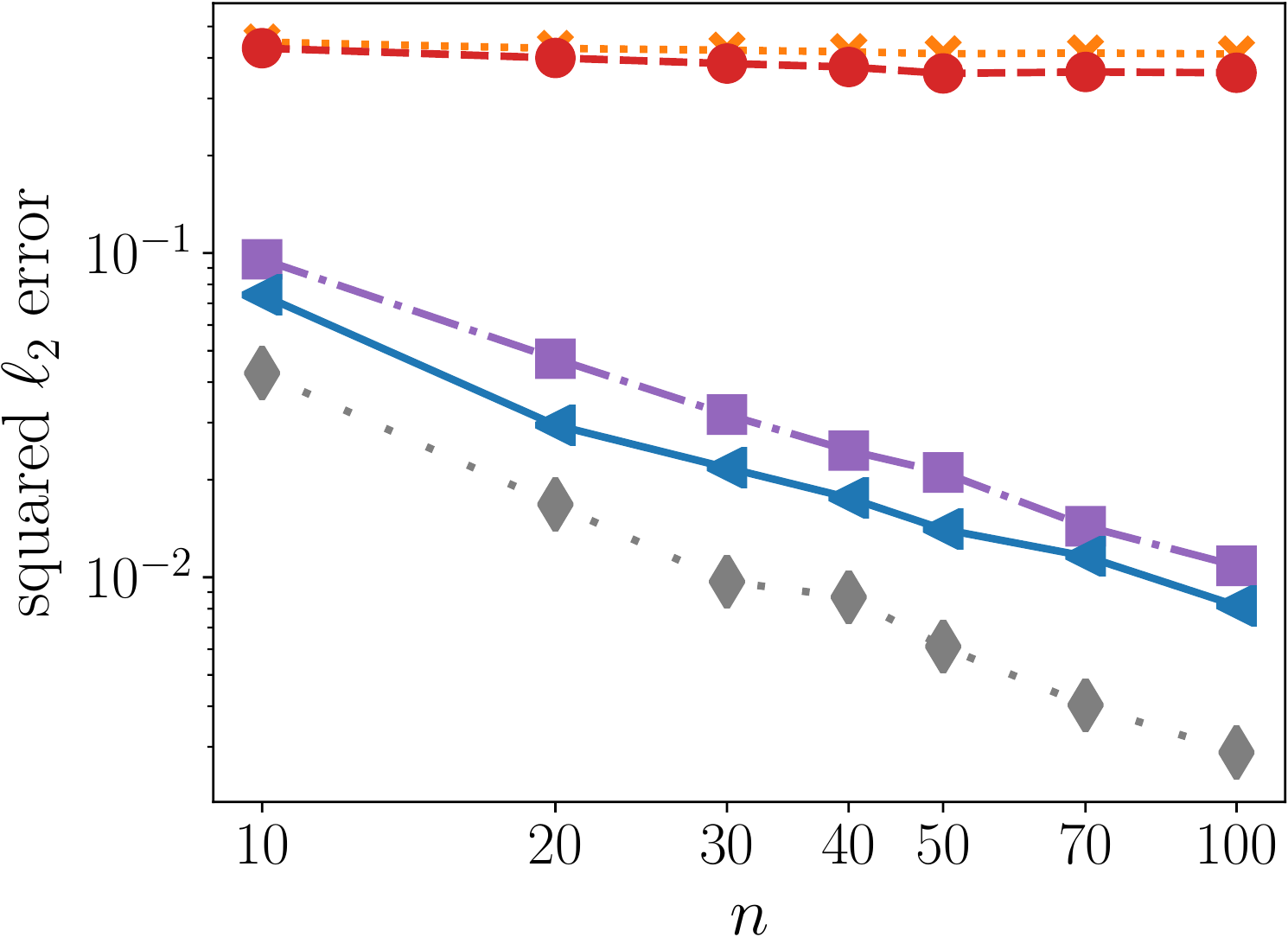}
    \includegraphics[width=\figurewidth]{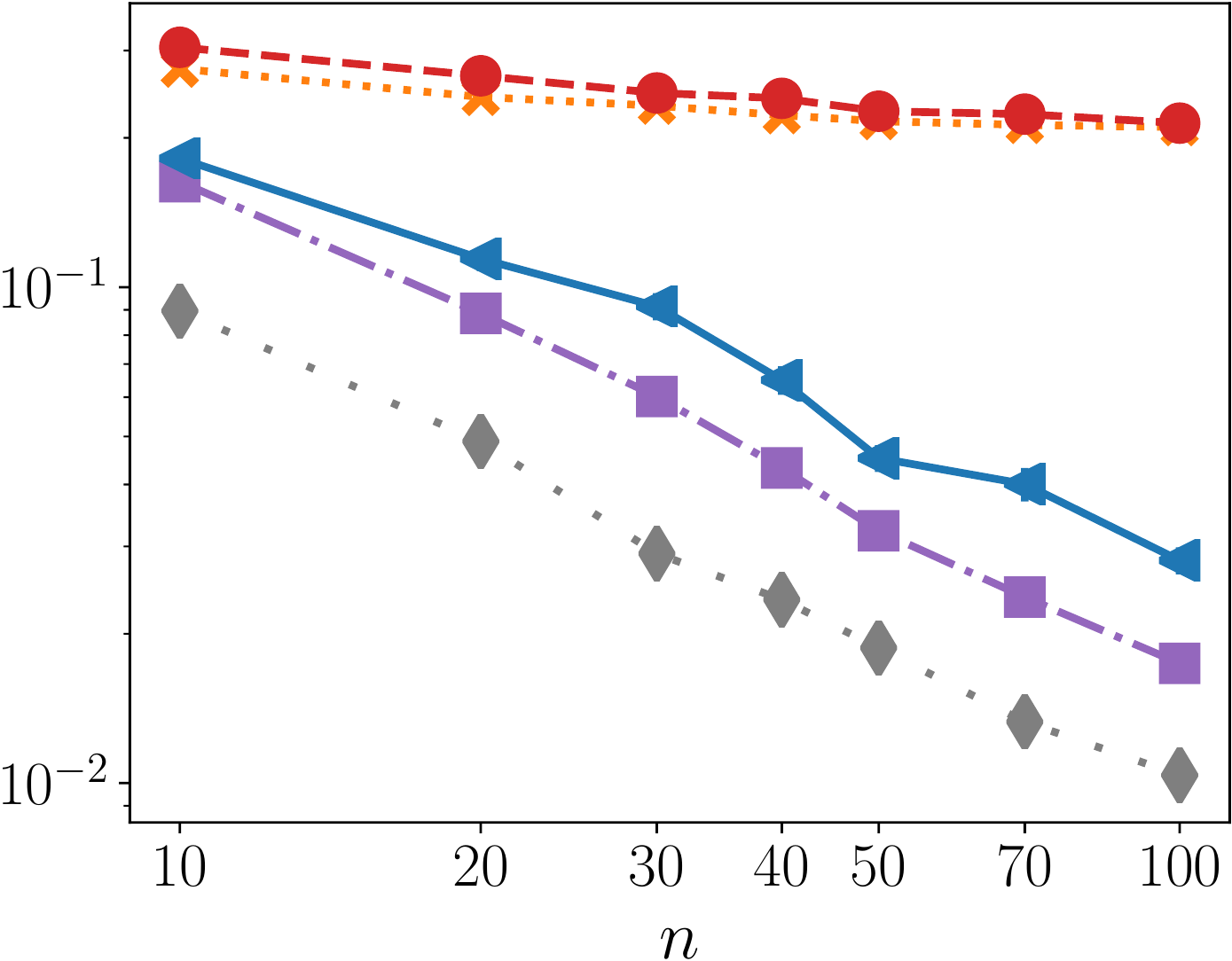}
    \includegraphics[width=\figurewidth]{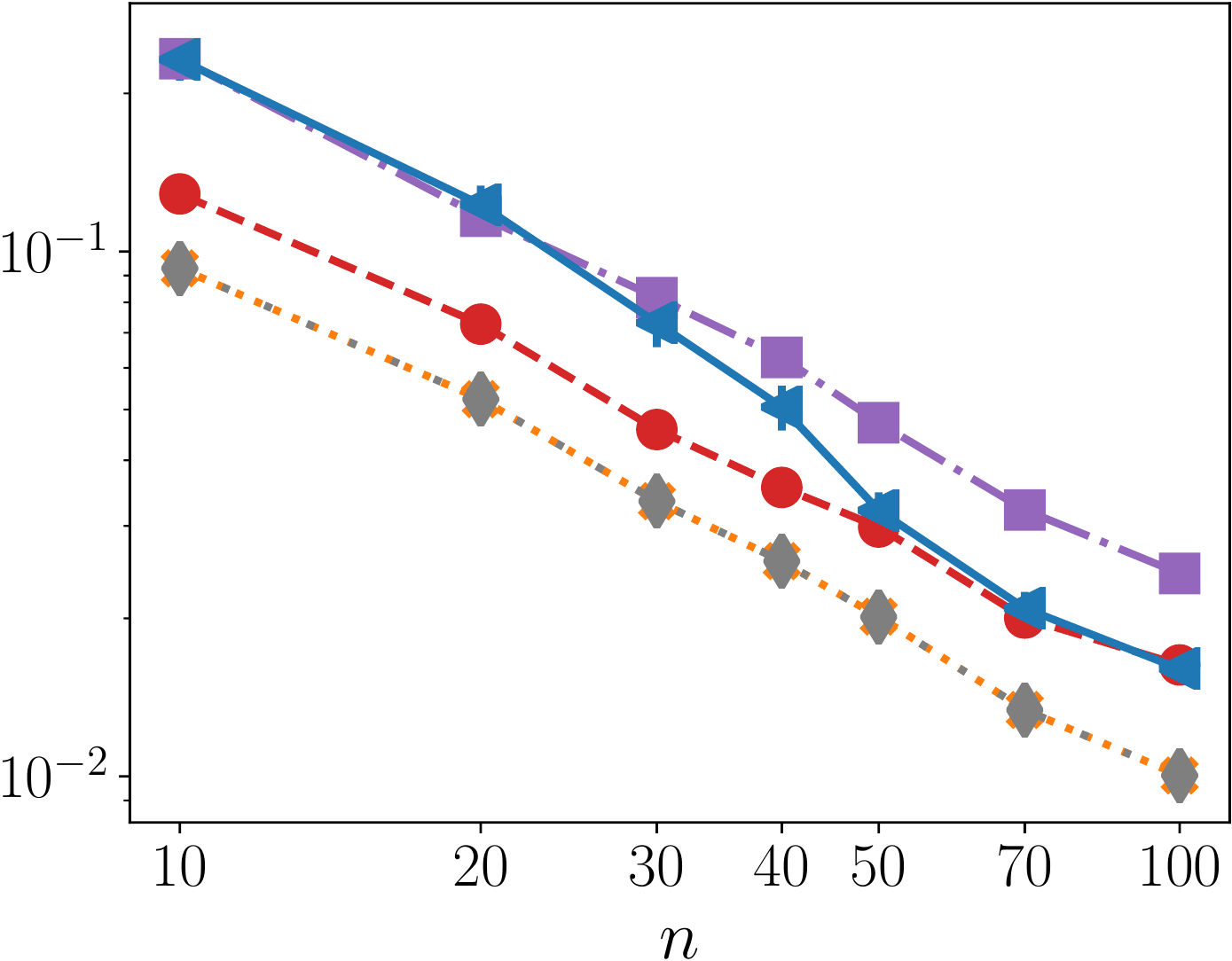}
}

\caption{\label{fig:vary_d_group_order_full}
    The performance of our estimator (with cross-validation and with the best fixed $\reg$) for various values of $\numstudent$, compared to the mean, median and \subsampling estimators.
}
\end{figure*}

%% file: text_fig_mode_bias_noise.tex
    \begin{minipage}[t]{\figurewidthlong}
        \centering
            Only bias\\$(\gaussianwidthbias=1, \gaussianwidthnoise=0)$
    \end{minipage}
    \begin{minipage}[t]{\figurewidth}
        \centering
            Both bias and noise\\$(\gaussianwidthbias=0.5, \gaussianwidthnoise=0.5)$
    \end{minipage}
    \begin{minipage}[t]{\figurewidth}
        \centering
            Only noise\\$(\gaussianwidthbias=0, \gaussianwidthnoise=1)$
    \end{minipage}


%% file: text_fig_lambda.tex
\begin{figure*}
\centering

\input{text_fig_mode_bias_noise}

    \includegraphics[width=\figurewidthlong]{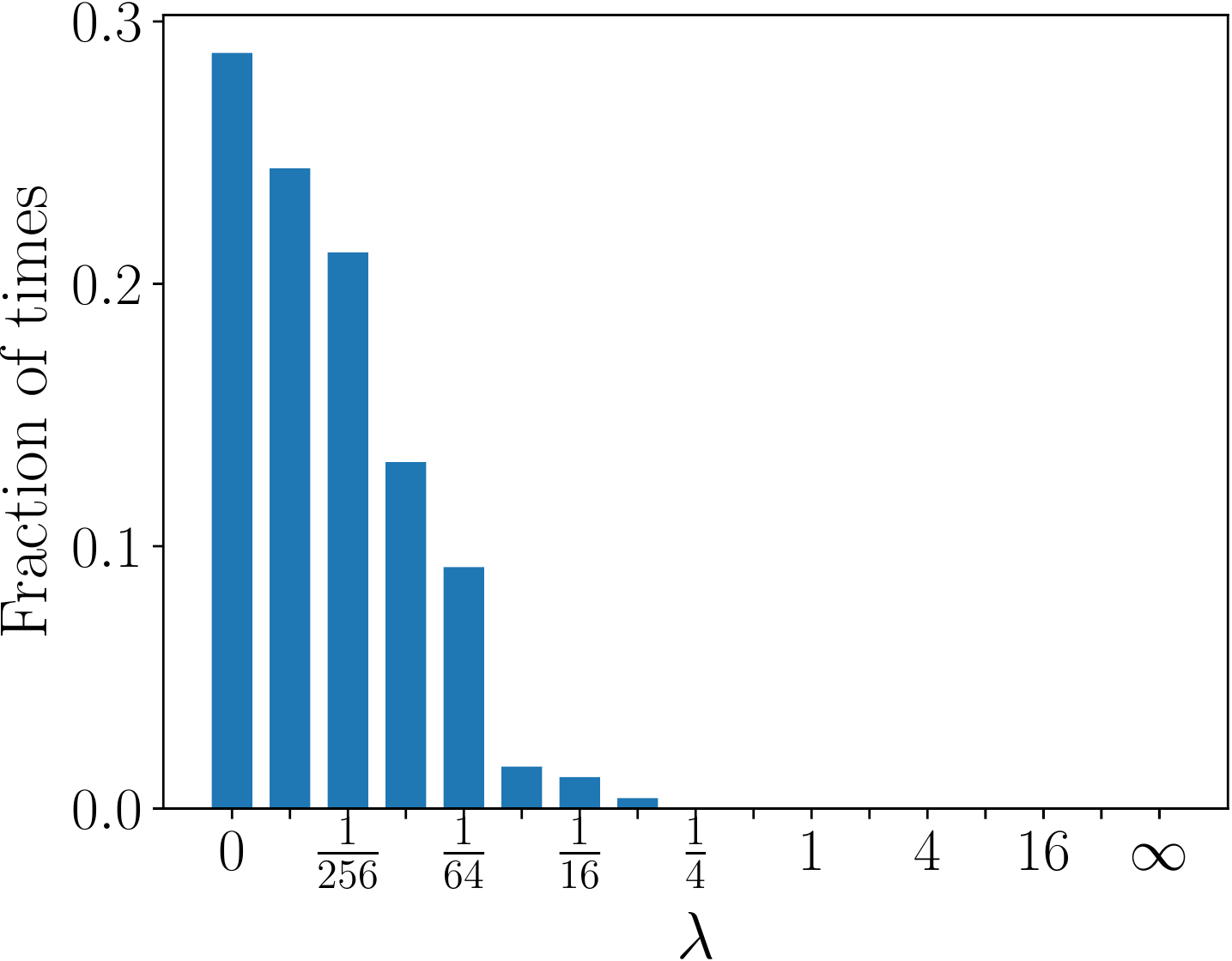}
    \includegraphics[width=\figurewidth]{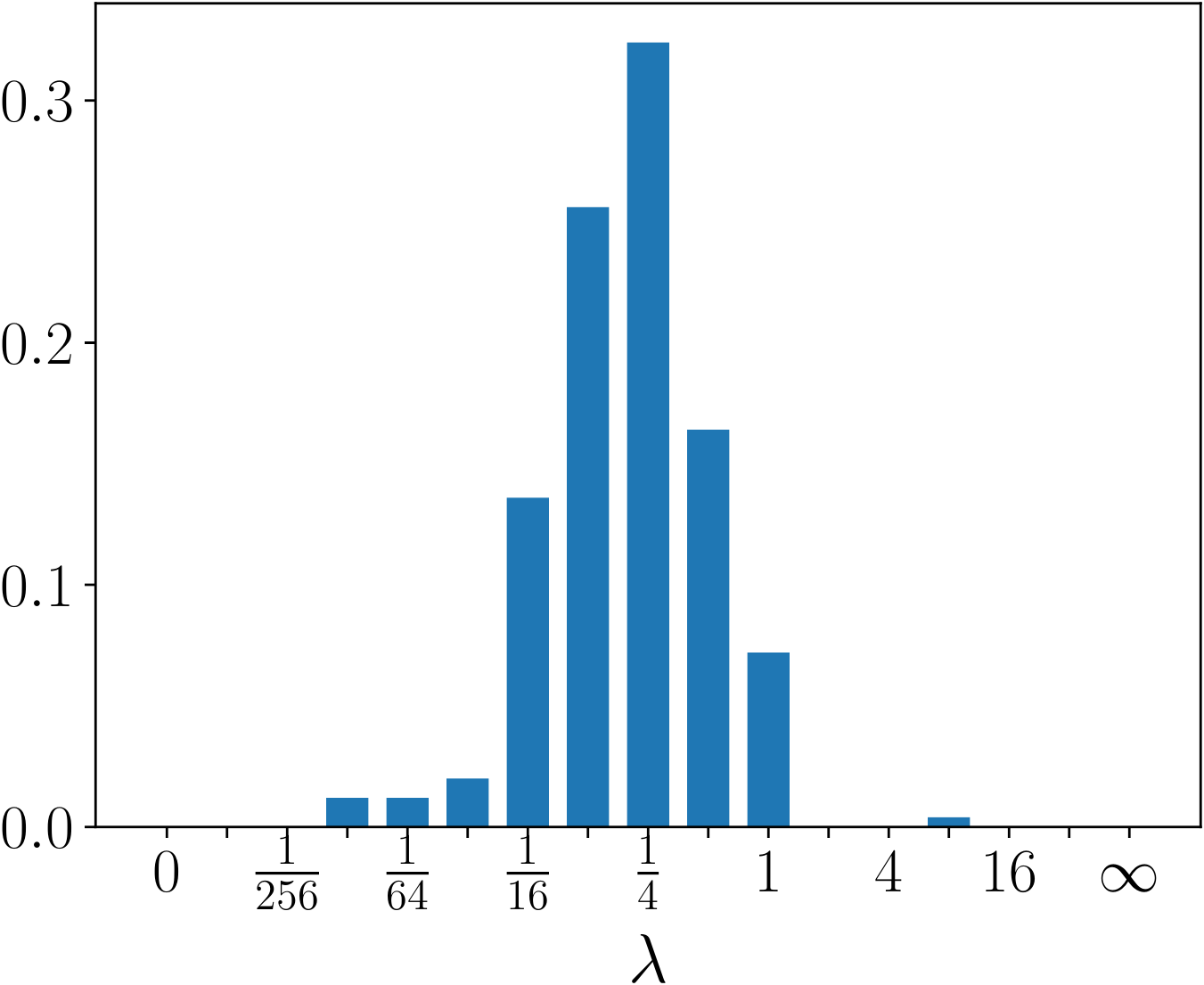}
    \includegraphics[width=\figurewidth]{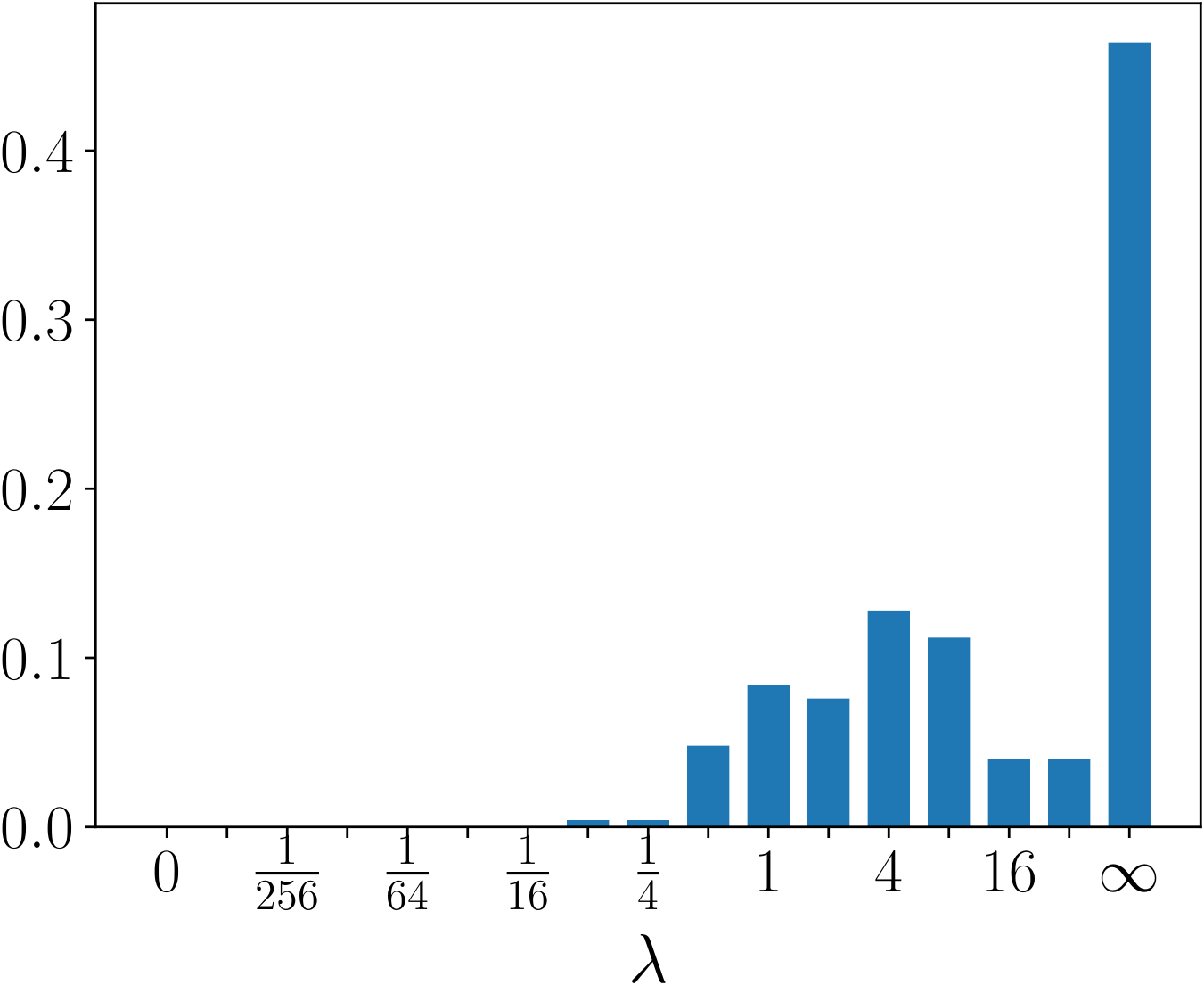}
    
     \caption{\label{fig:lambda}
        The histogram on the fraction of times each value of $\reg$ is chosen by cross-validation. Cross-validation is able to choose the value of $\reg$ adaptive to different amounts of bias and noise.
    }

\end{figure*}

%% file: text_fig_vary_n.tex
\begin{figure*}
\centering

\includegraphics[width=0.6\linewidth]{figures/legend.pdf}

\vspace{1mm}

\input{text_fig_mode_bias_noise}

\subfloat[Non-interleaving total ordering]{\label{float:vary_n_noninterleaving}
    \centering
    \includegraphics[width=\figurewidthlong]{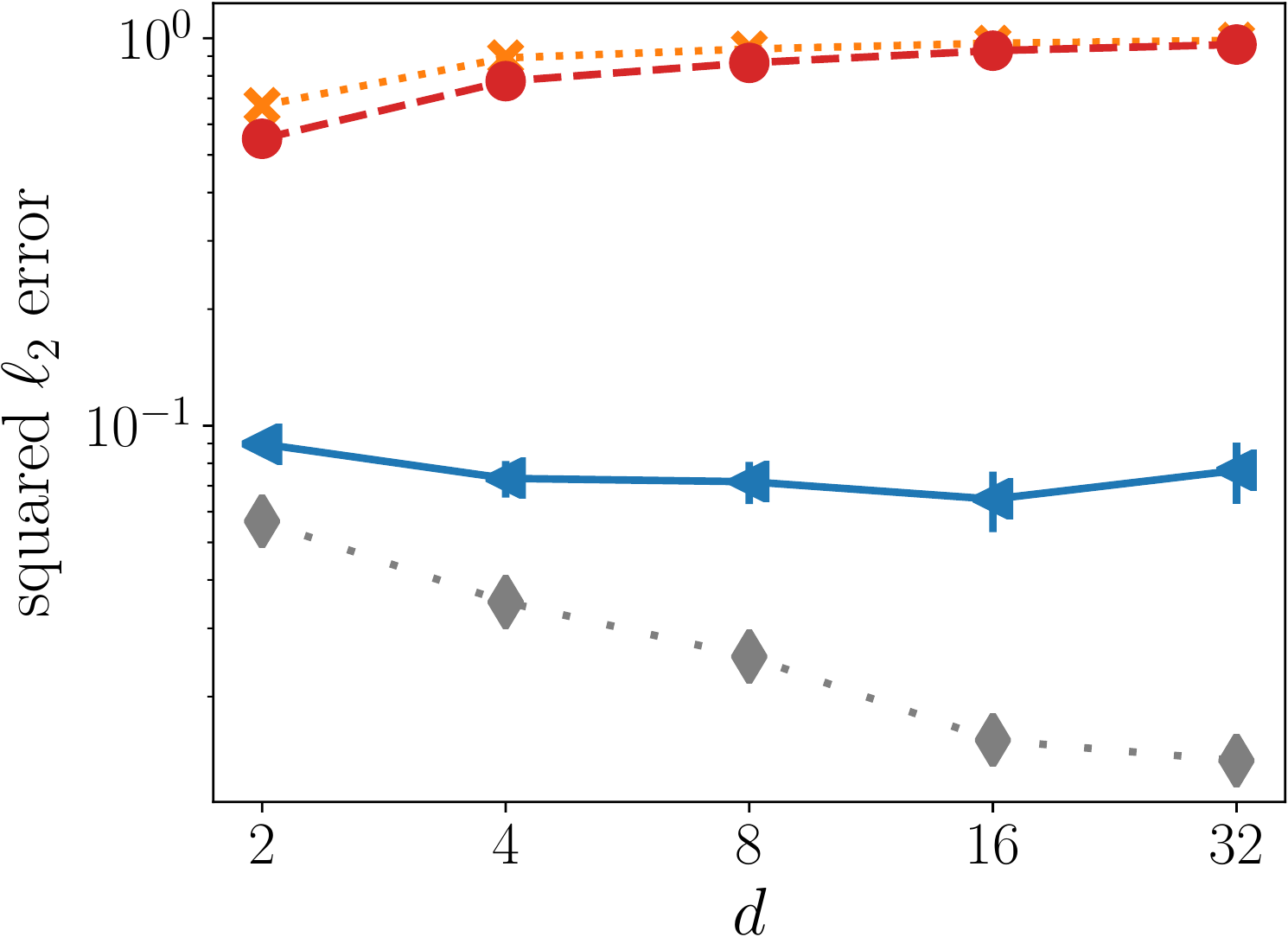}
    \includegraphics[width=\figurewidth]{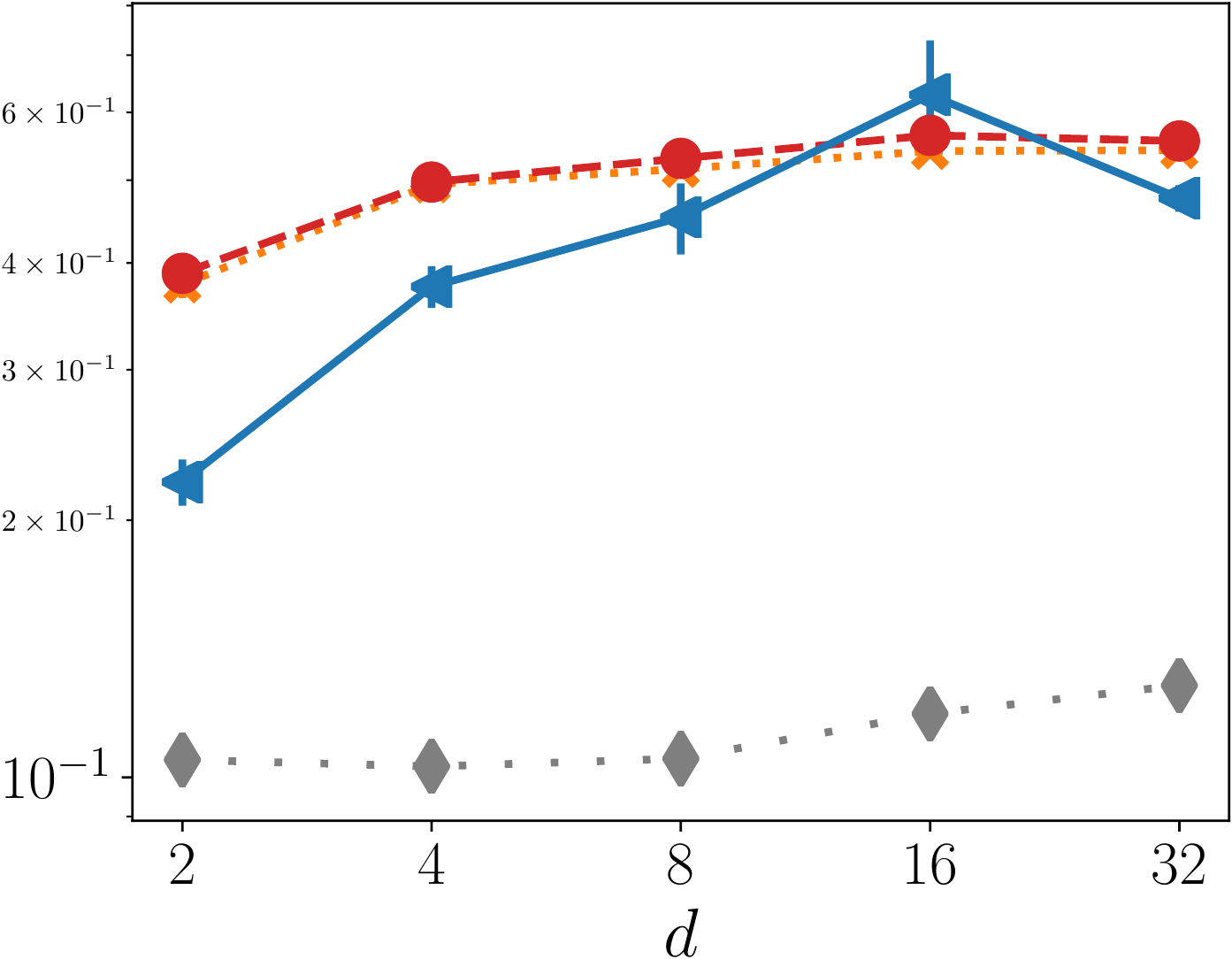}
    \includegraphics[width=\figurewidth]{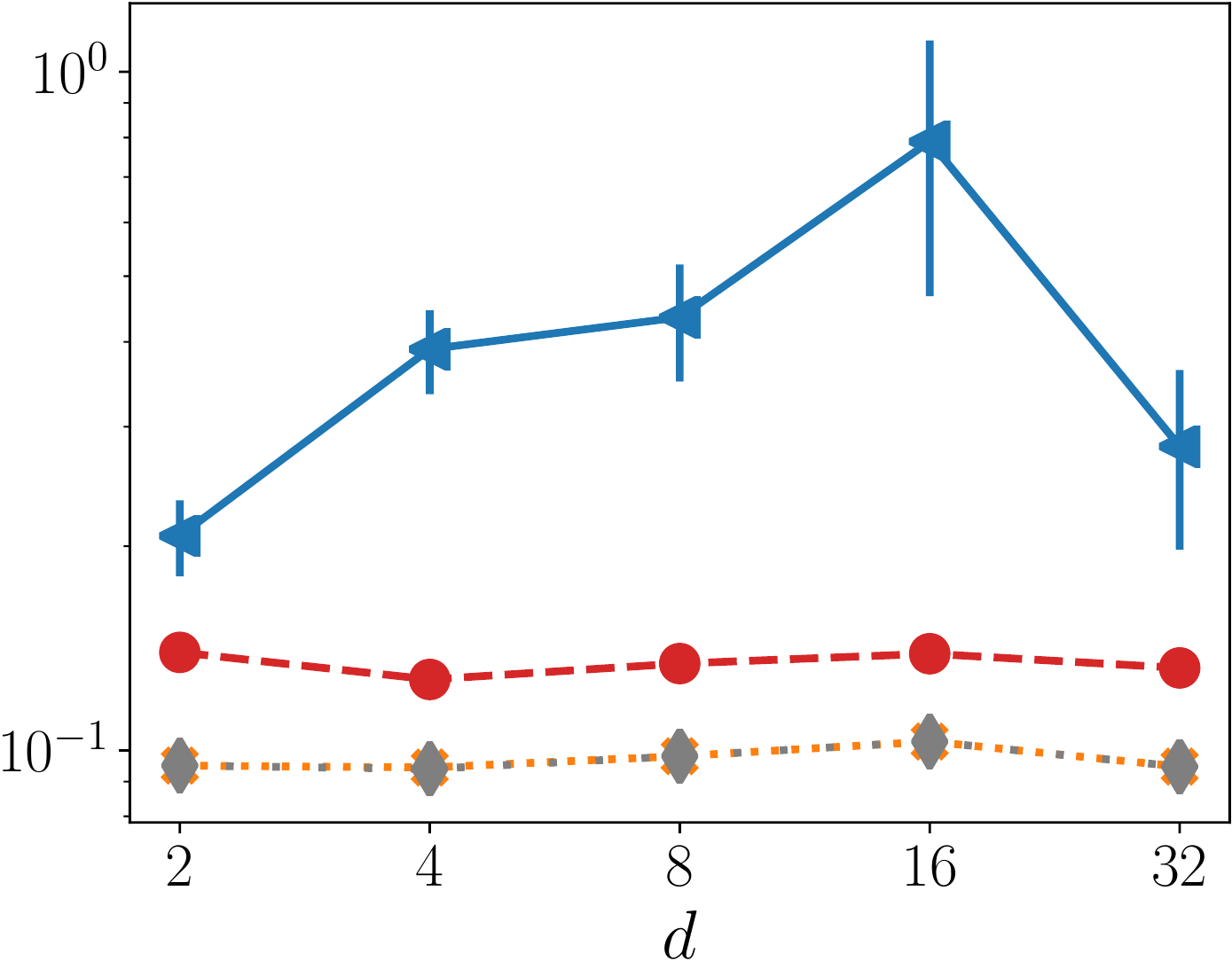}
}

\subfloat[Interleaving total ordering]{\label{float:vary_n_interleaving}
    \centering
    \includegraphics[width=\figurewidthlong]{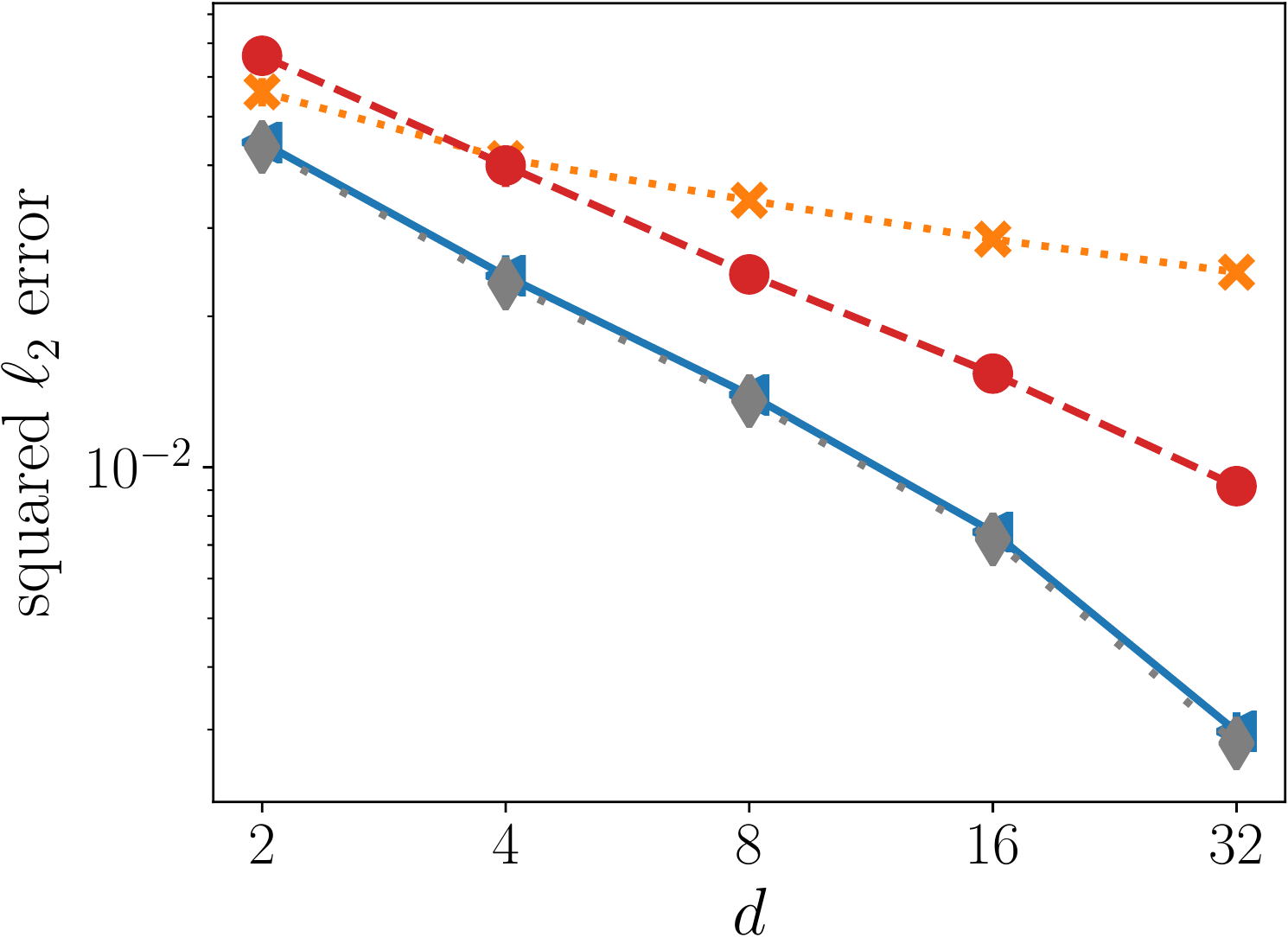}
    \includegraphics[width=\figurewidth]{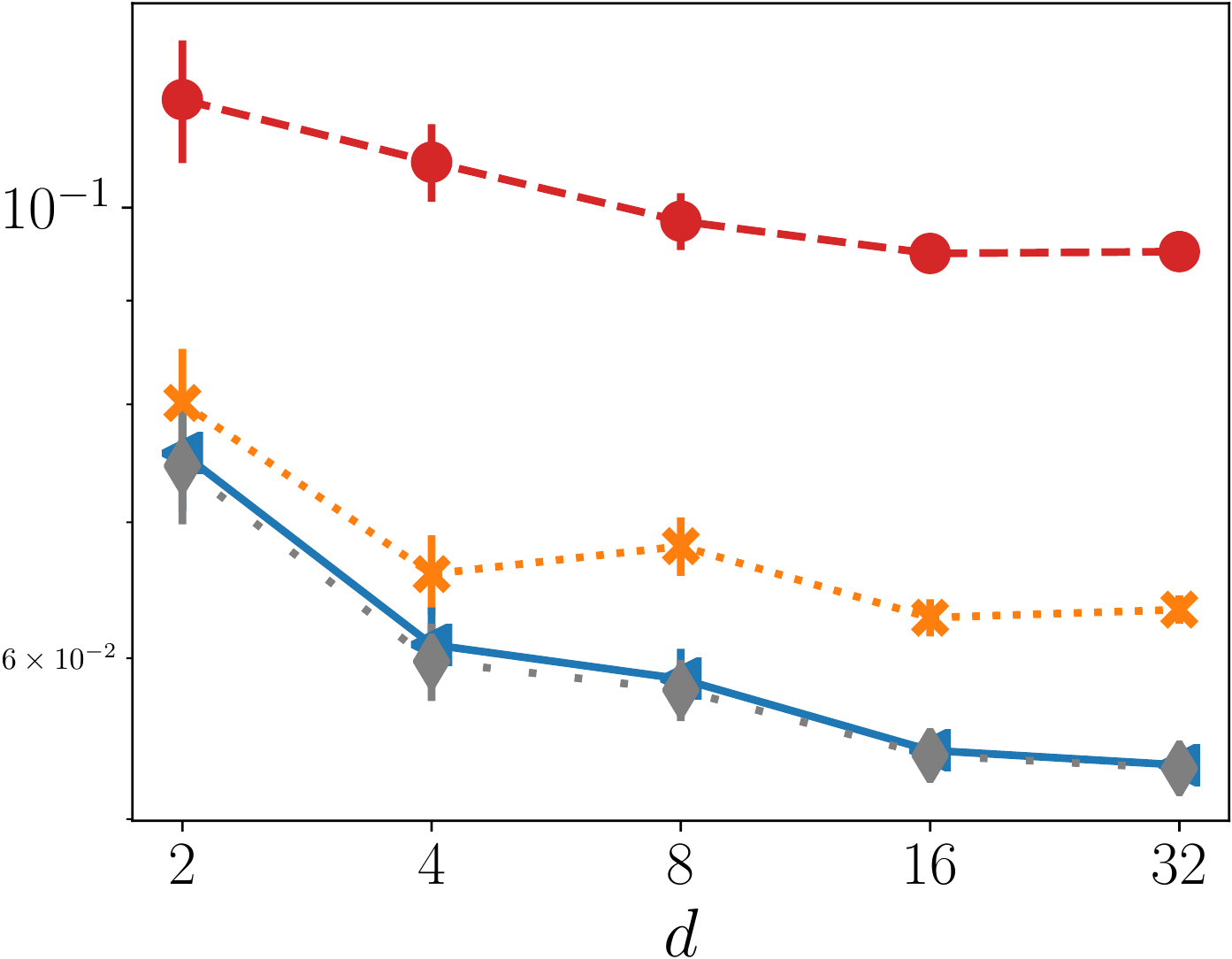}
    \includegraphics[width=\figurewidth]{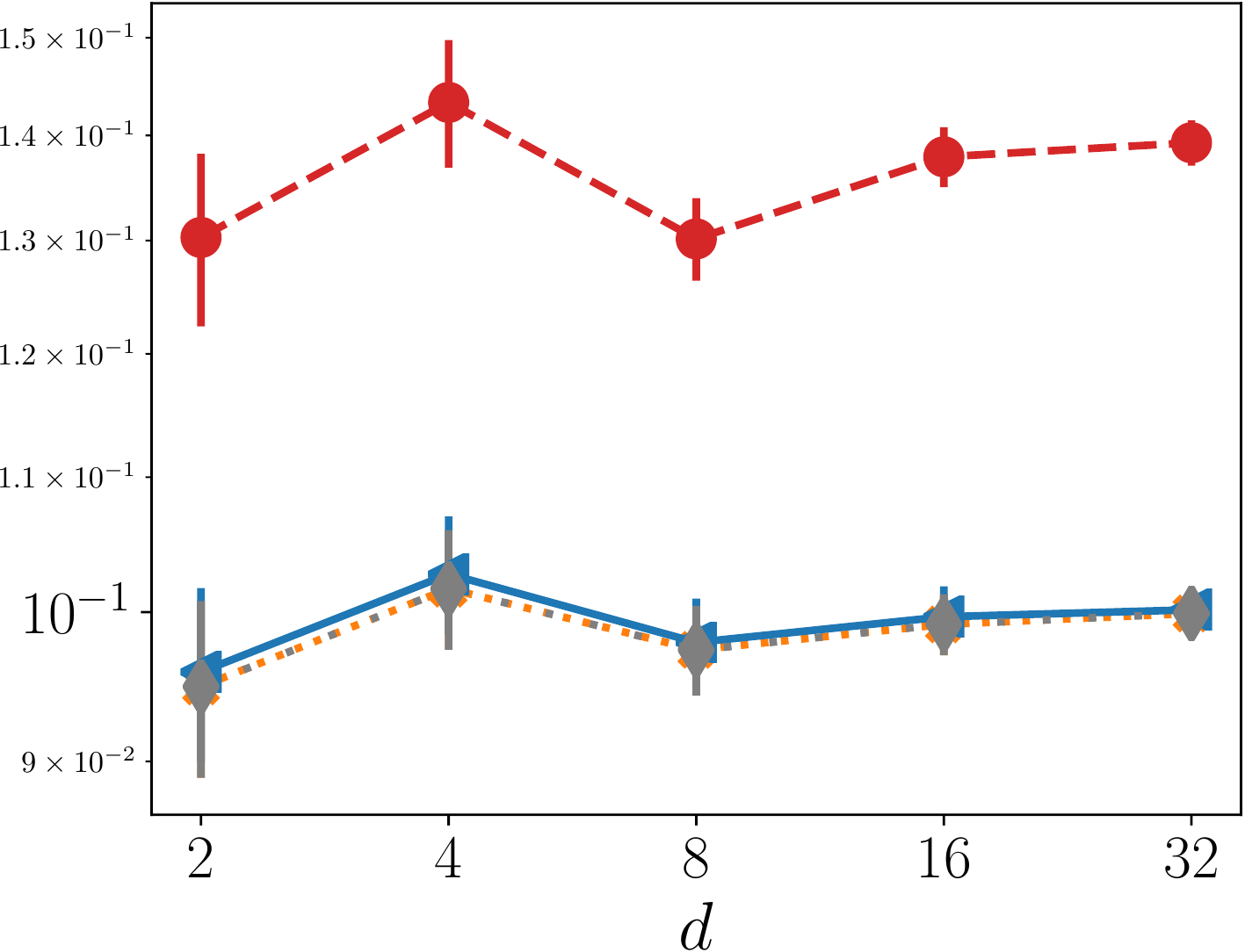}
}

\subfloat[Binary ordering]{\label{float:vary_n_binary}
    \centering
    \includegraphics[width=\figurewidthlong]{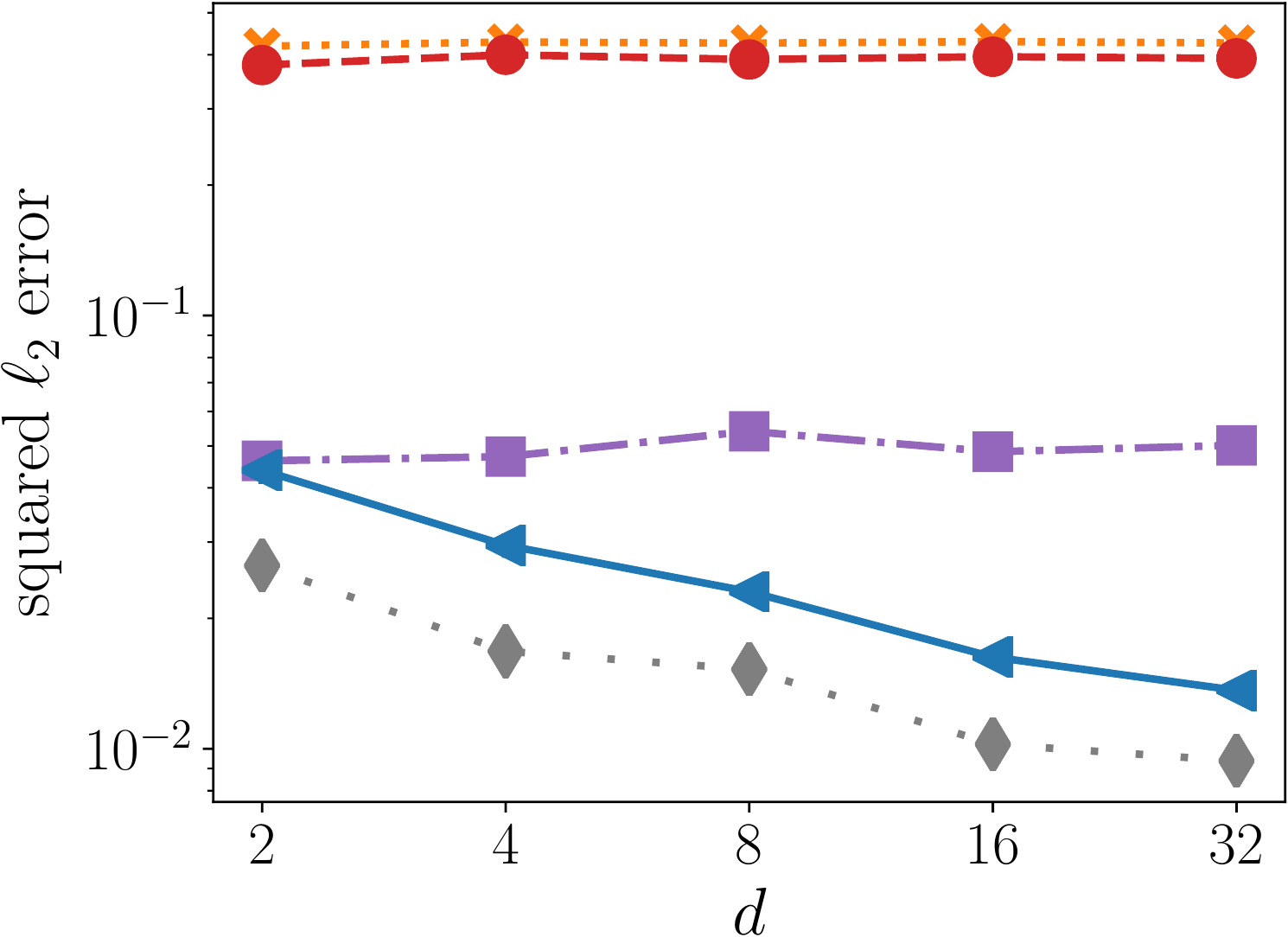}
    \includegraphics[width=\figurewidth]{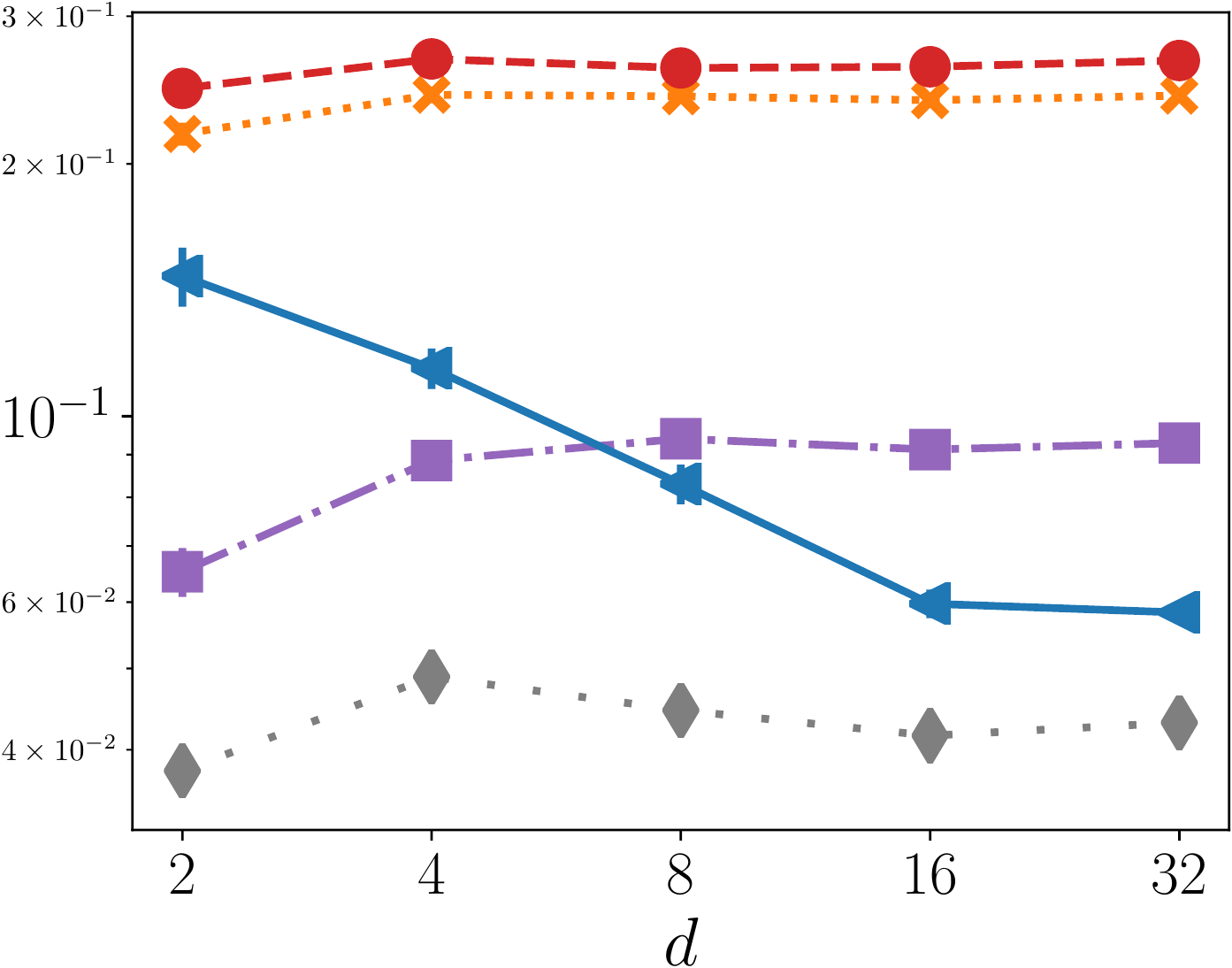}
    \includegraphics[width=\figurewidth]{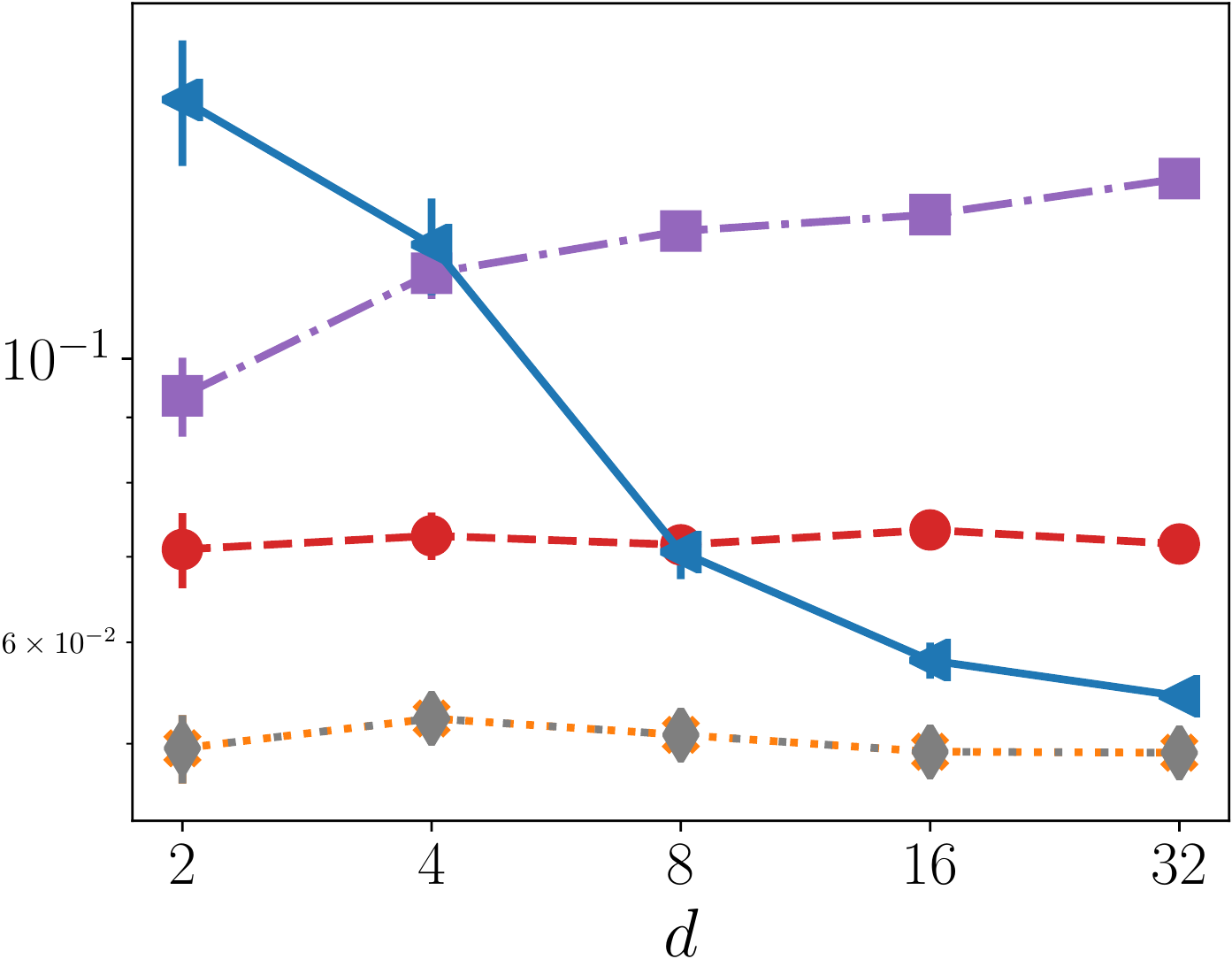}
}

 \caption{\label{fig:vary_n} The performance of our estimator (with cross-validation and with the best fixed $\reg$) for various values of $\numcourse$, compared to the mean, median, and \subsampling estimators.
 }
\end{figure*}

%% file: text_fig_tree.tex
\begin{figure*}
\centering

\includegraphics[width=0.7\linewidth]{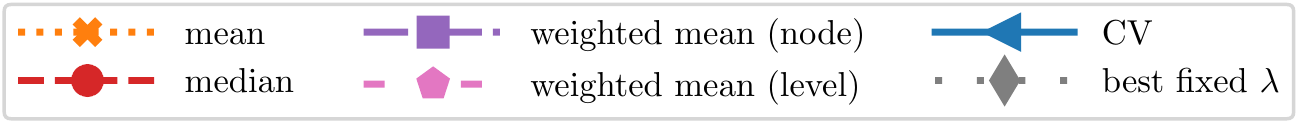}

\vspace{0cm}
\input{text_fig_mode_bias_noise}

\subfloat[Total binary tree]{\label{float:tree_single}
    \centering
    \includegraphics[width=\figurewidthlong]{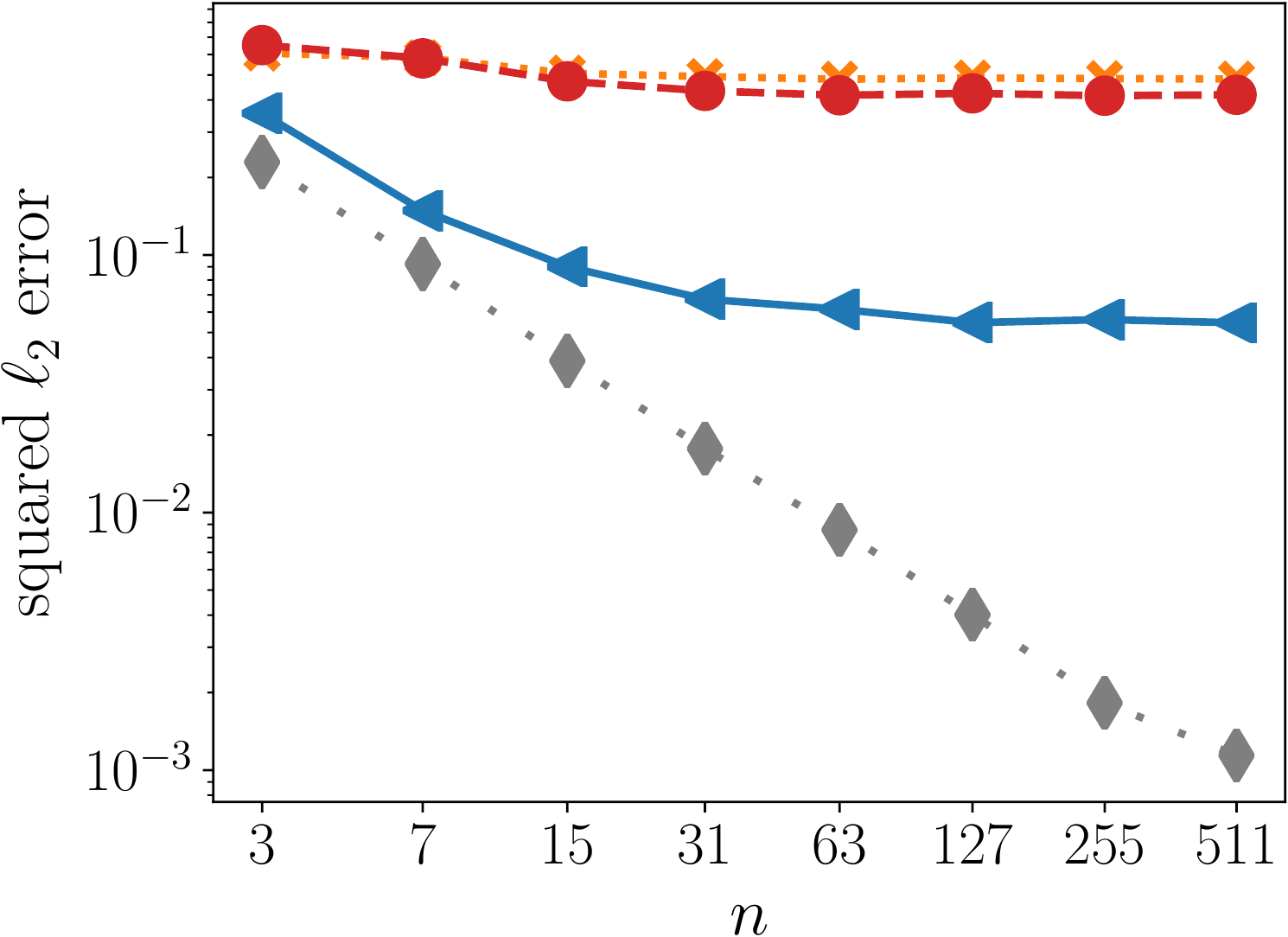}
    \includegraphics[width=\figurewidth]{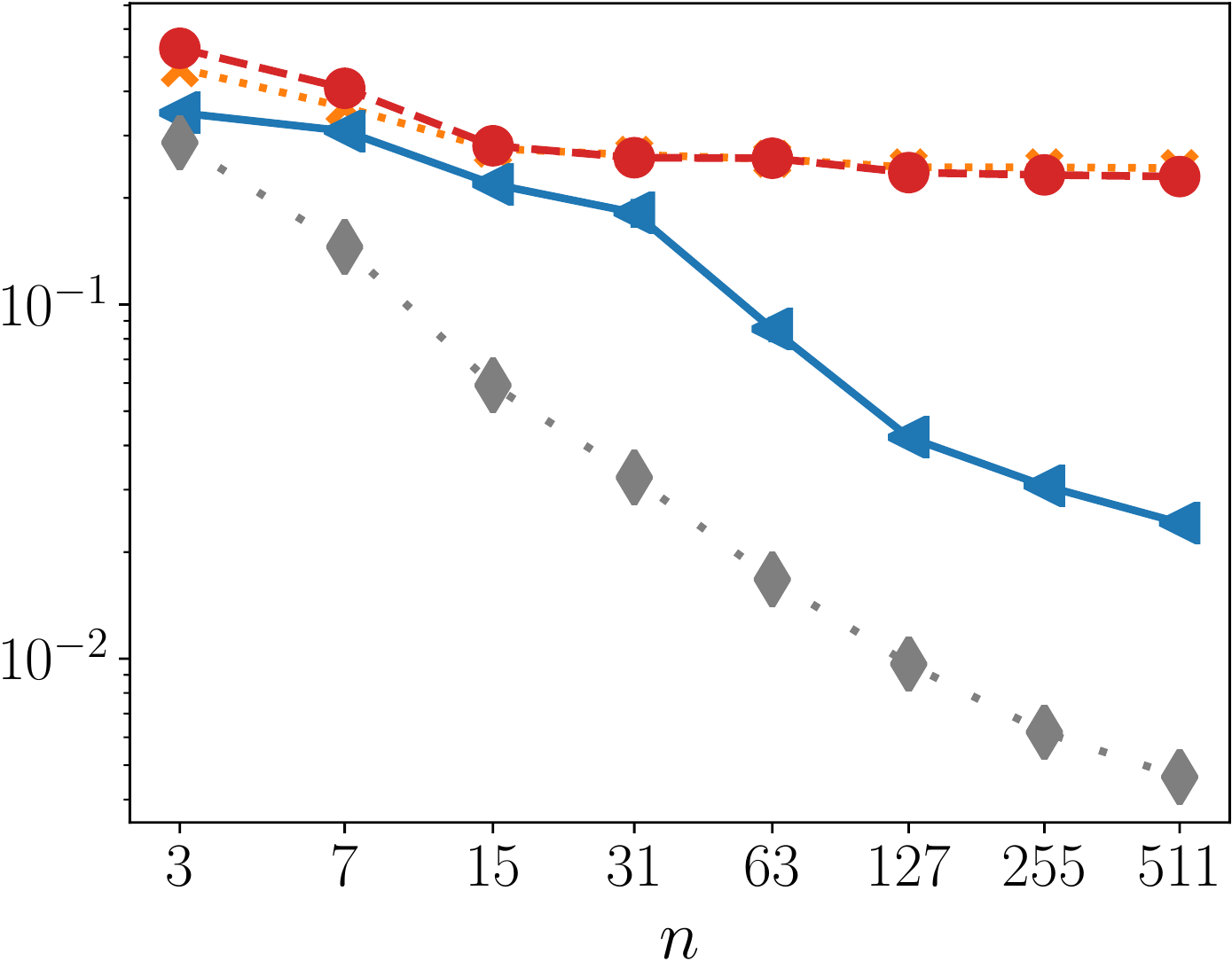}
    \includegraphics[width=\figurewidth]{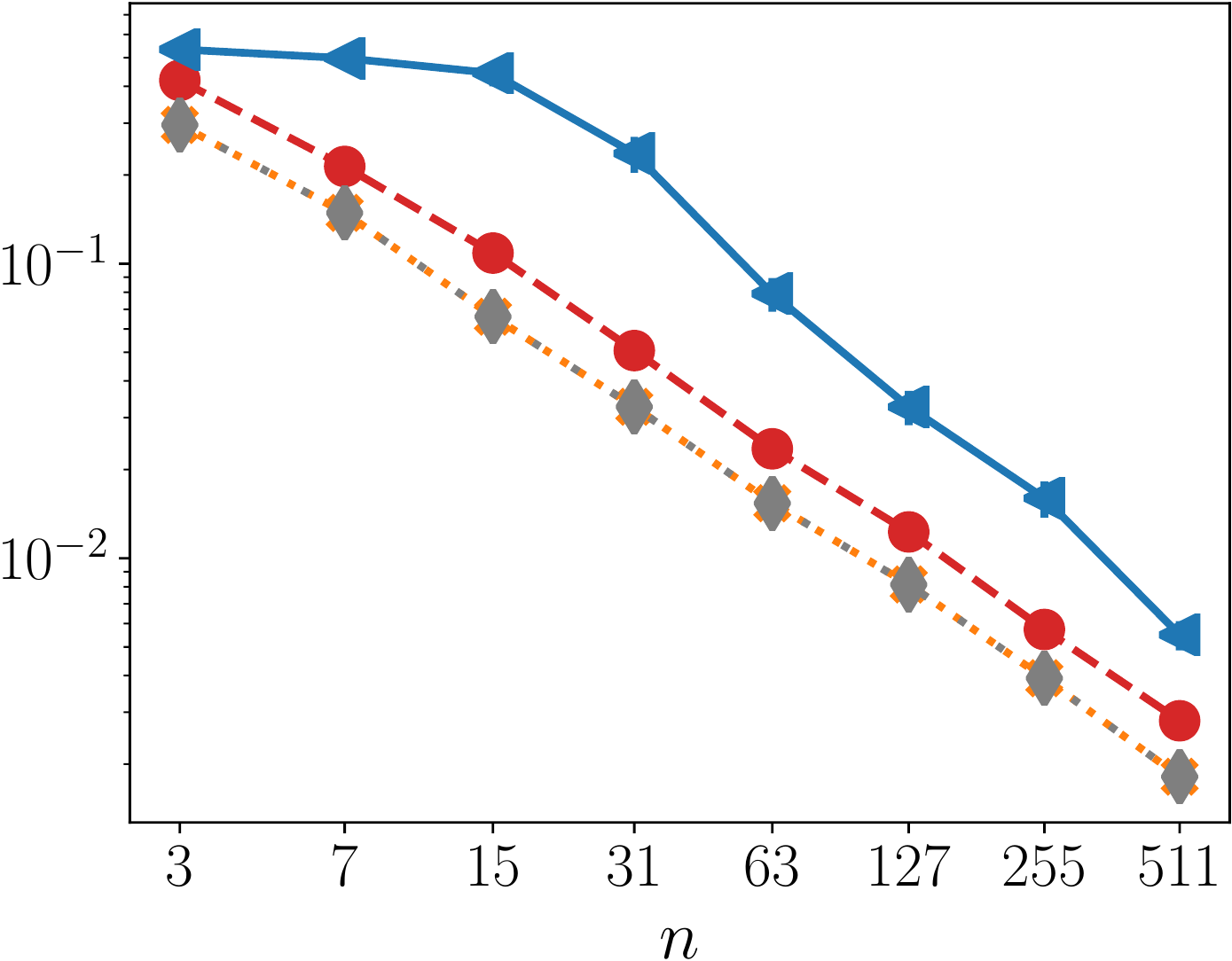}
}

\subfloat[Binary tree of $3$ levels]{\label{float:tree_level_3}
    \includegraphics[width=\figurewidthlong]{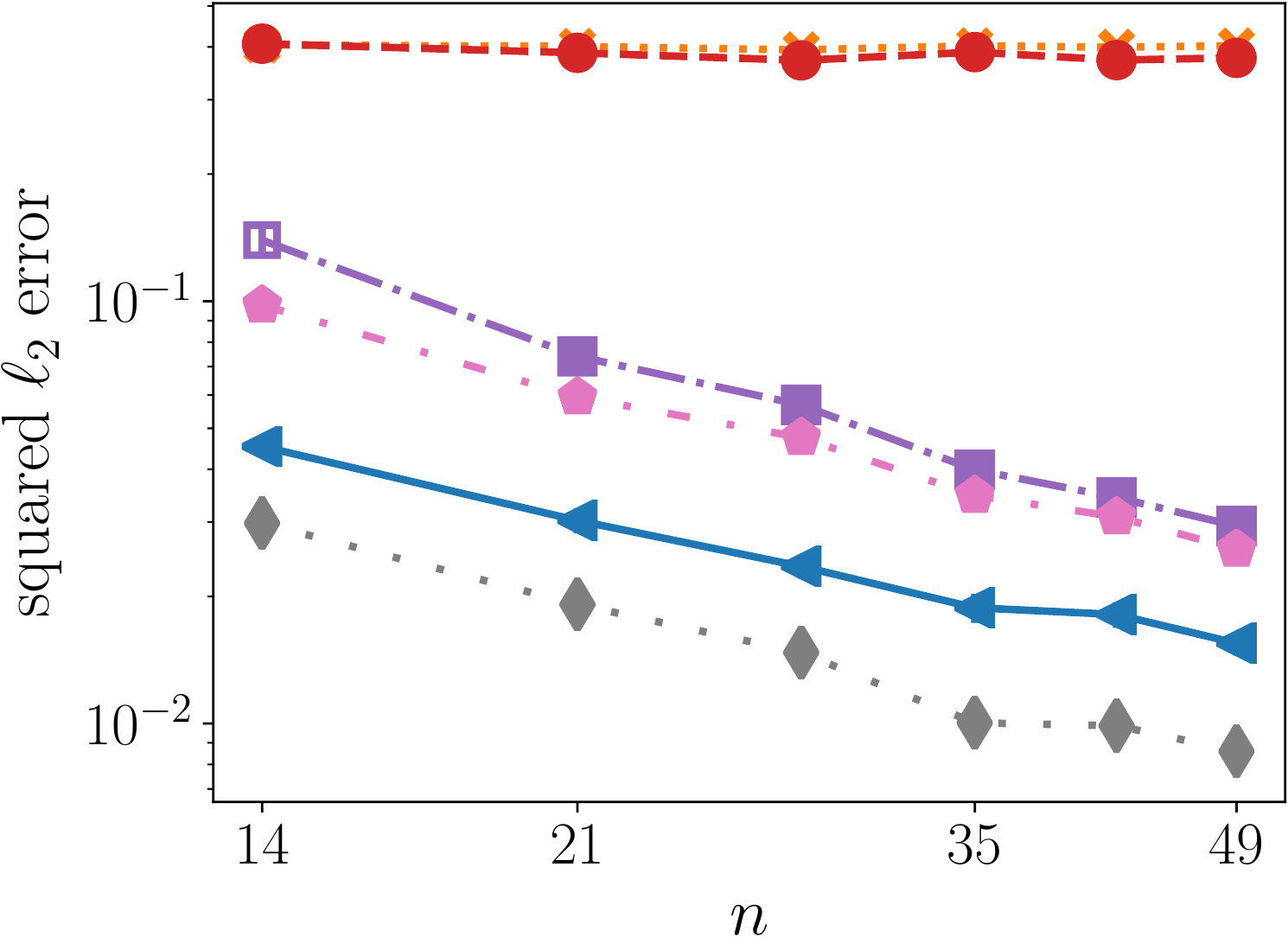}
    \includegraphics[width=\figurewidth]{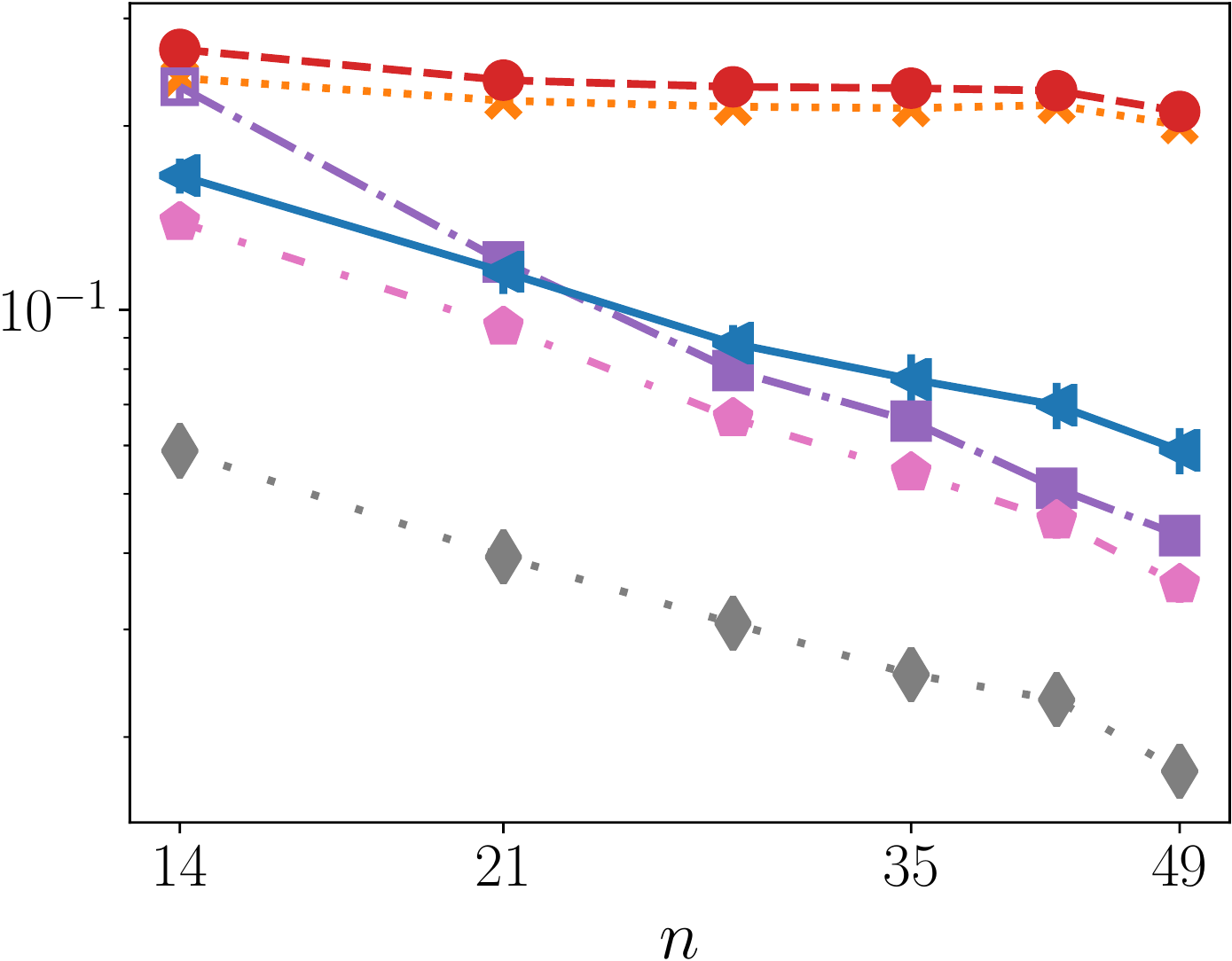}
    \includegraphics[width=\figurewidth]{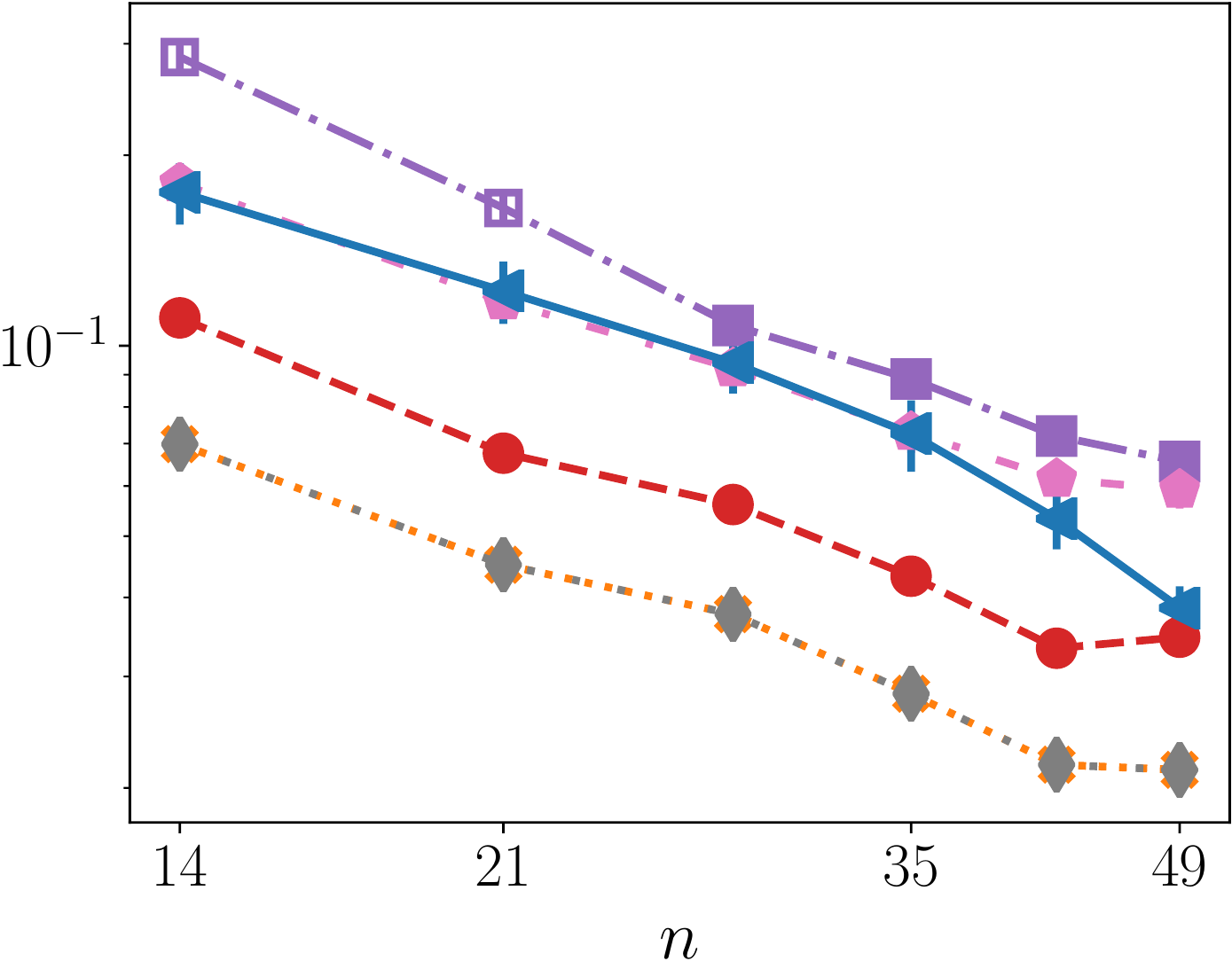}
}

 \caption{\label{fig:tree} The performance of our estimator (with cross-validation and with the best fixed $\reg$) compared to the mean, median, and two \subsampling estimators, under two types of partial orderings that are not group orderings.
 }
\end{figure*}

%% file: text_fig_indiana.tex
\begin{figure*}
\centering
    \subfloat[Overall]{
        \includegraphics[width=0.4\linewidth]{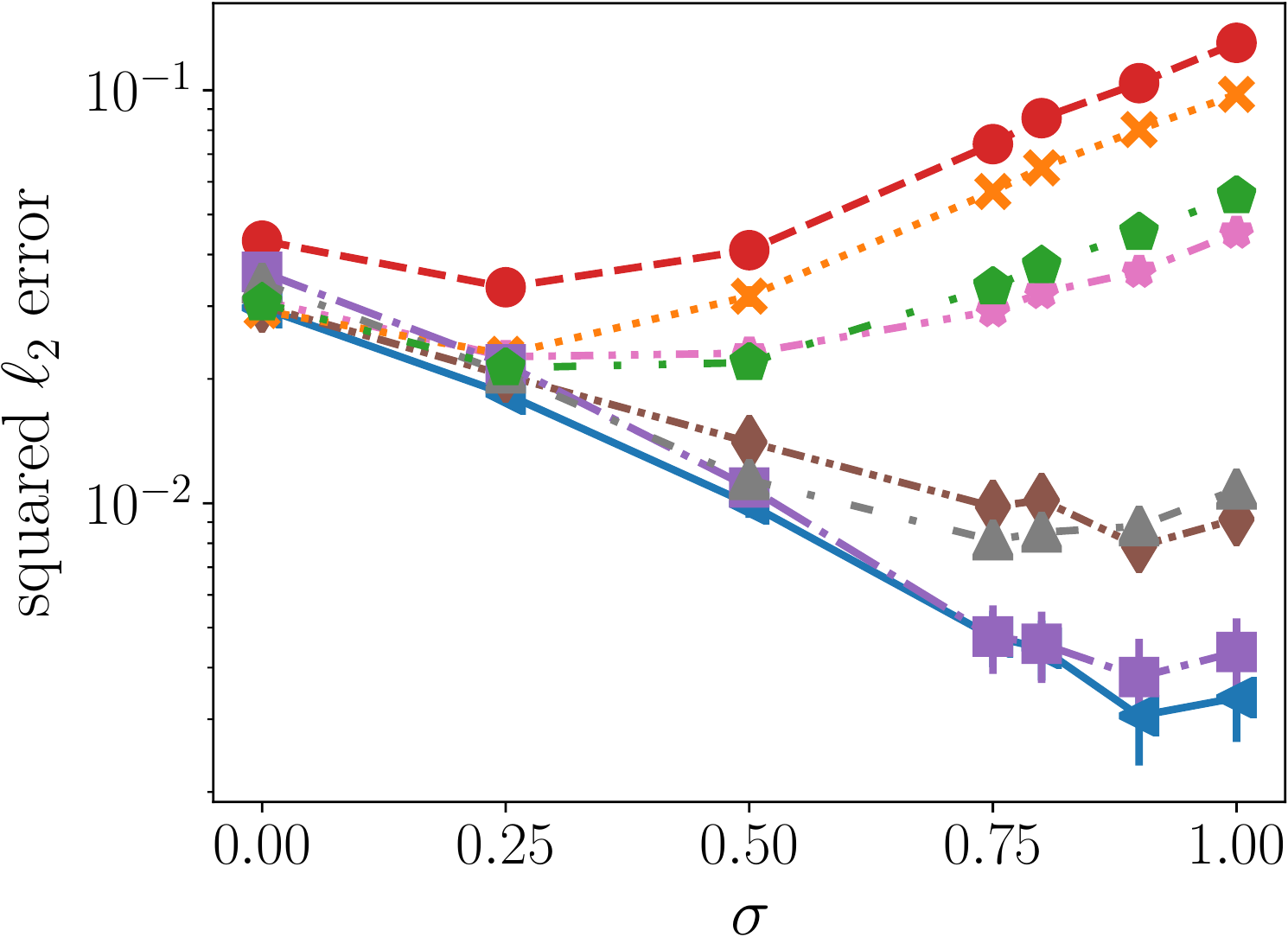}
        \hspace{0.5cm}
        \begin{minipage}[t]{0.5\linewidth}
            \raisebox{1.5cm}{\includegraphics[width=\linewidth]{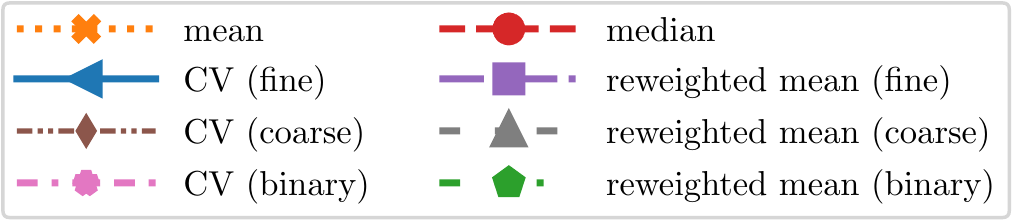}}
        \end{minipage}
    }\\
    \subfloat[Fine grades]{\label{float:indiana_fine_grades}
        \includegraphics[width=\figurewidthlong]{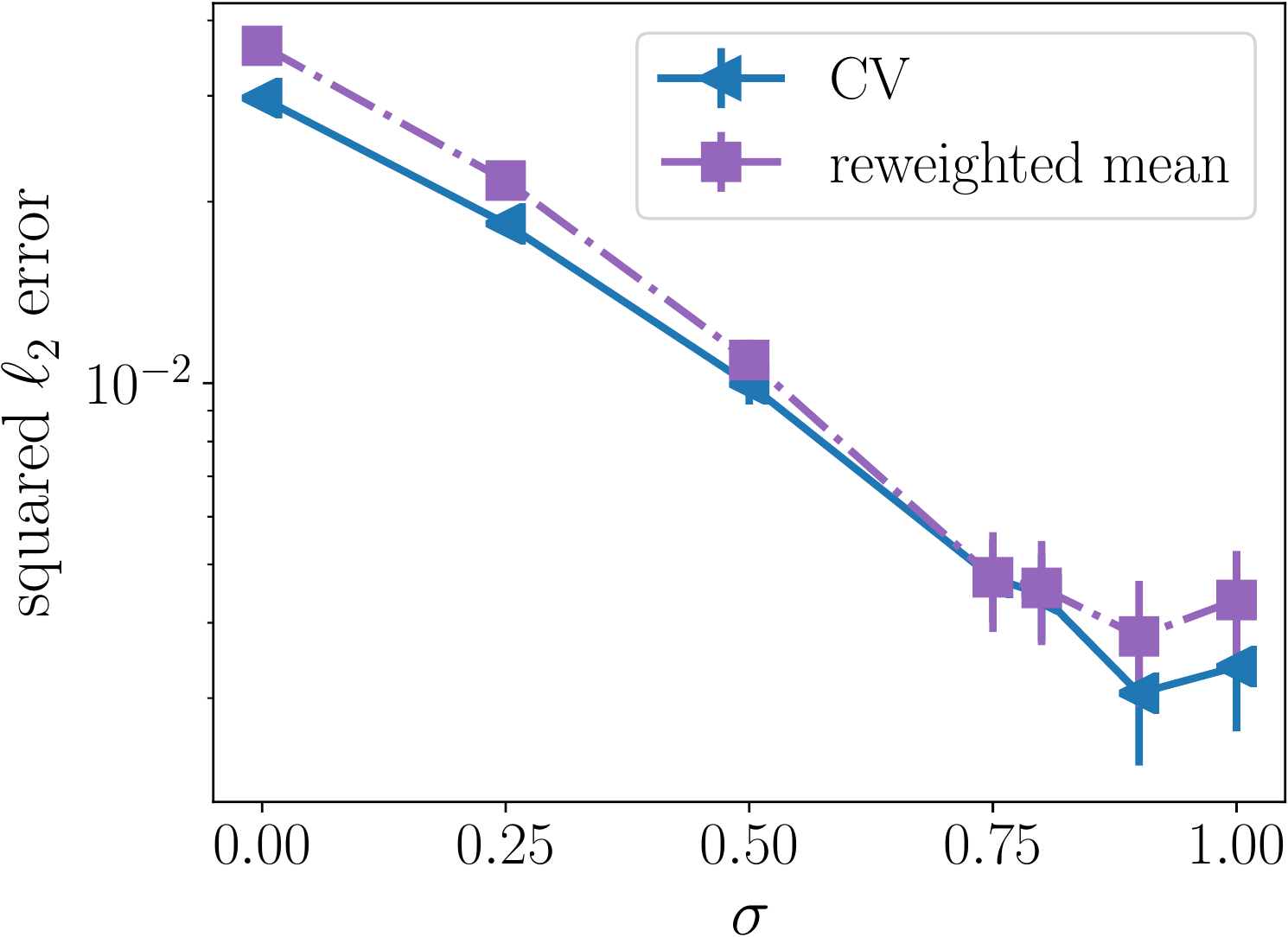}
    }
    \subfloat[Coarse grades]{\label{float:indiana_coarse_grades}
        \includegraphics[width=\figurewidth]{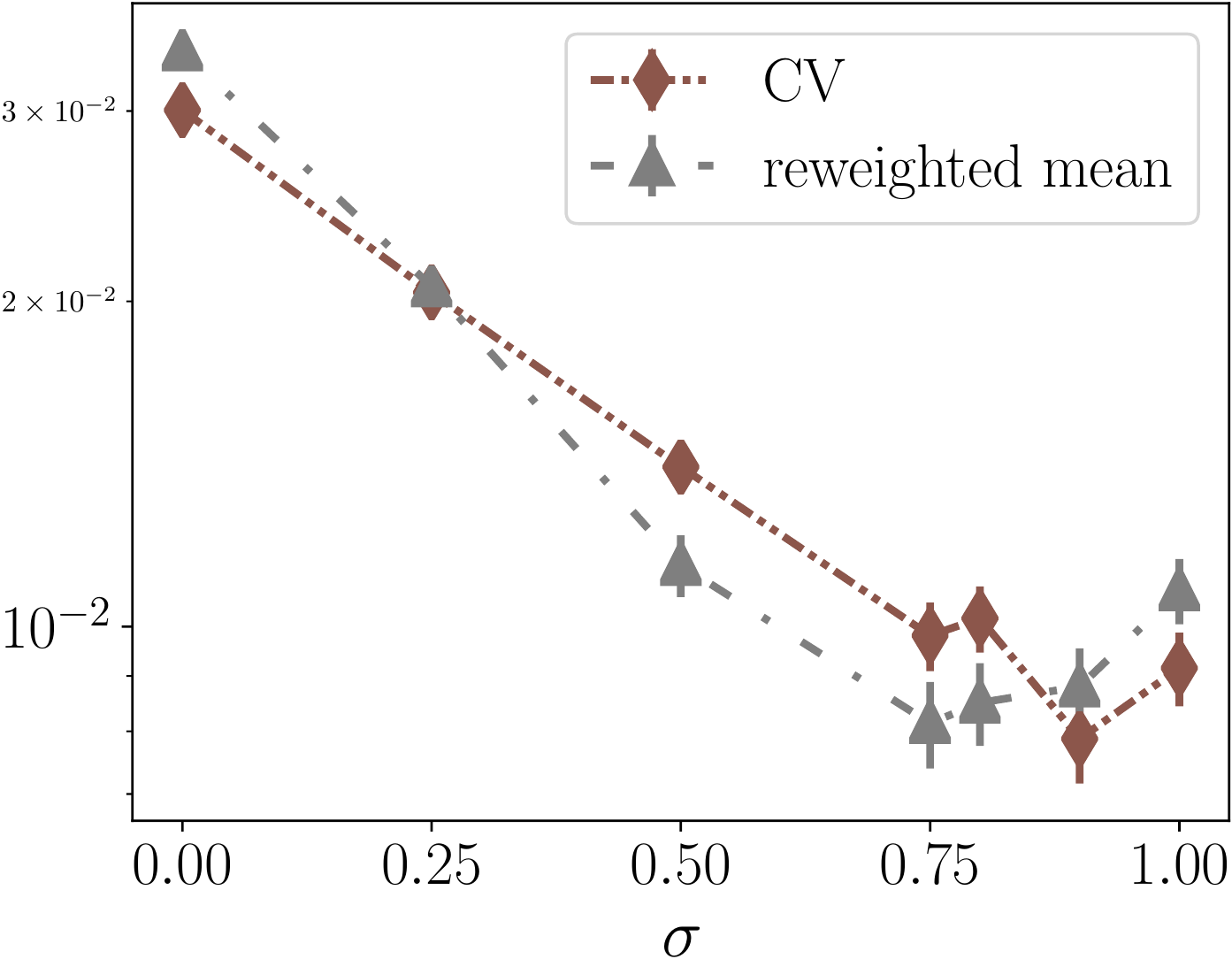}
    }
    \subfloat[Binary grades]{\label{float:indiana_binary_grades}
        \includegraphics[width=\figurewidth]{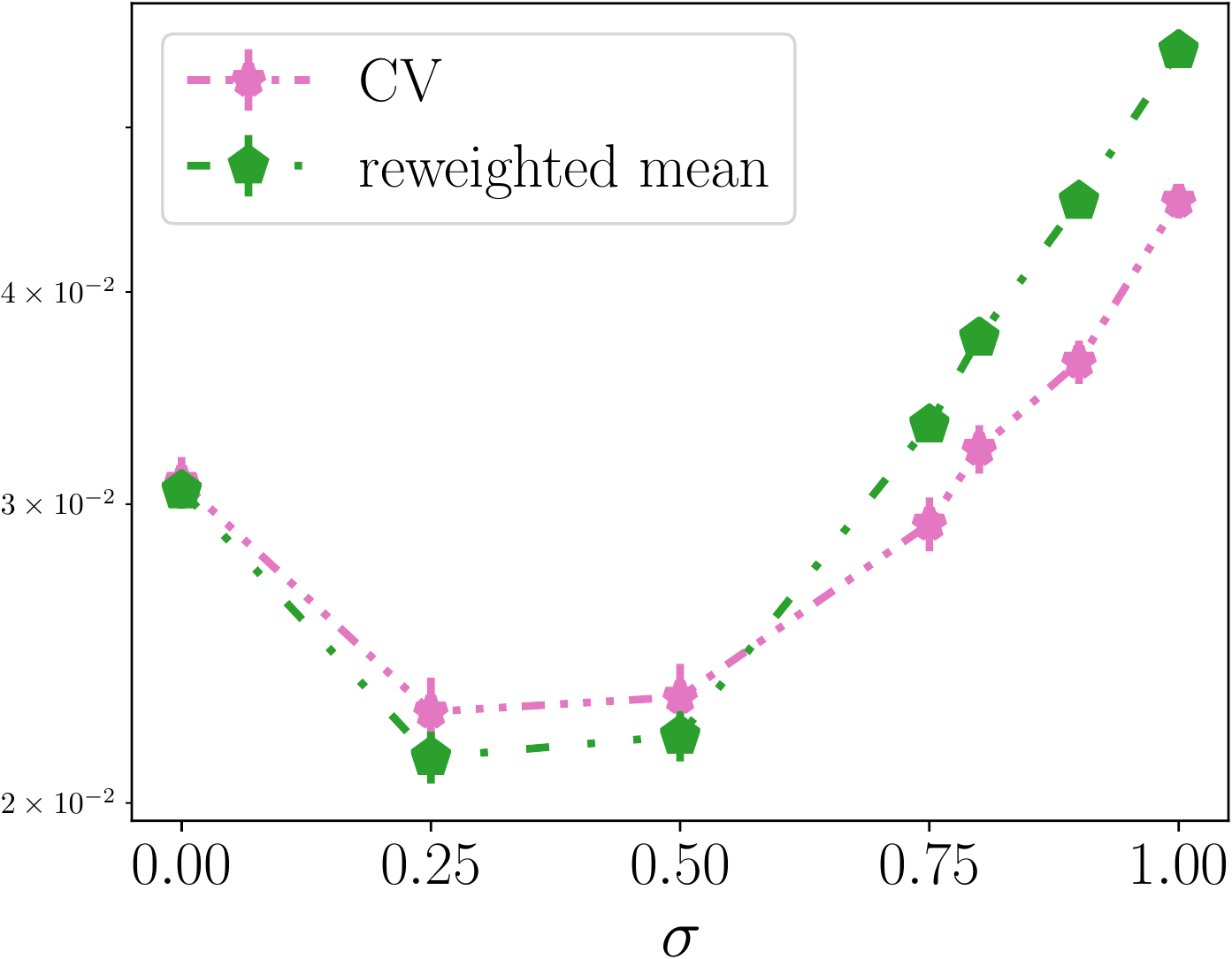}
    }

    \caption{\label{fig:indiana}
        The performance of our estimator (with \cv) on semi-synthetic grading data, compared to the mean, median and \subsampling estimators.
    }
\end{figure*}

%% file: text_appendix.tex
\vspace{1.5cm}
\noindent {\bf \Large Appendices}

\section{Auxiliary results} \label{app:auxilliary}

In this section, we present auxiliary theoretical results on comparing our estimator with the mean estimator (Appendix~\ref{app:mean}) and a \subsampling estimator that we introduce (Appendix~\ref{app:subsample}).  

\subsection{Comparison with the mean estimator} \label{app:mean}

Recall from Section~\ref{sec:experiment} that the mean estimator for estimating $\truemean$ is defined as $[\estmeanmean]_{\idxcourse} = \frac{1}{\numstudent}\sum_{\idxstudentscope} \obs_{\idxpair}$ for each class $\idxcoursescope$. Taking the mean ignores the bias, and hence it is natural to expect that this estimator does not perform well when the bias in the data is distributed unequally across classes. Intuitively, let us consider two classes of different quality. If students in a stronger class receive lower grades than students in a weaker class, then the bias induced by this distribution of grades may result in the mean estimator ranking the classes incorrectly. The following proposition formalizes this intuition and shows that the mean estimator indeed fails to compare the qualities of courses in the only-bias setting.

\begin{proposition}\label{prop:mean_not_consistent_comparision}
Suppose the assumptions~\ref{assumption:noise},~\ref{assumption:bias} and~\ref{assumption:d} hold and there is no noise, or equivalently $\gaussianwidthnoise=0$ in~\ref{assumption:noise}. Suppose the partial ordering satisfies any one of the conditions in Theorem~\ref{thm:consistency}:
    \begin{enumerate}[label={(\alph*)}]
        \item \label{part:mean_constant_fraction}
        any group ordering of $\numgroup$ groups with all $\const$-fractions, where $\const \in (0, \frac{1}{\numgroup})$ is a constant, or
        \item \label{part:mean_binary}
        any group ordering with $\numcourse=2$ courses and $\numgroup = 2$ groups, or
        \item \label{part:mean_total_order}
        any total ordering.
    \end{enumerate}
Then there exist a partial ordering that satisfies any one of the conditions~\ref{part:mean_constant_fraction} (with any number of groups $\numgroup\ge 2$),~\ref{part:mean_binary} or~\ref{part:mean_total_order}, true qualities $\truemean\in \reals^\numcourse$, a pair of courses $\idxcourse, \idxcoursealt \in [\numcourse]$, and an integer $\numstudentlb$ (dependent on the standard parameter $\gaussianwidthbias$ of the distribution of the bias and the number of groups $\numgroup$ in condition~\ref{part:mean_constant_fraction}), such that for all $\numstudent\ge \numstudentlb$, we have
\begin{align*}
    \Probbig{\sign\left([\estmeanmean]_\idxcourse - [\estmeanmean]_{\idxcoursealt}\right) = \sign(\truemean_\idxcourse - \truemean_{\idxcoursealt})}< 0.01.
\end{align*}
\end{proposition}
The proof of this result is provided in Appendix~\ref{app:proof_prop_mean_not_consistent}. Note that in condition~\ref{part:mean_constant_fraction} we require $\const\ne \frac{1}{\numgroup}$. This requirement is necessary because if $\const=\frac{1}{\numgroup}$, then the number of students in any course $\idxcoursescope$ and any group $\idxgroupscope$ has to be exactly $\const\numstudent$. In this case, the bias is evenly distributed across all courses, and in this case the mean estimator is consistent.
This negative result on comparing pairs of courses (combined with the fact that both model~\eqref{eq:model} and the mean estimator are shift invariant) implies the following negative result on estimation -- the mean estimator $\estmeanmean$ does not converge to the true $\truemean$ in probability. 

\begin{corollary}
\label{cor:mean_not_consistent_estimation}

Suppose the assumptions~\ref{assumption:noise},~\ref{assumption:bias} and~\ref{assumption:d} hold and there is no noise, or equivalently $\gaussianwidthnoise=0$ in~\ref{assumption:noise}. Consider any $\truemean\in \reals^\numcourse$. Suppose the partial ordering satisfies 
Then there exist a partial ordering that satisfies any one of the conditions~\ref{part:mean_constant_fraction},~\ref{part:mean_binary} or~\ref{part:mean_total_order}, and there exists a constant $\errbound>0$ such that for all $\numstudent\ge 1$ we have
    \begin{align*}
        \Probbig{\normtwo{\estmeanmean - \truemean}^2 < \errbound} < 0.01.
    \end{align*}
\end{corollary}
Recall that our estimator at $\reg=0$ is consistent in both comparing the quality of any pair of courses (Corollary~\ref{cor:consistency_comparison}) and estimating the qualities (Theorem~\ref{thm:consistency}). In contrast, the negative results in Proposition~\ref{prop:mean_not_consistent_comparision} and Corollary~\ref{cor:mean_not_consistent_estimation} show that the mean estimator is not consistent in comparison or estimation. Moreover, these negative results are stronger, in that they show the probability of correct comparison or estimation not only does not converge to $1$, but also can be arbitrarily small. The negative results on the mean estimator stem from the fact that the mean estimator completely ignores the fact that the bias is not evenly distributed across different courses. We remedy this issue by proposing a second baseline -- termed a \subsampling estimator in the following subsection.

\subsection{A \subsampling estimator}\label{app:subsample}
The second baseline, defined on group orderings only, re-weighs the observations to make the bias evenly distributed across courses, allowing to then take the mean. For each group $\idxgroup\in [\numgroup]$, denote $\blocklength_{\idxgroup, \textmin} \defn \min_{\idxcourse\in [\numcourse]} \blocklength_{\idxcourse\idxgroup}$ as the minimum number of students in group $\idxgroup$ among all courses. Denote $\setgroupidx = \{\idxgroup\in [\numgroup]: \blocklength_{\idxgroup, \textmin} > 0\}$ as the set of groups that appear in all courses. The \subsampling estimator consists of the following two steps.

\paragraph{Reweighting step} The estimator computes a weighted mean of each course $\idxcourse\in [\numcourse]$ as 
\begin{align}\label{eq:reweighting}
    [\estmeansubsample]_\idxcourse = \sum_{\idxgroup\in \setgroupidx} \frac{\blocklength_{\idxgroup, \textmin}}{\sum_{\idxgroupalt\in \setgroupidx} \blocklength_{\idxgroupalt, \textmin}}\sum_{\idxstudent: (\idxcourse, \idxstudent)\in \setgroup{\idxgroup}} \frac{\obs_\idxpair}{\blocklength_{\idxcourse\idxgroup}}.
\end{align}
Intuitively, the observations are reweighted in a way such that the bias distribution is balanced among courses. Specifically, for each course $\idxcourse\in[\numcourse]$ and each group $\idxgroup\in [\numgroup]$, this \subsampling estimator computes its group mean $\sum_{\idxstudent: (\idxcourse, \idxstudent)\in \setgroup{\idxgroup}} \frac{\obs_{\idxcourse\idxstudent}}{\blocklength_{\idxcourse\idxgroup}}$, and weighs the contribution of this group mean to the overall mean by the factor of $\frac{\blocklengthgroupmin{\idxgroup}}{\sum_{\idxgroupalt\in \setgroupidx} \blocklengthgroupmin{\idxgroupalt}}$. This reweighting can bee seen as the expected version of a sampling procedure, where for each course $\idxcourse\in [\numcourse]$ and each group $\idxgroup\in [\numgroup]$, we sample $\blocklengthgroupmin{\idxgroup}$ out of $\blocklength_{\idxcourse\idxgroup}$ observations so that the number of observations in group $\idxgroup$ is equal across all courses, and then take the mean on the sampled observations. Note that there are an infinite number choices for the weights to balance the biases, and the choice in~\eqref{eq:reweighting} motivated by sampling is quite natural. It has the property that if all courses have the same group distribution, then the reweighted mean reduces to sample mean.

\paragraph{Recentering step} We use the assumption that the bias and noise are centered, that is, $\sum_{\idxcourse\in [\numcourse]\idxstudent\in [\numstudent]}\Expect[\bias_{\idxcourse\idxstudent}]= 0$ and $\sum_{\idxcourse\in [\numcourse],\idxstudent\in [\numstudent]}\Expect[\noise_{\idxcourse\idxstudent}]= 0$. Under this assumption, we have
\begin{align}\label{eq:subsampling_recenter_equality}
    \frac{1}{\numstudent}\sum_{\idxcoursescope, \idxstudentscope} \Expect[\obs_{\idxpair}] =\frac{1}{\numstudent} \sum_{\idxcoursescope, \idxstudentscope}\Expect[\truemean_\idxcourse + \bias_{\idxpair} + \noise_\idxpair] = \sum_{\idxcoursescope}\truemean_\idxcourse.
\end{align}
Hence, we shift $\estmeansubsample$ by a constant such that the empirical version of~\eqref{eq:subsampling_recenter_equality} holds, that is, $\sum_{\idxcourse\in \numcourse} [\estmeansubsample]_\idxcourse =\frac{1}{\numstudent} \sum_{\idxcoursescope, \idxstudentscope} \obs_{\idxcourse\idxstudent}$. 
\begin{align}\label{eq:recentering}
    \estmeansubsample\leftarrow \estmeansubsample +\left(- \frac{1}{\numcourse} \sum_{\idxcourse\in [\numcourse]}[\estmeansubsample]_\idxcourse + \frac{1}{\numcourse\numstudent}\sum_{\idxcoursescope, \idxstudentscope} \obs_\idxpair \right)\vecone
\end{align}
This recentering step is necessary, because the expected mean of the bias over all courses after the reweighting step may not be $0$, as the reweighting step only aligns the bias across courses, but not necessarily to $0$. From~\eqref{eq:estimator_sum_equality_mean} in Lemma~\ref{lem:estimator_sum_equality}, our estimator also satisfies $\sum_{\idxcoursescope}\estmean_\idxcourse =\frac{1}{\numstudent}\sum_{\idxcoursescope,\idxstudentscope} \obs_{\idxcourse\idxstudent}$ for all $\reg\in [0, \infty]$, so this recentering also ensures a fair comparison with our estimator. Empirically we observe that the \subsampling estimator always performs better after the recentering step.

Note that \subsampling is undefined for total orderings. For group orderings with all constant fractions, \subsampling is also consistent. In this case, we present a simple example below, where our estimator at $\reg=0$ still performs better than \subsampling by a constant factor (uniform bias is assumed for analytical tractability).

\begin{proposition}\label{prop:uniform_example}
Suppose the number of courses is $\numcourse= 2$.
Suppose the number of groups is $\numgroup=2$, with a grade distribution of $(\blocklength_{11}, \blocklength_{12}) = ((\fraction\numstudent, (1-\fraction)\numstudent)$ and $(\blocklength_{21}, \blocklength_{22}) = ((1-\fraction)\numstudent, \fraction\numstudent)$ for some $\fraction\in (0, 1)$. Suppose there is no noise. Suppose bias in group $1$ is generated i.i.d. from $\uniform[-1, 0]$, and bias in group $2$ is generated i.i.d. from $\uniform[0, 1]$. Then the squared $\ell_2$-risk for the \subsampling estimator is $\estmeansubsample$ and for our estimator $\estmeanat{0}$ at $\reg=0$ is respectively
\begin{align*}
    \frac{1}{2}\Expect\norm*{\estmeansubsample - \truemean}_2^2 & = \frac{1}{24\numstudent} + \frac{1}{96\fraction(1-\fraction)\numstudent}\ge \frac{1}{12\numstudent}\\
    \frac{1}{2}\Expect\norm{\estmeanat{0}- \truemean}_2^2 & = \frac{1}{24\numstudent} + O\left(\frac{1}{\numstudent^2}\right).
\end{align*}
\end{proposition}
The proof of this result is provided in Appendix~\ref{app:proof_prop_uniform_example}.
Note that the risk  of our estimator is at most half of the error of \subsampling, if ignoring the higher-order term $O\left(\frac{1}{\numstudent^2}\right)$.

\section{Additional experimental details}\label{app:simulation}

In this section, we provide additional details for the experiments in Section~\ref{sec:experiment}.

\subsection{Implementation}\label{app:simulation_implementation}
We now discuss the implementation of our estimator. 

\paragraph{Solving the optimization (Line~\ref{line:compute_err_start} in Algorithm~\ref{alg:cv}):} We describe the implementation of solving the optimization~\eqref{eq:optimization} depending on the value of $\reg$.

\begin{itemize}
    \item \textbf{\boldmath{$\reg=\infty$}:} The estimator is computed as taking the mean of each course according to Proposition~\ref{prop:property_existence_at_infty}. 

    \item \textbf{\boldmath{$\reg \in (0, \infty)$}:}
    In the proof of Proposition~\ref{prop:uniqueness} we show that the objective~\ref{eq:model} is strictly convex in $(\meancourse, \biasmtx)$ on a convex domain. Hence, the problem is a QP with a unique solution. We solve for the QP using the CVXPY package. 

    \item \textbf{\boldmath{$\reg =0$}:}
    It can be shown that the objective~\eqref{eq:model} is still convex, but there may exist multiple solutions before the tie-breaking. We first obtain one solution of the QP using CVXPY, denoted $(\meancourse_0, \bias_0)$. The optimization~\eqref{eq:optimization} only has the first term, which is an $\ell_2$-projection from $\obs$ to the convex domain $\{\meancourse\vecone^T  + \bias: \meancourse\in \reals^\numcourse, \bias\in \reals^{\numcourse\times\numstudent}, \bias \text{ satisfies }\setpartialorder\}$. Hence, the value of $(\meancourse\vecone^T + \bias)$ is unique among all solutions $(\meancourse, \bias)$, and the set of solutions can be written as $\{(\meancourse, \bias): \meancourse=\meancourse_0 + u, \bias=\bias_0 - u\vecone^T, u\in \reals^\numcourse\}$. We implement the tie-breaking by solving $u$ using CVXPY, minimizing $\norm*{\bias}_F^2 = \norm{\bias_0 - u\vecone^T}_F^2$ subject to the ordering constraints on $\bias = \bias_0 - u\vecone^T$.
\end{itemize}

Finally, we discuss a speed-up technique for solving the QP. For total orderings, the number of constraints in $\setpartialorder$ is linear in the number of samples, whereas for general group orderings, the number of constraints in $\setpartialorder$ can become quadratic, making the QP solver slow. To speed up the optimization, it can be shown that for all \elements within any course and any group, the ordering of the estimated bias $\estbiasmtx$ at these \elements is the same as the ordering of the observations $\obsmtx$ at these \elements. Therefore, among the constraints in $\setpartialorder$ involving these \elements, we only keep the constraints that involve the maximum and the minimum \elements in this course and this group. Then we add the ordering of $\obsmtx$ at these \elements to the partial ordering $\setpartialorder$. This replacement reduces the number of constraints in $\setpartialorder$ and speeds up the QP solver.

\paragraph{Sampling a total ordering from the partial ordering $\setpartialorder$ (Line~\ref{line:sample_total_order} in Algorithm~\ref{alg:cv}):} When $\setpartialorder$ is a group ordering, sampling a total ordering uniformly at random is implemented by first sorting the \elements according to their group, and then permuting the them uniformly at random within each group.

When $\setpartialorder$ is a tree or a group tree, we sample a total ordering using the following procedure. We first take all \elements at the root of the tree, and place them in the total ordering as the lowest-ranked \elements (if there are multiple \elements at the root, then permute them uniformly at random in the total ordering). Consider each sub-tree consisting of a child node of the root and all its descendants. For the remaining positions in the total ordering, we assign these positions to the sub-trees uniformly at random. Then we proceed recursively to sample a total ordering for each sub-tree, and fill them back to their positions in the total ordering.

\paragraph{Interpolation (Line~\ref{line:interpolation} in Algorithm~\ref{alg:cv}):}
We sample $100$ total orderings to approximate the interpolation. 

\subsection{Extending the \subsampling estimator to tree orderings}\label{app:subsampling_tree}

We introduce the definitions of the two \subsampling estimators on tree orderings used in the simulation in Section~\ref{sec:exp_tree}. Note that the \subsampling estimator defined in Appendix~\ref{app:subsample} is with respect to the groups $\{\setgroup{\idxgroup}\}_{\idxgroupscope}$. We replace the groups in the \subsampling estimator by the following two partitions of the \elements.

\noindent \textbf{\Subsampling (node):} Each subset in the partition consists of all \elements in the same node of the tree.

\noindent \textbf{\Subsampling (level): } Each subset in the partition consists of all \elements on the same level of the tree.

\subsection{Extending our estimator and the \subsampling estimator to an unequal number of students per course}\label{app:experiment_unequal}

In the semi-synthetic experiment in Section~\ref{sec:experiment_indiana}, the number of students is unequal in different courses. We describe a natural extension of the \subsampling estimator and our estimator to this case.

First, we explain how to format the observations back to a matrix form. Denote $\numstudent_\idxcourse$ as the number of students in course $\idxcourse\in [\numcourse]$. Let $\numstudent=\max_{\idxcoursescope} \numstudent_\idxcourse$. Construct a matrix $\obsmtx\in \reals^{\numcourse\times \numstudent}$, where the first $\numstudent_\idxcourse$ \elements in each row $\idxcoursescope$ correspond to the observations in this course, and the values of the remaining \elements are set arbitrarily. Construct the set of observations $\set\in [\numcourse]\times [\numstudent]$, where the first $\numstudent_\idxcourse$ \elements in each row $\idxcoursescope$ are in $\set$. Estimation under an unequal number of students per course is equivalent to estimation given $\obsmtx$ (and its corresponding partial ordering $\setpartialorder$) restricted to the set $\set$. It remains to define the \subsampling estimator and our estimator restricted to any set $\set\in [\numcourse]\times [\numstudent]$.

\paragraph{The \subsampling estimator:}
In the definition of the the \subsampling estimator in Appendix~\ref{app:subsample}, the reweighting step is the same (only using the observations in $\set$). The recentering step restricted to $\set$ is defined as:
\begin{align*}
    \estmeansubsample\leftarrow \estmeansubsample +\left(- \sum_{\idxcourse\in [\numcourse]} \frac{\numstudent_\idxcourse}{\abs*{\set}}[\estmeansubsample]_\idxcourse + \frac{1}{\abs*{\set}}\sum_{\idxcoursescope, \idxstudentscope} \obs_\idxpair \right)\vecone
\end{align*}
Similar to Appendix~\ref{app:subsample}, after this recentering step, the \subsampling estimator satisfies the empirical version of an equality (Eq.~\eqref{eq:estimator_sum_equality_mean_variable_size} in Appendix~\ref{app:property_of_estimator}) that our estimator also satisfies.

\paragraph{Our estimator:} We extend Algorithm~\ref{alg:cv} naturally to being restricted to a set $\set$ as follows. In the data-splitting step, in Line~\ref{line:sample_total_order}, we replace the number of \elements from $\numcourse\numstudent$ to $\sum_{\idxcoursescope}\numstudent_\idxcourse$; in Lines~\ref{line:find_sub_order}-\ref{line:assign_end}, we replace the number of students from $\numstudent$ to $\numstudent_\idxcourse$, and only find the sub-ordering of the $\numstudent_\idxcourse$ \elements in $\set$. The validation step remains the same. 

%% file: text_proof.tex
\section{Proofs}\label{app:proof}

In this section, we provide proofs for all the theoretical claims made earlier. We begin by introducing some additional notation in Section~\ref{app:notation} which is used throughout the proofs. In Section~\ref{app:preliminaries}, we then provide certain preliminaries that are useful for the proofs. We then present the proofs in subsequent subsections.

For ease of notation, we ignore rounding throughout the proofs as it does not affect the claimed results.

\subsection{Notation}\label{app:notation}

\paragraph{Training-validation split $(\settrain, \setval)$:}
By Algorithm~\ref{alg:cv}, the number of \elements restricted to the set $\settrain$ or $\setval$ is the same for each course $\idxcourse$. Hence, we denote $\numstudenttrain$ and $\numstudentval$ as the number of students per course in $\settrain$ and $\setval$ respectively. Throughout the proofs, for simplicity we assume that $\numstudent$ is \emph{even}. In this case we have
\begin{align}\label{eq:size_train_val_half}
    & \numstudenttrain = \numstudentval=\frac{\numstudent}{2}.
\end{align}
All the proofs extend to the case where $\numstudent$ is odd under minor modifications.

We define the \elements in each course $\idxcoursescope$ restricted to $\settrain$ or $\setval$ as:
\begin{align*}
    & \settraincourse{\idxcourse} \defn \{\idxpairparen\in \settrain\}\\
    & \setvalcourse{\idxcourse} \defn \{\idxpairparen\in \setval\}.
\end{align*}
We slightly abuse the notation and say $\idxstudent\in \settraincourse{\idxcourse}$ if $\idxpairparen\in \settraincourse{\idxcourse}$. Likewise for $\setvalcourse{\idxcourse}$.

\paragraph{Group orderings:} Recall that from Definition~\ref{def:group_ordering} that $\setgroupof{\idxgroup}$ denotes the set of \elements in group $\idxgroupscope$. We define 
\begin{align*}
    & \settraingroup{\idxgroup} \defn \setgroupof{\idxgroup}\intersect \settrain\\
    & \setvalgroup{\idxgroup}\defn \setgroupof{\idxgroup}\intersect \setval.
\end{align*}
We denote the \elements of group $\idxgroup\in [\numgroup]$ in course $\idxcourse\in [\numcourse]$ restricted to $\setval$ as:
\begin{align*}
    \setcoursegroup{\idxcourse}{\idxgroup} \defn \setgroupof{\idxgroup}\intersect\setcourse{\idxcourse}.
\end{align*}
Furthermore, we define the \elements of $\setcoursegroup{\idxcourse}{\idxgroup}$ restricted to $\setval$ as
\begin{align*}
    \settraincoursegroup{\idxcourse}{\idxgroup} \defn \settraingroup{\idxgroup}\intersect \settraincourse{\idxcourse}\qquad
    \setvalcoursegroup{\idxcourse}{\idxgroup} \defn \setvalgroup{\idxgroup}\intersect \setvalcourse{\idxcourse}.
\end{align*} 
Again, we slightly abuse the notation and say $\idxstudent\in \setvalcoursegroup{\idxcourse}{\idxgroup}$ if  $\idxpairparen\in \setvalcoursegroup{\idxcourse}{\idxgroup}$.

We define $\blocklength_{\idxcourse\idxgroup}$ as the the number of students of group $\idxgroup \in [\numgroup]$ in course $\idxcourse \in [\numcourse]$.  We define $\blocklength_\idxgroup$ as the number of students of group $\idxgroup\in [\numgroup]$. We denote $\blocklength_{-\idxcourse, \idxgroup}$ as the number of students of group $\idxgroupscope$ and not in course $\idxcourse$. Namely,
\begin{subequations}\label{eq:def_blocklength}
\begin{align}
    & \blocklength_{\idxcourse\idxgroup}\defn \abs*{\setcoursegroup{\idxcourse}{\idxgroup}}\\
    & \blocklength_\idxgroup\defn \abs*{\setgroupof{\idxgroup}} = \sum_{\idxcoursescope} \blocklength_{\idxcourse\idxgroup}\\
    & \blocklength_{-\idxcourse, \idxgroup} \defn \abs*{\setgroupof{\idxgroup}\setminus \setcoursegroup{\idxcourse}{\idxgroup}}= \sum_{\idxcoursealt\ne \idxcourse} \blocklength_{\idxcoursealt\idxgroup}.
\end{align}
\end{subequations}
Furthermore, we define
\begin{subequations}\label{eq:def_blocklength_course_group_set}
\begin{align}
    & \blocklengthgrouptrain{\idxgroup} \defn \abs*{\settraingroup{\idxgroup}} \qquad
    \blocklengthgroupval{\idxgroup} \defn \abs*{\setvalgroup{\idxgroup}},\\
    & \blocklengthcoursegrouptrain{\idxcourse}{\idxgroup} \defn \abs*{\settraincoursegroup{\idxcourse}{\idxgroup}} \qquad
    \blocklengthcoursegroupval{\idxcourse}{\idxgroup} \defn \abs*{\setvalcoursegroup{\idxcourse}{\idxgroup}}.
\end{align}
\end{subequations}

\paragraph{Total ordering:} 
Consider the $\numcourse\numstudent$ \elements. We say that the \element $(\idxcourse, \idxstudent)$ is of rank $\rank\in [\numcourse\numstudent]$ if $\idxpairparen$ is the $\rank^\thcount$-smallest \element in among the $\numcourse\numstudent$ \elements.

We denote $\rank_\idxpair$ as the rank of each \element $\idxpairparen\in [\numcourse]\times [\numstudent]$. We denote $(\idxcourse_\rank, \idxstudent_\rank)$ as the \element of rank $\rank\in [\numcourse\numstudent]$.

\paragraph{Observations $\obsmtx$ and bias $\biasmtx$:} Denote the mean of all observations as 
\begin{align}\label{eq:def_mean_obs}
    \meanobs = \frac{1}{\numcourse\numstudent}\sum_{\idxcourse\in [\numcourse], \idxstudent\in [\numstudent]} \obs_{\idxcourse\idxstudent}. 
\end{align}
Denote the mean of the observations in any course $\idxcourse\in [\numcourse]$ as 
\begin{align}\label{eq:def_mean_obs_course}
    \meanobscourse{\idxcourse} = \frac{1}{\numstudent}\sum_{\idxstudent=1}^\numstudent\obs_{\idxcourse\idxstudent}.
\end{align}
Likewise we denote the mean of the bias in any course $\idxcoursescope$ as $\biascourse{\idxcourse}$.
We denote the mean of the bias of any course $\idxcoursescope$ as
\begin{align*}
    \meanbiasgroup{\idxgroup} = \frac{1}{\blocklength_\idxgroup}\sum_{\idxpairparen\in \setgroupof{\idxgroup}} \bias_{\idxpair}.
\end{align*}

Now restrict to group orderings. For any course $\idxcourse\in [\numcourse]$ and any group $\idxgroupscope$, denote the smallest and the largest observation in course $\idxcourse$ and group $\idxgroup$ as
\begin{subequations}\label{eq:def_obs_course_group_extremal}
\begin{align}
    \obscoursegroupmax{\idxcourse}{\idxgroup} \defn \max_{\idxstudent: (\idxcourse, \idxstudent)\in \setgroupof{\idxgroup}} \obs_{\idxcourse\idxstudent}\\
    \obscoursegroupmin{\idxcourse}{\idxgroup} \defn \min_{\idxstudent: (\idxcourse, \idxstudent)\in \setgroupof{\idxgroup}} \obs_{\idxcourse\idxstudent}
\end{align} 
\end{subequations}
We define $\biascoursegroupmax{\idxcourse}{\idxgroup}$ and $\biascoursegroupmin{\idxcourse}{\idxgroup}$ likewise. In addition, we define the smallest and the bias of any group $\idxgroupscope$ as
\begin{align}\label{eq:def_bias_group_extremal}
\begin{split}
    & \biasgroupmin{\idxgroup} = \min_{(\idxcourse, \idxstudent)\in \setgroupof{\idxgroup}} \bias_{\idxpair}\\
    & \biasgroupmax{\idxgroup} = \max_{(\idxcourse, \idxstudent)\in \setgroupof{\idxgroup}} \bias_{\idxpair}.
\end{split}
\end{align}

\paragraph{Statistics:}
We $\pdf$ as the p.d.f. of $\normal(0, 1)$. Denote $\cdf$ and $\invcdf$ as the corresponding c.d.f., and the inverse c.d.f., respectively. We slightly abuse notation and write $\Prob(X)$ as the p.d.f. of any continuous variable $X$.

For a set of i.i.d. random variables  $X_1, \ldots, X_n$, we denote $X^{(k)}$ as the $k^\thcount$ order statistics of $\{X_i\}_{i=1}^n$. We use the notation $X^{(k:n)}$ when we emphasize the sample size $n$.

Let $d\ge 2$ be any integer, and let $\totalorder$ be a total ordering of size $d$. We denote the monotonic cone with respect to $\totalorder$ as $\conemonotone\defn \left\{\theta\in \reals^d: \theta_{\totalorder(1)} \le \ldots \le \theta_{\totalorder(d)}\right\}$. For any vector $x\in \reals^d$, we denote the isotonic projection of $x$ as 
\begin{align}\label{eq:isotonic_proj}
    \proj_{\conemonotone}(x) \defn \argmin_{u\in \conemonotone_\totalorder} \normtwo{x-u}^2.
\end{align}
We denote $\conemonotone$ as the monotonic cone with respect to the identity ordering. 

\paragraph{Our estimator and the \cv algorithm:}
Recall from Line~\ref{line:compute_err_start} of Algorithm~\ref{alg:cv} that our estimator restricted to any set of elements $\set\subseteq[\numcourse]\times [\numstudent]$ is defined as the solution to:
\begin{align}\label{eq:optimization_restricted_general}
    \argmin_{ \meancourse\in \reals^\numcourse}
    \min_{\substack{
        \biasmtx\in \reals^{\numcourse\times \numstudent}\\
        \biasmtx \text{ satisfies } \setpartialorder}} 
    \norm*{\obsmtx - \meancourse\vecone^T - \biasmtx}_\set^2 + \reg \norm*{\biasmtx}_\set^2,
\end{align}
with the ties broken by minimizing $\normfro{\biasmtx}^2$.

We use the shorthand notation $(\estmean, \estbiasmtx)$ to denote the solution $(\estmeanat{\reg}, \estbiasat{\reg})$ to~\eqref{eq:optimization_restricted_general} when the value $\reg$ is clear from the context. Likewise we use the shorthand notation $\estbiasmtxvalat{\reg}$ to denote the interpolated bias $\estbiasmtxvalat{\reg}$ obtained in Line~\ref{line:interpolation} of Algorithm~\ref{alg:cv}.

Recall from Line~\ref{line:interpolation_per_total_order} in Algorithm~\ref{alg:cv} that we find the \element $(\idxcoursenn, \idxstudentnn)\in \settrain$ (or two \elements  $(\idxcoursenn_1, \idxstudentnn_1), (\idxcoursenn_2, \idxstudentnn_2)\in\settrain$) that is close to the considered \element $\idxpairparen\in\setval$ in any total ordering $\totalorder$. We call these one or two \elements from $\settrain$ as the ``nearest-neighbor'' of $\idxpairparen$ with respect to $\totalorder$, denoted $\setnn(\idxcourse, \idxstudent;\totalorder)$. Recall from Line~\ref{line:compute_err_end} in Algorithm~\ref{alg:cv} that $\errat{\reg}$ denotes the CV error at $\reg$.

Define the random variable $\setregoutside$ as the set 
\begin{align}\label{eq:def_set_reg_outside}
    \setregoutside \defn\{ \reg\in [0, \infty]:  \normtwo{\estmeanat{\reg}} > \errbound\}.
\end{align}
Under $\truemean=0$, the set $\setregoutside$ consists of the ``bad'' choices of $\reg$ whose estimate $\estmeanat{\reg}$ incurs a large squared $\ell_2$-error.

\paragraph{Taking the limit of $\numstudent\rightarrow \infty$:}
For ease of notation, we define the limit of taking $\numstudent\rightarrow \infty$ as follows. For example, in the statement of Theorem~\ref{thm:consistency}\ref{part:thm_consistency_constant_fraction}, we consider any fixed $\errbound > 0$. Then the notation
\begin{align}\label{eq:def_limit_n}
    \limn \Probbig{\normtwo{\estmeanat{0} - \truemean} < \errbound} = 1
\end{align}
is considered equivalent to the original statement of Theorem~\ref{thm:consistency}\ref{part:thm_consistency_constant_fraction} that for any $\probbound>0$, there exists an integer $\numstudentlb$, such that for every $\numstudent\ge \numstudentlb$ and every partial ordering satisfying the condition~\ref{part:thm_consistency_constant_fraction} we have
\begin{align*}
    \Probbig{\normtwo{\estmeanat{0} - \truemean} < \errbound} = 1.
\end{align*}
The notation~\eqref{eq:def_limit_n} has the alternative interpretation as follows. We construct a sequence of partial orderings $\{\setpartialorder_\numstudent\}_{\numstudent=1}^\infty$, where the partial ordering $\setpartialorder_\numstudent$ is on $\numcourse$ courses and $\numstudent$ students and satisfies the condition~\ref{part:thm_consistency_constant_fraction}. With $\numstudent$ students, the estimator $\estmeanat{0}$ is provided the partial ordering $\setpartialorder_\numstudent$. We consider any such fixed sequence $\{\setpartialorder_\numstudent\}_{\numstudent=1}^\infty$. Then the limit of $\numstudent\rightarrow \infty$ in~\eqref{eq:def_limit_n} is well-defined.

\subsection{Preliminaries}\label{app:preliminaries}

In this section we present preliminary results that are used in the subsequent proofs. 
Some of the preliminary results are defined based on a set of \elements $\set\subseteq [\numcourse]\times [\numstudent]$. We define the \elements in each course $\idxcoursescope$ as
\begin{align*}
    \setcourse{\idxcourse} \defn \{\idxpairparen\in \set\}.
\end{align*}
Again we say $\idxstudent\in \setcourse{\idxcourse}$ if $\idxpairparen\in \setcourse{\idxcourse}$. We define the number of \elements in each course $\idxcoursescope$ as $\numstudent_\idxcourse \defn \abs*{\set_\idxcourse}$.

Throughout the proofs, whenever a set $\set\subseteq [\numcourse]\times [\numstudent]$ is considered, \emph{we assume the set $\set$ satisfies $\numstudent_\idxcourse > 0$ for each $\idxcoursescope$} to avoid pathological cases. For ease of presentation, the order of the preliminary results does not exactly follow the sequential order that they are proved.

\subsubsection{Properties of the estimator}\label{app:property_of_estimator}

In this section we present a list of properties of our estimator. We start with the following proposition. This proposition shows the existence and uniqueness of the solution to our estimator~\eqref{eq:optimization_restricted_general} under its tie-breaking rule for any $\reg\in [0, \infty)$. That is, the estimator is well-defined on $\reg\in [0, \infty)$.

\begin{proposition}[Existence of the estimator at $\reg\in [0, \infty)$]\label{prop:uniqueness}
For any $\reg\in [0, \infty)$ and any $\set\subseteq[\numcourse]\times [\numstudent]$, there exists a unique solution to our estimator~\eqref{eq:optimization} under the tie-breaking rule, given any inputs $\obsmtx\in \reals^{\numcourse\times \numstudent}$ and any partial ordering $\setpartialorder$.
\end{proposition}
The proof of this result is provided in Appendix~\ref{app:proof_prop_uniqueness}.
Recall that the solution to~\eqref{eq:optimization_restricted_general} at $\reg=\infty$ is defined by taking the limit of $\reg \rightarrow \infty$ as:
\begin{subequations}\label{eq:limit_infty}
\begin{align}
    & \estmeanat{\infty} \defn \lim_{\reg\rightarrow \infty} \estmeanat{\reg}\\
    & \estbiasmtxat{\infty} \defn \lim_{\reg\rightarrow \infty} \estbiasmtxat{\reg}. 
\end{align}
\end{subequations}
The following proposition shows the existence of the solution~\eqref{eq:limit_infty}. That is, the limit in~\eqref{eq:limit_infty} is well-defined. This proposition is a generalization of Proposition~\ref{prop:property_existence_at_infty} to any set $\set\subseteq[\numcourse]\times [\numstudent]$, and its proof is a straightforward generalization of the proof of Proposition~\ref{prop:property_existence_at_infty} (Appendix~\ref{app:proof_prop_property_existence_at_infty}).

\begin{proposition}[Existence of the estimator at $\reg= \infty$]\label{prop:property_existence_at_infty_general}
    For any $\set\subseteq [\numcourse]\times[\numstudent]$, the solution $(\estmeanat{\infty}, \estbiasmtxat{\infty})$ defined in~\eqref{eq:limit_infty} exists. Moreover, we have
    \begin{align*}
        & [\estmeanat{\infty}]_\idxcourse = \frac{1}{\numstudent_\idxcourse} \sum_{\idxstudent\in \setcourse{\idxcourse}} \obs_{\idxcourse\idxstudent} \qquad \forall \idxcourse\in [\numcourse]\\
        & \estbiasmtxat{\infty} = 0.
    \end{align*}
\end{proposition}

The following lemma gives a relation between $\estmeanat{\reg}$ and $\estbiasmtxat{\reg}$ for any $\reg\in [0, \infty]$. This basic relation is used in proving multiple properties of the estimator to be presented subsequently in this section.
\begin{lemma}\label{lem:property_xi}
    For any $\reg \in [0, \infty]$, and any $\set\subseteq[\numcourse]\times [\numstudent]$, the solution $(\estmeanat{\reg}, \estbiasmtxat{\reg})$ to the estimator~\eqref{eq:optimization_restricted_general} satisfies
    \begin{align}\label{eq:property_xi_general_omega}
        \estmeanat{\reg}_\idxcourse = \frac{1}{\numstudent_\idxcourse}\sum_{\idxstudent\in \setcourse{\idxcourse}} \left(\obs_{\idxpair} - \estbiasat{\reg}_{\idxcourse\idxstudent}\right)\qquad \forall \idxcoursescope. 
    \end{align}
    In particular, in the special case of $\set = [\numcourse]\times [\numstudent]$, we have
    \begin{align}\label{eq:property_xi}
        \estmeanat{\reg}_\idxcourse = \frac{1}{\numstudent}\sum_{\idxstudentscope} \left(\obs_{\idxcourse\idxstudent} - \estbiasat{\reg}_{\idxcourse\idxstudent}\right) \qquad \forall \idxcoursescope.
    \end{align}
\end{lemma}
The proof of this result is provided in Appendix~\ref{app:proof_lem_property_xi}
The following property gives expressions of the sum of the \elements in $\estmean$ and the sum of the \elements in $\estbiasmtx$.

\begin{lemma}\label{lem:estimator_sum_equality}
    For any $\reg\in [0, \infty]$, any $\setscope$, the solution $(\estmeanat{\reg}, \estbiasmtxat{\reg})$ given any partial ordering $\setpartialorder$ and any observations $\obsmtx$ satisfies
    \begin{subequations}\label{eq:estimator_sum_equality_general}
        \begin{align}
        \sum_{(\idxcourse, \idxstudent)\in \set} \estbiasat{\reg}_{\idxcourse\idxstudent} & = 0\label{eq:estimator_sum_equality_bias_variable_size}\\
        \sum_{\idxcourse\in [\numcourse]} \numstudent_\idxcourse \estmeanat{\reg}_\idxcourse & =  \sum_{(\idxcourse, \idxstudent)\in \set} \obs_{\idxcourse\idxstudent}.\label{eq:estimator_sum_equality_mean_variable_size}
    \end{align}
    \end{subequations}
    In particular, in the special case of $\set= [\numcourse]\times [\numstudent]$, we have
    \begin{subequations}\label{eq:estimator_sum_equality}
        \begin{align}
        \sum_{\idxcourse\in [\numcourse], \idxstudent\in [\numcourse]}\estbiasat{\reg}_{\idxcourse\idxstudent} & = 0\label{eq:estimator_sum_equality_bias}\\
        \numstudent\sum_{\idxcoursescope}\estmeanat{\reg}_\idxcourse & = \sum_{\idxcoursescope, \idxstudentscope} \obs_{\idxcourse\idxstudent}.\label{eq:estimator_sum_equality_mean}
    \end{align}\end{subequations}
\end{lemma}
The proof of this result is provided in Appendix~\ref{app:proof_lem_estimator_sum_equality}.
The following property shows a shift-invariant property of our estimator. This property is used so that we assume $\truemean=0$ without loss of generality all the proofs.
\begin{proposition}[Shift-invariance of the estimator]\label{prop:property_shift_invariance}
Consider any $\setscope$, and any partial ordering $\setpartialorder$. Fix any $\reg\in [0, \infty]$. Let $(\estmeanat{\reg}, \estbiasmtxat{\reg})$ be the solution of our estimator for any observations $\obsmtx\in \reals^{\numcourse\times \numstudent}$ given $(\setpartialorder, \reg, \set)$. Consider any $\diffmeancourse\in \reals^\numcourse$. Then the solution of our estimator for the observations $\obsmtx + \diffmeancourse\vecone^T$ given  $(\setpartialorder, \reg, \set)$ is $(\estmeanat{\reg}+\diffmeancourse, \estbiasmtxat{\reg})$.
\end{proposition}
The proof of this result is provided in Appendix~\ref{app:proof_prop_property_shift_invariance}. Note that the observation model~\eqref{eq:model} is shift-invariant by definition. That is, consider any fixed $\biasmtx, \noisemtx\in \reals^{\numcourse\times\numstudent}$, denote the observations with $\truemean=0$ as $\obsmtx$. Then the observations with $\truemean = \diffmeancourse$ is $(\obsmtx + \diffmeancourse\vecone^T$). Hence, Proposition~\ref{prop:property_shift_invariance} implies the following corollary.
\begin{corollary}\label{cor:property_shift_invariance}
    Under the observation model~\eqref{eq:model}, consider any fixed bias $\biasmtx\in \reals^{\numcourse\times \numstudent}$ and noise $\noisemtx\in \reals^{\numcourse\times \numstudent}$. Suppose the solution of our estimator under $\truemean=0$ is $(\estmeanat{\reg}, \estbiasmtxat{\reg})$ given any $(\setpartialorder, \reg, \set)$. Then the solution under $\truemean = \diffmeancourse$ is $(\estmeanat{\reg} + \diffmeancourse, \estbiasmtxat{\reg})$.
\end{corollary}

Based on the result of Corollary~\ref{cor:property_shift_invariance}, it can be further verified that the cross-validation algorithm (Algorithm~\ref{alg:cv}) that uses our estimator is shift-invariant. Therefore, \emph{for all the proofs, we assume $\truemean=0$ without loss of generality}.

The following pair of lemmas (Lemma~\ref{lem:est_mean_err_implies_diff} and Lemma~\ref{lem:consistency_bound_pairwise_diff_to_err}) converts  between a bound on the difference of a pair of courses $\abs*{\estmean_\idxcourse - \estmean_{\idxcoursealt}}$ and a bound on $\norm*{\estmean}_2$. Lemma~\ref{lem:est_mean_err_implies_diff} is used in Theorem~\ref{thm:cv_bias_only} and Theorem~\ref{thm:cv_noise_only}; Lemma~\ref{lem:consistency_bound_pairwise_diff_to_err} is used in Theorem~\ref{thm:consistency}. Recall the notation $\setregoutside \defn\{ \reg\in [0, \infty]:  \normtwo{\estmeanat{\reg}} > \errbound\}$.

\begin{lemma}\label{lem:est_mean_err_implies_diff}
Suppose $\truemean=0$. Consider random $\settrain$ obtained by Algorithm~\ref{alg:cv}. Suppose the observations are generate from either:
\begin{enumerate}[label={(\alph*)}]
    \item \label{part:est_mean_err_implies_diff_only_bias}
    The bias is marginally distributed as $\normal(0, \gaussianwidthbias^2)$ following assumption~\ref{assumption:bias} and there is no noise, or
    \item \label{part:est_mean_err_implies_diff_only_noise}
    The noise is generated from $\normal(0, \gaussianwidthnoise^2)$ following assumption~\ref{assumption:noise}, and there is no bias.
\end{enumerate}
For any constant $\errbound > 0$, our estimator $\estmeanat{\reg}$ restricted to $\settrain$ satisfies
\begin{align*}
    \limn \Prob\left(\max_{\idxcourse, \idxcoursealt\in [\numcourse]}\left(\estmeanat{\reg}_\idxcourse - \estmeanat{\reg}_\idxcoursealt\right) > \frac{\errbound}{\sqrt{\numcourse}}, \quad\forall \reg\in \setregoutside\right) = 1,
\end{align*}
where the probability is taken over the randomness in the observations $\obsmtx$ and the training set $\settrain$.
\end{lemma}
The proof of this result is provided in Appendix~\ref{app:proof_lem_est_mean_err_implies_diff}.
%
\begin{lemma}\label{lem:consistency_bound_pairwise_diff_to_err}
    Suppose $\truemean=0$.
    Suppose the observations follow part~\ref{part:est_mean_err_implies_diff_only_bias} of Lemma~\ref{lem:est_mean_err_implies_diff}. Suppose the estimator is restricted to the set of either
    \begin{enumerate}[label={(\alph*)}]
        \item \label{part:diff_to_err_full}
        $\set = [\numcourse]\times [\numstudent]$, or
        \item \label{part:diff_to_err_train}
        random $\settrain$ obtained by Algorithm~\ref{alg:cv}.
    \end{enumerate}
    Fix any $\reg\in [0, \infty]$ and any $\errbound>0$. Suppose we have
    \begin{align}\label{eq:bound_pairwise_diff_whp}
        \lim_{\numstudent\rightarrow \infty}\Prob\left(\max_{\idxcourse, \idxcoursealt\in [\numstudent]}\abs*{\estmeanat{\reg}_\idxcourse - \estmeanat{\reg}_{\idxcoursealt}} < \errbound\right) = 1.
    \end{align}
    Then we have
    \begin{align*}
        \limn \Prob\left(\normtwo{\estmeanat{\reg}}< \errbound\right) = 1,
    \end{align*}
    where the probabilities are taken over the randomness in the observations $\obsmtx$ and (for part~\ref{part:diff_to_err_train}) in $\settrain$.
\end{lemma}
The proof of this result is provided in Appendix~\ref{app:proof_lem_consistency_bound_pairwise_diff_to_err}.
The following proposition gives a closed-form solution under $\numcourse=2$ courses and $\numgroup=2$ groups at $\reg=0$. This proposition is used for proving Theorem~\ref{thm:consistency}\ref{part:thm_consistency_binary} and Proposition~\ref{prop:uniform_example}. Recall the definitions of $\meanobs, \meanobscourse{\idxcourse}$, $\obscoursegroupmin{\idxcourse}{\idxgroup}$ and $\obscoursegroupmax{\idxcourse}{\idxgroup
}$ from~\eqref{eq:def_mean_obs},~\eqref{eq:def_mean_obs_course} and~\eqref{eq:def_obs_course_group_extremal}.
\begin{proposition}\label{prop:closed_form_solution}
Consider $\numcourse=2$ courses and any group ordering $\setpartialorder$ with $\numgroup=2$ groups. Let $\set=[\numcourse]\times [\numstudent]$. Suppose the bias $\biasmtx$ satisfies the partial ordering $\setpartialorder$, and there is no noise. Then the solution of our estimator~\eqref{eq:optimization} at $\reg=0$ has the closed-form expression $\estmeanat{0} = \meanobs + \begin{bmatrix}-1\\1\end{bmatrix}\cdot  \frac{\margin}{2}$,
    where \begin{align}\label{eq:closed_form_solution}
        \margin = \begin{cases}
            \obscoursegroupmin{2}{2} - \obscoursegroupmax{1}{1} & \text{if } \obscoursegroupmin{2}{2} - \obscoursegroupmax{1}{1} < \meanobs_2 - \meanobs_1\\
            \obscoursegroupmax{2}{1} - \obscoursegroupmin{1}{2} & \text{if } \obscoursegroupmax{2}{1} - \obscoursegroupmin{1}{2} > \meanobs_2 - \meanobs_1\\
            \meanobs_2 - \meanobs_1 & \text{o.w.}
        \end{cases}
        \end{align}
        If some of $\{\obscoursegroupmax{1}{1}, \obscoursegroupmax{2}{1}, \obscoursegroupmin{1}{2}, \obscoursegroupmin{2}{2}\}$ do not exist (i.e., when a certain course doesn't have students of a certain group), then the corresponding case in~\eqref{eq:closed_form_solution} is ignored.
\end{proposition}
The proof of this result is provided in Appendix~\ref{app:proof_prop_closed_form_solution}

\subsubsection{Order statistics}
This section presents a few standard properties of order statistics.

Consider $n$ i.i.d. random variables $\{\vargaussian_i\}_{i\in [n]}$ ordered as \begin{align*}
    \vargaussianorder{1} \le \ldots \le \vargaussianorder{n}.
\end{align*}
Define the maximal spacing as
\begin{align}
    \maxspacing_n \defn \max_{1 \le i \le n-1} (\vargaussianorder{i+1} - \vargaussianorder{i}).
\end{align} 
The following standard result from statistics states that the maximum difference between adjacent order statistics converges to $0$ for the Gaussian distribution.
\begin{lemma}\label{lem:maximal_spacing_order_stats}
Let $n > 1$ be any integer.  Let $\vargaussian_1, \ldots, \vargaussian_n$ be i.i.d. $\normal(0, 1)$.
Then for any $\errbound > 0$, we have \begin{align*}
    \limn \Prob(\maxspacing_n < \epsilon) =1.
\end{align*}
\end{lemma}
For completeness, the proof of this result is provided in Appendix~\ref{app:proof_lem_maximal_spacing_order_stats}.
Denote $\cdfinv$ as the inverse c.d.f. of $\normal(0, 1)$. The following standard result from statistics states that the order statistics converges to the inverse c.d.f.

\begin{lemma}
\label{lem:order_stats_consistent}
Let $X_1, \ldots , X_n$ be $\normal(0, 1)$. Fix constant $p \in (0, 1)$ and $\const \in \reals$. Let $\{k_n\}_{n=1}^\infty$ be a sequence such that $\frac{k_n}{n} = p + \frac{c}{\sqrt{n}} + o\left(\frac{1}{\sqrt{n}}\right)$. 
We have
\begin{align*}
    \vargaussianorder{k_n : n} \convprob \cdfinv(p).
\end{align*}
\end{lemma}
For completeness, the proof of this result is provided in Appendix~\ref{app:proof_lem_order_stats_consistent}.

The following standard result from statistics provides a simple bound on the maximum (and the minimum) of a set of i.i.d. Gaussian random variables.
\begin{lemma}\label{lem:maximal_gaussian}
    Let $X_1, \ldots, X_n$ be i.i.d. $\normal(0, \sigma^2)$. Then we have
    \begin{align*}
        & \limn \Prob\left(\max_{i\in [n]} X_i < 2\sigma\sqrt{\log n}\right) =1\\
        & \limn \Prob\left(\max_{i\in [n]} X_i - \min_{i\in [n]} X_i < 4\sigma\sqrt{\log n}\right) =1.
    \end{align*}
\end{lemma}

\subsubsection{Additional preliminaries}
In this section, we present several more additional preliminary results that are used in the subsequent proofs.

The following result considers the number of students under the all constant-fraction assumption given any training-validation split $(\settrain, \setval)$. Recall the definitions of $\blocklengthcoursegroup{\idxcourse}{\idxgroup}, \blocklengthgroup{\idxgroup}, \blocklengthcoursegroupval{\idxcourse}{\idxgroup}, \blocklengthgrouptrain{\idxgroup}$ and $\blocklengthgroupval{\idxgroup}$ from~\eqref{eq:def_blocklength} and~\eqref{eq:def_blocklength_course_group_set}.
\begin{lemma}\label{lem:bound_size}
    Assume $\blocklengthcoursegroup{\idxcourse}{\idxgroup}\ge 4$ for each $\idxcoursescope$ and $\idxgroupscope$. Consider any training-validation split $(\settrain, \setval)$ obtained by Algorithm~\ref{alg:cv}. Then we have the deterministic relations
    \begin{subequations}
    \begin{align}
       & \frac{\blocklengthcoursegroup{\idxcourse}{\idxgroup}}{4} \le \blocklengthcoursegroupval{\idxcourse}{\idxgroup} \le \frac{3\blocklengthcoursegroup{\idxcourse}{\idxgroup}}{4}\qquad \forall \idxcoursescope, \idxgroupscope\label{eq:bound_val_size_per_course}\\
       & \frac{\blocklengthcoursegroup{\idxcourse}{\idxgroup}}{4} \le \blocklengthcoursegrouptrain{\idxcourse}{\idxgroup} \le \frac{3\blocklengthcoursegroup{\idxcourse}{\idxgroup}}{4}\qquad \forall \idxcoursescope, \idxgroupscope\label{eq:bound_train_size_per_course}
    \end{align}
    \end{subequations}
    and
    \begin{subequations}
    \begin{align}
        & \frac{\blocklengthgroup{\idxgroup}}{4} \le \blocklengthgroupval{\idxgroup} \le \frac{3\blocklengthgroup{\idxgroup}}{4}\qquad \forall \idxgroupscope \label{eq:bound_val_size}\\
        & \frac{\blocklengthgroup{\idxgroup}}{4} \le \blocklengthgrouptrain{\idxgroup} \le \frac{3\blocklengthgroup{\idxgroup}}{4}\qquad \forall \idxgroupscope.\label{eq:bound_train_size}
    \end{align}
    \end{subequations}
\end{lemma}
The proof of this result is provided in Appendix~\ref{app:proof_lem_bound_size}.
The following result considers any total ordering. It states that the ranks of the adjacent \elements within $\settrain$, or the ranks of the adjacent \elements between $\settrain$ and $\setval$ differ by at most a constant. Formally, for any $1\le \idxgroup_1 < \idxgroup_2 \le \numcourse\numstudent$, the \element of rank $\idxgroup_1$ and the \element of rank $\idxgroup_2$ are said to be adjacent within $\settrain$, if both \elements are in $\settrain$, and \elements of ranks $\idxgroup_1 + 1$ through $\idxgroup_2 - 1$ are all in $\setval$. The two \elements are said be be adjacent between $\settrain$ and $\setval$, if one of the following is true:
\begin{itemize}
    \item The \elements of ranks $\idxgroup_1$ through $(\idxgroup_2-1)$ are in $\settrain$, and the \element of rank $\idxgroup_2$ is in $\setval$;
    \item The \elements of ranks $\idxgroup_1$ through $(\idxgroup_2-1)$ are in $\setval$, and the \element of rank $\idxgroup_2$ is in $\settrain$.
\end{itemize}
\begin{lemma}\label{lem:rank_difference_adjacent_elements}
    For any partition $(\settrain, \setval)$ obtained by Algorithm~\ref{alg:cv}, for any $1\le \idxgroup_1 < \idxgroup_2 \le \numcourse\numstudent$, suppose that the \element of rank $\idxgroup_1$ and the \element of rank $\idxgroup_2$ are
    \begin{enumerate}[label={(\alph*)}]
        \item \label{part:rank_difference_adjacent_within_train}
        adjacent within $\settrain$, or
        \item \label{part:rank_difference_adjacent_between_train_val}
        adjacent between $\settrain$ and $\setval$.
    \end{enumerate}
    Then we have
    \begin{align*}
        \idxgroup_2 - \idxgroup_1 \le 2\numcourse+1.
    \end{align*}
\end{lemma}
The proof of this result is provided in Appendix~\ref{app:proof_lem_rank_difference_adjacent_elements}.
The following lemma bounds the mean of the bias terms using standard concentration inequalities.
\begin{lemma}\label{lem:bias_bound_train}
    Consider any partial ordering $\setpartialorder$ and any random $\settrain$ obtained by Algorithm~\ref{alg:cv}. Suppose that the bias is marginally distributed as $\normal(0, 1)$ following assumption~\ref{assumption:bias}. For any $\errbound >0$, we have
    \begin{subequations}
    \begin{align}
        & \limn \Prob\left(\abs*{\frac{1}{\numstudenttrain} \sum_{\idxstudent\in \settraincourse{\idxcourse}} \bias_{\idxpair} - \frac{1}{\numstudent}\sum_{\idxstudentscope} \bias_\idxpair} < \errbound\right)= 1\qquad \forall \idxcoursescope,\label{eq:bias_bound_train_per_course}\\
        & \limn \Prob\left(\abs*{\frac{1}{\sizesettrain} \sum_{\idxpairparen\in \settrain} \bias_{\idxpair} } < \errbound\right) =1,\label{eq:bias_bound_train_all}
    \end{align}
    \end{subequations}
    where the probabilities are over the randomness in $\biasmtx$ and in $\settrain$.
\end{lemma}
The proof of this result is provided in Appendix~\ref{app:proof_lem_bias_bound_train}.

\subsection{Proof of Theorem~\ref{thm:consistency}}\label{app:proof_thm_consistency}

The proof follows notation in Appendix~\ref{app:notation} and preliminaries in Appendix~\ref{app:preliminaries}. By Corollary~\ref{cor:property_shift_invariance}, we assume $\truemean=0$ throughout the proof without loss of generality. We also assume without loss of generality that the standard deviation of the Gaussian bias is $\gaussianwidthbias= 1$. Given $\truemean=0$ and the assumption that there is no noise, model~\eqref{eq:model} reduces to
\begin{align}
    \obsmtx = \biasmtx.\label{eq:model_bias_consistency}
\end{align}
Recall that $\blocklength_{\idxcourse\idxgroup}$ denotes the number of observations in course $\idxcourse\in[\numcourse]$ of group $\idxgroup\in [\numgroup]$, and $\blocklength_\idxgroup$ denotes the number of observations of group $\idxgroup$ summed over all courses. For any positive constant $\const > 0$, we define the set $\setpairconstraint_\const$ as
\begin{align}\label{eq:consistency_def_set_pair_constraint}
 \setpairconstraint_\const \defn\; \left\{(\idxcourse, \idxcoursealt)\in [\numcourse]^2: \exists \idxgroup\in [\numgroup] \textrm{ such that } \frac{\blocklength_{\idxcourse\idxgroup}}{\blocklength_\idxgroup}, \frac{\blocklength_{\idxcourse',\idxgroup+1}}{\blocklength_{\idxgroup+1}} \ge \const\right\}.
\end{align}
In words, the definition~\eqref{eq:consistency_def_set_pair_constraint} says that for any pair of courses $(\idxcourse, \idxcoursealt)\in \setpairconstraint_\const$, we have that course $\idxcourse$ takes at least $\const$-fraction of observations in some group $\idxgroupscope$, and course $\idxcoursealt$ takes at least $\const$-fraction of observations in group $(\idxgroup+1)$. 

Before proving the three parts separately, we first state a few lemmas that are used for more than one part. The first lemma states that any $(\idxcourse, \idxcoursealt)\in \setpairconstraint_\const$ imposes a constraint on our estimator $\estmeanat{0}$ at $\reg=0$.

\begin{lemma}\label{lem:constraint_adjacent_group}
    Assume $\truemean=0$. Consider bias marginally distributed as $\normal(0, 1)$ following assumption~\ref{assumption:bias} and no noise. Let $\estmeanat{0}$ be the solution of our estimator at $\reg=0$. Fix any $\const > 0$. For any $(\idxcourse, \idxcoursealt) \in \setpairconstraint_\const$, we have that for any $\errbound>0$,
    \begin{align}\label{eq:consistency_pairwise_diff_by_constraint}
        \lim_{\numstudent\rightarrow \infty} \Probbig{\estmeanat{0}_{\idxcourse'} - \estmeanat{0}_\idxcourse < \errbound} =1.
    \end{align}
\end{lemma}
The proof of this result is provided in Appendix~\ref{app:proof_lem_constraint_adjacent_group}.
To state the next lemma, we first make the following definition of a ``cycle'' of courses.
\begin{definition}\label{def:cycle}
    Let $\lengthcycle \ge 2$ be an integer. We say that $(\idxcourse_1, \idxcourse_2, \ldots, \idxcourse_\lengthcycle)\in [\numcourse]^\lengthcycle$ is a ``cycle'' of courses with respect to $\setpairconstraint_\const$, if
\begin{subequations}\label{eq:consistency_cycle_condition}
    \begin{align}
    & (\idxcourse_\idxnode, \idxcourse_{\idxnode+1})\in \setpairconstraint_\const \qquad  \forall \idxnode\in [\lengthcycle-1],\\
    \textrm{and }\;
    & (\idxcourse_\lengthcycle, \idxcourse_1)\in \setpairconstraint_\const.
    \end{align}
    \end{subequations}
\end{definition}
The following lemma states that if there exists a cycle of courses, then the difference of the estimated quality $\estmean$ between any two courses in this cycle converges to $0$ in probability. 
\begin{lemma}\label{lem:consistency_cycle_to_diff}
    Fix any $\const > 0$. Suppose $\numcourse$ is a fixed constant. Let $(\idxcourse_1, \idxcourse_2, \ldots, \idxcourse_\lengthcycle)\in [\numcourse]^\lengthcycle$ for some $\lengthcycle\ge 2$ be a cycle with respect to $\setpairconstraint_\const$.
    Then for any $\errbound > 0$ we have
    \begin{align*}
    \lim_{\numstudent\rightarrow \infty} \Prob\left(\max_{\idxnode,\idxnodealt\in [\lengthcycle]} \abs*{\estmean_{\idxcourse_\idxnodealt} - \estmean_{\idxcourse_\idxnode}} < \errbound \right) = 1.
\end{align*}
\end{lemma}
The proof of this result is provided in Appendix~\ref{app:proof_lem_consistency_cycle_to_diff}.
Now we prove the three parts of Theorem~\ref{thm:consistency} respectively.

\subsubsection{Proof of part~\ref{part:thm_consistency_constant_fraction}}\label{app:proof_consistency_const_frac}
For clarity of notation, we denote the constant in the all constant-fraction assumption as $\constfrac$. Consider any $\idxcourse, \idxcoursealt\in [\numcourse]$ and any $\idxgroup\in [\numgroup-1]$. We have
\begin{align*}
    \frac{\blocklength_{\idxcourse\idxgroup}}{\blocklength_{\idxgroup}} \stackrel{\stepone}{\ge} \frac{\constfrac\numstudent}{\numcourse\numstudent} = \frac{\constfrac}{\numcourse},
\end{align*}
where step~\stepone is true by the all $\const$-fraction assumption from Definition~\ref{def:all_constant_fraction}. Hence, by the definition~\eqref{eq:consistency_def_set_pair_constraint} of $\setpairconstraint_\const$, we have $(\idxcourse, \idxcoursealt)\in \setpairconstraint_{\frac{\constfrac}{\numcourse}}$ for every $\idxcourse,\idxcoursealt\in [\numcourse]$. 
Hence, $(1, 2, \ldots, \numcourse)$ is a cycle with respect to $\setpairconstraint_\frac{\constfrac}{\numcourse}$ according to Definition~\ref{def:cycle}. Applying Lemma~\ref{lem:consistency_cycle_to_diff} followed by Lemma~\ref{lem:consistency_bound_pairwise_diff_to_err}\ref{part:diff_to_err_full} completes the proof.

\subsubsection{Proof of part~\ref{part:thm_consistency_binary}}

Without loss of generality we assume course 1 has more (or equal) students in group 1 than course 2, that is, we assume 
\begin{align}\label{eq:consistency_binary_blocklength_wlog}
    \blocklength_{11}\ge \blocklength_{21}.
\end{align}
Since we assume there are only two courses and two groups, we have 
\begin{align}\label{eq:consistency_binary_blocklength_wlog_two}
    \blocklength_{12} = \numstudent-\blocklength_{11} \le \numstudent- \blocklength_{21} = \blocklength_{22}.
\end{align} 

We fix any constant $\errbound > 0$. We now bound the probability that $\abs*{\estmean_2 - \estmean_1} < \errbound$. Specifically, we separately bound the probability of $\estmean_2 - \estmean_1 < \errbound$, and the probability of $\estmean_2 - \estmean_1 > -\errbound$. Finally, we invoke Lemma~\ref{lem:consistency_bound_pairwise_diff_to_err} to complete the proof. 

\paragraph{Bounding the probability of $\estmean_2 - \estmean_1 < \errbound$:}
By the definition~\eqref{eq:consistency_def_set_pair_constraint} of $\setpairconstraint_\const$, it can be verified that given~\eqref{eq:consistency_binary_blocklength_wlog}
 and~\eqref{eq:consistency_binary_blocklength_wlog_two} we have $(1, 2)\in \setpairconstraint_{0.5}$ (taking $\idxgroup=1$). By Lemma~\ref{lem:constraint_adjacent_group}, we have
\begin{align}\label{eq:consistency_binary_one_side_bound}
    \lim_{\numstudent\rightarrow \infty}\Prob(\estmean_2 - \estmean_1 < \errbound) = 1.
\end{align}

\paragraph{Bounding the probability of $\estmean_2 - \estmean_1 > -\errbound$:}
By the closed-form solution in Proposition~\ref{prop:closed_form_solution}, we have $\estmean_2 - \estmean_1 = \margin$ where $\margin$ is defined in~\eqref{eq:closed_form_solution} as
\begin{align}\label{eq:closed_form_solution_repeat}
    \margin = \begin{cases}
            \obscoursegroupmin{2}{2} - \obscoursegroupmax{1}{1} & \text{if } \obscoursegroupmin{2}{2} - \obscoursegroupmax{1}{1} < \meanobs_2 - \meanobs_1\\
            \obscoursegroupmax{2}{1} - \obscoursegroupmin{1}{2} & \text{if } \obscoursegroupmax{2}{1} - \obscoursegroupmin{1}{2} > \meanobs_2 - \meanobs_1\\
            \meanobs_2 - \meanobs_1 & \text{o.w.}
        \end{cases}
\end{align}
Recall from the model~\eqref{eq:model_bias_consistency} that $\obsmtx=\biasmtx$, and hence we have the deterministic relation $\obscoursegroupmin{2}{2} - \obscoursegroupmax{1}{1} = \biascoursegroupmin{2}{2} - \biascoursegroupmax{1}{1} \ge 0$ due to the assumption~\ref{assumption:bias} under the group ordering, and similarly we have the deterministic relation $\obscoursegroupmax{2}{1} - \obscoursegroupmin{1}{2} \le 0$. Consider the case of $\meanobs_2 - \meanobs_1 \ge 0$. In this case, only the first and the third cases in~\eqref{eq:closed_form_solution_repeat} are possible, and therefore we have $0 \le \margin \le \meanobs_2 - \meanobs_1$. Now consider the case of $\meanobs_2 - \meanobs_1 < 0$. In this case, only the second and the third cases in~\eqref{eq:closed_form_solution_repeat} are possible, and we have $\meanobs_2 - \meanobs_1 \le \margin \le 0$. Combining the two cases, we have the  relation
\begin{align}\label{eq:consistency_binary_induction}
    \estmean_2 - \estmean_1 = \margin > -\errbound \qquad\text{ if } \; \meanobs_2 - \meanobs_1 > -\errbound.
\end{align}
It suffices to bound the probability of $\meanobs_2 - \meanobs_1 > -\errbound$.

In what follows we show that $\limn \Prob(\meanobs_2 - \meanobs_1 > -\errbound) =1$. That is, we fix some small $\probbound >0$ and show that $\Prob(\meanobs_2 - \meanobs_1 > -\errbound) \ge 1-\probbound$ for all sufficiently large $\numcourse$. The intuition is that course 2 has more students in group $2$, which is the group of greater values of the bias. Since according to assumption~\ref{assumption:bias} the bias is assigned within each group uniformly at random, the set of observations in course 2 statistically dominates the set of observations in course 1. Therefore, $\meanobs_2$ should not be less than $\meanobs_1$ by a large amount.

We first condition on any fixed values of bias ranked as $\truebiasrank{1} \le \ldots \le \truebiasrank{2\numstudent}$ (since we assume the number of courses is $\numcourse=2$). Denote the mean of bias of group 1 as $\truemeanbiasgroup{1} = \frac{1}{\blocklength_1} \sum_{\idxrank=1}^{\blocklength_1} \truebiasrank{\idxrank}$ and the mean of bias of group 2 as $\truemeanbiasgroup{2} = \frac{1}{\blocklength_2} \sum_{\idxrank = \blocklength_1 +1}^{2\numstudent} \truebiasrank{\idxrank}$. Denote $\truebiasmaxminusmin\defn \truebiasrank{2\numstudent} - \truebiasrank{1}$ and denote $\biasmaxminusmin\defn \truebiasrank{2\numstudent} - \truebiasrank{1}$. By Hoeffding's inequality without replacement~\cite[Section~6]{hoeffding1963bound} on group 1 of course 1, we have
\begin{align*}
    \Prob\left[\abs*{\sum_{\idxstudent\in \setcoursegroup{1}{1}} \bias_{1\idxstudent} - \blocklength_{11} \truemeanbiasgroup{1} } \ge \truebiasmaxminusmin\sqrt{\blocklength_{11}\log \left(\frac{1}{\probbound}\right)}\;\middle|\; \truebiasmtx\right] 
    \le 2\exp\left(-\frac{2 \cdot \truebiasmaxminusmin^2 \blocklength\log(\frac{1}{\probbound})}{\blocklength\biasmaxminusmin^2}\right)= 2\probbound^2 \stackrel{\stepone}{\le }\frac{\probbound}{8},
\end{align*}
where~\stepone holds for any $\probbound\in (0, \frac{1}{16})$.
We apply Hoeffding's inequaltiy without replacement for any $\idxcourse\in \{1, 2\}$ and any $\idxgroup\in \{1, 2\}$. Using the fact that $\blocklength_{\idxcourse\idxgroup} \le \numstudent$ for any $\idxcourse\in \{1, 2\}$ and any $\idxgroup\in \{1, 2\}$, we have
\begin{align}\label{eq:consistency_binary_hoeffding}
    \Prob\left[\abs*{\sum_{\idxstudent\in \setcoursegroup{\idxcourse}{\idxgroup}} \bias_{\idxcourse\idxstudent} - \blocklength_{\idxcourse\idxgroup} \truemeanbiasgroup{\idxgroup} } \ge \truebiasmaxminusmin\sqrt{\numstudent\log \left(\frac{1}{\probbound}\right)}\;\middle|\; \truebiasmtx\right] \le \frac{\probbound}{8}.
\end{align}
Taking a union bound of~\eqref{eq:consistency_binary_hoeffding} over $\idxcourse\in \{1, 2\}$ and $\idxgroup\in \{1, 2\}$, we have that with probability at least $1-\frac{\probbound}{2}$,
\begin{align}
    \meanobs_2 - \meanobs_1 & = \frac{1}{\numstudent}\left(\sum_{\idxstudent\in \setcoursegroup{2}{1}}\bias_{2\idxstudent} + \sum_{\idxstudent\in \setcoursegroup{2}{2}}\bias_{2\idxstudent} - \sum_{\idxstudent\in \setcoursegroup{1}{1}}\bias_{1\idxstudent} - \sum_{\idxstudent\in \setcoursegroup{1}{2}}\bias_{1\idxstudent} \right)\nonumber\\
    & \stackrel{\stepone}{\ge} \frac{1}{\numstudent}\left( \blocklength_{21} \truemeanbiasgroup{1} +\blocklength_{22}\truemeanbiasgroup{2} - \blocklength_{11} \truemeanbiasgroup{1} - \blocklength_{12}\truemeanbiasgroup{2} -4 \truebiasmaxminusmin \sqrt{\numstudent\log\left(\frac{1}{\probbound}\right)}\right)\nonumber\\
    & =  \frac{1}{\numstudent}\left( (\blocklength_{21}-\blocklength_{11}) \truemeanbiasgroup{1} +(\blocklength_{22}-\blocklength_{12}) \truemeanbiasgroup{2}  -4 \truebiasmaxminusmin \sqrt{\numstudent\log\left(\frac{1}{\probbound}\right)}\right)\nonumber\\
    & \stackrel{\steptwo}{=} \frac{1}{\numstudent}\left( (\blocklength_{21}-\blocklength_{11}) (\truemeanbiasgroup{1}-\truemeanbiasgroup{2} ) -4 \truebiasmaxminusmin \sqrt{\numstudent\log\left(\frac{1}{\probbound}\right)}\right)\nonumber\\
    & \stackrel{\stepthree}{\ge} -4\truebiasmaxminusmin \sqrt{\frac{\log\left(\frac{1}{\delta}\right)}{\numstudent}},\label{eq:consistency_binary_row_mean_diff_bound}
\end{align}
where inequality~\stepone is true by~\eqref{eq:consistency_binary_hoeffding}, step~\steptwo is true because $\blocklength_{11} + \blocklength_{12} = \blocklength_{21} + \blocklength_{22}$ and hence $\blocklength_{21} - \blocklength_{11} = -(\blocklength_{22} - \blocklength_{12})$, and finally step~\stepthree is true by $\truemeanbiasgroup{1} \le \truemeanbiasgroup{2}$ due to the assumption~\ref{assumption:bias} of the bias and the group orderings.

Now we analyze the term $\biasmaxminusmin$ in~\eqref{eq:consistency_binary_row_mean_diff_bound}. By Lemma~\ref{lem:maximal_gaussian}, there exists integer $\numstudent_0$ such that for any $\numstudent\ge \numstudent_0$,
\begin{align}\label{eq:consistency_binary_maximal_gaussian}
    \Prob\left(
        \biasmaxminusmin \le 4\sqrt{\log2\numstudent}
    \right)  \ge 1- \frac{\probbound}{2}.
\end{align}
Let $\numstudent_1$ be a sufficiently large such that $\numstudent_1 \ge \numstudent_0$ and $16\sqrt{\log2\numstudent_1} \cdot \sqrt{\frac{\log(\frac{1}{\probbound})}{\numstudent_1}} < \errbound$. Then combining~\eqref{eq:consistency_binary_maximal_gaussian} with~\eqref{eq:consistency_binary_row_mean_diff_bound}, we have that for any $\numstudent\ge \numstudentlb$,
\begin{align}
    \Prob\left(\meanobs_2 - \meanobs_1 > -\errbound\right) & =\int_{\biasmtx\in \reals^{2\times \numstudent}} \Prob\left(\meanobs_2 - \meanobs_1 > -\errbound \given \biasmtx\right) \cdot \Prob(\biasmtx)\dd \biasmtx \nonumber\\
    & \ge \int_{\substack{\biasmtx\in \reals^{2\times\numstudent}: \\
    \biasmaxminusmin\le 4\sqrt{\log\numstudent}}} \Prob(\meanobs_2 - \meanobs_1 > -\errbound \given \biasmtx) \cdot \Prob(\biasmtx)\dd \biasmtx \nonumber\\
    & \stackrel{\stepone}{\ge} \left(1-\frac{\probbound}{2}\right) \cdot \Prob(\biasmaxminusmin \le 4\sqrt{\log2\numstudent})\nonumber\\
    & \stackrel{\steptwo}{\ge} \left(1-\frac{\probbound}{2}\right)^2 \ge 1 - \probbound,\label{eq:consistency_binary_mean_diff_bound}
\end{align}
where inequality~\stepone is true by~\eqref{eq:consistency_binary_row_mean_diff_bound} due to the choice of $\numstudent_1$, and inequality~\steptwo is true by~\eqref{eq:consistency_binary_maximal_gaussian}. Combining~\eqref{eq:consistency_binary_mean_diff_bound} with~\eqref{eq:consistency_binary_induction}, for any $\numstudent\ge \numstudent_1$, we have
\begin{align*}
    \Prob(\estmean_2 - \estmean_1 = \margin > -\errbound) \ge \Prob(\meanobs_2 - \meanobs_1 > -\errbound)\ge 1-\probbound.
\end{align*}
That is,
\begin{align}\label{eq:consistency_binary_one_side_bound_two}
    \lim_{\numstudent\rightarrow \infty}\Prob(\estmean_2 - \estmean_1 > -\errbound) = 1.
\end{align}

Finally, combining Step 1 and Step 2, we take a union bound of~\eqref{eq:consistency_binary_one_side_bound} and~\eqref{eq:consistency_binary_one_side_bound_two}, we have
\begin{align}\label{eq:consistency_binary_diff_bound}
    \limn \Probbig{\abs*{\estmean_2 - \estmean_1} < \errbound} = 1.
\end{align}
Given~\eqref{eq:consistency_binary_diff_bound}, we invoke Lemma~\ref{lem:consistency_bound_pairwise_diff_to_err} and obtain
\begin{align*}
    \limn \Probbig{\normtwo{\estmean}< \errbound} = 1,
\end{align*} 
completing the proof.

\subsubsection{Proof of part~\ref{part:thm_consistency_total_order}}\label{app:proof_consistency_total_order}

For total orderings, each observation forms its own group of size $1$ (that is, $\blocklengthgroup{\idxgroup}=1$ for all $\idxgroup\in [\numcourse\numstudent]$). A bias term belonging to group some $\idxgroup\in [\numcourse\numstudent]$ is equivalent to the bias term being rank $\idxgroup$. By the definition~\ref{eq:consistency_def_set_pair_constraint} of $\setpairconstraint_\const$, if course $\idxcourse$ contains rank $\idxgroup$ and course $\idxcoursealt$ contains rank $\idxgroup+1$ then we have $(\idxcourse, \idxcoursealt)\in \setpairconstraint_1$, because $\frac{\blocklength_{\idxcourse\idxgroup}}{\blocklengthgroup{\idxgroup}} = \frac{\blocklength_{\idxcoursealt, \idxgroup+1}}{\blocklengthgroup{\idxgroup+1}} = 1$ due to the total ordering. 

The proof consists of four steps:
\begin{itemize}
    \item In Step 1, we find a partition of the courses, where each subset in this partition consists of courses $\idxcourse$ whose estimated qualities $\estmean_\idxcourse$ are close to each other.
    
    \item In Step 2, we use this partition to analyze $\abs*{\estmean_{\idxcourse} - \estmean_{\idxcoursealt}}$.
    
    \item In Step 3, we upper-bound the probability that $\abs*{\estmean_{\idxcourse} - \estmean_{\idxcoursealt}}$ is large. If $\abs*{\estmean_{\idxcourse} - \estmean_{\idxcoursealt}}$ is large, then we construct an alternative solution according to the partition and derive a contradiction that $\estmean$ cannot be the optimal compared to the alternative solution.
    
    \item  In Step 4, we invoke Lemma~\ref{lem:consistency_bound_pairwise_diff_to_err} to convert the bound on $\abs*{\estmean_{\idxcourse} - \estmean_{\idxcoursealt}}$ to a bound on $\normtwo{\estmean}$.
\end{itemize}  

\paragraph{Step 1: Constructing the partition}
We describe the procedure to construct the partition of courses based on any given total ordering $\setpartialorder$. Without loss of generality, we assume that the minimal rank in course $\idxcourse$ is strictly less than the minimal rank in course $(\idxcourse+1)$ for every $\idxcourse\in [\numcourse-1]$. That is, we have
\begin{align}\label{eq:consistency_total_order_node_indexing_minimal_rank}
\min_{\idxstudentscope} \rank_\idxpair < \min_{\idxstudentscope} \rank_{\idxcourse+1, \idxstudent}\qquad \forall \idxcourse\in [\numcourse-1].
\end{align}

The partition is constructed in steps. We first describe the initialization of the partition. After the partition is initialized, we specify a procedure to ``merge'' subsets in the partition. We continue merging the subsets until there are no more subsets to merge according to a specified condition, and arrive at the final partition.

\paragraph{Initialization} We construct a directed graph of $\numcourse$ nodes, where each node $\idxcourse\in [\numcourse]$ represents course $\idxcourse$. We put a directed edge from node $\idxcourse$ to node $\idxcoursealt$ for every $(\idxcourse, \idxcoursealt)\in \setpairconstraint_1$. Let $\setsupernode_1, \ldots, \setsupernode_\numcourse\subseteq [\numcourse]$ be a partition of the $\numcourse$ nodes. We initialize the partition as $\setsupernode_\idxcourse = \{\idxcourse\}$ for all $\idxcourse\in [\numcourse]$. We also call each subset $\setsupernode_\idxcourse$ as a ``\supernode''.

\paragraph{Merging nodes}
We now merge the partition according to the following procedure. We find a cycle (of directed edges) in the constructed graph, such that the nodes (courses) in this cycle belong to at least two different \supernodes. If there are multiple such cycles, we arbitrarily choose one. We ``merge'' all the \supernodes involved in this cycle. Formally, we denote the \supernodes involved in this cycle as $\setsupernode_{\idxcourse_1}, \setsupernode_{\idxcourse_2}, \ldots, \setsupernode_{\idxcourse_\lengthcycle}$. To merge these \supernodes we construct a new \supernode $\setsupernode = \setsupernode_{\idxcourse_1}\union \setsupernode_{\idxcourse_2}\union \ldots \union \setsupernode_{\idxcourse_\lengthcycle}$. Then we remove the \supernodes $\setsupernode_{\idxcourse_1}, \setsupernode_{\idxcourse_2}, \ldots, \setsupernode_{\idxcourse_\lengthcycle}$ from the partition, and add the merged \supernode $\setsupernode$ to the partition.

We continue merging \supernodes, until there exist no such cycles that involve at least two different \supernodes. When we say we construct a partition we refer to this final partition after all possible merges are completed. 

\begin{figure}
    \centering
    \begin{minipage}[c]{0.5\linewidth}
    \centering
    \subfloat[The total ordering]{\label{float:supergraph_total_order}
        \includegraphics[width=0.99\linewidth]{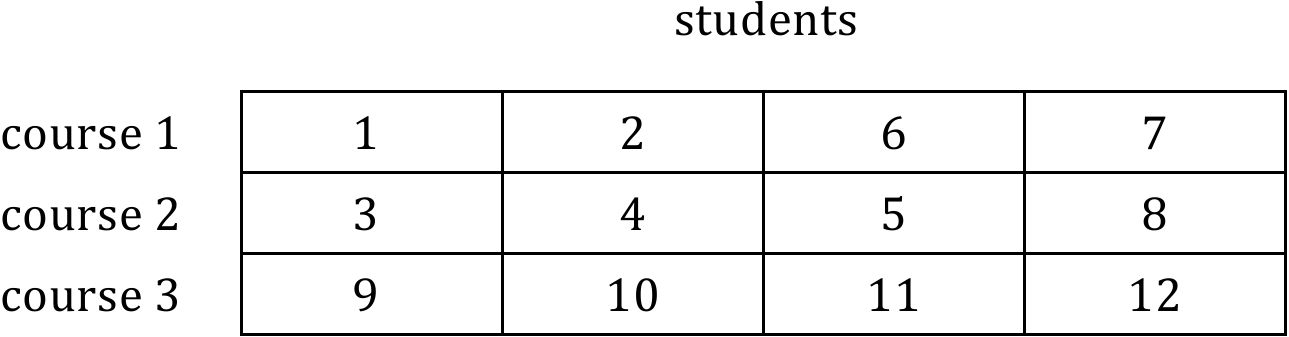}
    }
    \end{minipage}
    \hspace{1.25cm}
    \begin{minipage}[c]{0.3\linewidth}
    \centering
    \subfloat[The procedure of constructing the partition]{\label{float:supergraph_merge}
    \centering
        \includegraphics[width=0.99\linewidth]{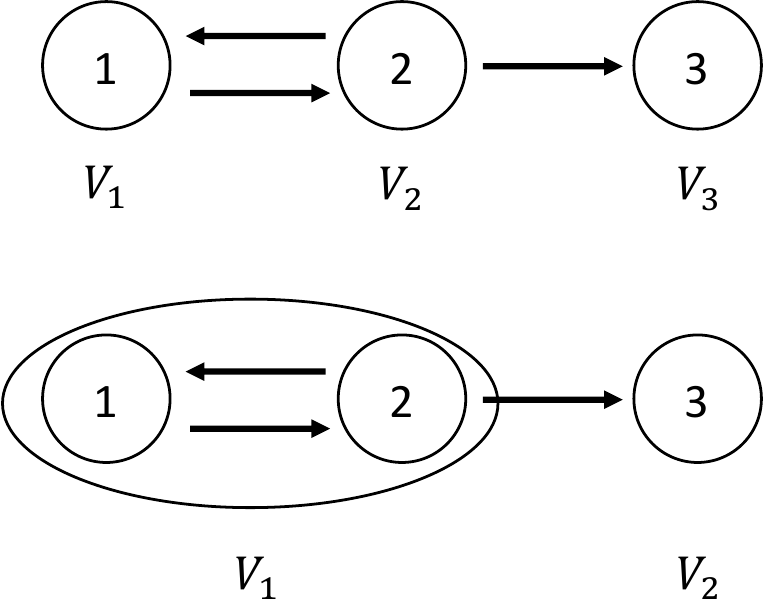}
    }
    \end{minipage}
    \caption{An example for constructing the partition of \supernodes.}
    \label{fig:supergraph}
\end{figure}

An example is provided in Fig.~\ref{fig:supergraph}. In this example we consider $\numcourse=3$ courses and $\numstudent=4$ students per course. We consider the total ordering in Fig.~\ref{fig:supergraph}\subref{float:supergraph_total_order}, where each integer in the table represents the rank of the corresponding \element with respect to this total ordering. The top graph of Fig.~\ref{fig:supergraph}\subref{float:supergraph_merge} shows the constructed graph and the initialized partition. At initialization there is a cycle between course $1$ and course $2$ (that belong to different \supernodes $\setsupernode_1$ and $\setsupernode_2$), so we merge the \supernodes $\setsupernode_1$ and $\setsupernode_2$ as shown in the bottom graph of Fig.~\ref{fig:supergraph}\subref{float:supergraph_merge}. At this point, there are no more cycles that involve more than one \supernode, so the bottom graph is the final constructed partition.

In what follows we state two properties of the partition. We define the length of a cycle as the number of edges in this cycle. The first lemma states that within the same \supernode, any two courses included in a cycle whose length is upper-bounded.
\begin{lemma}\label{lem:consistency_total_order_supergraph_single_connected}
    Consider the partition constructed from any total ordering $\setpartialorder$. Let $\setsupernode$ be any \supernode in this partition. Then for any $\idxcourse, \idxcoursealt\in \setsupernode$ with $\idxcourse\ne \idxcoursealt$, there exists a cycle whose length is at most $2(\numcourse-1)$, such that the cycle includes both course $\idxcourse$ and course $\idxcoursealt$. 
\end{lemma}
The proof of this result is provided in Appendix~\ref{app:proof_lem_consistency_total_order_supergraph_single_connected}.
The following lemma provides further properties on the constructed partition. We say that there exists an edge from \supernode $\setsupernode$ to $\setsupernode'$, if and only if there exists an edge from some node $\idxcourse\in \setsupernode$ to some node $\idxcoursealt\in \setsupernode'$. Denote $\numsupernodes$ as the number of \supernodes in the partition. Denote the \supernodes as  $\setsupernode_1, \ldots, \setsupernode_\numsupernodes$.
\begin{lemma}\label{lem:consistency_total_order_supergraph_line}
    Consider the partition constructed from any total ordering $\setpartialorder$. The \supernodes in this partition can be indexed in a way such that the only edges on the \supernodes are $(\setsupernode_{\idxnode}, \setsupernode_{\idxnode+1})$ for all $\idxnode\in [\numsupernodes-1]$. Under this indexing of \supernodes, the nodes within each \supernodes are consecutive, and increasing in the indexing of the \supernodes. That is, there exist integers $0= \idxcourse_1 < \idxcourse_2 <  \ldots < \idxcourse_{\numsupernodes+1} = \numcourse$, such that $\setsupernode_\idxnode = \{\idxcourse_{\idxnode}+1, \ldots, \idxcourse_{\idxnode+1}\}$ for each $\idxnode\in [\numsupernodes]$. 
    
    Moreover, for each $\idxnode\in [\numsupernodes]$, the ranks of \elements (with respect to the total ordering $\setpartialorder$) contained in the nodes of \supernode $\setsupernode_\idxnode$ are consecutive and increasing in the indexing of the \supernodes. That is, there exists integers $0 =  \rank_1 < \rank_2 \ldots < \rank_{\numsupernodes+1} = \numcourse\numstudent$, such that $\union_{\idxcourse\in \setsupernode_\idxnode}\union_{\idxstudentscope} \{\rank_{\idxpair}\} = \{\rank_\idxnode+1, \ldots, \rank_{\idxnode+1}\}$.
\end{lemma}
The proof of this result is provided in Appendix~\ref{app:proof_lem_consistency_total_order_supergraph_line}.
When we refer to a partition $(\setsupernode_1, \ldots, \setsupernode_\numsupernodes)$, we specifically refer to the indexing of the \supernodes that satisfies Lemma~\ref{lem:consistency_total_order_supergraph_line}. 

As an example, in Fig.~\ref{fig:supergraph} we have $\setsupernode_1 = \{1, 2\}$ and $\setsupernode_2 = \{3\}$. The ranks of \elements in $\setsupernode_1$ are  $\{1, \ldots, 8\}$, and the ranks of \elements in $\setsupernode_2$ are $\{9, \ldots, 12\}$.

\paragraph{Step 2: Analyzing $\abs*{\estmean_\idxcourse - \estmean_\idxcoursealt}$ using the partition} Our goal in Step 2 and Step 3 is to prove the that for any $\errbound > 0$, we have
\begin{align*}
    \lim_{\numstudent\rightarrow \infty} \prob\left(\max_{\idxcourse, \idxcoursealt\in [\numstudent]} \abs*{\estmean_\idxcoursealt - \estmean_\idxcourse} < \errbound\right) = 1.
\end{align*}
Equivalently, denote the ``bad'' event as 
\begin{align}\label{eq:consistency_total_order_def_bad_event}
\eventbad \defn \left\{\max_{\idxcourse, \idxcoursealt\in [\numstudent]} \abs*{\estmean_\idxcoursealt - \estmean_\idxcourse} > 4\numcourse^2\errbound\right\}.
\end{align}
The goal is to prove $
    \lim_{\numstudent\rightarrow \infty} \prob(\eventbad) = 0.$
In Step 2, we define some high-probability event (namely, $\event_1\intersect \event_2\intersect \event_3$ to be presented), and show that it suffices to prove 
\begin{align*}
    \limn \prob(\eventbad, \event_1\intersect \event_2\intersect \event_3) = 0.
\end{align*}
\paragraph{The event $\event_1$ bounds $\abs*{\estmean_\idxcoursealt - \estmean_\idxcourse}$ within each \supernode} We first bound $\abs*{\estmean_\idxcoursealt - \estmean_\idxcourse}$ for $\idxcourse, \idxcoursealt\in [\numcourse]$ within each \supernode. By Lemam~\ref{lem:consistency_total_order_supergraph_single_connected}, there exists a cycle of length at most $2(\numstudent-1)$ between any two courses $\idxcourse, \idxcoursealt$ within the same \supernode. Given assumption~\ref{assumption:d} that $\numstudent$ is a constant, by Lemma~\ref{lem:consistency_cycle_to_diff} we have that for each \supernode $\setsupernode$,
\begin{align}\label{eq:consistency_total_order_bound_within_supernode_individual}
    \lim_{\numstudent\rightarrow \infty} \Prob\left(\max_{\idxcourse, \idxcoursealt\in \setsupernode} \abs*{\estmean_\idxcourse - \estmean_\idxcoursealt} < \errbound \right) = 1.
\end{align}
Since the number of \supernodes is at most $\numcourse$, taking a union bound of~\eqref{eq:consistency_total_order_bound_within_supernode_individual} across all \supernodes in the partition, we have
\begin{align}\label{eq:consistency_total_order_bound_within_supernode}
    \lim_{\numstudent\rightarrow \infty} \Prob
        \left(\vphantom{\max_{\idxcourse, \idxcoursealt\in \setsupernode}}\right.
        \underbrace{
        \max_{\idxcourse, \idxcoursealt\in \setsupernode}
        \abs*{\estmean_\idxcourse - \estmean_\idxcoursealt} < \errbound, \quad \forall \setsupernode \text{ \supernode in the partition}}_{\event_1}
        \left.\vphantom{\max_{\idxcourse, \idxcoursealt\in \setsupernode}}\right)
     = 1.
\end{align}
We denote this event in~\eqref{eq:consistency_total_order_bound_within_supernode} as $\event_1$.

\paragraph{The event $\event_2$ bounds $\abs*{\estmean_\idxcoursealt - \estmean_\idxcourse}$ across \supernodes}
We then bound $\abs*{\estmean_\idxcoursealt - \estmean_\idxcourse}$ across different \supernodes. We consider adjacent \supernodes $\setsupernode_\idxnode$ and $\setsupernode_{\idxnode + 1}$ for any $\idxnode\in [\numsupernodes-1]$. By Lemma~\ref{lem:consistency_total_order_supergraph_line}, there exists an edge from $\setsupernode_\idxnode$ to $\setsupernode_{\idxnode + 1}$. That is, there exists $\idxcourse\in \setsupernode_\idxnode$ and $\idxcoursealt \in \setsupernode_{\idxnode+1}$ such that $(\idxcourse, \idxcoursealt)\in \setpairconstraint_1$. By Lemma~\ref{lem:constraint_adjacent_group}, we have
\begin{align}\label{eq:consistency_total_order_bound_adjacent_supernodes_single}
    \lim_{\numstudent\rightarrow \infty} \Prob\left(
        \estmean_\idxcoursealt - \estmean_\idxcourse < \errbound\right) = 1.
\end{align}
Since the number of \supernodes $\numsupernodes$ is at most $\numcourse$, taking a union bound of~\eqref{eq:consistency_total_order_bound_adjacent_supernodes_single} over all $\idxnode\in [\numsupernodes-1]$, we have
\begin{align}\label{eq:consistency_total_order_bound_adjacent_supernodes}
    \lim_{\numstudent\rightarrow \infty}\Prob
        \left(\vphantom{\min_{\idxcourse\in \setsupernode_\idxnode, \idxcoursealt\in \setsupernode_{\idxnode+1}} }\right.
        \underbrace{
            \min_{\idxcourse\in \setsupernode_\idxnode, \idxcoursealt\in \setsupernode_{\idxnode+1}} 
            \estmean_\idxcoursealt - \estmean_{\idxcourse} < \errbound, \quad \forall \idxnode\in [\numsupernodes-1]
        }_{\event_2}
        \left.\vphantom{\min_{\idxcourse\in \setsupernode_\idxnode, \idxcoursealt\in \setsupernode_{\idxnode+1}} }\right)
        = 1.
\end{align}
We denote this event in~\eqref{eq:consistency_total_order_bound_adjacent_supernodes} as $\event_2$.

\paragraph{Define $\event_3$:} Finally, we define $\event_3$ as the event that $\biasmtx$ is not a constant matrix. That is, 
\begin{align*}
    \event_3= \{\exists\idxcourse, \idxcoursealt\in [\numcourse], \idxstudent, \idxstudentalt\in [\numstudent]: \bias_{\idxpair} \ne \bias_{\idxcoursealt\idxstudentalt}\}.
\end{align*}
Since by assumption~\ref{assumption:bias} (setting $\gaussianwidthbias=1$) the bias terms $\{\bias_{\idxcourse\idxstudent}\}_{\idxcoursescope, \idxstudentscope}$ are marginally distributed as $\normal(0, 1)$, it is straightforward to see that the event $\event_3$ happens almost surely:
\begin{align}\label{eq:consistency_total_order_bound_almost_surely}
    \prob(\event_3) = 1.
\end{align}

\paragraph{Decompose $\eventbad$:} We decompose the bad event $\eventbad$ as
\begin{align}
    \Prob(\eventbad) & = \prob(\eventbad, \event_1\intersect\event_2\intersect\event_3) + \prob(\eventbad, \setcomplement{\event_1 \intersect \event_2\intersect\event_3}) \nonumber\\
    & \le \prob(\eventbad, \event_1\intersect\event_2\intersect \event_3) + \prob(\setcomplement{\event_1 \intersect \event_2\intersect\event_3}).\label{eq:consistency_total_order_event_bad_decompose}
\end{align}
Combining~\eqref{eq:consistency_total_order_bound_within_supernode},~\eqref{eq:consistency_total_order_bound_adjacent_supernodes} and~\eqref{eq:consistency_total_order_bound_almost_surely}, we have
\begin{align}\label{eq:consistency_total_order_event_bad_complement}
    \limn \Probbig{\setcomplement{\event_1 \intersect \event_2\intersect\event_3}} = \limn \prob(\setcomplement{\event_1}\union \setcomplement{\event_2}\union \setcomplement{\event_3}) \le \limn \left[ \prob(\setcomplement{\event_1}) + \prob(\setcomplement{\event_2})+\prob(\setcomplement{\event_3})\right] = 0.
\end{align}
Combining~\eqref{eq:consistency_total_order_event_bad_decompose} and~\eqref{eq:consistency_total_order_event_bad_complement}, in order to show $\lim_{\numstudent\rightarrow \infty} \Prob(\eventbad) = 0$ it suffices to show $\lim_{\numstudent\rightarrow \infty}\Prob(\eventbad, \event_1\intersect\event_2\intersect \event_3) = 0$. 

\paragraph{Step 3: Analyzing the event $\eventbad\intersect \event_1 \intersect \event_2\intersect \event_3$}
In this step, we analyze the event $\eventbad\intersect \event_1 \intersect \event_2\intersect \event_3$, and identify a new partition (namely, $\{\setsupernodeleft, \setsupernoderight\}$ to be defined) of the nodes. This new partition is used to drive a contradiction in Step 4.

First consider the case that the number of \supernodes is $\numsupernodes=1$. In this case $\event_1$ and $\eventbad$ gives a direct contradiction, and we have $\eventbad\intersect \event_1\intersect\event_2\intersect\event_3 = \emptyset$. We now analyze the case when the number of \supernodes is $\numsupernodes\ge 2$. We arbitrarily find one course from each \supernode and denote them as $\idxcourse_1\in \setsupernode_1, \ldots, \idxcourse_\numsupernodes\in \setsupernode_\numsupernodes$.

We condition on $\eventbad\intersect \event_1\intersect\event_2\intersect\event_3$. Recall that by definition~\eqref{eq:consistency_total_order_def_bad_event}, the event $\eventbad$ requires that there exists $\idxcourse, \idxcoursealt\in [\numcourse]$ such that 
\begin{align}
    \abs*{\estmean_\idxcoursealt - \estmean_\idxcourse} > 4\numcourse^2\errbound.\label{eq:consistency_total_order_bad_assumption}
\end{align} 
By the definition~\eqref{eq:consistency_total_order_bound_within_supernode} of $\event_1$, we have that $\idxcourse$ and $\idxcoursealt$ cannot be in the same \supernode. Hence, we assume $\idxcourse\in \setsupernode_\idxnode$ and $\idxcoursealt\in \setsupernode_\idxnodealt$, and assume $\idxnode < \idxnodealt$ without loss of generality. We bound $\estmean_{\idxcoursealt} - \estmean_{\idxcourse}$ as
\begin{align}
    \estmean_{\idxcoursealt} - \estmean_{\idxcourse} & = (\estmean_\idxcoursealt - \estmean_{\idxcourse_\idxnodealt}) + (\estmean_{\idxcourse_\idxnodealt} - \estmean_{\idxcourse_{\idxnodealt-1}}) + \ldots + (\estmean_{\idxcourse_{\idxnode+1}} - \estmean_{\idxcourse_\idxnode}) + (\estmean_{\idxcourse_\idxnode} - \estmean_{\idxcoursealt})\nonumber\\
    & \stackrel{\stepone}{<} 2\errbound + \numcourse\errbound < 4 \numcourse^2 \errbound,\label{eq:consistency_total_order_diff_one_side}
\end{align}
where~\stepone is true by events $\event_1$ and $\event_2$. Combining~\eqref{eq:consistency_total_order_bad_assumption} and~\eqref{eq:consistency_total_order_diff_one_side}, we must have $\estmean_\idxcoursealt - \estmean_{\idxcourse} <- 4\numcourse^2\errbound$, or equivalently 
\begin{align}\label{eq:consistency_total_order_must_have}
    \estmean_{\idxcourse} - \estmean_\idxcoursealt > 4\numcourse^2\errbound.
\end{align}
We decompose $\estmean_{\idxcourse} - \estmean_\idxcoursealt$ as
\begin{align}
    \estmean_{\idxcourse} - \estmean_\idxcoursealt & = (\estmean_\idxcourse - \estmean_{\idxcourse_\idxnode}) + (\estmean_{\idxcourse_\idxnode} - \estmean_{\idxcourse_{\idxnode+1}}) +\ldots + (\estmean_{\idxcourse_{\idxnodealt-1}} - \estmean_{\idxcourse_{\idxnodealt}}) + (\estmean_{\idxcourse_{\idxnodealt}} - \estmean_\idxcoursealt) \nonumber\\
    & \stackrel{\stepone}{<} 2\errbound + (\estmean_{\idxcourse_\idxnode} - \estmean_{\idxcourse_{\idxnode+1}}) +\ldots + (\estmean_{\idxcourse_{\idxnodealt-1}} - \estmean_{\idxcourse_{\idxnodealt}}),\label{eq:consistency_total_order_diff_decompose}
\end{align}
where~\stepone is due to event $\event_1$.
Combining~\eqref{eq:consistency_total_order_must_have} and~\eqref{eq:consistency_total_order_diff_decompose}, we have
\begin{align*}
    2\errbound + (\estmean_{\idxcourse_\idxnode} - \estmean_{\idxcourse_{\idxnode+1}}) +\ldots + (\estmean_{\idxcourse_{\idxnodealt-1}} - \estmean_{\idxcourse_{\idxnodealt}}) & > \estmean_\idxcourse-\estmean_{\idxcoursealt} > 4\numcourse^2\errbound \nonumber\\
    (\estmean_{\idxcourse_\idxnode} - \estmean_{\idxcourse_{\idxnode+1}}) +\ldots + (\estmean_{\idxcourse_{\idxnodealt-1}} - \estmean_{\idxcourse_{\idxnodealt}}) & > (4\numcourse^2 -2)\errbound > 3\numcourse^2 \errbound.
\end{align*}
Hence, we have
\begin{align}
    \numcourse\cdot \max\{ (\estmean_{\idxcourse_\idxnode} - \estmean_{\idxcourse_{\idxnode+1}}) , \ldots,  (\estmean_{\idxcourse_{\idxnodealt-1}} - \estmean_{\idxcourse_{\idxnodealt}})\} > 3\numcourse^2 \errbound \nonumber\\
    \max\{ (\estmean_{\idxcourse_\idxnode} - \estmean_{\idxcourse_{\idxnode+1}}) , \ldots,  (\estmean_{\idxcourse_{\idxnodealt-1}} - \estmean_{\idxcourse_{\idxnodealt}})\} > 3\numcourse \errbound.\label{eq:consistency_total_order_adjacent_diff_lower_bound}
\end{align}
Without loss of generality, we assume that in~\eqref{eq:consistency_total_order_adjacent_diff_lower_bound} we have integer $\idxnodesplit$ with $\idxnode \le \idxnodesplit< \idxnodealt$ such that 
\begin{align}\label{eq:consistency_total_order_largest_diff}
    \estmean_{\idxcourse_{\idxnodesplit}} - \estmean_{\idxcourse_{\idxnodesplit+1}} > 3\numcourse\errbound.
\end{align}
Now consider any $\idxnode, \idxnodealt\in [\numsupernodes]$ such that $\idxnode\le \idxnodesplit < \idxnodealt$, and for any $\idxcourse\in \setsupernode_{\idxnode}$ and $\idxcoursealt\in \setsupernode_\idxnodealt$, we have
\begin{align*}
    \estmean_\idxcourse - \estmean_{\idxcoursealt} & = (\estmean_\idxcourse - \estmean_{\idxcourse_\idxnode}) + (\estmean_{\idxcourse_\idxnode} - \estmean_{\idxcourse_{\idxnode+1}}) +\ldots  +(\estmean_{\idxcourse_\idxnodesplit} - \estmean_{\idxcourse_{\idxnodesplit+1}})  + \ldots + (\estmean_{\idxcourse_{\idxnodealt-1}} - \estmean_{\idxcourse_{\idxnodealt}}) + (\estmean_{\idxcourse_{\idxnodealt}} - \estmean_\idxcoursealt)\nonumber\\
    & \stackrel{\stepone}{>} -2\errbound + 3\numcourse \errbound - \numcourse\errbound> \errbound,
\end{align*}
where~\stepone is by events $\event_1$ and $\event_2$ combined with~\eqref{eq:consistency_total_order_largest_diff}. 
Equivalently, denote $\setsupernodeleft\defn \setsupernode_1 \union \ldots \union \setsupernode_{\idxnodesplit}$ and $\setsupernoderight\defn \setsupernode_{\idxnodesplit+1}\union \ldots \union \setsupernode_{\numsupernodes}$, we have \begin{align}\label{eq:super_node_diff_assumption}
    \estmean_\idxcourse - \estmean_{\idxcoursealt} > \errbound\qquad \forall\idxcourse\in \setsupernodeleft,\; \idxcoursealt \in \setsupernoderight.
\end{align}

\paragraph{Step 4: Showing $\Prob(\eventbad, \event_1\intersect\event_2\intersect\event_3) = 0$ by deriving a contradiction}
We consider any solution $(\estmean, \estbiasmtx)$ of our estimator at $\reg=0$ conditional on $\eventbad\intersect\event_1\intersect\event_2\intersect\event_3$, and derive a contradiction. Hence, we have $\Prob(\eventbad, \event_1 \intersect \event_2\intersect \event_3) = 0$.

\paragraph{Analyzing properties of $\estbiasmtx$}
By Lemma~\ref{lem:consistency_total_order_supergraph_line}, any bias term $\estbias_\idxpair$ for $\idxcourse\in \setsupernodeleft$ has a smaller rank than any bias term $\estbias_\idxpair$ for $\idxcourse\in \setsupernoderight$. 
Therefore, the mean of $\estbiasmtx$ over \elements in  $\setsupernodeleft$ is less than or equal to the mean of $\estbiasmtx$ over $\setsupernoderight$. That is, with the definition of $\estmeanbiasleft$ and $\estmeanbiasright$ as
\begin{subequations}\label{eq:def_mean_b_low_high}
\begin{align}
    & \estmeanbiasleft \defn \frac{1}{\abs*{\setsupernodeleft}\cdot \numstudent} \sum_{\idxcourse\in \setsupernodeleft}\sum_{\idxstudentscope} \estbias_\idxpair\\
    & \estmeanbiasright \defn \frac{1}{\abs*{\setsupernoderight}\cdot \numstudent} \sum_{\idxcourse\in \setsupernoderight}\sum_{\idxstudentscope} \estbias_\idxpair,
\end{align}
\end{subequations}
We have the deterministic relation $\estmeanbiasleft\le \estmeanbiasright$.

First consider the case of $\estmeanbiasleft = \estmeanbiasright$. Since $\estbiasmtx$ obeys the total ordering $\setpartialorder$, we have $\estbiasmtx=\const$ for some constant $\const$. Conditional on $\event_3$, it can be verified that for any $\const\in \reals$, the objective~\eqref{eq:optimization} attained at $(\estmean, \estbiasmtx)$ is strictly positive. Recall from the model~\eqref{eq:model_bias_consistency} that $\obsmtx=\biasmtx$. Hence, an objective~\eqref{eq:optimization} of $0$ can be attained by the solution $(0, \biasmtx)$. Contradiction to the assumption that $(\estmean, \estbiasmtx)$ is the minimizer of the objective.

Now we consider the case of $\estmeanbiasleft < \estmeanbiasright$. We have that either $\estmeanbiasleft < 0$ or $\estmeanbiasright > 0$ (or both). Without loss of generality we assume $\estmeanbiasright > 0$.

\paragraph{Constructing an alternative solution}
We now construct an alternative solution by increasing $\estmean_\idxcourse$ for every course $\idxcourse\in \setsupernoderight$ by a tiny amount, and prove for contradiction that this alternative solution is preferred by the tie-breaking rule of minimizing $\normfro{\biasmtx}^2$. We construct the alternative solution $(\estmean', \estbiasmtx')$ as
\begin{align}\label{eq:consistency_total_order_alternative_solution}
    \begin{split}
    \estmean'_\idxcourse & = 
    \begin{cases}
        \estmean_\idxcourse & \text{if } \idxcourse\in \setsupernodeleft\\
        \estmean_\idxcourse + \solutionincrease & \text{if }\idxcourse\in \setsupernoderight
    \end{cases}\\
    \estbiasmtx' & = \obsmtx-\estmean'\vecone^T,
    \end{split}
\end{align}
for some sufficiently small $\solutionincrease > 0$ whose value is specified later. Since $(\estmean, \estbiasmtx)$ is a solution, as discussed previously it has to attain an objective of $0$. By the construction~\eqref{eq:consistency_total_order_alternative_solution}, it can be verified that $(\estmean', \estbiasmtx')$ also attains an objective of $0$. In what remains for this step, we first show that the alternative solution $(\estmean', \estbiasmtx')$ satisfies all ordering constraints by the total ordering $\setpartialorder$. Then we show that $\normfro{\estbiasmtx'}^2<\normfro{\estbiasmtx}^2$, and therefore $(\estmean', \estbiasmtx')$ is preferred by the tie-breaking rule over $(\estmean, \estbiasmtx)$, giving a contradiction.

\paragraph{The alternative solution $(\estmean', \estbiasmtx')$ satisfies all ordering constraints in $\setpartialorder$}
Since both $(\estmean, \estbiasmtx)$ and $(\estmean', \estbiasmtx')$ attain an objective of $0$, we have the deterministic relation
\begin{align}\label{eq:consistency_total_order_alternative_solution_equality}
    \obs_\idxpair = \estmean_\idxcourse + \estbias_{\idxpair} = \estmean'_\idxcourse + \estbias'_{\idxpair}\qquad \forall \idxcourse\in [\numcourse], \idxstudent\in [\numstudent].
\end{align}
Consider any constraint $((\idxcourse, \idxstudent),(\idxcoursealt, \idxstudentalt))\in \setpartialorder$. If $\idxcourse, \idxcoursealt\in \setsupernodeleft$, then we have\begin{align*}
    \estbias'_{\idxcourse\idxstudent} - \estbias'_{\idxcoursealt\idxstudentalt} & = \obs_{\idxcourse\idxstudent} - \estmean'_\idxcourse - (\obs_{\idxcoursealt\idxstudentalt} - \estmean'_{\idxcoursealt})\\
    & =\obs_{\idxcourse\idxstudent} - \estmean_\idxcourse - (\obs_{\idxcoursealt\idxstudentalt} - \estmean_{\idxcoursealt})\\
    & =  \estbias_{\idxcourse\idxstudent} - \estbias_{\idxcoursealt\idxstudentalt} \stackrel{\stepone}{<} 0,
\end{align*}
where~\stepone is true because by assumption $(\estmean, \estbiasmtx)$ is the optimal solution, and hence $\estbiasmtx$ satisfies the ordering constraint of $\estbias_\idxpair \le \estbias_{\idxcoursealt\idxstudentalt}$. Similarly if $\idxcourse,\idxcoursealt\in \setsupernoderight$, then $(\estmean', \estbiasmtx')$ also satisfies this ordering constraint. Finally, consider the case where one of $\{\idxcourse, \idxcoursealt\}$ is in $\setsupernodeleft$ and the other is in $\setsupernoderight$. Due to Lemma~\ref{lem:consistency_total_order_supergraph_line} regarding the ranks combined with the definition of $(\setsupernodeleft, \setsupernoderight)$, it can only be the case that $\idxcourse\in \setsupernodeleft$ and $\idxcoursealt\in \setsupernoderight$. For any $\solutionincrease\in (0, \errbound)$, we have that conditional on $\eventbad\intersect \event_1 \intersect\event_2\intersect\event_3$,
\begin{align*}
    \estbias'_{\idxcourse\idxstudent} - \estbias'_{\idxcoursealt\idxstudentalt} & = (\obs_\idxpair - \estmean'_\idxcourse) - (\obs_{\idxcoursealt\idxstudentalt} - \estmean'_\idxcoursealt) \\
    & = (\bias_{\idxpair} - \estmean_{\idxcourse}) - (\bias_{\idxcoursealt\idxstudentalt} - \estmean_\idxcoursealt-\solutionincrease) \\
    & = (\bias_{\idxpair} - \bias_{\idxcoursealt\idxstudentalt}) + (\estmean_{\idxcoursealt} + \solutionincrease -\estmean_\idxcourse) \stackrel{\stepone}{<} 0,
\end{align*}
where~\stepone is true because the ordering constraint $(\idxpairparen, (\idxcoursealt, \idxstudentalt))$ gives  $\bias_{\idxpair} \le \bias_{\idxcoursealt\idxstudentalt}$. Moreover, we have $\estmean_\idxcoursealt - \estmean_\idxcourse  < -\errbound$ due to~\eqref{eq:super_node_diff_assumption}.
Hence, all ordering constraints are satisfied by the alternative solution $(\estmean', \estbiasmtx')$.

\paragraph{The alternative solution $(\estmean', \estbiasmtx')$ satisfies $\normfro{\estbiasmtx'} < \normfro{\estbiasmtx}$, thus preferred by tie-breaking}
Plugging in the construction~\eqref{eq:consistency_total_order_alternative_solution}, we compute $\normfro{\estbiasmtx'}^2$ as
\begin{align}
    \normfro{\estbiasmtx'}^2 & = \sum_{\idxcourse\in \setsupernodeleft}\sum_{\idxstudentscope} (\obs_{\idxcourse\idxstudent} - \estmean_\idxcourse)^2 +  \sum_{\idxcourse\in \setsupernoderight}\sum_{\idxstudentscope} (\obs_{\idxcourse\idxstudent} - \estmean_\idxcourse-\solutionincrease)^2\nonumber\\
    & \stackrel{\stepone}{=} \sum_{\idxcourse\in \setsupernodeleft}\sum_{\idxstudentscope} (\estbias_{\idxcourse\idxstudent})^2 +  \sum_{\idxcourse\in \setsupernoderight}\sum_{\idxstudentscope} (\estbias_{\idxcourse\idxstudent} - \solutionincrease)^2,\label{eq:consistency_total_order_norm_alt_solution}
\end{align}
where~\stepone is true by~\eqref{eq:consistency_total_order_alternative_solution_equality}.
Taking the partial derivative of~\eqref{eq:consistency_total_order_norm_alt_solution} with respect to $\solutionincrease$, we have
\begin{align}\label{eq:consistency_total_order_partial_deriv}
    \frac{\partial \normfro{\estbiasmtx'}^2}{\partial \solutionincrease} = 2\left( \abs*{\setsupernoderight}\cdot \numstudent\solutionincrease - \sum_{\idxcourse\in \setsupernoderight}\sum_{\idxstudentscope} \estbias_{\idxcourse\idxstudent} \right) = 2\abs*{\setsupernoderight}\cdot \numstudent(\solutionincrease -\estmeanbiasright).
\end{align}
By the assumption of $\estmeanbiasright > 0$, the partial derivative~\eqref{eq:consistency_total_order_partial_deriv} is strictly negative for any $\solutionincrease\in \left[0, \estmeanbiasright\right)$. Contradiction to the fact that $\estbiasmtx$ (corresponding to $\solutionincrease =0$) is the solution with the minimal Frobenius norm $\normfro{\estbiasmtx}^2$. Hence, $(\estmean, \estbiasmtx)$ cannot be a solution, and we have
\begin{align*}
    \prob(\eventbad, \event_1\intersect\event_2\intersect\event_3) = 0.
\end{align*}

\paragraph{Step 4: Invoking Lemma~\ref{lem:consistency_bound_pairwise_diff_to_err}}
Recall from Step 2 that $\limn \prob(\eventbad, \event_1\intersect \event_2\intersect \event_3) = 0$ implies $\lim_{\numstudent\rightarrow \infty}\prob(\eventbad) = 0$. Equivalently, for any $\errbound >0$ we have
\begin{align*}
    \lim_{\numstudent\rightarrow\infty}\Prob\left(\max_{\idxcourse, \idxcoursealt\in [\numcourse]} \abs*{\estmean_\idxcoursealt - \estmean_\idxcourse} < \errbound\right) = 1.
\end{align*}
Invoking Lemma~\ref{lem:consistency_bound_pairwise_diff_to_err} completes the proof.

\subsection{Proof of Proposition~\ref{prop:property_existence_at_infty}}\label{app:proof_prop_property_existence_at_infty}

    We denote $(\estmeanat{\infty}, \biasmtxat{\infty})$ as the values given by expression~\eqref{eq:properties_expression_at_infty}. We prove that
    \begin{align*}
        (\estmeanat{\infty}, \biasmtxat{\infty}) = \lim_{\reg\rightarrow \infty} (\estmeanat{\reg}, \estbiasmtxat{\reg}).
    \end{align*}

    Denote the minimal value of the first term in the objective~\eqref{eq:optimization} as
    \begin{align*}
        \minimalvalfirstterm\defn \min_{\substack{\meancourse\in \reals^\numcourse, \biasmtx\in \reals^{\numcourse\times \numstudent}\\ \biasmtx \text{ satisfies }\setpartialorder}} \norm*{\obsmtx - \meancourse\vecone^T - \biasmtx}_F^2.
    \end{align*}
    Denote $\valfirsttermatinfty$ as the value of the first term attained at $(\estmeanat{\infty}, \estbiasmtxat{\infty})$. By the definition of $\minimalvalfirstterm$ as the minimal value over the domain, we have $\valfirsttermatinfty \ge \minimalvalfirstterm$. We discuss the following two cases depending on the value of $\valfirsttermatinfty$.

   \paragraph{Case of $\valfirsttermatinfty = \minimalvalfirstterm$:} We have that $(\estmeanat{\infty}, \estbiasmtxat{\infty})$ is the solution for any $\reg\in (0, \infty)$, because it attains the minimal value separately for the two terms in the objective~\eqref{eq:optimization}. By Proposition~\ref{prop:uniqueness}, a unique solution exists for any $\reg \in (0, \infty)$. Hence, the limit $\lim_{\reg\rightarrow \infty} (\estmeanat{\reg}, \estbiasmtxat{\reg})$ exists and we have $(\estmeanat{\infty}, \estbiasmtxat{\infty}) = \lim_{\reg\rightarrow \infty} (\estmeanat{\reg}, \estbiasmtxat{\reg})$.
   
   \paragraph{Case of $\valfirsttermatinfty > \minimalvalfirstterm$:} We first show that $\lim_{\reg\rightarrow \infty} \estbiasmtxat{\reg} = 0$. That is, we show that for any $\epsilon > 0$, there exists some $\reg_0 >0$, such that $\normfro{\estbiasmtxat{\reg} }^2 < \epsilon$ for all $\reg\in (\reg_0, \infty)$.
   
   Take $\reg_0 = \frac{\valfirsttermatinfty - \minimalvalfirstterm }{\epsilon}$, and assume for contradiction that there exists some $\reg^* > \reg_0$ such that $\norm{\estbiasmtxat{\reg^*}}_F^2 > \epsilon$. The objective~\eqref{eq:optimization} (setting $\reg=\reg^*$) attained by $(\estmeanat{\reg^*}, \estbiasmtxat{\reg^*})$ is lower-bounded by
   \begin{align*}
       \normtwo{\obsmtx - \estmeanat{\reg^*} - \estbiasmtxat{\reg^*}}^2 + \reg^*\normfro{\estbiasmtxat{\reg^*}}^2 > \minimalvalfirstterm + \reg_0 \errbound
       > \minimalvalfirstterm + (\valfirsttermatinfty-\minimalvalfirstterm) = \valfirsttermatinfty.
   \end{align*}
   On the other hand, the objective attained by $(\estmeanat{\infty}, \estbiasmtxat{\infty})$ is $\valfirsttermatinfty$. Hence, $(\estmeanat{\infty}, \estbiasmtxat{\infty})$ attains a strictly smaller value of the objective than $(\estmeanat{\reg^*}, \estbiasmtxat{\reg^*})$ at $\reg=\reg^*$. Contradiction to the assumption that $(\estmeanat{\reg^*}, \estbiasmtxat{\reg^*})$ is the solution at $\reg=\reg^*$. Hence, we have $\lim_{\reg\rightarrow \infty} \estbiasmtxat{\reg} = 0$.
   
   Combining the fact that $\lim_{\reg\rightarrow \infty} \estbiasmtxat{\reg} = 0$ with the relation~\eqref{eq:property_xi} in Lemma~\ref{lem:property_xi} (at any $\reg\in [0, \infty)$), we have that for each $\idxcourse\in [\numcourse]$,
   \begin{align*}
       \estmeanat{\reg}_\idxcourse = \frac{1}{\numstudent}\sum_{\idxstudentscope} \left(\obs_{\idxcourse\idxstudent} - \estbiasat{\reg}_{\idxcourse\idxstudent}\right) \rightarrow \frac{1}{{\numstudent}}\sum_{\idxstudentscope} \obs_{\idxcourse\idxstudent}\qquad \text{ as }\reg\rightarrow \infty,
   \end{align*}
   completing the proof.

\subsection{Proof of Theorem~\ref{thm:cv_bias_only}}\label{app:proof_thm_cv_bias_only}

The proof follows notation in Appendix~\ref{app:notation} and preliminaries in Appendix~\ref{app:preliminaries}.
By Corollary~\ref{cor:property_shift_invariance}, we assume $\truemean = 0$ without loss of generality. We also assume without loss of generality that the standard deviation of the Gaussian bias distribution is $\gaussianwidthbias = 1$. Given $\truemean=0$ and the assumption that there is no noise, model~\eqref{eq:model} reduces to:
\begin{align}
    \obsmtx= \biasmtx.\label{eq:cv_bias_model}
\end{align}
Both part~\ref{part:cv_bias_only_constant_fraction} and part~\ref{part:cv_bias_only_total} consist of $3$ similar steps. We start with the first step, and proceed separately for the two remaining steps for the two parts.

\noindent\textbf{Step 1: Showing the consistency of our estimator at $\lambda=0$ restricted to the training set $\settrain$.}

In the first step, we show that our estimator is consistent under group orderings satisfying part~\ref{part:cv_bias_only_constant_fraction} and part~\ref{part:cv_bias_only_total}, on any fixed training set $\settrain\subseteq [\numcourse]\times [\numstudent]$ obtained by Algorithm~\ref{alg:cv}. Note that Theorem~\ref{thm:consistency}\ref{part:thm_consistency_constant_fraction} and Theorem~\ref{thm:consistency}\ref{part:thm_consistency_total_order} give the desired consistency result when the data is full observations $\set = [\numcourse]\times [\numstudent]$. It remains to extend the proof of Theorem~\ref{thm:consistency}\ref{part:thm_consistency_constant_fraction} and Theorem~\ref{thm:consistency}\ref{part:thm_consistency_total_order} to any $\settrain$ given by Algorithm~\ref{alg:cv}. The following theorem states that part~\ref{part:thm_consistency_constant_fraction} and part~\ref{part:thm_consistency_total_order} of Theorem~\ref{thm:consistency} still hold for the estimator~\eqref{eq:optimization_restricted_general} restricted to $\settrain$. We use $(\estmeanat{0}, \estbiasmtxat{0})$ to denote the solution to~\eqref{eq:optimization_restricted_general} restricted to $\settrain$ for the remaining of the proof of Theorem~\ref{thm:cv_bias_only}.

\begin{theorem}[Generalization of Theorem~\ref{thm:consistency} to any $\settrain$]
\label{thm:consistency_train}
Consider any fixed $\settrain\subseteq [\numcourse]\times [\numstudent]$ obtained by Algorithm~\ref{alg:cv}. Suppose the partial ordering is one of
\begin{enumerate}[label={(\alph*)}]
    \item \label{part:thm_consistency_train_const_frac}
        any group ordering satisfying the all $\const$-fraction assumption, or
    \item \label{part:thm_consistency_train_total_order}
        any total ordering.
\end{enumerate}
Then for any $\errbound>0$ and $\probbound>0$, there exists an integer $\numstudentlb$ (dependent on $\errbound, \probbound, \const, \numcourse$), such that for every $\numstudent\ge \numstudentlb$ and every partial ordering satisfying one of the conditions~\ref{part:thm_consistency_train_const_frac} or~\ref{part:thm_consistency_train_total_order}, the estimator $\estmeanat{0}$ (as the solution to~\eqref{eq:optimization_restricted_general} restricted to $\settrain$) satisfies
    \begin{align}
        \Prob\Big(\normtwo{\estmeanat{0} - \truemean} < \errbound\Big) \ge 1- \probbound.
    \end{align}
    Equivalently, for any $\errbound > 0$, we have
    \begin{align}
        \lim_{\numstudent\rightarrow \infty} \Prob\Big(\normtwo{\estmeanat{0} - \truemean} < \errbound\Big) =1.\label{eq:cv_only_bias_consistency_mean}
    \end{align}
\end{theorem}
The proof of this theorem is in Appendix~\ref{app:proof_thm_consistency_train}.
Now we consider the consistency of the bias term $\estbiasmtx$. Given the model~\eqref{eq:cv_bias_model}, the objective~\eqref{eq:optimization_restricted_general} at $\reg=0$ equals $0$ at the values of $(\estmean, \estbiasmtx) = (0, \biasmtx)$. Hence, objective~\eqref{eq:optimization_restricted_general} attains a value of $0$ at the solution $(\estmeanat{0}, \estbiasmtxat{0})$. Therefore, we have the deterministic relation $\obsmtx_\settrain = [\estmeanat{0}\vecone^T + \estbiasmtxat{0}]_\settrain$. For any $\idxpairparen\in \settrain$, we have
    \begin{align}\label{eq:cv_only_bias_group_order_bias_expression_at_zero}
        \estbiasat{0}_\idxpair  = \obsmtx_\idxpair - \estmeanat{0}_\idxcourse \stackrel{\stepone}{=} \bias_\idxpair - \estmeanat{0}_\idxcourse,
    \end{align}
    where equality~\stepone is true because of the model~\eqref{eq:cv_bias_model}. Combining~\eqref{eq:cv_only_bias_group_order_bias_expression_at_zero} with~\eqref{eq:cv_only_bias_consistency_mean}, we have that for any $\errbound > 0$,
    \begin{align}\label{eq:cv_only_bias_group_order_bias_converge_in_prob}
        \limn\Probbig{\abs*{\estbiasat{0}_\idxpair - \bias_\idxpair} < \errbound, \quad\forall \idxpairparen\in \settrain} = 1.
    \end{align}
    This completes Step 1 of the proof. The remaining two steps are presented separately for the two parts.

\subsubsection{Proof of part~\ref{part:cv_bias_only_constant_fraction}}
    
    We fix some constant $\errbound_1>0$ whose value is determined later.
    For clarity of notation, we denote the constant in the all constant-fraction assumption as $\constfrac$. 
    
    \noindent\textbf{Step 2: Computing the validation error at $\reg=0$}
    
    We first analyze the interpolated bias $\estbiasmtxvalat{0}$. Recall that $\settraingroup{\idxgroup}$ and $\setvalgroup{\idxgroup}$ denote the set of \elements of group $\idxgroup\in [\numgroup]$ in the training set $\settrain$ and the validation set $\setval$, respectively. By symmetry of the interpolation expression in Line~\ref{line:interpolation} of Algorithm~\ref{alg:cv} and Definition~\ref{def:group_ordering} of the group ordering, it can be verified that the interpolated bias $\estbiasinterp_\idxpair$ is identical for all \elements within any group $\idxgroup\in [\numgroup]$. That is, for each $\idxgroupscope$, we have
    \begin{align}\label{eq:cv_only_bias_const_frac_interp_bias_equal_within_group}
        \estbiasinterp_\idxpair = \estbiasinterp_{\idxcoursealt\idxstudentalt}, \text{ for any } \idxpairparen, (\idxcoursealt, \idxstudentalt)\in \setvalgroup{\idxgroup}.
    \end{align}
    Denote $\estbiasinterp_\idxgroup\defn \estbiasinterp_\idxpair$ for any $\idxpairparen\in \settraingroup{\idxgroup}$. By~\eqref{eq:cv_only_bias_const_frac_interp_bias_equal_within_group}, we have that $\estbiasinterpgroup{\idxgroup}$ is well-defined. Denote the random variables $\biasmeangrouptrain{\idxgroup}$ and $\biasmeangroupval{\idxgroup}$ as the mean of the (random) bias $\biasmtx$ in group $\idxgroupscope$, over $\settraingroup{\idxgroup}$ and $\setvalgroup{\idxgroup}$, respectively. Denote the random variable $\biasmeancoursegroupval{\idxcourse}{\idxgroup}$ as the mean of the (random) $\biasmtx$ of group $\idxgroupscope$ in course $\idxcoursescope$ over $\setval$. That is, we define
    \begin{align}
        \biasmeangrouptrain{\idxgroup} &\defn \frac{1}{\abs*{\settraingroup{\idxgroup}}}\sum_{\idxpairparen\in \settraingroup{\idxgroup}} \bias_{\idxpair}\label{eq:cv_only_bias_const_frac_def_mean_b_k_train}\\
        \biasmeangroupval{\idxgroup} &\defn \frac{1}{\abs*{\setvalgroup{\idxgroup}}}\sum_{\idxpairparen\in \setvalgroup{\idxgroup}} \bias_{\idxpair}\\
        \biasmeancoursegroupval{\idxcourse}{\idxgroup} &\defn \frac{1}{\abs*{\setvalcoursegroup{\idxcourse}{\idxgroup}}}\sum_{\idxstudent\in \setvalcoursegroup{\idxcourse}{\idxgroup}} \bias_{\idxpair}\label{eq:cv_only_bias_const_frac_def_mean_b_ik_val}.
    \end{align}
    Denote $\estbiasmeangrouptrain{\idxgroup}$ likewise as the mean of the estimated bias $\estbiasmtx$ over $\settraingroup{\idxgroup}$.
    Given $\obsmtx=\biasmtx$ from model~\eqref{eq:cv_bias_model}, the validation error at $\reg=0$ is computed as:
    \begin{align}
        \errat{0} & = \frac{1}{\sizesetval} \sum_{\idxpairparen\in \setval} \left(\obs_\idxpair - \estmeanat{0}_\idxcourse - \estbiasinterp_\idxpair\right)^2 \nonumber\\
        & = \frac{1}{\sizesetval}\sum_{\idxcourse\in [\numcourse], \idxgroup\in [\numgroup]} 
        \sum_{\idxstudent\in \setvalcoursegroup{\idxcourse}{\idxgroup}} \left(\bias_\idxpair - \estmeanat{0}_\idxcourse - \estbiasinterp_\idxgroup\right)^2.\label{eq:cv_only_bias_const_frac_val_err_at_zero}
    \end{align}
    We first analyze the term $\estbiasinterp_\idxgroup$ in~\eqref{eq:cv_only_bias_const_frac_val_err_at_zero}. The following lemma shows that the interpolation procedure in Algorithm~\ref{alg:cv} ensures that $\estbiasinterp_\idxgroup$ is close to $\estbiasmeangrouptrain{\idxgroup}$, the mean of the estimated bias over $\settraingroup{\idxgroup}$.
    
    \begin{lemma}\label{lem:cv_only_bias_bound_interp_mean}
        Consider any group ordering $\setpartialorder$ that satisfies the all $\constfrac$-fraction assumption, and any $\settrain\subseteq [\numcourse]\times [\numstudent]$ obtained by Algorithm~\ref{alg:cv}. Then for any $\reg\in [0, \infty]$ we have the deterministic relation:
        \begin{align*}
            \abs*{\estbiasinterpgroup{\idxgroup} - \estbiasmeangrouptrain{\idxgroup}}
            \le \frac{12}{\constfrac\numcourse\numstudent} \cdot \max_{\idxpairparen\in \settrain}\abs*{\estbias_\idxpair}\qquad \forall \idxgroupscope.
        \end{align*}
    \end{lemma}
    The proof of this result is provided in Appendix~\ref{app:proof_lem_cv_only_bias_bound_interp_mean}.
    Combining Lemma~\ref{lem:cv_only_bias_bound_interp_mean} with the consistency~\eqref{eq:cv_only_bias_group_order_bias_converge_in_prob} of $\estbiasmtxat{0}$ from Step 1 and a bound on $\max_{\idxpairparen\in \settrain}\abs*{\bias_\idxpair}$ from Lemma~\ref{lem:maximal_gaussian}, we have the following lemma.
    \begin{lemma}\label{cor:cv_only_bias_bound_interp_mean}
    Under the same condition as Lemma~\ref{lem:cv_only_bias_bound_interp_mean}, the interpolated bias at $\reg=0$ satisfies
    \begin{align*}
            \limn \Probbig{\abs*{\estbiasinterpgroup{\idxgroup} - \biasmeangrouptrain{\idxgroup}} < \errbound, \quad \forall \idxgroupscope} = 1.
        \end{align*}
    \end{lemma}
     The proof of this result is provided in Appendix~\ref{app:proof_cor_cv_only_bias_bound_interp_mean}.
     Recall that $\meanbiasgroup{\idxgroup}$ denotes the the mean of the bias of any group $\idxgroupscope$. The following lemma gives concentration inequality results that the quantities $\biasmeancoursegroupval{\idxcourse}{\idxgroup}$ and $\biasmeangrouptrain{\idxgroup}$ are close to $\biasgroup{\idxgroup}$. Note that this lemma is on the bias $\biasmtx$ and does not involve any estimator.
    \begin{lemma}\label{lem:cv_only_bias_const_frac_hoeffding}
        Consider any group ordering $\setpartialorder$ that satisfies the all $\constfrac$-fraction assumption. Consider any fixed training-validation split $(\settrain, \setval)$ obtained by Algorithm~\ref{alg:cv}. For any $\errbound >0$, we have
        \begin{subequations}\label{eq:cv_only_bias_const_frac_bound_diff_mean_train}
        \begin{align}
            & \limn \Probbig{\abs*{\biasmeancoursegroupval{\idxcourse}{\idxgroup} - \meanbiasgroup{\idxgroup}} < \errbound, \quad \forall \idxcoursescope, \idxgroupscope}= 1\label{eq:cv_only_bias_const_frac_bound_diff_block_and_interp}\\
            & \limn \Probbig{\abs*{\biasmeangrouptrain{\idxgroup} - \meanbiasgroup{\idxgroup} } < \errbound, \quad \forall \idxgroupscope}= 1.\label{eq:cv_only_bias_const_frac_bound_dif_group_train}
    \end{align}
    \end{subequations}
    \end{lemma}
    The proof of this result is provided in Appendix~\ref{app:proof_lem_cv_only_bias_const_frac_hoeffding}.
    Combining Lemma~\ref{cor:cv_only_bias_bound_interp_mean} and~\eqref{eq:cv_only_bias_const_frac_bound_diff_mean_train} from Lemma~\ref{lem:cv_only_bias_const_frac_hoeffding} with a union bound, we have the following corollary.
    \begin{corollary}\label{cor:cv_only_bias_const_frac_bound}
        Consider any group ordering $\setpartialorder$ that satisfies the all $\constfrac$-fraction assumption. Consider any fixed $\settrain\subseteq [\numcourse]\times [\numstudent]$ obtained by Algorithm~\ref{alg:cv}. For any $\errbound>0$, the interpolated bias at $\reg=0$ satisfies
        \begin{align*}
            \limn \Probbig{\abs*{\biasmeancoursegroupval{\idxcourse}{\idxgroup} - \estbiasinterpgroup{\idxgroup}} < \errbound, \quad \forall \idxcoursescope, \idxgroupscope}=1.
        \end{align*}
    \end{corollary}
    Consider each $\idxcoursescope$ and $\idxgroupscope$. The terms in the validation error~\eqref{eq:cv_only_bias_const_frac_val_err_at_zero} involving course $\idxcourse$ and group $ \idxgroup$ are:
    \begin{align*}
         \errat{0}_{\idxcourse\idxgroup} \defn  \frac{1}{\abs*{\setval}} \sum_{\idxstudent\in \setvalcoursegroup{\idxcourse}{\idxgroup}} \left(\bias_\idxpair - \estmeanat{0}_\idxcourse - \estbiasinterpgroup{\idxgroup}\right)^2
         & = \frac{1}{\abs*{\setval}} \left[
            \sum_{\idxstudent \in \setvalcoursegroup{\idxcourse}{\idxgroup}} \left(\bias_\idxpair - \estbiasinterpgroup{\idxgroup}\right)^2
            +\abs*{\setvalcoursegroup{\idxcourse}{\idxgroup}}\cdot \estmean_\idxcourse^2 - 2\sum_{\idxstudent\in \setvalcoursegroup{\idxcourse}{\idxgroup}} \left(\bias_\idxpair - \estbiasinterpgroup{\idxgroup}\right) \estmean_\idxcourse
         \right]\\
         & \stackrel{\stepone}{=}
            \underbrace{\frac{1}{\abs*{\setval}} \sum_{\idxstudent \in \setvalcoursegroup{\idxcourse}{\idxgroup}} \left(\bias_\idxpair - \estbiasinterpgroup{\idxgroup}\right)^2}_{\term_1}
            +\underbrace{ \frac{\abs*{\setvalcoursegroup{\idxcourse}{\idxgroup}}}{\abs*{\setval}} \estmean_\idxcourse^2}_{\term_2}
            - \underbrace{\frac{2\abs*{ \setvalcoursegroup{\idxcourse}{\idxgroup}}}{\abs*{\setval}}  \cdot (\biasmeancoursegroupval{\idxcourse}{\idxgroup} - \estbiasinterpgroup{\idxgroup}) \estmean_\idxcourse}_{\term_3},
        \end{align*}
        where~\stepone is true by the definition~\eqref{eq:cv_only_bias_const_frac_def_mean_b_ik_val} of $\biasmeancoursegroupval{\idxcourse}{\idxgroup}$. We now consider the three terms $\term_1, \term_2$ and $\term_3$ (dependent on $\idxcourse$ and $\idxgroup$), respectively. \paragraph{Term $\term_2$:} By the convergence~\eqref{eq:cv_only_bias_consistency_mean} of $\estmeanat{0}$ in Theorem~\ref{thm:consistency_train}\ref{part:thm_consistency_train_const_frac}, we have
        \begin{align}\label{eq:cv_bias_const_frac_term_two}
            \limn \Prob\left(\term_2 \le \frac{\abs*{\setvalcoursegroup{\idxcourse}{\idxgroup}}}{\abs*{\setval}}\errbound_1^2, \quad \forall \idxcoursescope, \idxgroupscope\right) =1.
        \end{align}
       \paragraph{Term $\term_3$:} We have
       \begin{align*}
           \term_3 \le 2\frac{\abs*{\setvalcoursegroup{\idxcourse}{\idxgroup}}}{\sizesetval}\cdot \abs*{\biasmeancoursegroupval{\idxcourse}{\idxgroup} - \estbiasinterpgroup{\idxgroup}}\cdot \abs*{\estmean_\idxcourse}
           \le 2\abs*{\biasmeancoursegroupval{\idxcourse}{\idxgroup} - \estbiasinterpgroup{\idxgroup}}\cdot \abs*{\estmean_\idxcourse}.
       \end{align*}
       By combining the convergence~\eqref{eq:cv_only_bias_consistency_mean} of $\estmeanat{0}$ in Theorem~\ref{thm:consistency_train}\ref{part:thm_consistency_train_const_frac} and Corollary~\ref{cor:cv_only_bias_const_frac_bound} with a union bound, we have 
        \begin{align}\label{eq:cv_bias_const_frac_term_three}
            \limn \Probbig{\term_3 \le \frac{2\abs*{\setvalcoursegroup{\idxcourse}{\idxgroup}} }{\sizesetval}\errbound_1^2, \quad\forall \idxcoursescope,\idxgroupscope}= 1.
        \end{align}
        
        \paragraph{Term $\term_1$:} We have
        \begin{align*}
         \term_1 = \frac{1}{\abs*{\setval}}\sum_{\idxstudent\in \setvalcoursegroup{\idxcourse}{\idxgroup}} \left(\bias_\idxpair - \estbiasinterpgroup{\idxgroup}\right)^2
         & = \frac{1}{\abs*{\setval}}\sum_{\idxstudent \in \setvalcoursegroup{\idxcourse}{\idxgroup}} \left(\bias_\idxpair - \biasmeancoursegroupval{\idxcourse}{\idxgroup} +\biasmeancoursegroupval{\idxcourse}{\idxgroup} - \estbiasinterpgroup{\idxgroup}\right)^2\\
         & = \frac{1}{\abs*{\setval}}
         \left[
            \sum_{\idxstudent \in \setvalcoursegroup{\idxcourse}{\idxgroup}} (\bias_\idxpair - \biasmeancoursegroupval{\idxcourse}{\idxgroup} )^2
            +\abs*{\setvalcoursegroup{\idxcourse}{\idxgroup}} \cdot (\biasmeancoursegroupval{\idxcourse}{\idxgroup} - \estbiasinterpgroup{\idxgroup})^2 +  2\sum_{\idxstudent\in \setvalcoursegroup{\idxcourse}{\idxgroup}} (\bias_\idxpair - \biasmeancoursegroupval{\idxcourse}{\idxgroup})(\biasmeancoursegroupval{\idxcourse}{\idxgroup} - \estbiasinterpgroup{\idxgroup})
        \right]\\
        & \stackrel{\stepone}{=} \frac{1}{\abs*{\setval}}
         \left[
            \sum_{\idxstudent \in \setvalcoursegroup{\idxcourse}{\idxgroup}} (\bias_\idxpair - \biasmeancoursegroupval{\idxcourse}{\idxgroup} )^2
            +\abs*{\setvalcoursegroup{\idxcourse}{\idxgroup}} \cdot (\biasmeancoursegroupval{\idxcourse}{\idxgroup} - \estbiasinterpgroup{\idxgroup})^2 \right]
    \end{align*}
    where inequality~\stepone holds because  $\sum_{\idxstudent\in \setvalcoursegroup{\idxcourse}{\idxgroup}} (\bias_\idxpair - \biasmeancoursegroupval{\idxcourse}{\idxgroup})=0$ by the definition~\eqref{eq:cv_only_bias_const_frac_def_mean_b_ik_val} of $\biasmeancoursegroupval{\idxcourse}{\idxgroup}$.
    By Corollary~\ref{cor:cv_only_bias_const_frac_bound}, we have
    \begin{align}\label{eq:cv_bias_const_frac_term_one}
        \limn \left(\term_1 < \frac{1}{\abs*{\setval}} \sum_{\idxstudent\in \setvalcoursegroup{\idxcourse}{\idxgroup}} (\bias_\idxpair - \biasmeancoursegroupval{\idxcourse}{\idxgroup} )^2 +\frac{\abs*{\setvalcoursegroup{\idxcourse}{\idxgroup}}}{\sizesetval}\errbound_1^2, \quad \forall \idxcoursescope, \idxgroupscope \right) = 1.
    \end{align}
    Combining the three terms from~\eqref{eq:cv_bias_const_frac_term_two}, ~\eqref{eq:cv_bias_const_frac_term_three} and~\eqref{eq:cv_bias_const_frac_term_one}, we bound $\erratcoursegroup{0}{\idxcourse}{\idxgroup}$ as
    \begin{align}\label{eq:cv_only_bias_const_frac_val_err_ik_bound}
        \limn \left(\erratcoursegroup{0}{\idxcourse}{\idxgroup} = \term_1 + \term_2 + \term_3 
        < \frac{1}{\abs*{\setval}} \sum_{\idxstudent\in \setvalcoursegroup{\idxcourse}{\idxgroup}} (\bias_\idxpair - \biasmeancoursegroupval{\idxcourse}{\idxgroup} )^2 + \frac{4\abs*{\setvalcoursegroup{\idxcourse}{\idxgroup}}}{\sizesetval} \errbound_1^2, \quad\forall \idxcoursescope, \idxgroupscope\right) = 1.
    \end{align}
    By the all $\constfrac$-fraction assumption, the number of groups is upper-bounded by a constant as $\numgroup\le \frac{1}{\constfrac}$. Taking a union bound of~\eqref{eq:cv_only_bias_const_frac_val_err_ik_bound} over $\idxcoursescope$ and $\idxgroupscope$, we have
    \begin{align}
        \limn \prob\left(\errat{0} = \sum_{\idxcoursescope, \idxgroupscope} \erratcoursegroup{0}{\idxcourse}{\idxgroup} < \frac{1}{\sizesetval} \sum_{\idxcoursescope, \idxgroupscope} \left[\sum_{\idxstudent\in \setvalcoursegroup{\idxcourse}{\idxgroup}} (\bias_\idxpair - \biasmeancoursegroupval{\idxcourse}{\idxgroup})^2 + 4\abs*{\setvalcoursegroup{\idxcourse}{\idxgroup}}\cdot \errbound_1^2 \right] \right) &=1 \nonumber\\
        \limn \prob\left(\errat{0} < \frac{1}{\sizesetval} \sum_{\idxcoursescope, \idxgroupscope} \sum_{\idxstudent\in \setvalcoursegroup{\idxcourse}{\idxgroup}} (\bias_\idxpair - \biasmeancoursegroupval{\idxcourse}{\idxgroup} )^2 + 4\errbound_1^2\right) & = 1.\label{eq:cv_only_bias_const_frac_val_err_atzero_bound}
    \end{align}
    This completes Step 2 of bounding the validation error at $\reg=0$.

    \noindent\textbf{Step 3: Computing the validation error at general $\reg\in \setregoutside$, and showing that it is greater than the validation error at $\reg=0$}
    
    Recall from~\eqref{eq:def_set_reg_outside} the definition of the random set $\setregoutside\defn \{\reg\in [0, \infty]: \normtwo{\estmeanat{\reg}} > \errbound \}$. In this step, we show that 
    \begin{align}\label{eq:cv_bias_const_frac_step_three_goal}
        \limn \Prob\left(\errat{\reg} > \errat{0}, \quad \forall \reg\in \setregoutside\right) = 1.
    \end{align}
    From~\eqref{eq:cv_bias_const_frac_step_three_goal}, we have that the estimated quality $\estmeanat{\regcv}$ by cross-validation satisfies
    \begin{align*}
        \limn \left(\regcv\not\in \setregoutside\right) = 1
    \end{align*}
    and consequently by the definition of $\setregoutside$
    \begin{align*}
        \limn \prob\left(\normtwo{\estmeanat{\regcv}} < \errbound\right) =1.
    \end{align*}
    It remains to prove~\eqref{eq:cv_bias_const_frac_step_three_goal}.
    
    \paragraph{Proof of~\eqref{eq:cv_bias_const_frac_step_three_goal}} For any $\idxcoursescope$ and $\idxgroupscope$, the terms in the validation error at any $\reg\in [0, \infty]$ involving course $\idxcourse$ and group $\idxgroup$ are computed as:
    \begin{align}
         \erratcoursegroup{\reg}{\idxcourse}{\idxgroup} = \frac{1}{\abs*{\setval}} \sum_{\idxstudent\in\setvalcoursegroup{\idxcourse}{\idxgroup}} \left(\bias_\idxpair - \estmeanat{\reg}_\idxcourse - \estbiasinterpatgroup{\reg}{\idxgroup}\right)^2
         & = \frac{1}{\abs*{\setval}}\sum_{\idxstudent \in\setvalcoursegroup{\idxcourse}{\idxgroup}} \left(\bias_\idxpair - \biasmeancoursegroupval{\idxcourse}{\idxgroup} + \biasmeancoursegroupval{\idxcourse}{\idxgroup} - \estmean_\idxcourse - \estbiasinterpgroup{\idxgroup}\right)^2 \nonumber\\
         & \stackrel{\stepone}{=} \frac{1}{\abs*{\setval}} \sum_{\idxstudent\in\setvalcoursegroup{\idxcourse}{\idxgroup}}  \left(\bias_\idxpair - \biasmeancoursegroupval{\idxcourse}{\idxgroup}\right)^2 + \underbrace{
            \frac{\abs*{\setvalcoursegroup{\idxcourse}{\idxgroup}}}{\abs*{\setval}}  \left(\biasmeancoursegroupval{\idxcourse}{\idxgroup} - \estmean_\idxcourse - \estbiasinterpgroup{\idxgroup}\right)^2
        }_{\term_{\idxcourse\idxgroup}},\label{eq:cv_bias_const_frac_err_at_reg}
    \end{align}
    where~\stepone is true because $\sum_{\idxstudent\in\setvalcoursegroup{\idxcourse}{\idxgroup}}  (\bias_\idxpair - \biasmeancoursegroupval{\idxcourse}{\idxgroup}) = 0$ by the definition~\eqref{eq:cv_only_bias_const_frac_def_mean_b_ik_val} of $\biasmeancoursegroupval{\idxcourse}{\idxgroup}$.
    Note that the first term in~\eqref{eq:cv_bias_const_frac_err_at_reg} is identical to the first term in~\eqref{eq:cv_only_bias_const_frac_val_err_ik_bound} from Step 2. We now analyze the second term $\term_{\idxcourse\idxgroup}$ in~\eqref{eq:cv_bias_const_frac_err_at_reg}. On the one hand, by Lemma~\ref{lem:est_mean_err_implies_diff}\ref{part:est_mean_err_implies_diff_only_bias}, we have
    \begin{align}\label{eq:cv_bias_const_frac_step_three_conditional_one}
        \lim_{\numstudent\rightarrow\infty} \Prob\left(\max_{\idxcourse, \idxcoursealt\in [\numcourse]} \estmean_\idxcourse - \estmean_\idxcoursealt > \frac{\errbound}{\sqrt{\numcourse}},\quad \forall \reg\in \setregoutside\right) = 1.
    \end{align}
    On the other hand, taking a union bound of~\eqref{eq:cv_only_bias_const_frac_bound_diff_block_and_interp} in Lemma~\ref{lem:cv_only_bias_const_frac_hoeffding} over $\idxcourse, \idxcoursealt\in [\numcourse]$, we have
    \begin{align}\label{eq:cv_bias_const_frac_step_three_conditional_two}
        \limn \Probbig{ \abs*{\biasmeancoursegroupval{\idxcourse}{\idxgroup} - \biasmeancoursegroupval{\idxcoursealt}{\idxgroup}} <\frac{\errbound}{2\sqrt{\numcourse}}, \quad \forall \idxcourse, \idxcoursealt\in [\numcourse],\idxgroupscope } = 1.
    \end{align}
    Conditional on~\eqref{eq:cv_bias_const_frac_step_three_conditional_one} and~\eqref{eq:cv_bias_const_frac_step_three_conditional_two}, for every $\reg\in \setregoutside$ and for every $\idxgroupscope$,
    \begin{align*}
        \max_{\idxcourse,\idxcoursealt\in [\numcourse]} \abs*{\left(\biasmeancoursegroupval{\idxcourse}{\idxgroup} - \estmean_\idxcourse - \estbiasinterpgroup{\idxgroup}\right) - \left(\biasmeancoursegroupval{\idxcoursealt}{\idxgroup} - \estmean_\idxcoursealt - \estbiasinterpgroup{\idxgroup}\right)} & = \max_{\idxcourse,\idxcoursealt\in [\numcourse]} \abs*{\left(\biasmeancoursegroupval{\idxcourse}{\idxgroup}- \biasmeancoursegroupval{\idxcoursealt}{\idxgroup}\right) - \left(\estmean_\idxcourse - \estmean_\idxcoursealt\right)}\\
        & \ge \max_{\idxcourse,\idxcoursealt\in [\numcourse]}(\estmean_\idxcourse - \estmean_\idxcoursealt) - \max_{\idxcourse,\idxcoursealt\in [\numcourse]}\abs*{\biasmeancoursegroupval{\idxcourse}{\idxgroup} - \biasmeancoursegroupval{\idxcoursealt}{\idxgroup}}\\
        & > \frac{\errbound}{\sqrt{\numcourse}} - \frac{\errbound}{2\sqrt{\numcourse}} = \frac{\errbound}{2\sqrt{\numcourse}}.
    \end{align*}
    Hence, conditional on~\eqref{eq:cv_bias_const_frac_step_three_conditional_one} and~\eqref{eq:cv_bias_const_frac_step_three_conditional_two},
    \begin{align}\label{eq:cv_only_bias_const_frac_max_of_pair}
        \max_{\idxcourse, \idxcoursealt\in [\numcourse]}\left\{(\biasmeancoursegroupval{\idxcourse}{\idxgroup} - \estmean_\idxcourse - \estbiasinterpgroup{\idxgroup} )^2, (\biasmeancoursegroupval{\idxcoursealt}{\idxgroup} - \estmean_\idxcoursealt - \estbiasinterpgroup{\idxgroup} )^2\right\}\ge \frac{\errbound^2}{16\numcourse} \qquad\forall \idxgroupscope,  \forall \reg\in \setregoutside.
    \end{align}
    Now consider the terms $\term_{\idxcourse\idxgroup}$. By~\eqref{eq:bound_val_size_per_course} from Lemma~\ref{lem:bound_size} combined with the all $\constfrac$-fraction assumption, we have
    \begin{align}\label{eq:cv_bias_const_frac_bound_size_val}
        \frac{\abs*{\setvalcoursegroup{\idxcourse}{\idxgroup}}}{\sizesetval} \ge \frac{1}{\sizesetval}\cdot \frac{\abs*{\setcoursegroup{\idxcourse}{\idxgroup}}}{4} \ge \frac{\constfrac\numstudent}{4\sizesetval} = \frac{\constfrac}{2\numcourse}.
    \end{align}
    Conditional on~\eqref{eq:cv_bias_const_frac_step_three_conditional_one} and~\eqref{eq:cv_bias_const_frac_step_three_conditional_two}, for every $\reg\in \setregoutside$ and $\idxcoursescope$,
    \begin{align*}
        \max_{\idxcourse, \idxcoursealt\in [\numcourse]} (\term_{\idxcourse\idxgroup} + \term_{\idxcoursealt\idxgroup}) & \stackrel{\stepone}{\ge} \frac{\constfrac}{2\numcourse} \left[\left(\biasmeancoursegroupval{\idxcourse}{\idxgroup} - \estmean_\idxcourse - \estbias_\idxgroup^\texttrain\right)^2 + \left(\biasmeancoursegroupval{\idxcoursealt}{\idxgroup} - \estmean_\idxcoursealt - \estbias_\idxgroup^\texttrain\right)^2\right] \\
        & \stackrel{\steptwo}{\ge} \frac{\constfrac}{2\numcourse}\frac{\errbound^2}{16\numcourse} = \frac{\constfrac\errbound^2}{32\numcourse^2},
    \end{align*}
    where inequality~\stepone is true by~\eqref{eq:cv_bias_const_frac_bound_size_val}, and inequality~\steptwo is true by~\eqref{eq:cv_only_bias_const_frac_max_of_pair}. Now consider the validation error $\errat{\reg}$. Conditional on~\eqref{eq:cv_bias_const_frac_step_three_conditional_one} and~\eqref{eq:cv_bias_const_frac_step_three_conditional_two}, for every $\reg\in \setregoutside$,
    \begin{align*}
        \errat{\reg} = \sum_{\idxcoursescope, \idxgroupscope} \erratcoursegroup{\reg}{\idxcourse}{\idxgroup}\nonumber
        & \stackrel{\stepone}{\ge} \frac{1}{\sizesetval}\sum_{\idxcoursescope, \idxgroupscope} \sum_{\idxstudent\in \setvalcoursegroup{\idxcourse}{\idxgroup}} (\bias_\idxpair - \biasmeancoursegroupval{\idxcourse}{\idxgroup})^2 + \sum_{\idxcoursescope,\idxgroupscope}(\term_{\idxcourse\idxgroup} + \term_{\idxcoursealt\idxgroup})\nonumber\\
        & > \frac{1}{\sizesetval}\sum_{\idxcoursescope, \idxgroupscope} \sum_{\idxstudent\in \setvalcoursegroup{\idxcourse}{\idxgroup}} (\bias_\idxpair - \biasmeancoursegroupval{\idxcourse}{\idxgroup})^2 + \frac{\constfrac\errbound^2}{32\numcourse^2},
    \end{align*}
    where inequality~\stepone is true by plugging in~\eqref{eq:cv_bias_const_frac_err_at_reg}. Hence,
    \begin{align}\label{eq:cv_only_bias_const_frac_val_err_at_reg_bound}
        \limn \left(\errat{\reg} > \frac{1}{\sizesetval}\sum_{\idxcoursescope, \idxgroupscope} \sum_{\idxstudent\in \setvalcoursegroup{\idxcourse}{\idxgroup}} (\bias_\idxpair - \biasmeancoursegroupval{\idxcourse}{\idxgroup})^2 + \frac{\constfrac\errbound^2}{32\numcourse^2}, \quad\forall \reg\in \setregoutside\right) = 1.
    \end{align}
    We set $\errbound_1$ to be sufficient small such that $4\errbound_1^2 < \frac{\constfrac\errbound^2}{32\numcourse^2}$. Taking a union bound of~\eqref{eq:cv_only_bias_const_frac_val_err_at_reg_bound} with ~\eqref{eq:cv_only_bias_const_frac_val_err_atzero_bound} from Step 2, we have
    \begin{align*}
        \lim_{\numstudent\rightarrow \infty}\Prob\left(\errat{\reg} > \errat{0}, \quad\forall \reg\in \setregoutside\right) = 1,
    \end{align*}
    completing the proof of~\eqref{eq:cv_bias_const_frac_step_three_goal}.

\subsubsection{Proof of part~\ref{part:cv_bias_only_total}}

    We fix some constant $\errbound_1>0$ whose value is determined later.
    Since the partial ordering $\setpartialorder$ is assumed to be a total ordering, we also denote it as $\totalorder$.
    
    \noindent\textbf{Step 2: Computing the validation error at $\reg=0$}
    
    For any \element $\idxpairparen\in \setval$, recall that $\setnn(\idxcourse, \idxstudent; \totalorder)\subseteq [\numcourse]\times [\numstudent]$ denotes the set (of size $1$ or $2$) of its nearest neighbors in the training set $\settrain$ with respect to the total ordering $\totalorder$. We use $\setnn(\idxcourse, \idxstudent)$ as the shorthand notation for $\setnn(\idxcourse, \idxstudent; \totalorder)$. For any $\reg\in [0, \infty]$, we define the mean of the estimated bias over the \nn set
    \begin{align*}
        \estbiasat{\reg}_{\setnn(\idxcourse, \idxstudent)} \defn \frac{1}{\abs*{\setnn(\idxcourse, \idxstudent)}}\sum_{(\idxcoursealt, \idxstudentalt)\in \setnn(\idxcourse, \idxstudent)} \estbiasat{\reg}_{\idxcoursealt\idxstudentalt}
    \end{align*}
    Similarly, we define
    \begin{align*}
        \bias_{\setnn(\idxcourse, \idxstudent)} \defn \frac{1}{\abs*{\setnn(\idxcourse, \idxstudent)}}\sum_{(\idxcoursealt, \idxstudentalt)\in \setnn(\idxcourse, \idxstudent)} \bias_{\idxcoursealt\idxstudentalt}.
    \end{align*}
    Since $\setpartialorder$ is a total ordering, the set of total orderings consistent with $\setpartialorder=\totalorder$ is trivially itself, that is, $\settotalorder = \{\totalorder\}$. Then in Line~\ref{line:interpolation} of Algorithm~\ref{alg:cv}, the interpolated bias for any \element $\idxpairparen\in \setval$ is $\estbiasinterpat{\reg}_\idxpair = \estbiasat{\reg}_{\setnn(\idxcourse, \idxstudent)}$.
    
    Recall from the model~\eqref{eq:cv_bias_model} that $\obsmtx=\biasmtx$. The validation error at $\reg=0$ is computed as:
     \begin{align}
        \errat{0} & = \frac{1}{\abs*{\setval}} \sum_{\idxpairparen \in \setval} \left(\bias_\idxpair - \estbiasat{0}_{\setnn(\idxcourse, \idxstudent)} - \estmeanat{0}_{\idxcourse}\right)^2 \nonumber\\
        & \le \frac{1}{\abs*{\setval}}\sum_{\idxpairparen \in \setval} \left(\abs*{\bias_\idxpair -\bias_{\setnn(\idxcourse, \idxstudent)}} + \abs*{\bias_{\setnn(\idxcourse, \idxstudent)} - \estbiasat{0}_{\setnn(\idxcourse, \idxstudent)}} + \abs*{\estmeanat{0}_{\idxcourse}}\right)^2.\label{eq:cv_only_bias_total_order_err_at_zero}
    \end{align}
    We consider the three terms inside the summation in~\eqref{eq:cv_only_bias_total_order_err_at_zero} separately.
    For the first term $\abs*{\bias_\idxpair -\bias_{\setnn(\idxcourse, \idxstudent)}}$, combining Lemma~\ref{lem:rank_difference_adjacent_elements}\ref{part:rank_difference_adjacent_between_train_val} with Lemma~\ref{lem:maximal_spacing_order_stats}, we have
    \begin{align}\label{eq:cv_only_bias_total_order_err_at_zero_term_one}
        \lim_{\numstudent\rightarrow \infty} \Probbig{ \abs*{\bias_\idxpair -\bias_{\setnn(\idxcourse, \idxstudent)}} < \errbound_1, \quad \forall \idxpairparen\in \setval } = 1
    \end{align} 
    For the second term $\abs{\bias_{\setnn(\idxcourse, \idxstudent)} - \estbiasat{0}_{\setnn(\idxcourse, \idxstudent)}}$, we have $\abs{\bias_{\setnn(\idxcourse, \idxstudent)} - \estbiasat{0}_{\setnn(\idxcourse, \idxstudent)}} \le \max_{\idxcoursescope, \idxstudentscope}\abs{ \bias_\idxpair - \estbiasat{0}_\idxpair}$. By the consistency~\eqref{eq:cv_only_bias_group_order_bias_converge_in_prob} of $\estbiasmtxat{0}$ from Step 1, we have
    \begin{align}\label{eq:cv_only_bias_total_order_err_at_zero_term_two}
        \lim_{\numstudent\rightarrow \infty} \Prob\left(\abs{\bias_{\setnn(\idxcourse, \idxstudent)} - \estbiasat{0}_{\setnn(\idxcourse, \idxstudent)}} < \errbound_1, \quad \forall \idxpairparen\in \setval\right) = 1.
    \end{align}
    For the third term $\estmeanat{0}_\idxcourse$, by~\eqref{eq:cv_only_bias_consistency_mean} in Theorem~\ref{thm:consistency_train}\ref{part:thm_consistency_train_total_order}, we have 
    \begin{align}\label{eq:cv_only_bias_total_order_err_at_zero_term_three}
        \limn \Probbig{\abs*{\estmean_{\idxcourse}} <\errbound_1, \quad \forall \idxcoursescope} = 1.
    \end{align}
    Taking a union bound over the three terms~\eqref{eq:cv_only_bias_total_order_err_at_zero_term_one},~\eqref{eq:cv_only_bias_total_order_err_at_zero_term_two} and~\eqref{eq:cv_only_bias_total_order_err_at_zero_term_three} and plugging them back to~\eqref{eq:cv_only_bias_total_order_err_at_zero}, the validation error at $\reg=0$ satisfies
    \begin{align}\label{eq:cv_only_bias_total_order_err_at_zero_final}
        \lim_{\numstudent\rightarrow \infty} \Probbig{\errat{0} \le 9\errbound_1^2}= 1.
    \end{align}

    \noindent\textbf{Step 3: Computing the validation error at general $\reg\in \setregoutside$, and showing that it is greater than the validation error at $\reg=0$}
    
    Recall the definition $\setregoutside\defn \{\reg\in [0, \infty]: \normtwo{\estmeanat{\reg}}> \errbound \}$. In this step, we establish
    \begin{align*}
        \limn(\regcv\not\in \setregoutside) = 1.
    \end{align*}
    By Lemma~\ref{lem:est_mean_err_implies_diff}\ref{part:est_mean_err_implies_diff_only_bias} combined with the assumption that $\numcourse=2$, we have
    \begin{align}\label{eq:cv_noise_total_order_diff_outside}
        \limn \Probbig{ \underbrace{\abs*{\estmeanat{\reg}_1 - \estmeanat{\reg}_2} > \frac{\errbound}{\sqrt{2}},\quad \forall \reg\in \setregoutside}_{\event}} = 1.
    \end{align}
    We denote the the event in~\eqref{eq:cv_noise_total_order_diff_outside} as $\event$. We define 
    \begin{subequations}\label{eq:def_setreg_diff_large_two}
    \begin{align}
        & \setreg_{2>1} \defn \left\{\reg\in [0, \infty]: \estmeanat{\reg}_2 -\estmeanat{\reg}_1 > \frac{\errbound}{\sqrt{2}}\right\}\\
        & \setreg_{1>2}\defn \left\{\reg\in [0, \infty]: \estmeanat{\reg}_1 -\estmeanat{\reg}_2 > \frac{\errbound}{\sqrt{2}}\right\}.
    \end{align}
    \end{subequations}
Then we have
    \begin{align}\label{eq:cv_bias_total_order_set_inclusion}
        \setregoutside\subseteq 
        \setreg_{2>1} \union \setreg_{1>2}\;\given\; \event.
    \end{align}
    We first analyze $\setreg_{2>1}$. We discuss the following two cases, depending on the comparison of the mean of the bias for the two courses. 
    
    \noindent \textbf{Case 1: $\sum_{\idxstudentscope} \bias_{1\idxstudent} \ge \sum_{\idxstudentscope} \bias_{2\idxstudent}$}
    
    We denote the event that Case 1 happens as $\event_1\defn \{ \sum_{\idxstudentscope}\bias_{1\idxstudent} \ge \sum_{\idxstudentscope} \bias_{2\idxstudent} \}$.
    In this case, our goal is to show
    \begin{align}\label{eq:cv_bias_total_order_case_one_goal}
        \limn \Probbig{\regcv \not\in \setregoutside\intersect \setreg_{2>1} , \event_1} = \limn (\event_1).
    \end{align}
    To show~\eqref{eq:cv_bias_total_order_case_one_goal} it suffices to prove
    \begin{align*}
        \limn\Probbig{\setregoutside\intersect \setreg_{2>1} = \emptyset,\event_1} = \limn \prob(\event_1).
    \end{align*}
    We separately discuss the cases of $\reg=\infty$ and $\reg\not\in \infty$.
    
    \paragraph{Showing $\infty \not \in \setregoutside\intersect \setreg_{2>1}$:}
    
    Denote the mean of the bias in each course in the training set $\settrain$ as $\biascoursetrain{\idxcourse} \defn \frac{1}{\numstudenttrain} \sum_{\idxstudent\in \settraincourse{\idxcourse}} \bias_{\idxpair}$ for $\idxcourse\in \{1, 2\}$.
    By~\eqref{eq:bias_bound_train_per_course} in Lemma~\ref{lem:bias_bound_train}, we have
    \begin{subequations}\label{eq:cv_only_bias_total_order_hoeffding_train}
    \begin{align}
        & \lim_{\numstudent\rightarrow \infty} \Prob\left(\biascoursetrain{1} - \frac{1}{\numstudent}\sum_{\idxstudentscope} \bias_{1\idxstudent} < -\frac{\errbound}{8}\right)  = 0\\
        & \limn \Prob\left(\biascoursetrain{2} - \frac{1}{\numstudent}\sum_{\idxstudentscope} \bias_{2\idxstudent} > \frac{\errbound}{8}\right) = 0
    \end{align}
    \end{subequations}
    Taking a union bound of~\eqref{eq:cv_only_bias_total_order_hoeffding_train}, we have
    \begin{align}
        \lim_{\numstudent\rightarrow\infty} \Prob
        \left(\vphantom{\sum_{\idxstudent\in [\numstudent]}}\right.
            \underbrace{
                \biascoursetrain{1} - \biascoursetrain{2} > \frac{1}{\numstudent}\sum_{\idxstudent\in [\numstudent]}( \bias_{1\idxstudent} -\bias_{2\idxstudent}) - \frac{\errbound}{4}
            }_{\event'}
        \left.\vphantom{\sum_{\idxstudent\in [\numstudent]}}\right)
        = 1.\label{eq:cv_only_bias_total_order_case_one_hoeffding}
    \end{align}
    Denote this event in~\eqref{eq:cv_only_bias_total_order_case_one_hoeffding} as $\event'$. Hence, we have
    \begin{align}\label{cv_only_bias_total_order_case_one_hoeffding_condition}
        \left.\biastrain_1 - \biastrain_2 > -\frac{\errbound}{4} \;\middle|\; (\event', \event_1)\right.
    \end{align}
    Recall from Proposition~\ref{prop:property_existence_at_infty_general} that we have our estimator at $\reg=\infty$ equals to the sample mean per course. That is, $\estmeanat{\infty} = \begin{bmatrix}\biastrain_1\\\biastrain_2\end{bmatrix}$. Hence, we have
    \begin{align*}
        \left.\estmeanat{\infty}_2 - \estmeanat{\infty}_1 < \frac{\errbound}{4} \;\middle| \; (\event', \event_1).\right.
    \end{align*}
    By the definition of $\setreg_{2>1}$, we have
    \begin{align}\label{eq:cv_bias_total_order_case_one_case_inf}
        \infty\not\in \setregoutside\intersect\setreg_{2>1} \;|\; (\event', \event_1).
    \end{align}
    
    \paragraph{Showing $\reg \not \in \setregoutside\intersect \setreg_{2>1}$ for general $\reg\in [0, \infty)$:}
    
    As an overview, we assume there exists some $\reg\in \setregoutside\intersect \setreg_{2>1}\setminus \{\infty\}$ and derive a contradiction.

    Denote the mean of the bias in the training set $\settrain$ as $\biastrain \defn \frac{1}{\sizesetval}\sum_{\idxpairparen\in \setval} \bias_\idxpair = \frac{\biascoursetrain{1} + \biascoursetrain{2}}{2}$.
    Since $\reg\in \setreg_{2>1}$, we have $\estmeanat{\reg}_2 - \estmeanat{\reg}_1 > \frac{\errbound}{\sqrt{2}}$. By~\eqref{eq:estimator_sum_equality_mean_variable_size} in Lemma~\ref{lem:estimator_sum_equality}, we have
    \begin{align*}
        \estmeanat{\reg_1} + \estmeanat{\reg_2} = 2\biastrain,
    \end{align*}
    and hence $\estmeanat{\reg}$ can be reparameterized as
    \begin{align}\label{eq:cv_bias_total_order_x_reparameterize}
        \estmeanat{\reg} = \biastrain + \solutionincrease\begin{bmatrix}-1\\1\end{bmatrix},\text{ for some } \solutionincrease > \frac{\errbound}{2\sqrt{2}}.
    \end{align}
    The following lemma gives a closed-form formula for $\ell_2$-regularized isotonic regression. Recall that $\conemonotone$ denotes the monotonic cone, and the isotonic projection for any $y\in \reals^d$ is defined in~\eqref{eq:isotonic_proj} as $\proj_\conemonotone(y) = \argmin_{u\in \conemonotone} \normtwo{y-u}^2$.
    \begin{lemma}\label{lem:iso_regularized_obj_close_form}
        Consider any $y\in \reals^d$ and any $\reg \in [0, \infty)$. Then we have
        \begin{align}
            \min_{u\in \conemonotone} \left(\normtwo{y-u}^2 + \reg \normtwo{u}^2\right) = \frac{1}{1+\reg}\normtwo{y - \proj_\conemonotone(y)}^2 + \frac{\reg}{1 + \reg} \normtwo{y}^2.\label{eq:iso_regularized_obj_close_form}
        \end{align}
    \end{lemma}
    The proof of this result is provided in Appendix~\ref{app:proof_lem_iso_regularized_obj_close_form}.
    We denote the objective~\eqref{eq:optimization_restricted_general} under any fixed $\meancourse\in \reals^\numcourse$ as
    \begin{align}
        \objgivenx(\meancourse)& \defn \min_{\biasmtx \text{ obeys } \totalorder} \norm*{\obsmtx - \meancourse\vecone^T - \biasmtx}^2_\settrain + \reg\norm*{\biasmtx}_\settrain^2 \nonumber\\
        & \stackrel{\stepone}{=} \frac{1}{1 + \reg} \underbrace{
            \norm*{(\obsmtx - \meancourse \vecone^T) - \proj_\totalorder(\obsmtx- \meancourse \vecone^T)}_\settrain^2
        }_{\objgivenx_1(\meancourse)}+ \frac{\reg}{1+\reg} \underbrace{\norm*{\obsmtx-\meancourse\vecone^T}_\settrain^2}_{\objgivenx_2(\meancourse)},\label{eq:iso_regularized_obj_decompose}
    \end{align}
    where equality~\stepone is true by~\eqref{eq:iso_regularized_obj_close_form} in Lemma~\ref{lem:iso_regularized_obj_close_form}.
    We now construct an alternative estimate $\estmean' = \biastrain\begin{bmatrix}1\\1\end{bmatrix}$, and show that
    \begin{align*}
        \objgivenx(\estmean) > \objgivenx(\estmean')\qquad \forall\reg\in \setregoutside\intersect\setreg_{2>1}\setminus\{\infty\}.
    \end{align*}
    We consider the two terms $\objgivenx_1(\meancourse)$ and $\objgivenx_2(\meancourse)$ in~\eqref{eq:iso_regularized_obj_decompose} separately.
    
    \paragraph{Term $\objgivenx_1$:} Recall from the model~\eqref{eq:cv_bias_model} that $\obsmtx=\biasmtx$. Hence, $\obsmtx$ satisfies the total ordering $\totalorder$, and hence $\obsmtx - \estmean'\vecone^T = \obsmtx - \biastrain \begin{bmatrix}1\\1\end{bmatrix}\vecone_\numstudent^T$ satisfies the total ordering $\totalorder$. That is,
    \begin{align*}
        \proj_\totalorder(\obsmtx-\estmean'\vecone^T) = \obsmtx-\estmean'\vecone^T.
    \end{align*}
    Hence,
    \begin{align}
        0 = \objgivenx_1(\estmean')
        \le \objgivenx_1(\estmeanat{\reg})\qquad \forall \reg\in [0, \infty].\label{eq:cv_only_bias_total_order_case_one_term_one}
    \end{align}
    \paragraph{Term $\objgivenx_2$: } We have
    \begin{align*}
        \objgivenx_2(\estmean) - \objgivenx_2(\estmean') & = \norm{\obsmtx - \estmeanat{\reg}\vecone^T}_\settrain^2 - \norm{\obsmtx - \estmean'\vecone^T}_\settrain^2 \nonumber\\
        & = \sum_{\idxstudent\in \settraincourse{1}} (\bias_{1\idxstudent} - \estmeanat{\reg}_1)^2 + \sum_{\idxstudent\in \settraincourse{2}} (\bias_{2\idxstudent} - \estmeanat{\reg}_2)^2 - \left[\sum_{\idxstudent\in \settraincourse{1}} (\bias_{1\idxstudent} - \estmean'_1)^2 + \sum_{\idxstudent\in \settraincourse{2}} (\bias_{2\idxstudent} - \estmean'_2)^2\right] \nonumber\\
        & = \numstudenttrain \left[2\biascoursetrain{1} (\estmean'_1 - \estmeanat{\reg}_1) + 2\biascoursetrain{2}(\estmean'_2 - \estmeanat{\reg}_2) + ( (\estmeanat{\reg}_1)^2 - (\estmean'_1)^2) + ((\estmeanat{\reg}_2)^2 - (\estmean'_2)^2)\right] \nonumber\\
        & =\numstudenttrain [2\solutionincrease(\biastrain_1 - \biastrain_2) + 2\solutionincrease^2] \nonumber\\
        & = 2\numstudenttrain\solutionincrease (\biastrain_1 - \biastrain_2 + \solutionincrease) \stackrel{\stepone}{>} 0 \;\given \; (\event', \event_1),
    \end{align*}
    where inequality~\stepone is true by combining~\eqref{cv_only_bias_total_order_case_one_hoeffding_condition} with~\eqref{eq:cv_bias_total_order_x_reparameterize}. Hence, we have
    \begin{align}
        \left.\objgivenx_2(\estmean) > \objgivenx_2(\estmean'), \quad \forall \reg\in \setregoutside\intersect \setreg_{2>1}\setminus\{\infty\} \;\middle| \; (\event', \event_1).\right.
        \label{eq:cv_only_bias_total_order_case_one_term_two_final}
    \end{align}
    Combining the term $\objgivenx_1$ from~\eqref{eq:cv_only_bias_total_order_case_one_term_one} and the term $\objgivenx_2$ from~\eqref{eq:cv_only_bias_total_order_case_one_term_two_final}, we have
    \begin{align*}
        \left.\objgivenx(\estmeanat{\reg}) > \objgivenx(\estmean'), \quad \forall \reg\in \setregoutside\intersect \setreg_{2>1}\setminus\{\infty\} \;\middle| \; (\event', \event_1).\right.
    \end{align*}
    Contradiction to the assumption that $\estmeanat{\reg}$ is optimal. Hence, we have
    \begin{align}\label{eq:cv_bias_total_order_case_one_case_not_inf}
        \regcv\not\in \setregoutside \intersect \setreg_{2>1}\setminus\{\infty\}  \;|\; (\event', \event_1).
    \end{align}
    Combining the cases of $\reg= \infty$ from~\eqref{eq:cv_bias_total_order_case_one_case_inf} and $\reg\ne\infty$ from~\eqref{eq:cv_bias_total_order_case_one_case_not_inf}, we have
    \begin{align*}
        \regcv\not\in \setregoutside \intersect \setreg_{2>1} \;|\; (\event', \event_1).
    \end{align*}
    Hence,
    \begin{align}
        \prob\left(\regcv\not\in \setregoutside \intersect \setreg_{2>1}, \event_1\right)& \ge \prob(\event', \event_1)\nonumber\\
        & = \prob(\event_1) - \prob(\event_1\intersect\setcomplement{\event'})\nonumber\\
        & \ge \prob(\event_1) - \prob(\setcomplement{\event'})\label{eq:cv_bias_total_order_case_one_prob}
    \end{align}
    Taking the limit of~\eqref{eq:cv_bias_total_order_case_one_prob}, we have
    \begin{align}
        \limn \prob\left(\regcv\not\in \setregoutside \intersect \setreg_{2>1}, \event_1\right) \stackrel{\stepone}{=} \limn \prob(\event_1),\label{eq:cv_bias_total_order_case_one_final}
    \end{align}
    where~\stepone is true by~\eqref{eq:cv_only_bias_total_order_case_one_hoeffding}.

    \noindent\textbf{Case 2: $\sum_{\idxstudentscope} \bias_{1\idxstudent} < \sum_{\idxstudentscope} \bias_{2\idxstudent}$}
    
    Denote the event that Case 2 happens as $\event_2\defn \left\{\sum_{\idxstudentscope} \bias_{1\idxstudent}< \sum_{\idxstudentscope} \bias_{2\idxstudent} \right\}$. 
    Our goal is to find a set of \elements on which the validation error is large. For any constant $\const > 0$, we define the set:
    \begin{align}\label{eq:cv_only_bias_total_order_def_set_pair}
        \setpairindices_\const \defn \{(\idxstudent, \idxstudent')\in [\numstudent]^2: 0 < \bias_{2\idxstudentalt} - \bias_{1\idxstudent} < \const\}.
    \end{align}
    Let $\const' > 0$ be a constant. Denote $\eventnonoverlaplinearval{\const'}{\const}$ as the event that there exists distinct values $(\idxstudent_1, \ldots, \idxstudent_{\const'\numstudent})$ and distinct values $(\idxstudent_1', \ldots, \idxstudent_{\const'\numstudent}')$, such that $(\idxstudent_k, \idxstudent_k')\in \setpairindices_\const \intersect \setval$ for all $k\in [\const'\numstudent]$. That is, the set $\setpairindices_\const\intersect\setval$ contains a subset of size at least $\const'\numstudent$ of pairs $(\idxstudent, \idxstudentalt)$, such that each \element $\bias_{1\idxstudent}$ and $\bias_{2\idxstudentalt}$ appears at most once in this subset. We denote this subset as $\setpairindicesnonoverlap$.

    The following lemma bounds the probability that $\eventnonoverlaplinearval{\const'}{\const}$ happens under case $\event_2$.
    
    \begin{lemma}\label{lem:index_pair_proportional_size_val}
        Suppose $\numcourse=2$. Assume the bias is distributed according to assumption~\ref{assumption:bias} with $\gaussianwidthbias= 1$. For any $\const > 0$, there exists a constant $\const' > 0$ such that
        \begin{align*}
            \limn \Prob\left(\eventnonoverlaplinearval{\const'}{\const}\intersect \event_2\right) = \limn \Prob(\event_2).
        \end{align*}
    \end{lemma}
    The proof of this result is provided in Appendix~\ref{app:proof_lem_index_pair_proportional_size_val}.
    Now consider the the validation error contributed by the pairs in the set $\setpairindicesnonoverlap$. We have
    \begin{align}
        \errat{\reg} \le \frac{1}{\sizesetval}\sum_{(\idxstudent, \idxstudentalt)\in \setpairindicesnonoverlap} \left[ \left(\bias_{1\idxstudent} - \estbiasat{\reg}_{\setnn(1,\idxstudent)} - \estmeanat{\reg}_1\right)^2 + \left(b_{2\idxstudentalt} - \estbiasat{\reg}_{\setnn(2,\idxstudentalt)} - \estmeanat{\reg}_2\right)^2 \right].\label{eq:cv_bias_total_order_case_two_val_err_at_reg}
    \end{align}
    We consider each individual term $(\idxstudent, \idxstudentalt)\in \setpairindicesnonoverlap$.  On the one hand, we have $\bias_{1\idxstudent} < \bias_{2\idxstudent'}$ by the definition~\eqref{eq:cv_only_bias_total_order_def_set_pair} of $\setpairindices_\const$. Therefore, the \element $(1,\idxstudent)$ is ranked lower than $(2, \idxstudent')$ in the total ordering $\settotalorder$. According to Algorithm~\ref{alg:cv}, it can be verified that their interpolated bias satisfies
    \begin{align}\label{eq:cv_only_bias_total_order_relation_b}
       \estbiasinterpat{\reg}_{\setnn(1,\idxstudent)} \le \estbiasinterpat{\reg}_{\setnn(2,\idxstudentalt)}\qquad\forall \reg\in [0, \infty].
    \end{align}
    On the other hand, we have
    \begin{align}\label{eq:cv_only_bias_total_order_relation_b_x}
        \left.\bias_{1\idxstudent} - \estmean_1 - (\bias_{2\idxstudentalt} - \estmean_2) = (\bias_{1\idxstudent} - \bias_{2\idxstudentalt}) + (\estmean_2 - \estmean_1)  \stackrel{\stepone}{>} -\frac{\errbound}{2} + \frac{\errbound}{\sqrt{2}} = \frac{\errbound}{5}, \quad\forall \reg\in \setregoutside\intersect \setreg_{2>1}\;\middle|\; (\eventnonoverlaplinearval{\const'}{\frac{\errbound}{2}}, \event),  \right.
    \end{align}
    where \stepone is true by the definition of $\setpairindices_\const$ in~\eqref{eq:cv_only_bias_total_order_def_set_pair} (setting $\const= \frac{\errbound}{2}$), and the definition~\ref{eq:def_setreg_diff_large_two} of $\setreg_{2>1}$.
    Combining~\eqref{eq:cv_only_bias_total_order_relation_b} and~\eqref{eq:cv_only_bias_total_order_relation_b_x}, we have that for all $(\idxstudent, \idxstudentalt)\in \setpairindicesnonoverlap$:
    \begin{align}
        \left(\bias_{1\idxstudent} - \estbiasinterpat{\reg}_{\setnn(1,\idxstudent)} - \estmeanat{\reg}_1\right)^2 + \left(\bias_{2\idxstudentalt} - \estbiasinterpat{\reg}_{\setnn(2,\idxstudentalt)} - \estmeanat{\reg}_2\right)^2 & \ge \min_{\substack{u_1, u_2\in \reals\\ u_1 \le u_2}} \min_{\substack{v_1, v_2 \in \reals\\v_1 - v_2 > \frac{\errbound}{5}}} (v_1- u_1)^2 +(v_2 - u_2)^2 \nonumber\\
        & > \left.\frac{\errbound^2}{50}, \quad\forall \reg\in \setregoutside\intersect\setreg_{2>1} \;\middle|\; (\eventnonoverlaplinearval{\const'}{\frac{\errbound}{2}},\label{eq:cv_bias_total_order_case_two_val_err_at_reg_pair} \event).\right.
    \end{align}
    Conditional on $\eventnonoverlaplinearval{\const'}{\frac{\errbound}{2}}$, there are at least $\const'\numstudent$ such non-overlapping pairs. Plugging~\eqref{eq:cv_bias_total_order_case_two_val_err_at_reg_pair} to~\eqref{eq:cv_bias_total_order_case_two_val_err_at_reg}, the validation error is lower-bounded as 
    \begin{align}\label{eq:cv_bias_total_order_case_two_err_at_reg_condition}
        \left.\errat{\reg} \ge \frac{1}{\sizesetval} \const'\numstudent\cdot \frac{\epsilon^2}{50}
        \ge \frac{2}{\numcourse\numstudent}\const'\numstudent \cdot \frac{\errbound^2}{50} = \frac{\const' \errbound^2}{25\numcourse},\quad\forall\reg\in \setregoutside\intersect\setreg_{2>1} \;\middle|\; (\eventnonoverlaplinearval{\const'}{\frac{\errbound}{2}}, \event).\right.
    \end{align}
    Setting the constant $\errbound_1$ to be a sufficiently small constant such that $9\errbound_1^2 < \frac{\const'\errbound^2}{25\numcourse}$, we have
    \begin{align}
        \prob \left(\errat{\reg} \ge \errat{0}, \quad\forall \reg\in \setregoutside\intersect\setreg_{2>1}, \event_2 \right) & \ge \prob\left(\errat{\reg} > \frac{\const'\errbound^2}{25\numcourse} > 9\errbound_1^2 > \errat{0}, \quad \forall \reg\in \setregoutside\intersect\setreg_{2>1}, \event_2\right)\nonumber\\
        & \ge \prob\left(\errat{\reg} > \frac{\const'\errbound^2}{25\numcourse}, \event_2\right) - \prob\left(\errat{0} > 9\errbound_1^2, \event_2\right)\nonumber\\
        & \stackrel{\stepone}{\ge} \prob\left(\eventnonoverlaplinearval{\const'}{\frac{\errbound}{2}}, \event, \event_2\right) - \prob\left(\errat{0} > 9\errbound_1^2\right)\\
        & = \prob\left(\eventnonoverlaplinearval{\const'}{\frac{\errbound}{2}}, \event\right) - \prob\left(\eventnonoverlaplinearval{\const'}{\frac{\errbound}{2}}, \event, \setcomplement{\event_2}\right) - \prob\left(\errat{0} > 9\errbound_1^2\right),\label{eq:cv_bias_case_two_prob_err_greater}
    \end{align}
    where~\stepone is true by~\eqref{eq:cv_bias_total_order_case_two_err_at_reg_condition}.
    Taking the limit of $\numstudent\rightarrow \infty$ in~\eqref{eq:cv_bias_case_two_prob_err_greater}, we have
    \begin{align*}
        \limn \prob\left(\errat{\reg} \ge \errat{0}, \quad\forall \reg\in \setregoutside\intersect\setreg_{2>1}, \event_2\right) = \limn \prob(\event_2).
    \end{align*}
    and \steptwo is true by combining Lemma~\ref{lem:index_pair_proportional_size_val},~\eqref{eq:cv_noise_total_order_diff_outside} and~\eqref{eq:cv_only_bias_total_order_err_at_zero_final} from Step 2. Equivalently,
    \begin{align}\label{eq:cv_bias_total_order_case_two_final}
        \limn \prob\left(\regcv\not\in \setregoutside\intersect \setreg_{2>1}, \event_2\right) = 1.
    \end{align}

    Finally, combining the two cases from~\eqref{eq:cv_bias_total_order_case_one_final} and~\eqref{eq:cv_bias_total_order_case_two_final}, we have
    \begin{subequations}\label{eq:cv_bias_total_order_reg_outside_two_cases}
    \begin{align}
        \limn \prob\left(\regcv\not\in \setregoutside\intersect\setreg_{2>1}\right) & = \limn \prob\left(\regcv\not\in \setregoutside\intersect\setreg_{2>1}, \event_1\right) + \limn \prob\left(\regcv\not\in \setregoutside\intersect\setreg_{2>1}, \event_2\right)\nonumber \\
        & = \limn \prob(\event_1) + \limn \prob(\event_2) = 1. 
    \end{align}
    By a symmetric argument on the set $\setreg_{1>2}$, we have
    \begin{align}
        \limn \prob\left(\regcv\not\in \setregoutside\intersect\setreg_{1>2}\right) & = 1.
    \end{align}
    \end{subequations}
    Hence, we have\begin{align*}
        \limn \prob\left(\regcv\not\in \setregoutside\right) & \ge \limn \prob\left(\regcv\not\in \setregoutside, \event\right) \\
        & \stackrel{\stepone}{\ge} \limn \prob\left(\regcv\not\in \setregoutside\intersect \setreg_{1>2}, \event\right) + \limn \prob\left(\regcv\not\in \setregoutside\intersect \setreg_{2>1}, \event\right)\\
        & \ge \limn \prob\left(\regcv\not\in \setregoutside\intersect \setreg_{1>2}\right) + \prob\left(\regcv\not\in \setregoutside\intersect \setreg_{2>1}\right) - 2 \limn \prob(\setcomplement{\event})\stackrel{\steptwo}{=} 1,
    \end{align*}
    where inequality~\stepone is true by~\eqref{eq:cv_bias_total_order_set_inclusion}, and equality~\steptwo is true by combining~\eqref{eq:cv_bias_total_order_reg_outside_two_cases} with~\eqref{eq:cv_noise_total_order_diff_outside}. This completes the proof.

\subsection{Proof of Theorem~\ref{thm:cv_noise_only}}\label{app:proof_thm_cv_noise_only}

The proof follows notation in Appendix~\ref{app:notation} and preliminaries in Appendix~\ref{app:preliminaries}. Similar to the proof of Theorem~\ref{thm:cv_bias_only}, without loss of generality we assume $\truemean=0$ and the standard deviation of the Gaussian noise is $\gaussianwidthnoise=1$. Under this setting, the model~\eqref{eq:model} reduces to:
\begin{align}
    \obsmtx= \noisemtx.\label{eq:cv_noise_model}
\end{align}
The proof consists of $3$ steps that are similar to the steps in Theorem~\ref{thm:cv_bias_only}. Both part~\ref{part:cv_noise_only_constant_fraction} and part~\ref{part:cv_noise_only_total} share the same first two steps as follows. We fix some constants $\errbound_1, \errbound_2 > 0$, whose values are determined later.

\noindent \textbf{Step 1: Showing the consistency of our estimator at $\lambda=\infty$ restricted to the training set $\settrain$} 
    
    By Proposition~\ref{prop:property_existence_at_infty_general}, our estimator $\estmeanat{\infty}$ at $\reg=\infty$ is identical to taking the sample mean of each course. By the model~\eqref{eq:cv_noise_model}, conditional on any training-validation split $(\settrain, \setval)$ given by Algorithm~\ref{alg:cv}, each observation is i.i.d. noise of $\normal(0, 1)$. Recall from~\eqref{eq:size_train_val_half} that the number of observations in each course restricted to the training set $\settrain$ is $\numstudenttrain = \frac{\numstudent}{2}$. Given the assumption~\ref{assumption:d} that the number of courses $\numcourse$ is a constant, sample mean on the training set $\settrain$ is consistent. That is,
    \begin{align}\label{eq:cv_only_noise_sample_mean_consistency}
        \limn\Probbig{\norminf{\estmeanat{\infty}} < \errbound_1} = 1.
    \end{align}
    By Proposition~\ref{prop:property_existence_at_infty_general}, we have $\estbiasmtxat{\infty} = 0$.
    
\noindent\textbf{Step 2: Computing the validation error at $\reg=\infty$}
    
    Recall from Algorithm~\ref{alg:cv} that the interpolated bias $\estbiasinterp_\idxpair$ for any \element $\idxpairparen\in \setval$ is computed as the mean of the estimated bias $\estbiasmtx$ from its nearest neighbor set in the training set $\settrain$. Since the estimated bias is $\estbiasmtxat{\infty}=0$, the interpolated bias is $\estbiasmtxvalat{\infty}= 0$. Recall the model~\eqref{eq:cv_noise_model} of $\obsmtx=\noisemtx$. The validation error at $\reg=\infty$ is computed as
    \begin{align}
         \errat{\infty} = \frac{1}{\abs*{\setval}} \sum_{\idxpairparen\in \setval} \left(\obs_{\idxpair} - \estmeanat{\infty}_\idxcourse - \estbiasinterpat{\infty}_\idxpair\right)^2 & =\frac{1}{\abs*{\setval}} \sum_{ \idxpairparen\in \setval} \left(\noise_{\idxpair} - \estmeanat{\infty}_\idxcourse\right)^2\nonumber \\
        & = \frac{1}{\sizesetval}
            \left[\vphantom{\sum_{\idxpairparen\in \setval}}\right.
            \underbrace{
                \sum_{\idxpairparen\in \setval} {\noise_\idxpair}^2}_{\term_1}
            - 2\underbrace{\sum_{\idxpairparen\in \setval} \noise_{\idxpair}\estmeanat{\infty}_\idxcourse}_{\term_2} + \underbrace{\sum_{\idxpairparen\in \setval} (\estmeanat{\infty}_\idxcourse)^2}_{\term_3}
        \left.\vphantom{\sum_{\idxpairparen\in \setval}}\right].\label{eq:cv_only_noise_err_at_infty}
    \end{align}
     We consider the three terms $\term_1, \term_2$ and $\term_3$ in~\eqref{eq:cv_only_noise_err_at_infty} separately. For the term $\term_1$, we have $\Expect[\noise_\idxpair^2] = \gaussianwidthnoise^2 = 1$. The number of samples is $\sizesetval = \numcourse\numstudentval = \numcourse \frac{\numstudent}{2}$. By Hoeffding's inequality we have
     \begin{align}\label{eq:cv_only_noise_const_frac_sample_mean_hoffding_one}
         \limn \Prob\left(\frac{1}{\abs*{\setval}} \sum_{\idxpairparen\in \setval} \noise_\idxpair^2 < 1 + \errbound_1\right) = 1.
     \end{align}
     For the term $\term_2$, we have $\Expect[\noise_\idxpair] = 0$. By Hoeffding's inequality and a union bound over $\idxcoursescope$ we have
     \begin{align}\label{eq:cv_only_noise_const_frac_sample_mean_hoffding_two}
         \limn \Prob\left(\frac{1}{\abs*{\setval}} \abs*{\sum_{\idxstudent\in \setvalcourse{\idxcourse}} \noise_{\idxpair} } < \errbound_1, \quad\forall \idxcoursescope\right) = 1.
     \end{align}
     Combining~\eqref{eq:cv_only_noise_const_frac_sample_mean_hoffding_two} with the consistency result~\eqref{eq:cv_only_noise_sample_mean_consistency} on $\estmeanat{\infty}$ from Step 1, we have
     \begin{align}\label{eq:cv_only_noise_const_frac_sample_mean_term_two}
         \limn \prob\left(\frac{1}{\sizesetval}\abs*{\term_2} < \numcourse\errbound_1^2\right)= 1.
     \end{align}
     For the term $\term_3$, we have
     \begin{align}\label{eq:cv_only_noise_sample_mean_term_three_decompose}
         \frac{1}{\sizesetval} \term_3 \le \max_{\idxcoursescope}\abs*{\estmean_\idxcourse}^2.
     \end{align}
     Combining~\eqref{eq:cv_only_noise_sample_mean_term_three_decompose} with the consistency result~\eqref{eq:cv_only_noise_sample_mean_consistency} on $\estmeanat{\infty}$ from Step 1, we have
     \begin{align}\label{eq:cv_only_noise_sample_mean_term_three}
         \limn \Prob\left(\frac{1}{\sizesetval} \term_3  < \errbound_1^2\right) = 1.
     \end{align}
     Taking a union bound of the terms $\term_1, \term_2$ and $\term_3$ from~\eqref{eq:cv_only_noise_const_frac_sample_mean_hoffding_one},~\eqref{eq:cv_only_noise_const_frac_sample_mean_term_two} and~\eqref{eq:cv_only_noise_sample_mean_term_three} and plugging them back to~\eqref{eq:cv_only_noise_err_at_infty}, we have
     \begin{align}\label{eq:cv_noise_err_at_infty_bound}
         \limn \Prob\left(\errat{\infty}\le (1 + \errbound_1) + \numcourse\errbound_1^2 + \errbound_1^2 = 1 + \errbound_1 + (\numcourse+1)\errbound_1^2 \right) = 1.
     \end{align}
     
     \textbf{Step 3 (preliminaries): Computing the validation error at general $\reg\in \setregoutside$, and showing that it is greater than the validation error at $\reg=\infty$}
     
     We set up some preliminaries for this step that are shared between part~\ref{part:cv_noise_only_constant_fraction} and part~\ref{part:cv_noise_only_total}. Then we discuss the two parts separately.
    
    Recall from~\eqref{eq:def_set_reg_outside} the definition of $\setregoutside\defn \{\reg\in [0, \infty]: \normtwo{\estmeanat{\reg}} >\errbound \}$. In this step, we show that 
    \begin{align}\label{eq:cv_noise_err_at_reg_greater}
        \limn \Prob\left(\errat{\reg} > \errat{\infty}, \quad \forall \reg\in \setregoutside\right) =1.
    \end{align}
    Then from~\eqref{eq:cv_noise_err_at_reg_greater} we have
    \begin{align*}
        \limn \left(\regcv \not\in \setregoutside\right) = 1,
    \end{align*}
    yielding the result of Theorem~\ref{thm:cv_noise_only}. It is sufficient to establish~\eqref{eq:cv_noise_err_at_reg_greater}.

    We now give some additional preliminary results for this step.
    By Lemma~\ref{lem:est_mean_err_implies_diff}, we have
    \begin{align}\label{eq:cv_only_noise_const_frac_assumption_mean_diff}
        \lim_{\numstudent\rightarrow \infty}\Prob
            \underbrace{\left(\max_{\idxcourse, \idxcoursealt\in [\numcourse]} \estmean_\idxcourse - \estmean_\idxcoursealt > \frac{\errbound}{\sqrt{\numcourse}}, \quad \forall \reg\in \setregoutside\right)}_{\event} = 1.
    \end{align}
    We denote this event in~\eqref{eq:cv_only_noise_const_frac_assumption_mean_diff} as $\event$.
    
    Both parts also use the following lemma that bounds the magnitude of the estimated bias $\estbiasmtx$ given some value of $\estmean$.
    
    \begin{lemma}\label{lem:bound_b_whp_given_x}
        Let $\set\subseteq [\numcourse]\times [\numstudent]$ be any non-empty set. For any $\reg \in [0, \infty]$, the solution $(\estmeanat{\reg}, \estbiasmtxat{\reg})$ restricted to the set $\set$ satisfies the deterministic relation
        \begin{align}\label{eq:bound_b_whp_given_x}
            \max_{\idxpairparen\in \set} \abs*{\estbiasat{\reg}_\idxpair} \le \max_{\idxpairparen\in \set} \abs*{\obs_\idxpair} + \norminf{\estmeanat{\reg}}.
        \end{align}
    \end{lemma}
    The proof of this result is provided in Appendix~\ref{app:proof_lem_bound_b_whp_given_x}.
    Now we proceed differently for Step 3 for part~\ref{part:cv_noise_only_constant_fraction} and part~\ref{part:cv_noise_only_total}.

\subsubsection{Proof of part~\ref{part:cv_noise_only_constant_fraction}}

    \paragraph{Step 3 (continued):}
    For clarity of notation, we denote the constant in the single constant-fraction as $\constfrac$.
    
    We analyze the validation error at any $\reg\in \setregoutside$ similar to Step 2. The difference is that Step 2 (at $\reg=\infty$) uses the consistency of $\estmeanat{\infty}$ from Step 1 on to bound the validation error. However, $\estmeanat{\reg}$ may not be consistent for any general $\reg\in \setregoutside$. Hence, we consider the following two subsets of $\setregoutside$ depending on the value of $\estmean$. 
    
    Similar to the proof of Theorem~\ref{thm:cv_bias_only}\ref{part:cv_bias_only_constant_fraction}, by Algorithm~\ref{alg:cv} the interpolated bias for \elements in each group $\idxgroupscope$ is identical for all $\idxpairparen\in \setvalgroup{\idxgroup}$. That is,
    \begin{align}\label{eq:cv_noise_bias_interp_equal_within_group}
        \estbiasinterp_\idxpair = \estbiasinterp_{\idxcoursealt\idxstudentalt}\qquad \forall \idxpairparen, (\idxcoursealt, \idxstudentalt)\in \setvalgroup{\idxgroup}.
    \end{align}
    We denote the interpolated bias for group $\idxgroup$ as $\estbiasinterp_\idxgroup\defn \estbiasinterp_\idxpair$ for $\idxpairparen\in \setvalgroup{\idxgroup}$.

 \paragraph{Case 1: } $\setreg_1 \defn \left\{\reg\in [0, \infty]: \max_{\idxcourse, \idxcoursealt\in [\numcourse]} \estmean_\idxcourse - \estmean_\idxcoursealt > 8\sqrt{\frac{\numcourse}{\constfrac}}\right\}$.
    
    Let $\idxgroupconstfrac\in [\numgroup]$ be a group that satisfies the single $\constfrac$-fraction assumption. By the definition of $\setreg_1$ we have 
        $\max_{\idxcourse, \idxcoursealt\in [\numcourse]} \left[(\estmean_\idxcourse + \estbiasinterpgroup{\idxgroupconstfrac})- (\estmean_\idxcoursealt + \estbiasinterpgroup{\idxgroupconstfrac})\right] > 8\sqrt{\frac{\numcourse}{\constfrac}}$ for any $\reg\in \setreg_1$,
    which implies that 
    \begin{align}\label{eq:cv_only_noise_const_frac_case_one_max}
        \max_{\idxcoursescope}\abs*{\estmean_\idxcourse + \estbiasinterpgroup{\idxgroupconstfrac}} > 4\sqrt{\frac{\numcourse}{\constfrac}}\qquad\forall\reg\in \setreg_1.
    \end{align}
    Combining~\eqref{eq:bound_val_size_per_course} from Lemma~\ref{lem:bound_size} with the single $\constfrac$-fraction assumption, one can see 
    \begin{align}\label{eq:cv_noise_const_frac_size_bound}
        \blocklengthcoursegroupval{\idxcourse}{\idxgroupconstfrac} \ge \frac{\blocklengthcoursegroup{\idxcourse}{\idxgroupconstfrac}}{4} > \frac{\constfrac\numstudent}{4}.
    \end{align}
    Given~\eqref{eq:cv_noise_const_frac_size_bound}, by Hoeffding's inequality we have
    \begin{subequations}\label{eq:cv_only_noise_const_frac_hoeffding_sign}
    \begin{align}
        \lim_{\numstudent\rightarrow\infty} \Prob\left(\sum_{\idxstudent\in \setvalcoursegroup{\idxcourse}{\idxgroupconstfrac}} \indicator\{\noise_{\idxpair} > 0\}  \ge \frac{\constfrac\numstudent}{12}\right) & = 1\\
        \lim_{\numstudent\rightarrow\infty} \Prob\left(\sum_{\idxstudent\in \setvalcoursegroup{\idxcourse}{\idxgroupconstfrac}} \indicator\{\noise_{\idxpair} < 0\}  \ge \frac{\constfrac\numstudent}{12}\right) & = 1.
    \end{align}
    \end{subequations}
    We denote the event
    \begin{align}\label{eq:cv_noise_const_frac_event_noise_sign}
        \event_1 \defn \left\{\sum_{\idxstudent\in \setvalcoursegroup{\idxcourse}{\idxgroupconstfrac}} \indicator\{\noise_{\idxpair} > 0\} \ge \frac{\constfrac\numstudent}{12}, \quad \forall \idxcoursescope\right\} \intersect \left\{\sum_{\idxstudent\in \setvalcoursegroup{\idxcourse}{\idxgroupconstfrac}} \indicator\{\noise_{\idxpair} <0\} \ge \frac{\constfrac\numstudent}{12}, \quad \forall \idxcoursescope\right\}.
    \end{align}
    Given that $\numcourse$ is a constant by the assumption~\ref{assumption:d}, taking~\eqref{eq:cv_only_noise_const_frac_hoeffding_sign} with a union bound over $\idxcoursescope$, we have
    \begin{align}\label{eq:cv_only_noise_const_frac_hoeffding_prob}
        \lim_{\numstudent\rightarrow\infty} \Prob(\event_1) = 1.
    \end{align}
    Let $\idxcourse^*$ be a random variable (as a function of $\reg$) defined as $\idxcourse^* \defn \argmax_{\idxcourse\in [\numcourse]} \abs*{\estmean_\idxcourse + \estbiasinterp_{\idxgroupconstfrac}}$ where the tie is broken arbitrarily. Conditional on $\event_1$, for any $\reg\in \setreg_1$ we have the deterministic relation
    \begin{align}
        \errat{\reg} = \frac{1}{\sizesetval}\sum_{\idxgroupscope}\sum_{\idxpairparen\in \setvalgroup{\idxgroup}} \left(\noise_\idxpair - \estmeanat{\reg}_\idxcourse - \estbiasinterpatgroup{\reg}{\idxgroup}\right)^2
        & \ge \frac{1}{\sizesetval} \sum_{\idxpairparen\in \setvalgroup{\idxgroupconstfrac}} (\noise_\idxpair - \estmean_\idxcourse - \estbiasinterpgroup{\idxgroupconstfrac})^2\nonumber\\
        & \ge \frac{1}{\sizesetval} \sum_{\idxstudent\in \setvalcoursegroup{\idxcourse^*}{\idxgroupconstfrac}} (\noise_{\idxcourse^*\idxstudent} - \estmean_{\idxcourse^*} - \estbiasinterpgroup{\idxgroupconstfrac})^2\nonumber\\
        & \stackrel{\stepone}{\ge} \frac{1}{\sizesetval} \frac{\constfrac\numstudent}{12} \left(4\sqrt{\frac{\numcourse}{\constfrac}}\right)^2\nonumber\\
        & = \left.\frac{2}{\numcourse\numstudent}\cdot \frac{\constfrac\numstudent}{12}\frac{16\numcourse}{\constfrac} =  \frac{8}{3},\quad\forall \reg\in \setreg_1 \;\middle|\; \event_1.\right.\label{eq:cv_only_noise_const_frac_case_one_prob_conditional}
    \end{align}
    where~\stepone is true by~\eqref{eq:cv_only_noise_const_frac_case_one_max} and the definition~\eqref{eq:cv_noise_const_frac_event_noise_sign} of $\event_1$.
    Combining~\eqref{eq:cv_only_noise_const_frac_case_one_prob_conditional} with~\eqref{eq:cv_only_noise_const_frac_hoeffding_prob}, we have
    \begin{align}\label{eq:cv_only_noise_const_frac_case_one_final}
        \lim_{\numstudent\rightarrow \infty}\Prob\left(\errat{\reg} \ge \frac{4}{3}, \quad \forall \reg\in \setreg_1\right) \ge \Prob\left(\event_1\right) =1.
    \end{align}
    
    \paragraph{Case 2:} $\setreg_2 = \setregoutside\intersect \left\{\reg\in [0, \infty]: \max_{\idxcourse, \idxcoursealt\in [\numcourse]} \estmean_\idxcourse-\estmean_\idxcoursealt \le 8\sqrt{\frac{\numcourse}{\constfrac}}\right\}$.
    
    Note that we have $\setregoutside\subseteq \setreg_1\union \setreg_2$ by the definition of $\setreg_1$ and $\setreg_2$. We decompose the validation error as:
    \begin{align}
        \errat{\reg} & = \frac{1}{\sizesetval} \sum_{\idxgroupscope}\sum_{\idxpairparen\in \setvalgroup{\idxgroup}} \left(\noise_{\idxpair} - \estmeanat{\reg}_\idxcourse - \estbiasinterpatgroup{\reg}{\idxgroup}\right)^2\nonumber \\
        & = \frac{1}{\sizesetval}\left[
                \sum_{\idxpairparen\in \setval} \noise_\idxpair^2
            - 2\sum_{\idxgroupscope}\sum_{\idxpairparen\in \setvalgroup{\idxgroup}} \noise_{\idxpair} \left(\estmeanat{\reg}_\idxcourse+\estbiasinterpatgroup{\reg}{\idxgroup}\right)
            + \sum_{\idxgroupscope}\sum_{\idxpairparen\in \setvalgroup{\idxgroup}} \left(\estmeanat{\reg}_\idxcourse + \estbiasinterpatgroup{\reg}{\idxgroup}\right)^2\right]\nonumber\\
        & = \frac{1}{\sizesetval}
        \left[\vphantom{\sum_{\idxpairparen\in \setval}}\right.
            \underbrace{
                \sum_{\idxpairparen\in \setval} \noise_\idxpair^2}_{\term_1}
            - 2\underbrace{\sum_{\idxpairparen\in \setval} \noise_{\idxpair} \estmeanat{\reg}_\idxcourse}_{\term_2}
            +2\underbrace{\sum_{\idxgroupscope}\sum_{\idxpairparen\in \setvalgroup{\idxgroup}}\noise_\idxpair \estbiasinterpatgroup{\reg}{\idxgroup}}_{\term_3}
            +\underbrace{\sum_{\idxgroupscope}\sum_{\idxpairparen\in \setvalgroup{\idxgroup}} \left(\estmeanat{\reg}_\idxcourse + \estbiasinterpatgroup{\reg}{\idxgroup}\right)^2}_{\term_4}
        \left.\vphantom{\sum_{\idxpairparen\in \setval}}\right].\label{eq:cv_only_noise_cnst_frac_err_decompose}
    \end{align}
    We analyze the four terms $\term_1, \term_2, \term_3$ and $\term_4$ in~\eqref{eq:cv_only_noise_cnst_frac_err_decompose} separately. 
    
    \paragraph{Term $\term_1$:} Similar to~\eqref{eq:cv_only_noise_const_frac_sample_mean_hoffding_one} from Step 2, by Hoeffding's inequality we have
    \begin{align}\label{eq:cv_only_noise_const_frac_err_at_reg_hoffding_one}
         \lim_{\numstudent\rightarrow \infty} \Prob\left(\frac{1}{\abs*{\setval}} \sum_{\idxpairparen\in \setval} \noise_\idxpair^2 > 1 - \errbound_2\right) = 1.
     \end{align}
    \paragraph{Term $\term_2$:} Recall that  $\numcourse$ is a constant by the assumption~\ref{assumption:d}. Similar to~\eqref{eq:cv_only_noise_const_frac_sample_mean_hoffding_two} from Step 2, by Hoeffding with a union bound over $\idxcoursescope$, we have
    \begin{align}\label{eq:cv_noise_const_frac_err_at_reg_term_two_hoeffding}
        \lim_{\numstudent\rightarrow \infty} \Prob\underbrace{
            \left(\frac{1}{\sizesetval} \abs*{\sum_{\idxstudent\in \setvalcourse{\idxcourse}} \noise_\idxpair} < \errbound,\quad \forall \idxcoursescope \right)
        }_{\event_2} =1.
    \end{align}
    Denote this event in~\eqref{eq:cv_noise_const_frac_err_at_reg_term_two_hoeffding} as $\event_2$.

    We now bound $\norminf{\estmean}$. By Hoeffding's inequality, on the training $\settrain$ we have:
    \begin{align}\label{eq:cv_noise_const_frac_err_at_reg_term_two_hoeffding_overall}
        \limn \prob\underbrace{
            \left(\frac{1}{\sizesettrain}\abs*{\sum_{\idxpairparen\in \settrain} \noise_\idxpair} < \sqrt{\frac{1}{\numcourse\constfrac}}\right)
        }_{\event_2'}= 1.
    \end{align}
    Plugging~\eqref{eq:estimator_sum_equality_mean_variable_size} in Lemma~\ref{lem:estimator_sum_equality} to~\eqref{eq:cv_noise_const_frac_err_at_reg_term_two_hoeffding_overall}, we have
    \begin{align}\label{eq:cv_only_noise_const_frac_case_two_hoeffding}
        \left.\abs*{\sum_{\idxcoursescope} \estmeanat{\reg}_\idxcourse} = \frac{1}{\numstudenttrain} \abs*{\sum_{\idxpairparen\in \settrain} \noise_\idxpair} < \sqrt{\frac{\numcourse}{\constfrac}}\qquad \forall \reg\in \setreg_2,\quad\text{conditional on }\event_2'.\right.
    \end{align}
    Combining~\eqref{eq:cv_only_noise_const_frac_case_two_hoeffding} with the definition of $\setreg_2$, we have
    \begin{align}\label{eq:cv_only_nose_const_frac_x_inf_bound}
        \left.\norminf{\estmean} \le 8\sqrt{\frac{\numcourse}{\constfrac}}\qquad \forall\reg\in \setreg_2\;\middle|\; \event_2'.\right.
    \end{align}
    To see~\eqref{eq:cv_only_nose_const_frac_x_inf_bound}, assume for contradiction that~\eqref{eq:cv_only_nose_const_frac_x_inf_bound} does not hold. Consider the case of $\estmean_{\idxcourse^*} > 8\sqrt{\frac{\numcourse}{\constfrac}}$ for some $\idxcourse^*\in [\numcourse]$. Then by the definition of $\setreg_2$, we have $\estmean_\idxcourse> 0$ for all $\idxcoursescope$. Then we have $\abs*{\sum_{\idxcoursescope} \estmean_\idxcourse} > 8\sqrt{\frac{\numcourse}{\constfrac}}$. Contradiction to~\eqref{eq:cv_only_noise_const_frac_case_two_hoeffding}. A similar argument applies if $\estmean_{\idxcourse^*} < -8\sqrt{\frac{\numcourse}{\constfrac}}$. Hence,~\eqref{eq:cv_only_nose_const_frac_x_inf_bound} holds. 
    
    Finally, combining~\eqref{eq:cv_only_nose_const_frac_x_inf_bound} with~\eqref{eq:cv_noise_const_frac_err_at_reg_term_two_hoeffding}, we have:
    \begin{align}\label{eq:cv_noise_const_frac_err_at_reg_term_two_condition}
        \frac{1}{\sizesetval} \abs*{\term_2} & = \frac{1}{\sizesetval} \abs*{\sum_{\idxpairparen\in \setval} \noise_\idxpair \estmean_\idxcourse} \\
        & \le \frac{\numcourse}{\sizesetval}
        \max_{\idxcoursescope}\abs*{\sum_{\idxpairparen\in \setval} \noise_\idxpair}\cdot \norminf{\estmean} < 8\numcourse\sqrt{\frac{\numcourse}{\constfrac}}\errbound_2\qquad\forall \reg\in \setreg_2, \quad\text{conditional on }(\event_2, \event_2').
    \end{align}
    Hence, we have
    \begin{align*}
        \limn \prob\left(\frac{1}{\sizesetval} \abs*{\term_2} < 8\numcourse\sqrt{\frac{\numcourse}{\constfrac}}\errbound,\quad\forall\reg\in \setreg_2\right) \ge \limn \prob\left(\event_2 \intersect\event_2'\right)\stackrel{\stepone}{=} 1,
    \end{align*}
    where~\stepone is true by~\eqref{eq:cv_noise_const_frac_err_at_reg_term_two_hoeffding} and~\eqref{eq:cv_noise_const_frac_err_at_reg_term_two_hoeffding_overall}.
    
    \paragraph{Term $\term_3$:} We use the following standard result derived from statistics.
    \begin{lemma}\label{lem:gaussian_width}
        Consider any fixed $d \ge 1$. Let $Z\sample \normal(0, I_d)$. Then we have
        \begin{align*}
            \lim_{d\rightarrow \infty}\Prob\left(\sup_{\substack{
                \normtwo{\theta} = 1\\
                \theta_1 \le \ldots \le \theta_d}
            } \theta^T Z\le d^{\frac{1}{4}} \right) = 1.
        \end{align*}
    \end{lemma}
    For completeness, the proof of this lemma is in Appendix~\ref{app:proof_gaussian_width}.
    We now explain how to apply Lemma~\ref{lem:gaussian_width} on $\estbiasmtxval_\settrain$.
    
    \paragraph{The ordering of $\estbiasmtxval$:} Take any arbitrary total ordering $\totalorder\in \settotalorder$ that is consistent with the partial ordering $\setpartialorder$. Recall from~\eqref{eq:cv_noise_bias_interp_equal_within_group} that the interpolated bias within each group $\idxgroupscope$ is identical, so $\estbiasmtxval$ satisfies the total ordering $\totalorder$.
    
    \paragraph{Bounding $\norm{\estbiasmtxval}_\settrain$:}
    We bound each $\estbiasinterpgroup{\idxgroup}$. Recall that each $\estbiasinterpgroup{\idxgroup}$ is a mean of $\estbiasmtx$ on its nearest-neighbor set. Hence, we have
    \begin{align}\label{eq:bound_b_whp_given_x_repeat}
        \max_{\idxgroupscope}\abs{\estbiasinterpgroup{\idxgroup}}\le \max_{\idxpairparen\in \settrain} \abs*{\estbiasat{\reg}_\idxpair} \stackrel{\stepone}{\le} \max_{\idxpairparen\in \settrain} \abs*{\obs_\idxpair} + \norminf{\estmeanat{\reg}} \qquad\forall \reg\in [0, \infty],
    \end{align}
    where~\stepone is true by~\eqref{eq:bound_b_whp_given_x} in Lemma~\ref{lem:bound_b_whp_given_x}.
    We consider the term $\max_{\idxpairparen\in \setval}\abs{\obs_\idxpair}$ on the RHS of~\eqref{eq:bound_b_whp_given_x_repeat}. Recall from the model~\eqref{eq:cv_noise_model} that $\obsmtx=\noisemtx$. Hence, we have
    \begin{align}\label{eq:cv_only_noise_const_frac_max_gaussian}
        \limn \Prob\underbrace{
            \left(\max_{\idxpairparen\in \setval}\abs*{\obs_\idxpair} < 2\sqrt{\log\numcourse\numstudent}\right)
        }_{\event_2''}\stackrel{\stepone}{=} 1,
    \end{align}
    where~\stepone is true by Lemma~\ref{lem:maximal_gaussian}. Plugging~\eqref{eq:cv_only_noise_const_frac_max_gaussian} and the bound on $\norminf{\estmean}$ from~\eqref{eq:cv_only_nose_const_frac_x_inf_bound} to~\eqref{eq:bound_b_whp_given_x_repeat}, we have that conditional on $\event_2''$ and $\event_2'$,
    \begin{align*}
        \max_{\idxgroupscope} \abs{\estbiasinterpgroup{\idxgroup}}
        &{\le } \max_{\idxpairparen\in \settrain} \abs{\obs_\idxpair} + \norminf{\estmeanat{\reg}}\\
        & {\le} \left.2\sqrt{\log\numcourse\numstudent} + 8\sqrt{\frac{\numcourse}{\constfrac}}\qquad\forall \reg\in \setreg_2 \;\middle|\; (\event_2', \event_2'').\right.
    \end{align*}
    Hence, we have
    \begin{align*}
        \left.\norm{\estbiasmtxval}_\settrain \le \sqrt{\sizesettrain}\cdot  \max_{\idxgroupscope} \abs*{\estbiasinterpgroup{\idxgroup}} \le \sqrt{\numcourse\numstudentval}\left(2\sqrt{\log\numcourse\numstudent} + 8\sqrt{\frac{\numcourse}{\constfrac}} \right) \qquad\forall \reg\in \setreg_2 \;\middle|\; (\event_2', \event_2'').\right.
    \end{align*}
    and therefore
    \begin{align}\label{eq:cv_noise_const_frac_bound_b_interp}
        \limn \prob\left( \norm{\estbiasmtxval}_\settrain \le \sqrt{\numcourse\numstudentval}\left(2\sqrt{\log\numcourse\numstudent} + 8\sqrt{\frac{\numcourse}{\constfrac}} \right), \quad\forall \reg\in \setreg_2\right) \ge \limn \prob(\event_2'\intersect\event_2'')= 1.
    \end{align}
    \paragraph{Applying Lemma~\ref{lem:gaussian_width}:}
    For the term $\term_3$, for any constant $C > 0$, we have
    \begin{align}\label{eq:cv_noise_const_frac_term_three_decompose}
        \Prob\left(
        \abs*{\term_3} < C(\numcourse\numstudenttrain)^\frac{1}{4},\quad\forall \reg\in \setreg_2\right) 
        \ge \Prob
        \left(\vphantom{\abs*{\frac{\term_3}{C}}}\right.
            \underbrace{
                \left\{\abs*{\frac{\term_3}{C}} < (\numcourse\numstudenttrain)^\frac{1}{4},\quad\forall \reg\in \setreg_2\right\}
            }_{\event_3} \intersect
            \underbrace{
                \left\{\norm*{\frac{\estbiasmtxval}{C}}_\settrain \le 1,\quad \forall \reg\in \setreg_2\right\}}_{\event_4}
        \left.\vphantom{\abs*{\frac{\term_3}{C}}}\right)
    \end{align}
    We have
    \begin{align}\label{eq:cv_noise_const_frac_term_three_event_relation}
        \Prob(\setcomplement{\event_3\intersect \event_4}) = \Prob(\setcomplement{\event_4}) + \Prob(\setcomplement{\event_3}\intersect \event_4) 
    \end{align}
    Setting $C =\sqrt{\numcourse\numstudentval}\left(2\sqrt{\log\numcourse\numstudent} + 8\sqrt{\frac{\numcourse}{\constfrac}} \right)$, by~\eqref{eq:cv_noise_const_frac_bound_b_interp} we have 
    \begin{align}\label{eq:cv_noise_const_frac_term_three_event_one}
        \Prob(\setcomplement{\event_4}) = 0.
    \end{align}
    Applying Lemma~\ref{lem:gaussian_width} on $\frac{\estbiasmtxval_\settrain}{C}$, we have
    \begin{align}\label{eq:cv_noise_const_frac_term_three_event_two}
        \limn \Prob(\setcomplement{\event_3} \intersect \event_4)  = 0.
    \end{align}
    Plugging~\eqref{eq:cv_noise_const_frac_term_three_event_one} and~\eqref{eq:cv_noise_const_frac_term_three_event_two} to~\eqref{eq:cv_noise_const_frac_term_three_event_relation}, we have
    \begin{align}\label{eq:cv_noise_const_frac_term_three_event_zero}
        \limn \Prob(\setcomplement{\event_3 \intersect\event_4}) = 0.
    \end{align}
    Combining~\eqref{eq:cv_noise_const_frac_term_three_event_zero} with~\eqref{eq:cv_noise_const_frac_term_three_decompose}, we have
    \begin{align*}
        \limn \Prob\left(\abs*{\term_3} < C(\numcourse\numstudenttrain)^\frac{1}{4} = (\numcourse\numstudenttrain)^\frac{3}{4} \left(2\sqrt{\log \numcourse\numstudent} + 8\sqrt{\frac{\numcourse}{\constfrac}}\right),\quad\forall \reg\in \setreg_2\right) = 1.
    \end{align*}
    Hence, we have
    \begin{align}\label{eq:cv_only_noise_const_frac_case_two_term_three}
         \lim_{\numstudent\rightarrow \infty }\Prob\left( \frac{1}{\sizesetval} \abs*{\term_3} < \errbound_2\right) = 1.
    \end{align}
    
    \paragraph{Term $\term_4$:} Recall that $\idxgroupconstfrac$ denotes a group $\idxgroupconstfrac$ that satisfies the single $\constfrac$-fraction assumption. By the definition of $\event$ from~\eqref{eq:cv_only_noise_const_frac_assumption_mean_diff}, we have 
    \begin{align}
        \left.\max_{\idxcourse, \idxcoursealt\in [\numcourse]} (\estmean_\idxcourse + \estbiasinterpgroup{\idxgroupconstfrac}) - (\estmean_{\idxcoursealt} + \estbiasinterpgroup{\idxgroupconstfrac}) > \frac{\errbound}{\sqrt{\numcourse}} \qquad\forall \reg\in \setreg_2, \;\middle|\;\event.\right.
    \end{align}
    Therefore, we have
    \begin{align}\label{eq:cv_noise_const_frac_term_four_k_bound}
        \left.\max_{\idxcourse, \idxcoursealt\in [\numcourse]}\left[(\estmean_\idxcourse + \estbiasinterpgroup{\idxgroupconstfrac})^2+ (\estmean_{\idxcoursealt} + \estbiasinterpgroup{\idxgroupconstfrac})^2\right] > \frac{\errbound^2}{4\numcourse}\qquad\forall \reg\in \setreg_2\;\middle|\;\event.\right.
    \end{align}
    We bound the term $\term_4$ as
    \begin{align*}
        \left.\frac{1}{\sizesetval}\term_4 \ge \frac{1}{\sizesetval} \sum_{\idxpairparen\in \setvalgroup{\idxgroupconstfrac}} (\estmean_\idxcourse+\estbiasinterpgroup{\idxgroupconstfrac})^2 \stackrel{\stepone}{\ge} \frac{2}{\numcourse\numstudent}\cdot  \frac{\constfrac\numstudent}{4}\cdot \frac{\errbound^2}{4\numcourse} = \frac{\constfrac\errbound^2}{8\numcourse^2} \qquad\forall\reg\in \setreg_2\;\middle|\;\event,\right.
    \end{align*}
    where~\stepone is true by combining~\eqref{eq:cv_noise_const_frac_size_bound} and~\eqref{eq:cv_noise_const_frac_term_four_k_bound}. Hence,
    \begin{align}\label{eq:cv_only_noise_const_frac_case_two_term_four}
        \Prob\left(\term_4 \ge \frac{\constfrac\errbound^2}{8\numcourse^2} \quad\forall\reg\in \setreg_2\right) \ge \Prob\left(\event\right)= 1.
    \end{align}
    
    \paragraph{Putting things together:} Plugging the four terms from~\eqref{eq:cv_only_noise_const_frac_err_at_reg_hoffding_one},~\eqref{eq:cv_noise_const_frac_err_at_reg_term_two_hoeffding},~\eqref{eq:cv_only_noise_const_frac_case_two_term_three} and~\eqref{eq:cv_only_noise_const_frac_case_two_term_four} respectively back to~\eqref{eq:cv_only_noise_cnst_frac_err_decompose}, we have
    \begin{align}\label{eq:cv_only_noise_const_frac_case_two_final}
        \lim_{\numstudent\rightarrow \infty}\Prob\left(\errat{\reg} > (1-\errbound_2) + 8\numcourse\sqrt{\frac{\numcourse}{\constfrac}}\errbound_2 + \errbound_2 + \frac{\constfrac\errbound^2}{8\numcourse^2} , \quad \forall \reg\in\setreg_2 \right) = 1.
    \end{align}
    Finally, combining the two cases from~\eqref{eq:cv_only_noise_const_frac_case_one_final} and~\eqref{eq:cv_only_noise_const_frac_case_two_final}, we have
    \begin{align}\label{eq:cv_noise_const_frac_err_at_reg_final}
        \lim_{\numstudent\rightarrow \infty} \Prob\left(\errat{\reg} \ge \frac{8}{3}\minstat\left( 1 + 16\numcourse\sqrt{\frac{\numcourse}{\constfrac}}\errbound_2 + \frac{\constfrac\errbound^2}{8\numcourse^2}\right), \quad\forall \reg\in \setregoutside\right) =1.
    \end{align}
    Recall from~\eqref{eq:cv_noise_err_at_infty_bound} that the validation error at $\reg=\infty$ is bounded as
    \begin{align}
        \lim_{\numstudent\rightarrow \infty} \Prob\left(\errat{\infty}\le  1 + \errbound_1 + (\numcourse+1)\errbound_1^2 \right) = 1.\label{eq:cv_noise_err_at_infty_bound_recall}
    \end{align} 
    Combining~\eqref{eq:cv_noise_const_frac_err_at_reg_final} and~\eqref{eq:cv_noise_err_at_infty_bound_recall} with choices of $(\errbound_1, \errbound_2)$ (dependent on $\errbound, \numcourse, \constfrac$) such that $\frac{8}{3}\minstat\left( 1 + 16\numcourse\sqrt{\frac{\numcourse}{\constfrac}}\errbound_2 + \frac{\constfrac\errbound^2}{8\numcourse^2}\right) >  1 + \errbound_1 + (\numcourse+1)\errbound_1^2$, we have
    \begin{align*}
        \limn \prob\left(\errat{\infty} > \errat{0},\quad \forall \reg\in\setregoutside\right) = 1,
    \end{align*}
    completing the proof.

\subsubsection{Proof of part~\ref{part:cv_noise_only_total}}

    For clarity of notation, we denote the constant in the constant-fraction interleaving assumption as $\constfrac$.
    Since $\setpartialorder$ is a total ordering, we also denote it as $\totalorder$.
    
    \paragraph{Step 3 (continued):}
    Combining~\eqref{eq:estimator_sum_equality_mean_variable_size} with Hoeffding's inequality, we have
    \begin{align}\label{eq:cv_noise_total_order_sum_estmean_equality}
        \lim_{\numstudent\rightarrow \infty} \Prob
        \left(\vphantom{\abs*{\sum_{\idxpairparen\in \settrain}}}\right.
        \underbrace{
            \abs*{\estmean_1 + \estmean_2} = \frac{1}{\numstudenttrain}\abs*{\sum_{\idxpairparen\in \settrain} \noise_\idxpair} < \errbound\minstat \frac{16}{\sqrt{\constfrac}}, \quad \forall \reg\in \setregoutside 
        }_{\event_1}
        \left.\vphantom{\abs*{\sum_{\idxpairparen\in \settrain}}}\right)
        = 1.
    \end{align}
    We denote this event in~\eqref{eq:cv_noise_total_order_sum_estmean_equality} as $\event_1$.

    \paragraph{Analyzing the number of interleaving points} 
    Let $\setinterleaving\subseteq [2\numstudent-1]$ denotes the interleaving points. Recall that $(\idxcourse_\rank, \idxstudent_\rank)$ denotes \element of rank $\rank$, and $\rank_\idxpair$ denotes the rank of the \element $\idxpairparen$. We slightly abuse the notation to say $\idxpairparen\in \setinterleaving$ if $\rank_\idxpair\in \setinterleaving$, and also for other definitions of subsets of interleaving points later in the proof. Denote $\setinterleaving_\idxcourse\subseteq \setinterleaving$ as the set of interleaving points in course $\idxcourse\in \{1, 2\}$:
    \begin{align*}
        \setinterleaving_\idxcourse = \setinterleaving\intersect \{\rank\in [2\numstudent-1]: \idxcourse_\rank = \idxcourse\}.
    \end{align*}
    Denote $\setinterleavingval_\idxcourse$ as the set of interleaving points in $\setinterleaving_\idxcourse$ that are  in the validation set:
    \begin{align*}
        \setinterleavingval_\idxcourse = \setinterleaving_\idxcourse \intersect  \setval.
    \end{align*}
    We define $\setinterleavingpairs$ as a set of pairs of interleaving points as:
    \begin{align*}
        \setinterleavingpairs \defn \{ (\rank, \rank')\in [2\numstudent-1]^2: \rank\in \setinterleavingval_1,\; \rank' \in \setinterleavingval_2, \;\rank < \rank'\}.    
    \end{align*}
    Define $\event_\const$ as the event that there exists distinct values $(\rank_1, \rank_1', \ldots, \rank_{\const\numstudent}, \rank_{\const\numstudent}')$ such that $(\rank_\idxrank, \rank_{\idxrank}')\in \setinterleavingpairs$ for all $\idxrank\in [\const\numstudent]$. That is, $\setinterleavingpairs$ includes $\const\numstudent$ distinct pairs where each interleaving point appears at most once. We define $\setinterleavingpairs'$ likewise as
    \begin{align*}
         \setinterleavingpairs' \defn \{ (\rank, \rank')\in [2\numstudent-1]^2: \rank\in \setinterleavingval_2,\; \rank' \in \setinterleavingval_1, \;\rank < \rank'\}. 
    \end{align*}
    and define $\event'_\const$ likewise.
    
    The following lemma bounds the probability of the event $\event_{\frac{1}{36}}$ and $\event_{\frac{1}{36}}'$.
    \begin{lemma}\label{lem:cv_only_noise_total_order_size_set_interleaving}
        Suppose $\numcourse = 2$. Then we have
        \begin{align*}
            \lim_{\numstudent\rightarrow \infty} \Prob
                \left(\event_{\frac{1}{36}}\intersect \event_{\frac{1}{36}}'\right)= 1.
        \end{align*}
    \end{lemma}
    The proof of this result is provided in Appendix~\ref{app:proof_lem_cv_only_noise_total_order_size_set_interleaving}.
    Denote $\setinterleaving^+$ as the set of the half of the highest interleaving points and $\setinterleaving^-$ as the set of the half of the lowest interleaving points. That is, we define
    \begin{align*}
        \setinterleaving^+ & \defn \setinterleaving \intersect \{\rank\in [2\numstudent-1]: \rank > \median(\setinterleaving)\}\\
        \setinterleaving^- & \defn \setinterleaving \intersect \{\rank\in [2\numstudent-1]: \rank < \median(\setinterleaving)\}.
    \end{align*}
    Furthermore, for $\idxcourse\in \{1, 2\}$, we define
    \begin{align*}
        \setinterleavingvalplus_\idxcourse & \defn \setinterleaving^+ \intersect\setinterleaving_\idxcourse \intersect \setval\\
        \setinterleavingvalminus_\idxcourse & \defn \setinterleaving^- \intersect\setinterleaving_\idxcourse \intersect \setval.
    \end{align*}
    The following lemma lower-bounds the size of $\setinterleavingvalplus_\idxcourse$ and $\setinterleavingvalminus_\idxcourse$.
    \begin{lemma}\label{lem:cv_only_noise_total_order_size_half_set}
        We have
        \begin{align*}
            \lim_{\numstudent\rightarrow \infty}\Prob\underbrace{
                \left(\abs*{T} \ge \frac{\constfrac\numstudent}{36}, \quad \forall T\in \{\setinterleaving_1^{\textval+}, \setinterleaving_1^{\textval-}, \setinterleaving_2^{\textval+}, \setinterleaving_2^{\textval-}\}\right)
            }_{\event_2} = 1.
        \end{align*}
    \end{lemma}
    The proof of this result is provided in Appendix~\ref{app:proof_lem_cv_only_noise_total_order_size_half_set}.
    We denote this event in Lemma~\ref{lem:cv_only_noise_total_order_size_half_set} as $\event_2$.
    
    \paragraph{Bounding the validation error}
    Similar to part~\ref{part:cv_noise_only_constant_fraction}, we discuss the following two cases depending on the value of $\estmean$.
    
    \paragraph{Case 1: $\setreg_1 = \setreg_\errbound\intersect \left\{\reg\in [0, \infty]: \estmeanat{\reg}_1 < -\frac{32}{\sqrt{\constfrac}} \right\}$}
    It can be verified that due to~\eqref{eq:cv_noise_total_order_sum_estmean_equality}, we have
    \begin{align}\label{eq:cv_noise_total_order_case_one_assumption}
        \left.\estmeanat{\reg}_1 < -\frac{32}{\sqrt{\constfrac}} < \frac{16}{\sqrt{\constfrac}} < \estmeanat{\reg}_2 \quad\forall \reg\in \setreg_1 \;\middle|\; \event.\right.
    \end{align} 
    By Hoeffding's inequality combined with Lemma~\ref{lem:cv_only_noise_total_order_size_half_set}, we have
    \begin{subequations}\label{eq:cv_noise_total_order_case_one_hoeffding}
    \begin{align}
        \lim_{\numstudent\rightarrow\infty} \prob\left(\sum_{\idxpairparen\in \setinterleavingvalminus_1}\indicator\{\noise_{\idxpair} > 0\} > \frac{\constfrac\numstudent}{96}\right) = 1\label{eq:cv_noise_total_order_case_one_hoeffding_minus}\\
        \lim_{\numstudent\rightarrow\infty} \prob\left(\sum_{\idxpairparen\in \setinterleavingvalplus_2} \indicator\{\noise_{\idxpair} < 0\} > \frac{\constfrac\numstudent}{96}\right) = 1.\label{eq:cv_noise_total_order_case_one_hoeffding_plus}
    \end{align}
    \end{subequations}
    Denote the event
    \begin{align*}
        \event_3 \defn \left\{\sum_{\idxpairparen\in \setinterleavingvalminus_1}\indicator\{\noise_{\idxpair} > 0\} > \frac{\constfrac\numstudent}{96}\right\} \intersect \left\{\sum_{\idxpairparen\in \setinterleavingvalplus_2}\indicator\{\noise_{\idxpair} < 0\} > \frac{\constfrac\numstudent}{96}
        \right\}.
    \end{align*}
    Taking a union bound of~\eqref{eq:cv_noise_total_order_case_one_hoeffding}, we have
    \begin{align}\label{eq:cv_noise_total_order_case_one_hoeffding_union}
        \limn \prob(\event_3) = 1.
    \end{align}
    We slightly abuse the notation and denote $\estbiasinterp_\rank$ as the value of the interpolated bias on the \element of rank $\rank$. That is, we define $\estbiasinterp_\rank \defn \estbiasinterp_{\idxcourse_\rank\idxstudent_\rank}$. It can be verified that $\estbiasinterp_\rank$ is non-decreasing in $\rank$ due to the nearest-neighbor interpolation in Algorithm~\ref{alg:cv}. Hence, $\estbiasinterp_\rank \le 0$ for all $\rank\in \setinterleaving^-$ or $\estbiasinterp_\rank \ge 0$ for all $\rank\in \setinterleaving^+$.

    First consider the case $\estbiasinterp_\rank \le 0$ for all $\rank\in \setinterleavingminus$. We bound the validation error at $\reg\in \setreg_1$ as:
    \begin{align}
        \errat{\reg} & \ge \frac{1}{\sizesetval} \sum_{\idxpairparen\in \setinterleavingvalminus_1} \left(\noise_{\idxpair} - \estmeanat{\reg}_1 - \estbiasinterpat{\reg}_{\idxpair}\right)^2\\
        & \stackrel{\stepone}{\ge} \left.\frac{1}{\sizesetval}\cdot \abs*{\setinterleavingvalminus_1} \cdot \left(0 + \frac{16}{\sqrt{\constfrac}} + 0\right)^2 \stackrel{\stepone}{\ge} \frac{1}{\numstudent} \frac{\constfrac\numstudent}{96}\frac{256}{\constfrac} =\frac{8}{3},\quad\forall \reg\in \setreg_1 \;\middle|\; (\event_1, \event_2, \event_3),\right.
    \end{align}
    where~\stepone is true by~\eqref{eq:cv_noise_total_order_case_one_assumption} and the definition of $\event_3$, and~\steptwo is true by the definition of  $\event_2$. Hence, we have
    \begin{subequations}\label{eq:cv_noise_case_one_final}
    \begin{align}
        \limn\left(\errat{\reg} \ge \frac{8}{3} \quad\forall \reg\in \setreg_1, \{\estbiasinterp_\rank \le 0 \text{ for all }\rank\in \setinterleavingminus\}\right) \stackrel{\stepone}{\ge} \Prob\left(\estbiasinterp_\rank \le 0 \text{ for all }\rank\in \setinterleavingminus\right),
    \end{align}
    where~\stepone is true by~\eqref{eq:cv_noise_total_order_sum_estmean_equality}, Lemma~\ref{lem:cv_only_noise_total_order_size_half_set} and~\eqref{eq:cv_noise_total_order_case_one_hoeffding_union}. By a similar argument, we have
    \begin{align}
        \limn\left(\errat{\reg} \ge \frac{8}{3} \quad\forall \reg\in \setreg_1, \{\estbiasinterp_\rank \ge 0 \text{ for all }\rank\in \setinterleavingplus\}\right) \ge \Prob\left(\estbiasinterp_\rank \ge 0 \text{ for all }\rank\in \setinterleavingplus\right),
    \end{align}
    \end{subequations}
    Summing over~\eqref{eq:cv_noise_case_one_final}, we have
    \begin{align}\label{eq:cv_noise_total_order_case_one_final}
        \limn \Prob\left(\errat{\reg} \ge \frac{8}{3},\quad\forall \reg\in \setreg_1\right) = 1.
    \end{align}

\paragraph{Case 2: $\setreg_2 = \setreg_\errbound\intersect\left\{\reg\in [0, \infty]: \estmeanat{\reg}_1 > -\frac{32}{\sqrt{\constfrac}}\right\}$} 

It can be verified that due to~\eqref{eq:cv_noise_total_order_sum_estmean_equality}, we have
\begin{align}\label{eq:cv_noise_total_order_case_two_assumption}
    -\frac{32}{\sqrt{\constfrac}} < \{\estmean_1,\estmean_2\} < \frac{48}{\sqrt{\constfrac}}.
\end{align}
Similar to Case 2 in part~\ref{part:cv_noise_only_constant_fraction}, we decompose the validation error at $\reg\in \setreg_2$ as
\begin{align*}
    \errat{\reg} & = \frac{1}{\sizesetval} \sum_{\idxpairparen\in \setval} \left(\noise_{\idxpair} - \estmeanat{\reg}_\idxcourse - \estbiasinterpat{\reg}_{\idxpair}\right)^2\\
    & = \frac{1}{\sizesetval}
        \left[\vphantom{\sum_{\idxpairparen\in \setval}}\right.
        \underbrace{\sum_{\idxpairparen\in \setval} \noise_\idxpair^2}_{\term_1} -2 \underbrace{\sum_{\idxpairparen\in \setval} \noise_\idxpair \estmeanat{\reg}_\idxcourse}_{\term_2} - 2
        \underbrace{\sum_{\idxpairparen} \noise_\idxpair \estbiasinterpat{\reg}_\idxpair}_{\term_3} + 
        \underbrace{\sum_{\idxpairparen} \left(\estmeanat{\reg}_\idxcourse+\estbiasinterpat{\reg}_\idxpair\right)^2}_{\term_4}
        \left.\vphantom{\sum_{\idxpairparen\in \setval}}\right].
\end{align*}
Given that $\norminf{\estmean}$ is bounded by a constant by~\eqref{eq:cv_noise_total_order_case_two_assumption}, the analysis of the terms $\term_1, \term_2$ and $\term_3$ follows the proof in part~\ref{part:cv_noise_only_constant_fraction}. We have
\begin{subequations}\label{eq:cv_only_noise_total_order_case_two_terms}
\begin{align}
    & \limn \prob\left(\frac{1}{\sizesetval} \term_1 > 1-\errbound_2\right) = 1.\\
    & \limn \prob\left(\frac{1}{\sizesetval} \sum_{\idxpairparen\in \setval} \abs*{\term_2} < \frac{96}{\sqrt{\constfrac}}\errbound_2 \right) = 1.\\
    & \limn \prob\left(\frac{1}{\sizesetval} \abs*{\term_3} < \errbound_2\right) = 1.
\end{align}
\end{subequations}
Now we consider the last term $\term_4$. Recall from~\eqref{eq:cv_only_noise_const_frac_assumption_mean_diff} that 
\begin{align*}
    \left.\abs*{\estmean_2 - \estmean_1} > \frac{\errbound}{\sqrt{2}} \quad\forall \reg\in \setreg_2 \;\middle|\; \event.\right.
\end{align*}
First consider the case of $\setreg_{2>1}\defn \left\{\reg\in [0, \infty]: \estmeanat{\reg}_2- \estmeanat{\reg}_1 > \frac{\errbound}{\sqrt{2}}\right\}$. Consider any $(\rank, \rank')\in \setinterleavingpairs$. By the definition of $\setinterleavingpairs$ we have $\rank < \rankalt$. Hence, we have $\estbiasinterp_\rank \le \estbiasinterp_{\rankalt}$ due to the nearest-neighbor interpolation in Algorithm~\ref{alg:cv}. Hence, we have $\estmean_2 +\estbiasinterp_{\rankalt} - (\estmean_1 + \estbiasinterp_\rank) > \frac{\errbound}{\sqrt{2}}$ and consequently
\begin{align*}
        \left.(\estmean_1 + \estbiasinterp_\rank)^2 + (\estmean_2 +\estbiasinterp_{\rankalt})^2 > \frac{\epsilon^2}{8}\quad\forall\reg\in \setreg_2 \intersect\setreg_{2>1} \;\middle|\; \event.\right.
\end{align*}
We bound the term $\term_4$ as:
\begin{subequations}\label{eq:cv_noise_total_order_case_two_term_four_bound_conditional}
\begin{align}
    \frac{1}{\sizesetval}\term_4 & \ge \frac{1}{\sizesetval}\sum_{(\rank, \rankalt)\in \setinterleavingpairs} \left[(\estmean_1 + \estbiasinterp_\rank)^2 + (\estmean_2 +\estbiasinterp_{\rankalt})^2\right] \nonumber\\
    & \stackrel{\stepone}{\ge} \left.\frac{1}{2\numstudent}\cdot\frac{\constfrac\numstudent}{36}\cdot\frac{\errbound^2}{8} = \frac{\constfrac\errbound^2}{576} \qquad\forall \reg\in \setreg_2  \intersect \setreg_{2>1} \;\middle|\; (\event_{\frac{1}{36}}, \event),\right.
\end{align}
where inequality~\stepone is true by the definition of $\event_{\frac{1}{36}}$.
Define $\setreg_{1>2}\defn \left\{\reg\in [0, \infty]: \estmeanat{\reg}_1 - \estmeanat{\reg}_2 >\frac{\errbound}{\sqrt{2}}\right\}$. With a similar argument, we have
\begin{align}
    \left.\frac{1}{\sizesetval}\term_4 \ge \frac{\constfrac\errbound^2}{576},\quad\forall\reg\in \setreg_2 \intersect\setreg_{1>2} \;\middle|\; (\event'_{\frac{1}{36}}, \event).\right.
\end{align}
\end{subequations}
Combining~\eqref{eq:cv_noise_total_order_case_two_term_four_bound_conditional}, we have
\begin{align*}
    \left.\frac{1}{\sizesetval}\term_4 \ge \frac{\constfrac\errbound^2}{576},\quad\forall\reg\in \setreg_2  \;\middle|\; (\event_{\frac{1}{36}}, \event'_{\frac{1}{36}}, \event).\right.
\end{align*}
By Lemma~\ref{lem:cv_only_noise_total_order_size_set_interleaving} and~\eqref{eq:cv_only_noise_const_frac_assumption_mean_diff}, we have
\begin{align}\label{eq:cv_only_noise_total_order_case_two_term_four}
    \limn\prob\left(\frac{1}{\sizesetval}\term_4 \ge \frac{\constfrac\errbound^2}{576},\quad\forall\reg\in \setreg_2\right) \ge \limn \prob\left( \event_{\frac{1}{36}}, \event'_{\frac{1}{36}}, \event\right) =1.
\end{align}
\paragraph{Putting things together:} Combining the four terms from~\eqref{eq:cv_only_noise_total_order_case_two_terms} and~\eqref{eq:cv_only_noise_total_order_case_two_term_four}, we have
\begin{align}\label{eq:cv_noise_total_order_case_two_final}
    \limn \prob\left(\errat{\reg} > 1- \errbound_2 -\frac{128}{\sqrt{\constfrac}}\errbound_2 - 2\errbound_2 + \frac{\constfrac\errbound^2}{576} = 1 - \left(3 + \frac{128}{\sqrt{\constfrac}}\right)\errbound_2 + \frac{\constfrac\errbound^2}{576}, \quad\forall \reg\in\setreg_2\right) = 1.
\end{align}
Combining the two cases from~\eqref{eq:cv_noise_total_order_case_one_final} and~\eqref{eq:cv_noise_total_order_case_two_final}, we have
\begin{align}\label{eq:cv_noise_total_order_err_at_reg_final}
    \limn \prob\left(\errat{\reg} > \frac{8}{3}\minstat \left[ 1 - \left(3 + \frac{128}{\sqrt{\constfrac}}\right)\errbound_2 + \frac{\constfrac\errbound^2}{576}\right],\quad\forall\reg\in\setreg_2\right) = 1.
\end{align}
Recall from~\eqref{eq:cv_noise_err_at_infty_bound} that the validation error at $\reg=\infty$ is bounded as (taking $\numcourse=2$):
\begin{align}\label{eq:cv_noise_total_order_err_at_infty_bound_recall}
    \lim_{\numstudent\rightarrow \infty} \Prob\left(\errat{\infty}\le  1 + \errbound_1 + 3\errbound_1^2 ,\quad\forall\reg\in \setregoutside\right) = 1.
\end{align}
Combining~\eqref{eq:cv_noise_total_order_err_at_reg_final} and~\eqref{eq:cv_noise_total_order_err_at_infty_bound_recall} with choices of $(\errbound_1, \errbound_2)$ (dependent on $\errbound, \constfrac$) such that $\frac{8}{3}\minstat \left[ 1 - \left(3 + \frac{128}{\sqrt{\constfrac}}\right)\errbound_2 + \frac{\constfrac\errbound^2}{576}\right] >  1 + \errbound_1 + 3\errbound_1^2$, we have
    \begin{align*}
        \limn \prob\left(\errat{\infty} > \errat{0},\quad \forall \reg\in\setregoutside\right) = 1,
    \end{align*}
    completing the proof.

\subsection{Proof of Proposition~\ref{prop:mean_not_consistent_comparision}}\label{app:proof_prop_mean_not_consistent}

To prove the claimed result, we construct partial orderings that satisfy each of the conditions~\ref{part:mean_constant_fraction},~\ref{part:mean_binary}, and~\ref{part:mean_total_order} separately, and show that the mean estimator fails under each construction. 
Intuitively, the mean estimator does not account for any bias, so we construct partial orderings where the mean of the bias differs significantly across courses, and show that the mean estimator fails on these construction. Without loss of generality we assume that the standard deviation parameter for the Gaussian distribution of the bias is $\gaussianwidthbias=1$. 

\subsubsection{Proof of part~\ref{part:mean_constant_fraction}}

We first construct a partial ordering that satisfies the condition~\ref{part:mean_constant_fraction}, and then bound the mean of each course to derive the claimed result. For clarity of notation, we denote the constant in the all constant-fraction assumption as $\constfrac$. 
 
\paragraph{Constructing the partial ordering:}
Recall from Definition~\ref{def:all_constant_fraction} that the all $\constfrac$-fraction assumption requires that each course $\idxcoursescope$ has at least $\blocklength_{\idxcourse\idxgroup} \ge \constfrac\numstudent$ students in each group $\idxgroup \in [\numgroup]$. Let $\const_0 = 1-\constfrac\numgroup$. Due to the assumption that $\constfrac \in (0, \frac{1}{\numgroup})$, we have that $\const_0 > 0$ is a constant. We construct the following group ordering $\setpartialorder$, where the number of students in each course from each group is specified as
 \begin{itemize}
     \item \textbf{Course 1:}  The course has $(\constfrac+\const_0)\numstudent$ students from group $1$, and $\constfrac \numstudent$ students from each remaining group $\idxgroup \in \{2, \ldots, \numgroup\}$. That is,
     \begin{subequations}\label{eq:mean_construction}
     \begin{align}
         \blocklength_{1\idxgroup} = 
         \begin{cases}
            (\constfrac + \const_0) \numstudent & \text{if }\idxgroup=1\\
            \constfrac\numstudent & \text{if }2\le \idxgroup\le\numgroup.
         \end{cases}
     \end{align}
      
     \item \textbf{Course 2:} The course has $( \constfrac+\const_0 ) \numstudent$ students from group $\numgroup$, and $\constfrac \numstudent$ students from each remaining group $\idxgroup \in [\numgroup-1]$. That is,
     \begin{align}
        \blocklength_{2\idxgroup} = \begin{cases}
         (\constfrac + \const_0)\numstudent  & \text{if }1 \le \idxgroup\le \numgroup-1\\
         \constfrac\numstudent. & \text{if } \idxgroup=\numgroup.
         \end{cases}
     \end{align}
     \item \textbf{Course $\idxcourse \ge 3$}:
     The course has an equal number of students from each group $\idxgroupscope$. That is, for every $3\le \idxcourse\le \numcourse$,
     \begin{align*}
         \blocklength_{\idxcourse\idxgroup} = \frac{\numstudent}{\numgroup}\qquad \forall \idxgroupscope.
     \end{align*}
    \end{subequations}
 \end{itemize}
It can be seen that this construction of the group ordering $\setpartialorder$ is valid, satisfying the equality $\sum_{\idxgroupscope}\blocklength_{\idxcourse\idxgroup} = \numstudent$ for each $\idxcoursescope$. Moreover, the group ordering $\setpartialorder$ satisfies the all $\constfrac$-fraction assumption.
Intuitively, course $1$ contains more students associated with negative bias (from group $1$), and course $2$ contains more students associated with positive bias (from group $\idxgroup$). The mean estimator underestimates the quality of course $1$, and overestimates the quality of course $2$. We construct some true qualities $\truemean$ with $\truemean_1 > \truemean_2$, whose values are specified later in the proof.

\paragraph{Bounding the mean of each course:}
Denote the mean of the bias in any course $\idxcourse\in \{1, 2\}$ of group $\idxgroupscope$ as $\biasmeancoursegroup{\idxcourse}{\idxgroup}\defn \frac{1}{\blocklengthcoursegroup{\idxcourse}{\idxgroup}} \sum_{\idxstudent\in \setcoursegroup{\idxcourse}{\idxgroup}}\bias_\idxpair$. Similar to the proof of Lemma~\ref{lem:cv_only_bias_const_frac_hoeffding} (see Appendix~\ref{app:proof_consistency_const_frac} for its statement and Appendix~\ref{app:proof_lem_cv_only_bias_const_frac_hoeffding} for its proof), due to assumptions~\ref{assumption:bias} and~\ref{assumption:d} we establish the following lemma.
\begin{lemma}\label{lem:mean_hoeffding}
    Consider any group ordering $\setpartialorder$ that satisfies the all $\constfrac$-fraction assumption. For any $\errbound>0$, we have
    \begin{align*}
        \limn \Probbig{\underbrace{
            \abs*{\biasmeancoursegroup{\idxcourse}{\idxgroup} - \meanbiasgroup{\idxgroup}} < \errbound, \quad \forall \idxcoursescope, \idxgroupscope}_{\event_1}} = 1.
    \end{align*}
\end{lemma}
Denote this event in Lemma~\ref{lem:mean_hoeffding} as $\event_1.$
Recall that $\blocklengthgroup{\idxgroup}$ denotes the number of students in each group $\idxgroupscope$. From the construction of the group ordering $\setpartialorder$, we have $\blocklength_0\defn \blocklengthgroup{1} = \blocklengthgroup{\numgroup} = (2\constfrac+\const_0 + \frac{\numcourse-2}{\numgroup})\numstudent$.
Recall that $\bias^{(\idxrank)}$ denotes the $\idxrank^\thcount$ order statistics of $\{\bias_{\idxpair}\}_{\idxcoursescope, \idxstudentscope}$. By the assumption~\ref{assumption:bias} of the bias and the construction of the partial ordering $\setpartialorder$, the group $1$ contains the $\blocklengthgroup{1}$ lowest bias terms, $\{\bias^{(1)}, \ldots, \bias^{(\blocklength_0)}\}$, and the group $\numgroup$ contains the $\blocklengthgroup{\numgroup}$ highest bias terms, $\{\bias^{(\numcourse\numstudent-\blocklength_0 + 1)}, \ldots, \bias^{(\numcourse\numstudent)}\}$. Hence, we have
\begin{align*}
    & \meanbiasgroup{1} < \frac{\bias^{(\frac{\blocklength_0}{2})} + \biasorder{\blocklength_0}}{2}\\
    & \meanbiasgroup{\numgroup} > \frac{
        \biasorder{\numcourse\numstudent - \blocklength_0} + \biasorder{\numcourse\numstudent-\frac{\blocklength_0}{2}}
    }{2}.
\end{align*}
By the convergence of the order statistics from Lemma~\ref{lem:order_stats_consistent}, it can be shown that there exists some constant $\const > 0$ (dependent on $\numcourse, \numgroup$ and $\constfrac $), such that 
\begin{align}\label{eq:mean_group_diff}
    \limn \Probbig{\underbrace{\meanbiasgroup{\numgroup} - \meanbiasgroup{1} > \const}_{\event_2}} = 1.
\end{align}
Denote this event in~\eqref{eq:mean_group_diff} as $\event_2$.
The mean estimator is computed as
\begin{subequations}\label{eq:mean_expression_one_two}
\begin{align}
    [\estmeanmean]_1 =\truemean_1+ \frac{1}{\numstudent}\sum_{\idxgroupscope} \blocklengthcoursegroup{1}{\idxgroup}\biasmeancoursegroup{1}{\idxgroup}\\
     [\estmeanmean]_2 =\truemean_2 + \frac{1}{\numstudent}\sum_{\idxgroupscope} \blocklengthcoursegroup{2}{\idxgroup}\biasmeancoursegroup{2}{\idxgroup}
\end{align}
\end{subequations}
Taking the difference on~\eqref{eq:mean_group_diff}, conditional on $\event_1$ and $\event_2$,
\begin{align}
    [\estmeanmean]_2 - [\estmeanmean]_1 & = (\truemean_2 - \truemean_1) + \frac{1}{\numstudent}\sum_{\idxgroupscope}(\blocklengthcoursegroup{2}{\idxgroup}\biasmeancoursegroup{2}{\idxgroup} - \blocklengthcoursegroup{1}{\idxgroup}\biasmeancoursegroup{1}{\idxgroup})\nonumber\\
    &\stackrel{\stepone}{>} (\truemean_2 - \truemean_1) + \frac{1}{\numstudent}\sum_{\idxgroupscope}(\blocklengthcoursegroup{2}{\idxgroup}\meanbiasgroup{\idxgroup} - \blocklengthcoursegroup{1}{\idxgroup}\meanbiasgroup{\idxgroup})- 2\errbound\nonumber\\
    & \stackrel{\stepone}{=}(\truemean_2 - \truemean_1) + \const_0 (\biasgroup{\numgroup}-\meanbiasgroup{1})-2\errbound\nonumber\\
    & \stackrel{\stepthree}{>} (\truemean_2 - \truemean_1) + \const_0\const-2\errbound.\label{eq:mean_const_frac_mean_diff}
\end{align}
where inequality~\stepone is true by the event $\event_1$, and equality~\stepone is true by plugging in the construction of the group ordering from~\eqref{eq:mean_construction}, and inequality~\stepthree is true by the definition~\eqref{eq:mean_group_diff} of $\event_2$. We set $\errbound = \frac{\const_0\const}{4}$, and set $\truemean_1 = \frac{\const_0\const}{2}$ and $\truemean_2 = 0$. Then by~\eqref{eq:mean_const_frac_mean_diff} we have
\begin{align}\label{eq:mean_const_frac_sign}
    \Prob(\left[\estmeanmean]_2 - [\estmeanmean]_1 > 0\right) = 1.
\end{align}
Combining~\eqref{eq:mean_const_frac_sign} with the fact that $\truemean_2 - \truemean_1 < 0$, completing the proof of part~\ref{part:mean_constant_fraction}.

\subsubsection{Proof of part~\ref{part:mean_binary}}
To construct the partial ordering, we set $\numgroup = 2$ and $\numcourse=2$ in construction we used for part~\ref{part:mean_constant_fraction}. This completes the proof of part~\ref{part:mean_binary}.

\subsubsection{Proof of part~\ref{part:mean_total_order}}
We construct a total ordering where the bias obeys the following order (same as the ``non-interleaving'' total ordering described in Section~\ref{sec:exp_vary_n}):
\begin{align*}
    \bias_{11} \le \ldots \le \bias_{1\numstudent} \le \bias_{21} \le \ldots \le \bias_{2\numstudent}\le \ldots \le \bias_{\numcourse1}\le \ldots \le \bias_{\numcourse\numstudent}.
\end{align*}
In this construction, course $1$ contains the $\numstudent$ students with the lowest bias, and course $\numcourse$ contains the $\numstudent$ students with the highest bias. Recall that $\biascourse{\idxcourse}$ denotes the mean of the bias in course $\idxcoursescope$. We have
\begin{align*}
    & \biascourse{1} = \frac{1}{\numstudent} \sum_{\idxstudentscope} \bias_{1\idxstudent}< \frac{\biasorder{\frac{\numstudent}{2}} + \biasorder{\numstudent}}{2}\\
    & \biascourse{\numgroup}= \frac{1}{\numstudent} \sum_{\idxstudentscope} \bias_{2\idxstudent} > \frac{\biasorder{\numcourse\numstudent -\frac{\numstudent}{2}} + \biasorder{\numcourse\numstudent}}{2}.
\end{align*}
Similar to part~\ref{part:mean_constant_fraction}, by Lemma~\ref{lem:order_stats_consistent}, there exists a positive constant $\const > 0$ (dependent on $\numcourse$), such that
\begin{align*}
    \limn \prob\left(\biascourse{\numgroup} - \biascourse{1} > \const\right) = 1.
\end{align*}
Let $\truemean_1 = \const$ and $\truemean_2 = 0$.
We have
\begin{align}\label{eq:mean_total_order_sign}
    \limn \prob([\estmeanmean]_\numgroup - [\estmeanmean]_1 =\truemean_2 - \truemean_1 + \biascourse{2} - \biascourse{1} >0) = 1.
\end{align}
Combining~\eqref{eq:mean_total_order_sign} with the fact that $\truemean_1 > \truemean_\numgroup$ completes the proof of part~\ref{part:mean_total_order}.

\subsection{Proof of Proposition~\ref{prop:uniform_example}}\label{app:proof_prop_uniform_example}

By Corollary~\ref{cor:property_shift_invariance}, we assume $\truemean=0$ without loss of generality. Denote the bias of course 1 as $\{\unifneg_\idxstudent\}_{\idxstudent\in [\fraction\numstudent]}$ in group $1$, and $\{\unifpos_\idxstudent\}_{\idxstudent\in [(1-\fraction)\numstudent]}$ in group 2. Denote the bias of course 2 as $\{\unifneg'_\idxstudent\}_{\idxstudent\in [(1-\fraction)\numstudent]}$ in group 1 and $\{\unifpos'_\idxstudent\}_{\idxstudent\in [\fraction\numstudent]}$ in group 2. We have $\unifneg_\idxstudent, \unifneg'_\idxstudent\sample \uniform[-1, 0]$ and $\unifpos_\idxstudent, \unifpos'_\idxstudent \sample \uniform[0,1]$. 
Denote the mean of $\{\unifneg_\idxstudent\}, \{\unifpos_\idxstudent\}, \{\unifneg'_\idxstudent\}$ and $\{\unifpos'_\idxstudent\}$ as $\meanunifneg, \meanunifpos, \meanunifneg'$ and $\meanunifpos'$ respectively. We prove the claimed result respectively for the \subsampling estimator (Appendix~\ref{app:proof_uniform_subsampling}) and for our estimator at $\reg=0$ (Appendix~\ref{app:proof_uniform_our_estiamtor}). Both parts use the following standard result regarding the uniform distribution.
\begin{lemma}\label{lem:uniform_square}
    Let $\rvunif_1, \ldots, \rvunif_n$ be i.i.d. $\uniform[0, 1]$, we have 
    \begin{align*}
        \Expect\left(\frac{\sum_{i=1}^n \rvunif_i}{n}\right)^2 = \frac{1}{4} + \frac{1}{12n}.
    \end{align*}
\end{lemma}

\subsubsection{The \subsampling estimator}\label{app:proof_uniform_subsampling}

We follow the definition of the \subsampling estimator defined in Appendix~\ref{app:subsample}.
In the reweighting step, by~\eqref{eq:reweighting} we have
\begin{align}\label{eq:uniform_example_reweighting}
    \estmeansubsample = \frac{1}{2}\begin{bmatrix}
        \meanunifneg + \meanunifpos\\
       \meanunifneg' + \meanunifpos'
    \end{bmatrix}.
\end{align}
In the recentering step, by~\eqref{eq:recentering} we have
\begin{align}
    \estmeansubsample & \leftarrow \estmeansubsample +\left(- \frac{1}{2} \sum_{\idxcourse\in\{1, 2\}}[\estmeansubsample]_\idxcourse + \frac{1}{2\numstudent}\sum_{\idxcourse\in \{1, 2\}, \idxstudentscope} \obs_\idxpair \right)\vecone\nonumber\\
    & = \estmeansubsample + \left(-\frac{[\estmeansubsample]_1 + [\estmeansubsample]_2}{2} + \frac{\fraction\numstudent\meanunifneg + (1-\fraction)\numstudent\meanunifpos + (1-\fraction)\numstudent\meanunifneg' + \fraction\numstudent\meanunifpos'}{2\numstudent}\right)\vecone \nonumber\\
    & = \frac{[\estmeansubsample]_1 - [\estmeansubsample]_2}{2}\begin{bmatrix}
            1\\
            -1
        \end{bmatrix} + \left( \frac{\fraction\meanunifneg + (1-\fraction)\meanunifpos + (1-\fraction)\meanunifneg' + \fraction\meanunifpos'}{2}\right)\vecone \nonumber\\
    & \stackrel{\stepone}{=} \frac{ \meanunifneg + \meanunifpos-\meanunifneg' - \meanunifpos'}{4}\begin{bmatrix}
        1\\
        -1
    \end{bmatrix} + \left( \frac{\fraction\meanunifneg + (1-\fraction)\meanunifpos + (1-\fraction)\meanunifneg' + \fraction\meanunifpos'}{2}\right)\vecone,\label{eq:uniform_recentering}
\end{align}
where equality~\stepone is true by plugging in~\eqref{eq:uniform_example_reweighting} from the reweighting step.
By symmetry, we have $\Expect[\estmeansubsample]_1^2 = \Expect[\estmeansubsample]_2^2$, so we only consider course $1$. By~\eqref{eq:uniform_recentering}, we have
\begin{align*}
    \Expect[\estmeansubsample]_1^2 & \stackrel{\stepone}{=} \Expect\left(\frac{\meanunifneg +\meanunifpos - \meanunifneg'- \meanunifpos' }{4}\right)^2 + \Expect\left(\frac{\fraction\meanunifneg + (1-\fraction)\meanunifpos + (1-\fraction)\meanunifneg' + \fraction\meanunifpos'}{2}\right)^2\\
    & {=} \frac{1}{16} \Expect\left[\meanunifneg'^2 + \meanunifpos'^2 + \meanunifneg^2 + \meanunifpos^2 -4\cdot\frac{1}{2}\frac{1}{2}\right]\\
    &\qquad\qquad+ \frac{1}{4} \Expect\left[(1-\fraction)^2\meanunifneg'^2 + \fraction^2\meanunifpos'^2 + \fraction^2\meanunifneg^2 + (1-\fraction)^2\meanunifpos^2 - 2\left(\frac{\fraction^2}{4} + \frac{(1-\fraction)^2}{4}\right)\right]\\
    & = \frac{1}{8} \Expect\left[\meanunifneg^2 + \meanunifpos^2 - \frac{1}{2} \right] + \frac{1}{2}\Expect\left[\fraction^2  \meanunifneg^2 + (1-\fraction)^2 \meanunifpos^2 - \frac{\fraction^2 + (1-\fraction)^2}{4}\right]\\
    & \stackrel{\steptwo}{=} \frac{1}{8}\left[\frac{1}{4} + \frac{1}{12\fraction\numstudent} + \frac{1}{4} + \frac{1}{12(1-\fraction)\numstudent} -\frac{1}{2}\right] + \frac{1}{2}\Expect\left[\frac{\fraction^2}{4} + \frac{\fraction^2}{12\fraction\numstudent} + \frac{(1-\fraction)^2}{4} + \frac{(1-\fraction)^2}{12(1-\fraction)\numstudent} - \frac{\fraction^2 + (1-\fraction)^2}{4}\right]\\
    & =\frac{1}{96\numstudent}\left(\frac{1}{\fraction} + \frac{1}{1-\fraction}\right) + \frac{1}{24\numstudent}\\
    & = \frac{1}{24\numstudent} + \frac{1}{96\fraction(1-\fraction)\numstudent}.
\end{align*}
where~\stepone is true because it can be verified by algebra that $\Expect\left[
    \left(\frac{\meanunifneg + \meanunifpos-\meanunifneg' - \meanunifpos' }{4}\right)
    \left(\frac{\fraction\meanunifneg + (1-\fraction)\meanunifpos + (1-\fraction)\meanunifneg' + \fraction\meanunifpos'}{2}\right)  \right] = 0$, and~\steptwo is true by Lemma~\ref{lem:uniform_square}. Finally, we have
    \begin{align*}
        \frac{1}{2}\Expect\norm*{\estmeansubsample}]_2^2 = \frac{1}{2}\left(\Expect[\estmeansubsample]_1^2] + \Expect[\estmeansubsample]_2^2\right) =\Expect[\estmeansubsample]_1^2
        =\frac{1}{24\numstudent} + \frac{1}{96\fraction(1-\fraction)\numstudent}\ge \frac{1}{24\numstudent} + \frac{1}{24\numstudent} = \frac{1}{12\numstudent},
    \end{align*}
    where the inequality holds because $\fraction(1-\fraction) \le \frac{1}{4}$ for every $\fraction\in (0, 1)$.

\subsubsection{Our estimator at \texorpdfstring{$\reg = 0$}{\unichar{"03BB}=0}}\label{app:proof_uniform_our_estiamtor}

Recall from Proposition~\ref{prop:closed_form_solution} that for $\numcourse=2$ courses and $\numgroup=2$ groups, our estimator at $\reg=0$ has the closed-form expression
$\estmeanat{0} = \meanobs + \begin{bmatrix}-1\\1\end{bmatrix}\cdot  \frac{\margin}{2}$, where \begin{align}\label{eq:uniform_closed_form_r}
        \margin = \begin{cases}
            \obscoursegroupmin{2}{2} - \obscoursegroupmax{1}{1} & \text{if } \obscoursegroupmin{2}{2} - \obscoursegroupmax{1}{1} < \meanobs_2 - \meanobs_1\\
            \obscoursegroupmax{2}{1} - \obscoursegroupmin{1}{2} & \text{if } \obscoursegroupmax{2}{1} - \obscoursegroupmin{1}{2} > \meanobs_2 - \meanobs_1\\
            \meanobs_2 - \meanobs_1 & \text{o.w.}
        \end{cases}
\end{align}
By~\eqref{eq:uniform_closed_form_r}, we have
\begin{align}\label{eq:uniform_err}
    \frac{1}{2}\Expect\normtwo{\estmeanat{0}}^2 = \frac{1}{2}\Expect\left[\left(\meanobs - \frac{\margin}{2}\right)^2 + \left(\meanobs + \frac{\margin}{2}\right)^2\right] = \Expect[\meanobs^2] + \frac{1}{4}\Expect[\margin^2].
\end{align}
We analyze the two terms in~\eqref{eq:uniform_err} separately.

\paragraph{Term of $\Expect[\meanobs^2]$}:
For ease of notation, we denote the random variables
\begin{align*}
    & \{\tildeu_\idxstudent\}_{\idxstudent\in[\numstudent]} \defn  \{U_\idxstudent\}_{\idxstudent\in [\fraction\numstudent]}\union \{U'_\idxstudent\}_{\idxstudent\in [(1-\fraction)\numstudent]}\\
    & \{\tildev_\idxstudent\}_{\idxstudent\in[\numstudent]} \defn \{V_\idxstudent\}_{\idxstudent\in [(1-\fraction)\numstudent]}\union \{V'_\idxstudent\}_{\idxstudent\in [\fraction\numstudent]}
\end{align*}
Then $\{\tildeu_\idxstudent\}_{\idxstudent\in[\numstudent]}$ is i.i.d. $\uniform[-1, 0]$ and $\{\tildev_\idxstudent\}_{\idxstudent\in[\numstudent]}$ is i.i.d. $\uniform[0, 1]$.
We have
\begin{align}
    \Expect[\meanobs^2] & = \Expect\left(\frac{\sum_{\idxgeneral\in[\numstudent]} \tildeu_i + \sum_{\idxgeneral\in[\numstudent]} \tildev_i}{2\numstudent}\right)^2\nonumber\\
    & = \frac{1}{4\numstudent^2}\Expect\left[\sum_{\idxgeneral\in[\numstudent]} \tildeu_i^2 + \sum_{\idxgeneral\in[\numstudent]} \tildev_i^2 + 2\sum_{\idxgeneral\in[\numstudent],\idxgeneralalt\in [\numstudent]}\tildeu_\idxgeneral \tildev_\idxgeneralalt + \sum_{\idxgeneral\in [\numstudent]} \sum_{\idxgeneralalt\ne \idxgeneral} \tildeu_\idxgeneral \tildeu_\idxgeneralalt + \sum_{\idxgeneral\in [\numstudent]} \sum_{\idxgeneralalt\ne \idxgeneral} \tildev_\idxgeneral \tildev_\idxgeneralalt\right]\nonumber\\
    & = \frac{1}{4\numstudent^2}\left[\frac{\numstudent}{3} + \frac{\numstudent}{3} + 2 \numstudent^2 \left(-\frac{1}{4}\right) + \numstudent(\numstudent-1) \frac{1}{4} + \numstudent(\numstudent-1) \frac{1}{4}\right]\nonumber\\
    & = \frac{1}{24\numstudent}.\label{eq:uniform_term_one}
\end{align}

\paragraph{Term of $\Expect[\margin^2]$:} To analyze the term $\Expect[\margin^2]$, we use the following standard result from statistics. 
\begin{lemma}\label{lem:uniform_extremal}
    Let $\rvunif_1, \ldots, \rvunif_n\sample \uniform[0, 1]$. Let $\rvunifmin = \min_{i\in [n]} \rvunif_i$. We have
    \begin{align*}
        & \Expect[\rvunifmin] = \frac{1}{n+1}\\
        & \Expect[\rvunifmin^2] = \frac{2}{(n+1)(n+2)}.
    \end{align*}
\end{lemma}

We define 
\begin{align*}
    & \unifnegmax \defn \max_{\idxstudent\in [\fraction\numstudent]} \unifneg_\idxstudent\\
    & \unifposmin \defn \min_{\idxstudent\in [(1-\fraction)\numstudent]} \unifpos_\idxstudent,
\end{align*}
and define $\unifnegmax'$ and $\unifposmin'$ likewise.
By~\eqref{eq:uniform_closed_form_r} it can be verified that we have the deterministic relation
\begin{align*}
    \abs*{\margin} & \le (\obscoursegroupmin{2}{2} - \obscoursegroupmax{1}{1})\maxstat(\obscoursegroupmin{1}{2} - \obscoursegroupmax{2}{1}) \\
    & \stackrel{\stepone}{=} (\unifposmin' - \unifnegmax)\maxstat(\unifposmin - \unifnegmax') \\
    & \le \unifposmin' - \unifnegmax+\unifposmin - \unifnegmax',
\end{align*}
where equality~\stepone is true by the assumption that there is no noise and the assumption of $\truemean = 0$. Therefore,
\begin{align}
    \Expect[\margin^2] & \le \Expect\left[ (\unifposmin'- \unifnegmax) + (\unifposmin- \unifnegmax')\right]^2 \nonumber\\
    & =\underbrace{
        \Expect(\unifposmin'-\unifnegmax)^2}_{\term_1} +
        \underbrace{\Expect(\unifposmin-\unifnegmax')^2}_{\term_2} +
        2\underbrace{\Expect(\unifposmin'-\unifnegmax)(\unifposmin-\unifnegmax')}_{\term_3}.\label{eq:uniform_expect_r}
\end{align}
We consider the three terms $\term_1, \term_2$ and $\term_3$ separately. For the term $\term_1$, by Lemma~\ref{lem:uniform_extremal} we have
\begin{align*}
    \term_1 & =
     \Expect [\unifposmin']^2 + \Expect[\unifnegmax^2] - 2\Expect[ \unifposmin'\unifnegmax]\\
     & =2\cdot \frac{2}{(\fraction\numstudent+1)(\fraction\numstudent+2)} + 2 \cdot\frac{1}{(\fraction\numstudent+1)^2} \le\frac{6}{\fraction^2\numstudent^2}.
    \end{align*}
Likewise, for the term $\term_2$ we have
\begin{align*}
    \term_2 \le \frac{6}{(1-\fraction)^2\numstudent^2}.
\end{align*}
For the term $\term_3$, by Lemma~\ref{lem:uniform_extremal} we have
\begin{align*}
    \term_3 = \frac{2}{\fraction\numstudent+1} \cdot \frac{2}{(1-\fraction)\numstudent+1} \le \frac{4}{\fraction(1-\fraction)\numstudent^2}.
\end{align*}
Plugging the three terms back to~\eqref{eq:uniform_expect_r}, we have
\begin{align}\label{eq:uniform_term_two}
    \Expect[\margin^2] \le \frac{6}{\fraction^2\numstudent^2} + \frac{6}{(1-\fraction)^2\numstudent^2} + \frac{8}{\fraction(1-\fraction)\numstudent^2} = \frac{\const}{\numstudent^2},
\end{align}
for some constant $\const > 0$.

Finally, plugging~\eqref{eq:uniform_term_one} and~\eqref{eq:uniform_term_two} back to~\eqref{eq:uniform_err}, we have
\begin{align*}
    \frac{1}{2}\Expect\normtwo{\estmeanat{0}} \le \frac{1}{24\numstudent} + \frac{\const}{4\numstudent^2},
\end{align*}
completing the proof.

\subsection{Proof of preliminaries}
In this section, we present the proofs of the preliminary results presented in Appendix~\ref{app:preliminaries}.

\subsubsection{Proof of Proposition~\ref{prop:uniqueness}}\label{app:proof_prop_uniqueness}
    To avoid clutter of notation, we first prove the case for $\set=[\numcourse]\times [\numstudent]$, and then comment on the general case of $\setscope$. 
    
    Now consider $\set=[\numcourse]\times [\numstudent]$, where our estimator~\eqref{eq:optimization_restricted_general} reduces to~\eqref{eq:optimization}. We separately consider the cases of $\reg=0$ and $\reg\in (0, \infty)$.
    
    \paragraph{Case of $\reg=0$ } The objective~\eqref{eq:optimization} becomes
    \begin{align}\label{eq:properties_optimization_rewrite}
        \min_{ \meancourse\in \reals^\numcourse}
    \min_{\substack{
        \biasmtx\in \reals^{\numcourse\times \numstudent}\\
        \biasmtx \text{ satisfies } \setpartialorder}} 
    \norm*{\obsmtx - \meancourse\vecone^T - \biasmtx}_F^2 = \min_{\substack{\meanplusbiasmtx\in \reals^{\numcourse\times \numstudent} \\ \meanplusbiasmtx\in \{\meancourse \vecone^T + \biasmtx \given \meancourse\in \reals^{\numcourse},\, \biasmtx\in \reals^{\numcourse\times \numstudent},\, \biasmtx \text{ satisfies }\setpartialorder\}}} \norm*{\obsmtx - \meanplusbiasmtx}_F^2.
    \end{align}
    It can be verified that the set $\{ \meancourse \vecone^T + \biasmtx \given \meancourse\in \reals^{\numcourse},\, \biasmtx\in \reals^{\numcourse\times \numstudent},\, \biasmtx \text{ satisfies }\setpartialorder\}$ is a closed convex set. 
    By the Projection Theorem~\cite[Proposition 1.1.9]{bertsekas2009convex}, a unique minimizer $\meanplusbiasmtxopt$ to the RHS of~\eqref{eq:properties_optimization_rewrite} exists. Therefore, the set of minimizers to the LHS of~\eqref{eq:properties_optimization_rewrite} can be written as $\{(\meancourse, \meanplusbiasmtxopt - \meancourse\vecone^T) \given \meancourse\in \reals^\numcourse\}$. The tie-breaking rule minimizes the Frobenius norm $\norm*{\biasmtx}_F^2$. That is, we solve
    \begin{align}\label{eq:properties_obj_minimal_norm}
        \min_{\meancourse\in \reals^\numcourse} \norm*{\meanplusbiasmtxopt - \meancourse\vecone^T}_F^2.
    \end{align}
    It can be verified that a unique solution to~\eqref{eq:properties_obj_minimal_norm} exists, because the objective is quadratic in $\meancourse$. Hence, the tie-breaking rule defines a unique solution $(\meancourse, \biasmtx)$.
    
    \paragraph{Case of $\reg\in (0, \infty)$}
    
    It can be verified that the objective~\eqref{eq:optimization} is strictly convex in $(\meancourse,\biasmtx)$.
    Therefore, there exists at most one minimizer~\cite[Proposition 3.1.1]{bertsekas2009convex}. 
    
    It remains to prove that there exists a minimizer. It is straightforward to see that the objective is continuous in $(\meancourse, \biasmtx)$. We now prove that the objective is coercive on $\{(\meancourse, \biasmtx): \meancourse\in \reals^{\numcourse}, \biasmtx\in \reals^{\numcourse\times \numstudent}, \biasmtx\text{ satisfies } \setpartialorder\}$. That is, for any constant $\constcoercive >0$, there exists a constant $\normcoercive >0$, such that the objective at $(\meancourse, \biasmtx)$ is greater than $\constcoercive$ for all $(\meancourse, \biasmtx)$ in the domain $\{(\meancourse, \biasmtx): \meancourse\in \reals^{\numcourse}, \biasmtx\in \reals^{\numcourse\times \numstudent}, \biasmtx\text{ satisfies } \setpartialorder\}$ with 
    \begin{align}\label{eq:coercive_assumption}
        \normtwo{\meancourse}^2 + \normfro{\biasmtx}^2 > \normcoercive
    \end{align}
    Given coercivity, invoking Weierstrass’ Theorem~\cite[Proposition 3.2.1]{bertsekas2009convex} completes the proof. 
    
    We set 
    \begin{align}\label{eq:coercive_set}
        \normcoercive = \numcourse\left[\left(1 + \frac{1}{\sqrt{\reg}}\right) \sqrt{\constcoercive} + \max_{\idxcoursescope, \idxstudentscope}{\obsmtx} \right]^2 + \frac{1}{\reg} \constcoercive.
    \end{align}
    We discuss the following two cases depending on the value of $\norm*{\biasmtx}_F^2$.

    \paragraph{Case of $\norm*{\biasmtx}_F^2 \ge \frac{\constcoercive}{\reg}$} The second term of the objective~\eqref{eq:optimization_restricted_general} is lower-bounded as $\reg\norm*{\biasmtx}_F^2 \ge \constcoercive$. Hence, the objective~\eqref{eq:optimization} is at least $\constcoercive$.
    
    \paragraph{Case of $\norm*{\biasmtx}_F^2 < \frac{\constcoercive}{\reg}$:} 
    Combining~\eqref{eq:coercive_assumption} and~\eqref{eq:coercive_set}, we have \begin{align*}
        \normtwo{\meancourse}^2 > \normcoercive - \normfro{\biasmtx}^2 > \numcourse\left[(1 + \frac{1}{\sqrt{\reg}}) \sqrt{\constcoercive} + \max_{\idxcoursescope, \idxstudentscope}\obs_\idxpair \right]^2.
    \end{align*}
    Hence, there exists some $\idxcourse^*\in [\numcourse]$ such that 
    \begin{align}
        \abs*{\meancourse_{\idxcourse^*}} > (1 + \frac{1}{\sqrt{\reg}}) \sqrt{\constcoercive} + \max_{\idxcoursescope, \idxstudentscope}\obs_\idxpair.\label{eq:coercive_bound_x}
    \end{align}
    Consider the $(\idxcourse^*, \idxstudent)$ entry in the matrix $(\obsmtx - \meancourse\vecone^T - \biasmtx)$ for any $\idxstudent\in [\numstudent]$. We have
    \begin{align*}
        \abs*{(\obsmtx - \meancourse\vecone^T - \biasmtx)_{\idxcourse^*\idxstudent}} & \ge \abs*{\meancourse_{\idxcourse^*}} -\abs*{\obs_{\idxcourse^*\idxstudent}} - \abs*{\bias_{\idxcourse^*\idxstudent}}\\
        & \ge \abs*{\meancourse_{\idxcourse^*}} - \max_{\idxcoursescope, \idxstudentscope}\obs_\idxpair - \norm*{\biasmtx}_F\\
        & \stackrel{\stepone}{>} \left(1 + \frac{1}{\sqrt{\reg}}\right)\sqrt{\constcoercive} -\sqrt{\frac{\constcoercive}{\reg}} = \sqrt{\constcoercive},
    \end{align*}
    where~\stepone is true by~\eqref{eq:coercive_bound_x} and the assumption of the case that $\normfro{\biasmtx}^2 < \frac{1}{\reg} \constcoercive$.
    Hence, the second term in the objective~\eqref{eq:optimization} is lower-bounded by
    \begin{align*}
        \norm*{\obsmtx - \meancourse\vecone^T - \biasmtx}_F^2 & \ge \abs*{(\obsmtx - \meancourse\vecone^T - \biasmtx)_{\idxcourse^*\idxstudent}}^2 > \constcoercive,
    \end{align*}
    and therefore the objective~\eqref{eq:optimization} is greater than $\constcoercive$.
    
    Combining the two cases depending on $\normfro{\biasmtx}^2$ completes the proof of the coercivity of the objective~\eqref{eq:optimization} in terms of $(\meancourse, \biasmtx)$. Invoking the Weierstrass’ Theorem~\cite[Proposition 3.2.1]{bertsekas2009convex} completes the proof of $\set = [\numcourse]\times [\numstudent]$.
    
    \paragraph{Extending the proof to general $\setscope$:} For general $\setscope$, by a similar argument the solution $(\estmean, \{\estbias_\idxpair\}_{\idxpairparen\in\set})$ exists and is unique. Note that the objective~\eqref{eq:optimization_restricted_general} is independent from $\{\bias_\idxpair\}_{\idxpairparen\not\in \set}$,so  we have $    \estbias_\idxpair = 0$ for each $\idxpairparen\not\in \set$. Hence, a unique solution $(\estmean, \estbiasmtx)$ to~\eqref{eq:optimization_restricted_general} exists for general $\set$.

\subsubsection{Proof of Lemma~\ref{lem:property_xi}}\label{app:proof_lem_property_xi}

    It is sufficient to prove the general version~\eqref{eq:property_xi_general_omega}. First consider $\reg=\infty$. It can be verified that the closed-form expression~\eqref{eq:properties_expression_at_infty} for the solution at $\reg=\infty$ satisfies the claimed relation~\eqref{eq:property_xi_general_omega}. 
    
    It remains to consider the case of $\reg\in [0, \infty)$.
    Given the value of the solution $\estbiasmtxat{\reg}$, we solve for $\estmeanat{\reg}$ by minimizing the first term of the objective~\eqref{eq:optimization} as
    \begin{align}\label{eq:property_xi_obj}
        \min_{\meancourse\in\reals^\numcourse } \norm{\obsmtx - \meancourse\vecone^T - \estbiasmtxat{\reg}}_F^2.
    \end{align}
    Writing out all the terms in~\eqref{eq:property_xi_obj} and completing the square yields the claimed relation~\eqref{eq:property_xi_general_omega}.

\subsubsection{Proof of Lemma~\ref{lem:estimator_sum_equality}}\label{app:proof_lem_estimator_sum_equality}

    It is sufficient to prove the general version~\eqref{eq:estimator_sum_equality_general}.
    First consider the case of $\reg=\infty$. It can be verified that the closed-form expression expressions~\eqref{eq:properties_expression_at_infty} for the solution at $\reg=\infty$ satisfies the claimed relations~\eqref{eq:estimator_sum_equality}.
    
    It remains to consider the case of $\reg\in [0, \infty)$. First we prove~\eqref{eq:estimator_sum_equality_bias_variable_size}.
    Assume for contradiction that $\sum_{(\idxcourse, \idxstudent)\in \set} \estbias_{\idxcourse\idxstudent}\ne 0$. Consider the set of alternative solutions $(\estmean_\contradictionsumbias, \estbiasmtx_\contradictionsumbias)$ parameterized by some $\contradictionsumbias\in \reals$ as 
    \begin{subequations}\label{eq:estimator_sum_equality_construct_alt}
    \begin{align}
        & \estmean_\contradictionsumbias = \estmean + \contradictionsumbias\vecone_\numcourse\\
        & \estbiasmtx_\contradictionsumbias = \estbiasmtx - \contradictionsumbias\vecone_\numcourse\vecone_\numstudent^T.
    \end{align}
    \end{subequations}
    Note that the original solution $(\estmean, \estbiasmtx)$ corresponds to $\contradictionsumbias=0$.
    
    Since $\estbiasmtx_\contradictionsumbias$ in~\eqref{eq:estimator_sum_equality_construct_alt} is obtained by subtracting all entries in the matrix by a constant $\contradictionsumbias$, the bias term $\estbias_\contradictionsumbias$ satisfies the partial ordering $\setpartialorder$ for any $\contradictionsumbias\in \reals$. Moreover, since by construction~\eqref{eq:estimator_sum_equality_construct_alt} the value of $(\estmean_\contradictionsumbias\vecone_\numcourse+\estbias_\contradictionsumbias)$ is the same for all $\contradictionsumbias\in \reals$, the first term in the objective~\eqref{eq:optimization} is equal for all $\contradictionsumbias\in \reals^\numcourse$. Now consider the second term $\norm{\estbiasmtx_\contradictionsumbias}_\set^2$. Writing out the terms in $\norm{\estbiasmtx_\contradictionsumbias}_\set^2$ and completing the square, we have  $\norm{\estbias_\contradictionsumbias}_\set^2$ is minimized at $\contradictionsumbias = \frac{1}{\abs*{\set}} \sum_{(\idxcourse, \idxstudent)\in \set} \estbias_{\idxcourse\idxstudent} \ne 0$. Contradiction to the assumption that the solution at $\contradictionsumbias = 0$ minimizes the objective, completing the proof of ~\eqref{eq:estimator_sum_equality_bias_variable_size}.
    
    Now we prove~\eqref{eq:estimator_sum_equality_mean_variable_size}. By~\eqref{eq:property_xi_general_omega} from Lemma~\ref{lem:property_xi} and summing over $\idxcourse\in [\numcourse]$, we have
    \begin{align*}
        \sum_{\idxcoursescope}\numstudent_\idxcourse \estmean_\idxcourse =\sum_{\idxcourse\in [\numcourse]} \sum_{\idxstudent \setcourse{\idxcourse}} ( \obs_{\idxcourse\idxstudent} - \estbias_{\idxcourse\idxstudent}) = \sum_{(\idxcourse, \idxstudent)\in \set} (\obs_{\idxcourse\idxstudent} - \estbias_{\idxcourse\idxstudent}) \stackrel{\stepone}{=} \sum_{(\idxcourse, \idxstudent)\in \set} \obs_{\idxcourse\idxstudent},
    \end{align*}
    where equality~\stepone is true by~\eqref{eq:estimator_sum_equality_bias_variable_size}, completing the proof of~\eqref{eq:estimator_sum_equality_mean_variable_size}.
   
\subsubsection{Proof of Proposition~\ref{prop:property_shift_invariance}}\label{app:proof_prop_property_shift_invariance}

First consider the case of $\reg=\infty$, the claimed result can be verified using the closed-form expressions~\eqref{eq:properties_expression_at_infty} at $\reg=\infty$. It remains to consider the case of any $\reg\in [0, \infty)$. Assume for contradiction that the solution at $\obsmtx + \diffmeancourse\vecone^T$ is not $(\estmean+\diffmeancourse, \estbiasmtx)$, but instead $(\estmean+\diffmeancourse + u, \estbiasmtx')$ for some non-zero $u \in \reals^\numcourse$. By the optimality of $(\estmean+\diffmeancourse + u, \estbiasmtx')$, we have
    \begin{align}
        \norm{(\obsmtx + \diffmeancourse\vecone^T) - (\estmean + \diffmeancourse + u)\vecone^T - \estbiasmtx'}_\set^2 + \reg\norm{\estbiasmtx'}_\set^2 & \le \norm{(\obsmtx + \diffmeancourse\vecone^T) - (\estmean + \diffmeancourse)\vecone^T - \estbiasmtx}_\set^2 + \reg\norm{\estbiasmtx}_\set^2 \label{eq:shift_invariance_obj_comparison_original}\\
        \norm{\obsmtx  - (\estmean + u)\vecone^T - \estbiasmtx'}_\set^2 + \reg\norm{\estbiasmtx'}_\set^2 & \le \norm{\obsmtx  - \estmean \vecone^T - \estbiasmtx}_\set^2 + \reg\norm{\estbiasmtx}_\set^2.\label{eq:shift_invariance_obj_comparison}
    \end{align}
    If strict inequality in~\eqref{eq:shift_invariance_obj_comparison} holds, then $(\estmean+u, \estbiasmtx')$ attains a strictly smaller objective on observations $\obsmtx$ given $(\setpartialorder, \reg, \set)$ than $(\estmean, \estbiasmtx)$. Contradiction to the assumption that $(\estmean, \estbiasmtx)$ is optimal on the observations $\obsmtx$. Otherwise, equality holds in~\eqref{eq:shift_invariance_obj_comparison} and hence in~\eqref{eq:shift_invariance_obj_comparison_original}. By the tie-breaking rule of the equality~\eqref{eq:shift_invariance_obj_comparison_original} on the observations $(\obsmtx + \diffmeancourse\vecone^T)$, we have
    \begin{align}\label{eq:shift_invariance_tie_brekaing}
        \norm{\estbiasmtx'}_\set^2 < \norm{\estbiasmtx}_\set^2,
    \end{align}
    Combining~\eqref{eq:shift_invariance_tie_brekaing} with the equality of~\eqref{eq:shift_invariance_obj_comparison} yields a contradiction to the assumption that $(\estmean, \estbiasmtx)$ is optimal on the observations $\obsmtx$, and hence is chosen by the tie-breaking rule over the alternative solution $(\estmean+u, \estbiasmtx')$.

\subsubsection{Proof of Lemma~\ref{lem:est_mean_err_implies_diff}}\label{app:proof_lem_est_mean_err_implies_diff}

    The proof relies on~\eqref{eq:estimator_sum_equality_mean_variable_size} from Lemma~\ref{lem:estimator_sum_equality}. Assume without loss of generality that $\truemean = 0$. We first show that on the RHS of~\eqref{eq:estimator_sum_equality_mean_variable_size}, we have that $\sum_{\idxpairparen\in \settrain} \obs_\idxpair$ converges to $0$ for random $\settrain$ obtained by Algorithm~\ref{alg:cv}.
    
    Fix some constant $\errbound_1 > 0$ whose value is determined later.
    
    \textbf{Part~\ref{part:est_mean_err_implies_diff_only_noise}:} For any fixed $\settrain$, by Hoeffding's inequality, we have
    \begin{subequations}\label{eq:property_mean_err_implies_diff_hoeffding}
    \begin{align}
        \limn \Prob\left(\abs*{\frac{1}{\abs*{\settrain}} \sum_{\idxpairparen\in \settrain} \obs_\idxpair} < \errbound_1 \right)  =1.
    \end{align}
    
    \textbf{Part~\ref{part:est_mean_err_implies_diff_only_bias}:} Given the assumption that $\truemean = 0$ and the assumption that there is no noise, we have $\obsmtx = \biasmtx$. By~\eqref{eq:bias_bound_train_all} from Lemma~\ref{lem:bias_bound_train}, we have 
    \begin{align}\label{eq:est_mean_err_implies_diff_only_bias_hoeffding}
        \limn \Prob\left(\abs*{\frac{1}{\abs*{\settrain}} \sum_{\idxpairparen\in \settrain} \obs_\idxpair} < \errbound_1 \right) =1.
    \end{align}
    \end{subequations}

    The rest of the proof is the same for both parts. Denote the event in~\eqref{eq:property_mean_err_implies_diff_hoeffding} as $\event$. We now condition on $\event$ and consider the LHS of~\eqref{eq:estimator_sum_equality_mean_variable_size}. By~\eqref{eq:size_train_val_half}, the number of students in each course  $\idxcoursescope$ is $\numstudenttrain =\frac{1}{2}\numstudent$. Consider any $\reg\in [0,\infty]\in \setregoutside$. By the definition of $\setregoutside$ we have $\normtwo{\estmeanat{\reg}} \ge \errbound$. There exists some $\idxcourse^*$ such that $\abs*{\estmean_{\idxcourse^*}} \ge \frac{\errbound}{\sqrt{\numcourse}}$. Assume without loss of generality that $\estmean_{\idxcourse^*} > \frac{\errbound}{\sqrt{\numcourse}}$. We now show that there exists some $\idxcoursealt$ such that  $\estmean_{\idxcoursealt} \le 0$. Assume for contradiction that $\estmean_\idxcourse > 0$ for all $\idxcourse\in [\numcourse]$. Then by~\eqref{eq:estimator_sum_equality_mean_variable_size}, we have
    \begin{align*}
        \sum_{\idxpairparen\in \settrain} \obs_\idxpair = \numstudenttrain\sum_{\idxcoursescope} \estmean_\idxcourse \ge \numstudenttrain \estmean_{\idxcourse^*} > \frac{\numstudent}{2} \frac{\errbound}{\sqrt{\numcourse}}.
    \end{align*}
    Therefore,
    \begin{align*}
         \frac{1}{\sizesettrain}\sum_{\idxpairparen\in \set} \obs_\idxpair= \frac{2}{\numcourse\numstudent}\frac{\numstudent}{3}\frac{\errbound}{\sqrt{\numcourse}} = \frac{2\errbound}{3\numcourse^{\frac{3}{2}}}.
    \end{align*}
    Setting $\errbound_1$ to be sufficiently small such that $\errbound_1 < \frac{2\errbound}{3\numcourse^{\frac{3}{2}}}$ yields a contradiction with $\event$. Hence, conditional on $\event$, there exists some $\idxcourse^*_2$ such that $\estmean_{\idxcourse^*_2}\le 0$. Therefore, $\max_{\idxcourse, \idxcoursealt\in [\numcourse]} (\estmean_\idxcourse - \estmean_\idxcoursealt) \ge \estmean_{\idxcourse^*} - \estmean_{\idxcourse^*_2} > \frac{\errbound}{\sqrt{\numcourse}}$. A similar argument applies to the case of $\estmean_{\idxcourse^*} < -\frac{\errbound}{\sqrt{\numcourse}}$. Hence, we have
    \begin{align}\label{eq:properties_estimator_mean_err_implies_diff_conditional}
        \left.\max_{\idxcourse, \idxcoursealt\in [\numcourse]} (\estmean_\idxcourse - \estmean_\idxcoursealt) > \frac{\errbound}{\sqrt{\numcourse}},\quad \forall \reg\in \setregoutside \;\middle|\;\event.\right.
    \end{align}
    Combining~\eqref{eq:properties_estimator_mean_err_implies_diff_conditional} with~\eqref{eq:property_mean_err_implies_diff_hoeffding}, we have
    \begin{align*}
        \limn \left(\max_{\idxcourse, \idxcoursealt\in [\numcourse]} (\estmean_\idxcourse - \estmean_\idxcoursealt), \quad\forall \reg\in \setregoutside\right) \ge \prob(\event) = 1,
    \end{align*}
    completing the proof.

\subsubsection{Proof of Lemma~\ref{lem:consistency_bound_pairwise_diff_to_err}}\label{app:proof_lem_consistency_bound_pairwise_diff_to_err}

We follow the proof of Lemma~\ref{lem:est_mean_err_implies_diff}, we assume $\truemean = 0$  without loss of generality. Then fix some constant $\errbound_1 > 0$, and estalish concentration inequalities on the RHS of~\eqref{eq:estimator_sum_equality_mean_variable_size}.

\paragraph{Part~\ref{part:diff_to_err_train}:}
Same as~\eqref{eq:est_mean_err_implies_diff_only_bias_hoeffding} from Lemma~\ref{lem:est_mean_err_implies_diff}, we have
\begin{subequations}\label{eq:property_pairwise_diff_to_err_hoeffding}
\begin{align}
    \limn \Prob\left(\abs*{\frac{1}{\abs*{\settrain}} \sum_{\idxpairparen\in \settrain} \obs_\idxpair} < \errbound_1 \right) =1.
\end{align}

\paragraph{Part~\ref{part:diff_to_err_full}:}
By Hoeffding's inequality, we have
\begin{align}
    \limn \Prob\left(\frac{1}{\numcourse\numstudent} \abs*{\sum_{\idxcoursescope, \idxstudentscope} \obs_\idxpair} < \errbound_1 \right) =1.
\end{align}
\end{subequations}
The rest of the proof is the same for both parts. Combining~\eqref{eq:property_pairwise_diff_to_err_hoeffding} with~\eqref{eq:estimator_sum_equality_mean_variable_size}, we have
\begin{align}\label{eq:pairwise_diff_to_err_condition_two}
    \limn \Prob\left(\abs*{\frac{1}{\numcourse}\sum_{\idxcoursescope} \estmean_\idxcourse} < \errbound_1 \right)=1.
\end{align}
Fix any value $\errbound > 0$. Denote $\event$ as the event that the events in both~\eqref{eq:bound_pairwise_diff_whp} and~\eqref{eq:pairwise_diff_to_err_condition_two} hold. By a union bound of~\eqref{eq:bound_pairwise_diff_whp} and~\eqref{eq:pairwise_diff_to_err_condition_two}, we have
\begin{align}\label{eq:pairwise_diff_to_err_union_whp}
    \limn (\event) = 1.
\end{align}
Condition on $\event$ and consider the value of $\estmeanat{\reg}_1$. First consider the case of $\estmean_1 > \errbound$, then by~\eqref{eq:bound_pairwise_diff_whp} we have $\estmean_\idxcourse > 0$ for each $\idxcoursescope$. Then
\begin{align*}
    \left.\frac{1}{\numcourse} \abs*{\sum_{\idxcoursescope} \estmean_\idxcourse} = \frac{1}{\numcourse} \sum_{\idxcoursescope} \estmean_\idxcourse > \frac{\epsilon}{\numcourse} \quad\middle|\; \estmean_1 > \errbound, \event\right.
\end{align*}
A similar argument applies to the case of e $\estmean_1 < -\errbound$, and we have
\begin{align*}
    \left.\frac{1}{\numcourse} \abs*{\sum_{\idxcoursescope} \estmean_\idxcourse} > \frac{\epsilon}{\numcourse} \quad\middle|\; \abs*{\estmean_1} > \errbound, \event\right.
\end{align*}
The same argument applies to each $\idxcoursescope$. We have
\begin{align*}
    \left.\frac{1}{\numcourse} \abs*{\sum_{\idxcoursescope} \estmean_\idxcourse} > \frac{\epsilon}{\numcourse} \quad\middle|\; \norminf{\estmean} > \errbound, \event\right.
\end{align*}
Taking a sufficiently small $\errbound_1$ such that $\errbound_1 < \frac{\errbound}{\numcourse}$ in~\eqref{eq:pairwise_diff_to_err_condition_two} yields a contradiction. Hence, we have
\begin{align}\label{eq:pairwise_diff_to_err_event_contradiction}
    \limn \prob(\norminf{\estmean} > \errbound, \event) = 0.
\end{align}
Hence,
\begin{align*}
    \limn \prob\left(\normtwo{\estmean} > \sqrt{\numcourse}\errbound \right) \le \limn \prob\left(\norminf{\estmean} > \errbound \right) \stackrel{\stepone}{=} \limn\left(\norminf{\estmean} > \errbound, \setcomplement{\event}\right)  \le \limn \prob(\setcomplement{\event})\stackrel{\steptwo}{=} 0,
\end{align*}
where inequality~\stepone is true by~\eqref{eq:pairwise_diff_to_err_event_contradiction} and~\steptwo is true by~\eqref{eq:pairwise_diff_to_err_union_whp},
completing the proof.

\subsubsection{Proof of Proposition~\ref{prop:closed_form_solution}}\label{app:proof_prop_closed_form_solution}

    Without loss of generality we assume $\truemean = 0$. By~\eqref{eq:estimator_sum_equality_mean} from Lemma~\ref{lem:estimator_sum_equality} with the assumption that $\numcourse=2$, we have $\frac{1}{2}(\estmean_1 + \estmean_2) = \meanobs$, and hence without loss of generality we parameterize $\estmean$ with some $\margin\in \reals$ as 
    \begin{align}\label{eq:x_parametrization}
        \estmean_\margin = \meanobs + \begin{bmatrix}-1\\1\end{bmatrix}\cdot  \frac{\margin}{2}
    \end{align}
    It remains to determine the value of $\margin$.

    Given $\truemean = 0$ and the assumption that there is no noise, we have $\obsmtx=\biasmtx$. By the assumption~\ref{assumption:bias} on the bias, we have $\biasmtx$ obeys the ordering constraints $\setpartialorder$. Hence, setting $(\estmean, \estbiasmtx) = (0, \biasmtx)$ gives an objective of $0$ in~\eqref{eq:optimization}. Hence, at the optimal solution $(\estmean_\margin, \estbiasmtx_\margin)$, the objective~\eqref{eq:optimization} equals $0$. At the optimal solution, we have
    \begin{align}\label{eq:b_parametrization}
        \estbiasmtx_\margin = \obsmtx - \estmean_\margin\vecone^T.
    \end{align}

The rest of the proof consists of two steps in determining the value of $\margin$. First, we find the set of $\margin$ such that $\estbiasmtx_\gamma$ satisfies the ordering constraint $\setpartialorder$. Then we find the optimal $\margin$ from this set that is chosen by tie-breaking, minimizing $\normfro{\estbiasmtx_\margin}^2$.

\paragraph{Step 1: Finding the set of $\gamma$ that satisfies the ordering constraint}
    Given $\obsmtx=\biasmtx$, for any $\margin\reals$ we have that $\estbiasmtx_\margin$ satisfies all ordering constraints in $\setpartialorder$ that are within the same course, that is, the ordering constraints in the form of $((\idxcourse, \idxstudent), (\idxcourse, \idxstudentalt))\in \setpartialorder$ with $\idxcourse\in \{1, 2\}$. Hence, we only need to consider ordering constraints involving both courses, that is, the ordering constraints in the form of $(\idxpairparen, (\idxcoursealt, \idxstudentalt))$ with $\{\idxcourse, \idxcoursealt\} = \{1, 2\}$. It can be verified that these constraints involving both courses are satisfied if and only if
    \begin{align}\label{eq:closed_form_ordering_constraints}
        \begin{cases}
            \obscoursegroupmax{1}{1} - \estmean_1 \le \obscoursegroupmin{2}{2} - \estmean_2\\
            \obscoursegroupmax{2}{1} - \estmean_2 \le \obscoursegroupmin{1}{2} - \estmean_1.
        \end{cases}
    \end{align}
    
    Plugging the parameterization~\eqref{eq:x_parametrization} of $\estmean_\margin$ into~\eqref{eq:closed_form_ordering_constraints}, we have
    \begin{align}\label{eq:set_gamma}
        \obscoursegroupmax{2}{1} - \obscoursegroupmin{1}{2} \le \gamma \le \obscoursegroupmin{2}{2}- \obscoursegroupmax{1}{1}.
    \end{align}
    
    Note that the range in~\eqref{eq:set_gamma} is always non-empty, because given $\obsmtx=\biasmtx$, we have $\obscoursegroupmax{1}{1} \le \obscoursegroupmin{1}{2}$ and $\obscoursegroupmax{2}{1} \le \obscoursegroupmin{2}{2}$ and hence $\obscoursegroupmax{2}{1} - \obscoursegroupmin{1}{2} \le \obscoursegroupmin{2}{2}- \obscoursegroupmax{1}{1}$.
    
    \paragraph{Step 2: Finding the optimal $\gamma$ from the range~\eqref{eq:set_gamma} minimizing $\normfro{\estbiasmtx_\margin}^2$}
    
    Using the parameterizations~\eqref{eq:x_parametrization} and~\eqref{eq:b_parametrization}, we write $\normfro{\estbiasmtx_\margin}^2$ as
    \begin{align}
        \normfro{\estbiasmtx_\margin}^2 & = \normfro{\obsmtx - \estmean_\margin\vecone^T}^2 \nonumber\\
        & \stackrel{\stepone}{=} \sum_{\idxstudentscope} \left( \obs_{1\idxstudent} - \meanobs+\frac{\margin}{2}\right)^2 + \sum_{\idxstudentscope} \left(\obs_{2j} - \meanobs- \frac{\margin}{2}\right)^2\label{eq:closed_form_before_complete_square}.
    \end{align}

    Writing out the terms in~\eqref{eq:closed_form_before_complete_square} and completing the square, we have that minimizing $\normfro{\estbias_\margin}^2$ is equivalent to minimizing the term:
    \begin{align}
        \frac{\numstudent}{2} \left(\margin - (\meanobs_2 - \meanobs_1)\right)^2\label{eq:closed_form_after_complete_square}
    \end{align}
    Combining~\eqref{eq:set_gamma} and~\eqref{eq:closed_form_after_complete_square} gives the yields expression~\eqref{eq:closed_form_solution} for the optimal $\margin$.

\subsubsection{Proof of Lemma~\ref{lem:maximal_spacing_order_stats}}\label{app:proof_lem_maximal_spacing_order_stats}
The lemma is a direct consequence of the following result (given that almost-sure convergence implying convergence in probability).
\begin{lemma}[Theorem 2 in~\cite{deheuvels1985spacing}]
Let $X_1, \ldots, X_n$ be i.i.d. $\gaussian(0, 1)$. We have
\begin{align*}
    \limsup_{n\rightarrow \infty} \frac{\sqrt{2\log n}}{\log\log n} \maxspacing_n = 1\quad \text{ almost surely},
\end{align*}
where $\log$ is the logarithm of base $2$.
\end{lemma}

\subsubsection{Proof of Lemma~\ref{lem:order_stats_consistent}}\label{app:proof_lem_order_stats_consistent}

Let $\pdf$ be the p.d.f. of $\normal(0, 1)$. Let $\cdf_n$ be the empirical c.d.f. and the empirical inverse c.d.f. of $n$ i.i.d. samples from $\normal(0, 1)$ and let $\cdfinv_n$ be the inverse of $\cdf_n$.

The claim is a straightforward combination of the following two lemmas.
The first lemma states that the empirical inverse c.d.f. converges to the true inverse c.d.f.
The second lemma states that order statistics converges to the empirical inverse c.d.f. 
\begin{lemma}[Example 3.9.21 of~\cite{vaart1996weak}, or Corollary 21.5 of~\cite{vaart1998asymptotic}]
Consider any fixed $p\in (0, 1) $. Assume that $\cdf$ is differentiable at $\cdf^{-1}(p)$ and $\pdf(\cdf^{-1}(p))> 0$. Then we have
\begin{align*}
\sqrt{n}\left[\cdf_n^{-1}(p) - \cdf^{-1}(p)\right] \convdist N\left(0, \frac{p(1-p)}{\pdf^2(\cdf^{-1}(p))}\right).
\end{align*}
\end{lemma}

\begin{lemma}[Lemma 21.7 in~\cite{vaart1998asymptotic}]
Fix constant $p \in (0, 1)$. Let $\{k_n\}_{n=1}^\infty$ be a sequence of integers such that $\frac{k_n}{n} = p + \frac{c}{\sqrt{n}} + o\left(\frac{1}{\sqrt{n}}\right)$ for some constant $\const$. Then
\begin{align*}
    \sqrt{n}\left[\vargaussianorder{k_n : n} - \cdfinv_n(p)\right] \convprob \frac{c}{\pdf (\cdfinv(p))}
\end{align*}
\end{lemma}

\subsubsection{Proof of Lemma~\ref{lem:bound_size}}\label{app:proof_lem_bound_size}

    We consider any fixed $\idxcoursescope, \idxgroupscope$, and any fixed total ordering $\totalorder_0$ generated by Line~\ref{line:sample_total_order} of Algorithm~\ref{alg:cv}. Note that the $\blocklengthcoursegroup{\idxcourse}{\idxgroup}$ \elements in $\setcoursegroup{\idxcourse}{\idxgroup}$ are consecutive with respect to the sub-ordering of $\totalorder_0$ restricted to course $\idxcourse$ in Line~\ref{line:find_sub_order} of Algorithm~\ref{alg:cv}. Then it can be verified from Line~\ref{line:assign_start}-\ref{line:assign_end} of Algorithm~\ref{alg:cv} that
    \begin{align}\label{eq:bound_val_size_per_course_cases}
        \frac{\blocklengthcoursegroup{\idxcourse}{\idxgroup}}{2}-1 \le \blocklengthcoursegroupval{\idxcourse}{\idxgroup} \le \frac{\blocklengthcoursegroup{\idxcourse}{\idxgroup}}{2} + 1,
    \end{align}
    It can be verified that~\eqref{eq:bound_val_size_per_course_cases} along with the assumption that $\blocklengthcoursegroup{\idxcourse}{\idxgroup}\ge 4$ yields~\eqref{eq:bound_val_size_per_course}. Summing~\eqref{eq:bound_val_size_per_course} over $\idxcoursescope$ yields~\eqref{eq:bound_val_size}. Finally, replacing the validation set $\setval$ by  the training set $\settrain$ in the proof of~\eqref{eq:bound_val_size_per_course} and~\eqref{eq:bound_val_size}  yields~\eqref{eq:bound_train_size_per_course} and~\eqref{eq:bound_train_size}, respectively.

\subsubsection{Proof of Lemma~\ref{lem:rank_difference_adjacent_elements}}\label{app:proof_lem_rank_difference_adjacent_elements}

We prove part~\ref{part:rank_difference_adjacent_within_train} and part~\ref{part:rank_difference_adjacent_between_train_val} together. Note that if the \element of rank $\idxgroup_1$ and the \element of rank $\idxgroup_2$ are adjacent within $\settrain$, or adjacent between $\settrain$ and $\setval$, the $(\idxgroup_2 - \idxgroup_1 - 1)$ \elements of ranks from $\idxgroup_1 + 1$ through $\idxgroup_2 - 1$ are within the same set (i.e., $\settrain$ or $\setval$). Assume for contradiction that $\idxgroup_2 - \idxgroup_1 \ge 2\numcourse+2$. Then the number of \elements from rank $\idxgroup_1+1$ through $\idxgroup_2 -1$ is at least $\idxgroup_2 - \idxgroup_1 - 1 \ge 2\numcourse+1$. Consider these \elements. There exists a course $\idxcourse^*$ such that the number of such \elements within this course is at least $3$. Given that these \elements have consecutive ranks, they are consecutive within course $\idxcourse^*$. Hence, two of these \elements in course $\idxcourse^*$ appear as the same pair of \elements in Line~\ref{line:assign_end} of Algorithm~\ref{alg:cv}. According to Line~\ref{line:assign_end} of Algorithm~\ref{alg:cv}, one \element in this pair is assigned to $\settrain$ and the other \element is assigned to $\setval$. Contradiction to the assumption that all of these \elements are from the same set.

\subsubsection{Proof of Lemma~\ref{lem:bias_bound_train}}\label{app:proof_lem_bias_bound_train}

\paragraph{Proof of~~\eqref{eq:bias_bound_train_per_course}:}
We consider any course $\idxcoursescope$. We first fix any value of $\biasmtx = \truebiasmtx$. Fix any $\totalorder_0$ of the $\numcourse\numstudent$ \elements (in Line~\ref{line:sample_total_order} of Algorithm~\ref{alg:cv}). Recall from Line~\ref{line:find_sub_order} of Algorithm~\ref{alg:cv} that the sub-ordering of the $\numstudent$ \elements in course $\idxcourse$ according to $\totalorder_0$ is denoted as $(\idxcourse, \idxstudent^{(1)}), \ldots, (\idxcourse, \idxstudent^{(\numstudent)})$.

Consider each pair $(\idxcourse, \idxstudent^{(2\position-1)})$ and $(\idxcourse, \idxstudent^{(2\position)})$ for $\position \in \left[\frac{\numstudent}{2}\right]$. Algorithm~\ref{alg:cv} randomly assigns one of the two \elements to the training set $\settrain$ uniformly at random. Denote $\randsign_\position$ as the the value from this pair that is assigned to training set. Then we have
    \begin{align*}
       \randsign_\position = \begin{cases}
            \truebias_{\idxcourse,\idxstudent^{(2\position-1)}} & \text{with probability }0.5\\
            \truebias_{\idxcourse, \idxstudent^{(2\position)}} & \text{with probability } 0.5.
        \end{cases}
    \end{align*}
    Denote $\biasmaxminusmin \defn \max_{\idxstudentscope} \bias_{\idxpair} - \min_{\idxstudentscope}\bias_{\idxpair}$ and denote $\truebiasmaxminusmin = \max_{\idxstudentscope}\truebias_{\idxpair} - \min_{\idxstudentscope}\truebias_\idxpair$. Recall from~\eqref{eq:size_train_val_half} that $\numstudenttrain=\frac{\numstudent}{2}$. Fix any $\probbound > 0$. By Hoeffding's inequality, there exists $\numstudent_1$ such that for all $\numstudent\ge \numstudent_1$,
    \begin{align*}
        \Prob\left(\abs*{\frac{1}{\numstudenttrain}\sum_{\position\in \left[\frac{\numstudent}{2}\right]} \randsign_\position  - \frac{1}{\numstudenttrain} \Expect[\randsign_\position]} < \truebiasmaxminusmin \sqrt{\frac{\log\numstudent }{\numstudent}} \;\middle|\; \biasmtx = \truebiasmtx\right) \ge 1-\frac{\probbound}{2}.
    \end{align*}
    Equivalently, for all $\numstudent\ge \numstudent_1$,
    \begin{align}\label{eq:bias_bound_train_per_course_hoeffding_conditional}
        \limn \Prob\left(\abs*{\frac{1}{\numstudenttrain} \sum_{\idxstudent\in \settraincourse{\idxcourse}} \bias_{\idxpair} - \frac{1}{\numstudent}\sum_{\idxstudentscope} \bias_\idxpair} < \truebiasmaxminusmin \sqrt{\frac{\log\numstudent }{\numstudent}} \;\middle|\; \biasmtx = \truebiasmtx\right)
        \ge 1- \frac{\probbound}{2}.
    \end{align}
    Now we analyze the term $\biasmaxminusmin$. By Lemma~\ref{lem:maximal_gaussian}, we have that there exists $\numstudent_2$ such that for all $\numstudent\ge \numstudent_2$,
    \begin{align}\label{eq:bias_bound_train_per_course_maximal}
        \Probbig{\biasmaxminusmin \le 4\sqrt{\log\numstudent}} \ge 1-\frac{\probbound}{2}.
    \end{align}
    Fix any $\errbound >0$. Take $\numstudent_0$ to be sufficiently large such that $\numstudent_0 \ge \max\{\numstudent_1, \numstudent_2\}$ and $\frac{4\log\numstudent_0}{\sqrt{\numstudent_0}} < \errbound$. We have that for all $\numstudent\ge \numstudent_0$,
    \begin{align*}
        \Prob\left(\abs*{\frac{1}{\numstudenttrain} \sum_{\idxstudent\in \settraincourse{\idxcourse}} \bias_{\idxpair} - \frac{1}{\numstudent}\sum_{\idxstudentscope} \bias_\idxpair} < \errbound\right)
        & =
        \int_{\truebiasmtx\in \reals^{\numcourse\times \numstudent}}\Prob\left(\abs*{\frac{1}{\numstudenttrain} \sum_{\idxstudent\in \settraincourse{\idxcourse}} \bias_{\idxpair} - \frac{1}{\numstudent}\sum_{\idxstudentscope} \bias_\idxpair} < \errbound \;\middle|\; \truebiasmtx\right)\cdot \prob(\truebiasmtx) \dd\truebiasmtx\\
        & \ge \int_{\substack{\truebiasmtx\in \reals^{\numcourse\times \numstudent}:\\
        \truebiasmaxminusmin\le 4\sqrt{\log\numstudent}}} \Prob\left(\abs*{\frac{1}{\numstudenttrain} \sum_{\idxstudent: \idxpairparen\in \settrain} \bias_{\idxpair} - \frac{1}{\numstudent}\sum_{\idxstudentscope} \bias_\idxpair} < \errbound \;\middle|\; \biasmtx\right)\cdot \prob(\truebiasmtx) \dd\truebiasmtx\\
        & \stackrel{\stepone}{\ge} \left(1-\frac{\probbound}{2}\right)\cdot\Prob\left(\biasmaxminusmin \le 4\sqrt{\log\numstudent}\right) \\
        & \stackrel{\steptwo}{\ge} \left(1-\frac{\probbound}{2}\right)^2 \ge 1-\probbound,
    \end{align*}
    where inequality~\stepone is true by~\eqref{eq:bias_bound_train_per_course_hoeffding_conditional} and inequality~\steptwo is true by~\eqref{eq:bias_bound_train_per_course_maximal}, completing the proof.

\paragraph{Proof of~\eqref{eq:bias_bound_train_all}:} 
By Hoeffding's inequality, we have that for any $\errbound > 0$,
\begin{align}\label{eq:bias_bound_train_all_hoeffding_mean}
    \limn \prob\left(\frac{1}{\numcourse\numstudent}\abs*{\sum_{\idxcoursescope, \idxstudentscope} \bias_{\idxpair}} < \errbound\right) = 1.
\end{align}
Recall from assumption~\ref{assumption:d} that $\numcourse$ is assumed to be a constant. Taking a union bound of~\eqref{eq:bias_bound_train_per_course} over $\idxcoursescope$ and~\eqref{eq:bias_bound_train_all_hoeffding_mean}, folloed by using the triangle inequality yields the claimed result.

\subsection{Proof of auxiliary results for Theorem~\ref{thm:consistency}}\label{app:proof_consistency_auxiliary}

In this section, we present the proofs of the auxiliary results for Theorem~\ref{thm:consistency}.

\subsubsection{Proof of Lemma~\ref{lem:constraint_adjacent_group}}\label{app:proof_lem_constraint_adjacent_group}

Fix any $\const > 0$ and fix any $(\idxcourse, \idxcoursealt)\in \setpairconstraint_\const$. Suppose $\idxgroup\in[\numgroup]$ satisfies the definition~\eqref{eq:consistency_def_set_pair_constraint} corresponding to $(\idxcourse, \idxcoursealt)$. We prove that for any $\errbound>0$ and $\probbound>0$, there exists some $\numstudentlb$ such that for all $\numstudent\ge \numstudentlb$,
    \begin{align*}
        \Probbig{\estmeanat{0}_{\idxcourse'} - \estmeanat{0}_\idxcourse < \errbound} \ge 1-\probbound.
    \end{align*}
    The proof consists of two steps. In the first step, we consider the rank of the maximum bias in course $\idxcourse$ of group $\idxgroup$ (that is, $\max_{\idxpairparen\in \setcoursegroup{\idxcourse}{\idxgroup}} \rank_\idxpair$), and the rank of the minimum bias in course $\idxcourse'$ of group $(\idxgroup+1)$ (that is, $\min_{\idxpairparen\in \setcoursegroup{\idxcoursealt}{\idxgroup+1}} \rank_\idxpair$). We bound the difference between these two ranks, and then bound the difference between the values of these two terms. In the second step, we show that the ordering constraint imposed by this pair of bias terms leads to the claimed bound~\eqref{eq:consistency_pairwise_diff_by_constraint} on $\estmeanat{0}_\idxcoursealt - \estmeanat{0}_\idxcourse$.

    \paragraph{Step 1: Bounding the difference of a pair of bias terms}
    
    Recall from~\eqref{eq:def_bias_group_extremal} that $\bias_{\idxgroup, \textmax}$ denotes the largest bias of group $\idxgroup$, and $\bias_{\idxgroup+1, \textmin}$ denotes the smallest bias of group $\idxgroup+1$. We denote the rank of $\bias_{\idxgroup, \textmax}$ as $\rank$. By the definition of group ordering, the value of $\rank$ is deterministic and we have $\rank = \sum_{\idxgroupalt=1}^\idxgroup\blocklengthgroup{\idxgroupalt}$.
    Then the rank of $\bias_{\idxgroup+1, \textmin}$ is $(\rank + 1)$.
    
    Recall that $\biascoursegroupmax{\idxcourse}{\idxgroup}$ denotes the largest bias in course $\idxcourse$ of group 
    $\idxgroup$, and $\biascoursegroupmin{\idxcourse}{\idxgroup}$ denotes the smallest bias in course $\idxcourse$ of group 
    $\idxgroup$. Let $\rankrand_\idxgroup$ be a random variable denoting the difference between the ranks of $\biasgroupmax{\idxgroup}$ and $\biascoursegroupmax{\idxcourse}{\idxgroup}$, and let  $\rankrand_{\idxgroup+1}$ be a random variable denoting the difference between the ranks of $\biasgroupmin{\idxgroup+1}$ and $\biascoursegroupmin{\idxcourse,}{\idxgroup+1}$. Equivalently, the ranks of $\bias_{\idxcourse\idxgroup,\textmax}$ and $\bias_{\idxcourse+1,\idxgroup+1,\textmin}$ are $(\rank - \rankrand_\idxgroup)$ and $(\rank+1 + \rankrand_{\idxgroup+1})$, respectively, and we have $\rankrand_\idxgroup, \rankrand_{\idxgroup+1} \ge 0$.
    
    Recall that the biases within a group are ordered uniformly at random among all courses. For any constant integer $\rank_0 > 0$, if we have $\rankrand_\idxgroup \ge \rank_0$, then the bias terms corresponding to ranks of $(\rank - \rank_0+1), \ldots, \rank$ are not assigned to course $\idxcourse$. Recall that $\blocklength_{-\idxcourse, \idxgroup} = \blocklength_\idxgroup - \blocklength_{\idxcourse\idxgroup}$ denotes the number of observations in group $\idxgroup$ that are not in course $\idxcourse$. We bound the random variable $\rankrand_\idxgroup$ as
    \begin{align}\label{eq:consistency_rank_bound_one}
        \Prob(\rankrand_\idxgroup \ge \rank_0)= \prod_{m=0}^{\rank_0 - 1} \frac{\blocklength_{-\idxcourse, \idxgroup} - m}{\blocklength_{\idxgroup}-m} < \left( \frac{\blocklength_{-\idxcourse, \idxgroup}}{\blocklength_{\idxgroup}}\right)^{\rank_0} \stackrel{\stepone}{\le} (1-\const)^{\rank_0},
    \end{align}
    where step~\stepone is true by the definition~\eqref{eq:consistency_def_set_pair_constraint} of $\setpairconstraint_\const$. Similarly we have
    \begin{align}\label{eq:consistency_rank_bound_two}
        \Prob(\rankrand_{\idxgroup+1} \ge \rank_0) \le (1-\const)^{\rank_0}.
    \end{align}
    Taking $\rank_0 = \frac{\log(\frac{4}{\probbound})}{\log(1-\const)}$ and taking a union bound of~\eqref{eq:consistency_rank_bound_one} and~\eqref{eq:consistency_rank_bound_two}, we have 
    \begin{align}\label{eq:consistency_rank_bound_combine}
        \Probbig{\rankrand_{\idxgroup} + \rankrand_{\idxgroup+1} < 2\rank_0} \ge \Probbig{\rankrand_{\idxgroup} < \rank_0, \rankrand_{\idxgroup+1} < \rank_0}
        \ge 1-2(1-\const)^{\rank_0} = 1 - \frac{\probbound}{2}.
    \end{align}
    By Lemma~\ref{lem:maximal_spacing_order_stats}, there exists $\numstudent_0$ such that for all $\numstudent \ge \numstudent_0$, we have 
    \begin{align}\label{eq:consistency_max_spacing}
        \Prob\left(\maxspacing < \frac{\epsilon}{2\rank_0+1}\right) > 1-\frac{\probbound}{2},
    \end{align} 
    where $\maxspacing$ is the maximum difference between a pair of bias terms of adjacent ranks, defined as $\maxspacing \defn \max_{\idxgeneral\in [\numcourse\numstudent-1]} \bias^{(\idxgeneral+1)} - \bias^{(\idxgeneral)}$.
    Taking a union bound of~\eqref{eq:consistency_max_spacing} with~\eqref{eq:consistency_rank_bound_combine}, we have that for all $\numstudent\ge \numstudentlb$ 
    \begin{align}
        \biascoursegroupmin{\idxcoursealt,}{\idxgroup+1} - \biascoursegroupmax{\idxcourse}{\idxgroup} &< \left[(\rank + 1 +\rankrand_{\idxgroup+1}) - (\rank - \rankrand_\idxgroup)+1\right]\cdot \maxspacing \nonumber\\
        & \le (2\rank_0 + 1)\maxspacing
        < \epsilon,\quad \text{ with probability at least } 1- \probbound.\label{eq:consistency_bias_bound}
    \end{align}
    
    Due to the assumption of no noise and the assumption of $\truemean=0$, the observation model~\eqref{eq:model} reduces to $\obsmtx = \biasmtx$. In particular, we have $\obscoursegroupmax{\idxcourse}{\idxgroup} = \biascoursegroupmax{\idxcourse}{\idxgroup}$ and $\obscoursegroupmin{\idxcoursealt,}{\idxgroup+1} = \biascoursegroupmin{\idxcoursealt,}{\idxgroup+1}$. Moreover, the solution $(\estmean, \estbiasmtx)=(0, \biasmtx)$ gives an objective~\eqref{eq:optimization} of $0$ at $\reg=0$ due to $\obsmtx=\biasmtx$. Therefore the solution $(\estmeanat{0}, \estbiasmtxat{0})$ by our estimator gives an objective of $0$, satisfying the deterministic relation $\obs_{\idxcourse\idxstudent} = \estmeanat{0}_\idxcourse + \estbiasat{0}_{\idxcourse\idxstudent}$. By definition of the group ordering, the group ordering includes the constraint requiring $\estbiasat{0}_{\idxcourse\idxgroup,\textmax} \le \estbiasat{0}_{\idxcoursealt,\idxgroup+1, \textmin}$. Therefore, this ordering constraint requires the solution $(\estmeanat{0}, \estbiasmtxat{0})$ to satisfy
    \begin{align}
        \estbiasat{0}_{\idxcoursealt, \idxgroup+1, \textmin} - \estbiasat{0}_{\idxcourse\idxgroup, \textmax} & = (\obscoursegroupmin{\idxcoursealt,}{\idxgroup+1} -\estmeanat{0}_\idxcoursealt)- (\obscoursegroupmax{\idxcourse}{\idxgroup} - \estmeanat{0}_\idxcourse)\nonumber\\
        & = (\biascoursegroupmin{\idxcoursealt,}{\idxgroup+1} -\estmeanat{0}_\idxcoursealt) - (\biascoursegroupmax{\idxcourse}{\idxgroup} - \estmeanat{0}_\idxcourse) \ge 0\label{eq:consistency_constraint_before_rearrange}
    \end{align}
    Rearranging~\eqref{eq:consistency_constraint_before_rearrange}and combining it with~\eqref{eq:consistency_bias_bound}, we have that for all $\numstudent\ge \numstudentlb$,
    \begin{align*}
        \Probbig{\estmeanat{0}_{\idxcoursealt} - \estmeanat{0}_\idxcourse & \le \bias_{\idxcoursealt,\idxgroup+1,\textmin} -  \bias_{\idxcourse\idxgroup,\textmax} {<} \errbound} \ge 1-\probbound,
    \end{align*}
    completing the proof.

\subsubsection{Proof of Lemma~\ref{lem:consistency_cycle_to_diff}}\label{app:proof_lem_consistency_cycle_to_diff}

First of all, we assume that $\lengthcycle \le \numcourse$ without loss of generality. This is because if $\lengthcycle > \numcourse$, then there exists a course $\idxcourse$ that appears twice in this cycle. We write the cycle as $(\idxcourse_1, \ldots, \idxcourse, \ldots, \idxcoursealt, \ldots, \idxcourse, \ldots, \idxcourse_\lengthcycle)$, where $\idxcoursealt\in [\numcourse]$ denotes some course appearing in between the two occurrences of $\idxcourse$. We obtain a shortened cycle by replacing the segment $(\idxcourse, \ldots,\idxcoursealt, \ldots \idxcourse)$ with a single $\idxcourse$. By shortening the cycle the set of courses that appear in this cycle remain the same. We keep shortening the cycle until $\lengthcycle\le \numcourse$.

Fix any $\errbound> 0$ and $\probbound > 0$. Recall from assumption~\ref{assumption:d} that $\numcourse$ is assumed to be a constant. By applying Lemma~\ref{lem:constraint_adjacent_group} on the $\lengthcycle$ pairs in~\eqref{eq:consistency_cycle_condition} of $\setpairconstraint_\const$, and taking a union bound over these $\lengthcycle$ pairs, we have that there exists $\numstudentlb$ such that for all $\numstudent\ge \numstudentlb$, with probability at least $1-\probbound$ we simultaneously have
\begin{align}
    \begin{split}\label{eq:consistency_const_frac_diff_cycle}
    & \estmean_{\idxnode_2} - \estmean_{\idxnode_1} < \frac{\epsilon}{\numcourse},\\
    & \estmean_{\idxnode_3} - \estmean_{\idxnode_2} < \frac{\epsilon}{\numcourse},\\
    & \quad \quad \vdots\\
    & \estmean_{\idxnode_\lengthcycle} - \estmean_{\idxnode_{\lengthcycle-1}} < \frac{\errbound}{\numcourse},\\
    & \estmean_{\idxnode_1} - \estmean_{\idxnode_\lengthcycle} < \frac{\epsilon}{\numcourse}.
    \end{split}
\end{align}

Consider any $\idxnode< \idxnodealt$ with $\idxnode, \idxnodealt\in [\lengthcycle]$. Conditional on~\eqref{eq:consistency_const_frac_diff_cycle} we have
\begin{align}\label{eq:consistency_const_frac_diff_cycle_one}
    \estmean_{\idxcourse_{\idxnodealt}} - \estmean_{\idxcourse_\idxnode} = (\estmean_{\idxcourse_\idxnodealt} - \estmean_{\idxcourse_{\idxnodealt-1}}) + \ldots + (\estmean_{\idxcourse_{\idxnode+1}} - \estmean_{\idxcourse_\idxnode}) < \errbound.
\end{align}
On the other hand, conditional on~\eqref{eq:consistency_const_frac_diff_cycle} we also have
\begin{align}
    \estmean_{\idxcourse_\idxnode} - \estmean_{\idxcourse_\idxnodealt}  & = (\estmean_{\idxcourse_\idxnode} - \estmean_{\idxcourse_{\idxnode-1}}) + \ldots + (\estmean_{\idxcourse_2} - \estmean_{\idxcourse_1})  + (\estmean_{\idxcourse_1} - \estmean_{\idxcourse_\lengthcycle}) + \ldots + (\estmean_{\idxcourse_{\idxnodealt+1}} - \estmean_{\idxcourse_{\idxnodealt}})< \errbound\label{eq:consistency_const_frac_diff_cycle_two}
\end{align}
Combining~\eqref{eq:consistency_const_frac_diff_cycle_one} and~\eqref{eq:consistency_const_frac_diff_cycle_two}, we have that for all $\numstudent\ge \numstudentlb$,
\begin{align*}
    \Probbig{\abs*{\estmean_{\idxcourse_\idxnodealt} - \estmean_{\idxcourse_\idxnode}} < \errbound,\quad \forall \idxnode,\idxnodealt\in [\lengthcycle] } \ge 1-\probbound.
\end{align*}
Equivalently,
\begin{align*}
    \lim_{\numstudent\rightarrow \infty} \Prob\left(\max_{\idxnode, \idxnodealt\in [\lengthcycle]} \abs*{\estmean_{\idxcoursealt} - \estmean_{\idxcourse}} < \errbound\right) = 1,
\end{align*}
completing the proof.

\subsubsection{Proof of Lemma~\ref{lem:consistency_total_order_supergraph_single_connected}}\label{app:proof_lem_consistency_total_order_supergraph_single_connected}

The proof consists of two steps. We first show that if there exists a cycle including the nodes $\idxcourse, \idxcoursealt\in \setsupernode$, then this cycle can be modified to construct a cycle of length at most $2(\numcourse-1)$ including $\idxcourse$ and $\idxcoursealt$. In the second step, we prove the existence of a cycle.

\paragraph{Constructing a cycle of length at most $2(\numcourse-1)$ given a cycle of arbitrary length} Fix any \supernode $\setsupernode$ and any $\idxcourse, \idxcoursealt\in \setsupernode$. We assume that there exists a cycle including the nodes $\idxcourse$ and $\idxcoursealt$. By the definition of a cycle, this cycle includes a directed path $\idxcourse\rightarrow \idxcoursealt$ and a directed path $\idxcoursealt\rightarrow\idxcourse$. If the directed path $\idxcourse\rightarrow \idxcoursealt$ has length greater than $(\numcourse-1)$, then there exists some course $\idxcourse''\in [\numcourse]$ (which may or may not equal to $\idxcourse$ or $\idxcoursealt$) that appears at least twice in this cycle. Then we decompose the path into three sub-paths of $\idxcourse\rightarrow \idxcourse''$, $\idxcourse''\rightarrow \idxcourse''$, and $\idxcourse'' \rightarrow \idxcoursealt$. We remove the sub-path $\idxcourse''\rightarrow \idxcourse''$, and concatenate the subpaths $\idxcourse\rightarrow \idxcourse''$ and $\idxcourse'' \rightarrow \idxcoursealt$, giving a new path $\idxcourse\rightarrow \idxcoursealt$ of strictly smaller length than the original path. We continue shortening the path until each course appears at most once in the path, and hence the path is of length at most $(\numcourse-1)$. Likewise we shorten the path  $\idxcoursealt\rightarrow \idxcourse$ to have length at most $(\numcourse-1)$. Finally, combining these two paths $\idxcourse\rightarrow \idxcoursealt$ and $\idxcoursealt\rightarrow\idxcourse$ gives a cycle of length at most $2(\numcourse-1)$, including nodes $\idxcourse$ and $\idxcoursealt$.

\paragraph{Existence of a cycle of arbitrary length} We prove the existence of a cycle including $\idxcourse$ and $\idxcoursealt$ by induction on the procedure that constructs the partition. At initialization, each \supernode contains a single course. The claim is trivially satisfied because for any \supernode $\setsupernode$ there do not exist $\idxcourse, \idxcoursealt\in \setsupernode$ with $\idxcourse\ne \idxcoursealt$. Now consider any merge step that merges \supernodes $\setsupernode_1, \ldots, \setsupernode_{\lengthcycle}$ for some $\lengthcycle\ge 2$ during the construction of the partition. By definition, the merge occurs because there is a cycle that includes at least one course from each of the \supernodes $\setsupernode_1, \ldots, \setsupernode_\lengthcycle$. We denote the course from $\setsupernode_\idxnode$ that is included the cycle as $\idxcourse_\idxnode\in \setsupernode_\idxnode$ for each $\idxnode\in [\lengthcycle]$. If there exist multiple courses from $\setsupernode_\idxnode$ included in the cycle, we arbitrarily choose one as $\idxcourse_\idxnode$). Denote the merged \supernode as $\setsupernode = \setsupernode_1 \union\ldots \union \setsupernode_\lengthcycle$. Now consider any two courses  $\idxcourse$ and $\idxcoursealt$ from the same \supernode.

First consider the case of $\idxcourse$ and $\idxcoursealt$ are from a \supernode that is not $\setsupernode$, then by the induction hypothesis there is a cycle including both $\idxcourse$ and $\idxcoursealt$. 

Now consider the case of $\idxcourse, \idxcoursealt\in \setsupernode$. We have that $\idxcourse\in \setsupernode_{\idxnode}$ and $\idxcoursealt\in \setsupernode_{\idxnodealt}$ for some $\idxnode, \idxnodealt\in [\lengthcycle]$. If $\idxnode = \idxnodealt$, then by the induction hypothesis there is a cycle that includes both $\idxnode$ and $\idxnodealt$. If $\idxnode\ne \idxnodealt$, then by the induction hypothesis, there is a directed path $\idxcourse\rightarrow \idxcourse_\idxnode$ within $\setsupernode_\idxnode$ (trivially if $\idxcourse = \idxcourse_\idxnode$), and a directed path $\idxcourse_\idxnodealt\rightarrow \idxcoursealt$ within $\setsupernode_\idxnodealt$ (trivially if $\idxcoursealt = \idxcourse_\idxnodealt$). Moreover, by the definition of $\idxcourse_\idxnode$ and $\idxcourse_\idxnodealt$, we have that $\idxcourse_\idxnode$ and $\idxcourse_\idxnodealt$ are included in a cycle. Hence, there exists a directed path $\idxcourse_\idxnode \rightarrow \idxcourse_\idxnodealt$. Concatenating the paths $\idxcourse\rightarrow \idxcourse_\idxnode$, $\idxcourse_\idxnode \rightarrow \idxcourse_\idxnodealt$ and $\idxcourse_\idxnodealt\rightarrow \idxcoursealt$ gives a path $\idxcourse\rightarrow \idxcoursealt$. Likewise there exists a path $\idxcoursealt\rightarrow \idxcourse$. Hence, for any $\idxcourse, \idxcoursealt\in \setsupernode$, there exists a cycle that includes both $\idxcourse$ and $\idxcoursealt$.

\subsubsection{Proof of Lemma~\ref{lem:consistency_total_order_supergraph_line}}\label{app:proof_lem_consistency_total_order_supergraph_line}
The proof consists of four steps. The first step gives a preliminary property on the graph, to be used in the later steps. The second step shows that each \supernode contains courses that are consecutive. The third step shows that the ranks of \elements in each \supernode are consecutive. The fourth step shows that the edges only exist between \supernodes that are adjacent in their indexing.

\paragraph{Step 1: There exists a path from any course $\idxcourse$ to any course $\idxcoursealt$ with $\idxcourse < \idxcoursealt$}
Denote the minimal rank in course $\idxcourse$ and in course $\idxcoursealt$ as $\rank$ and $\rankalt$, respectivly. By the assumption~\eqref{eq:consistency_total_order_node_indexing_minimal_rank}, we have $\rank < \rankalt$. We consider the courses corresponding to the \elements of ranks $\rank$ through $\rankalt$, denoted as $(\idxcourse_\rank, \ldots, \idxcourse_{\rankalt})$. For any integer $\idxrank\in \{\rank, \ldots, \rankalt-1\}$ if $\idxcourse_\idxrank\ne \idxcourse_{\idxrank+1}$, then by the definition of $\setpairconstraint_\const$ from~\eqref{eq:consistency_def_set_pair_constraint} we have $(\idxcourse_\idxrank, \idxcourse_{\idxrank+1})\in \setpairconstraint_1$ because these two \elements have consecutive ranks. Hence, there is an edge $\idxcourse_\idxrank\rightarrow \idxcourse_{\idxrank+1}$ by the construction of the graph. Concatenating all such edges $\{\idxcourse_{\idxrank}\rightarrow\idxcourse_{\idxrank+1}\}_{\idxrank\in \{\rank, \ldots, \rankalt-1\}: \idxcourse_\idxrank\ne \idxcourse_{\idxrank+1}\}}$ gives a path $\idxcourse\rightarrow \idxcoursealt$.

\paragraph{Step 2: Each \supernode contains consecutive nodes} We prove that the nodes within each \supernode are consecutive. That is, for each \supernode $\setsupernode$, there exist courses $\idxcourse, \idxcoursealt\in [\numcourse]$ with $\idxcourse<\idxcoursealt$ such that $\setsupernode= \{\idxcourse,\idxcourse+1, \ldots, \idxcoursealt\}$. It suffices to consider any course $\idxcourse''$ such that $\idxcourse< \idxcourse''< \idxcoursealt$ and show that $\idxcourse''\in \setsupernode$. Assume for contradiction that $\idxcourse''\not \in \setsupernode$. By Step 1, there exists a path $\idxcourse\rightarrow \idxcourse''$ and also a path $\idxcourse'' \rightarrow \idxcoursealt$. Since $\idxcourse, \idxcoursealt\in \setsupernode$, by Lemma~\ref{lem:consistency_total_order_supergraph_single_connected} there exists a path $\idxcoursealt\rightarrow \idxcourse$. Hence, by concatenating these three paths $\idxcourse\rightarrow \idxcourse'', \idxcourse''\rightarrow \idxcoursealt$ and $\idxcoursealt\rightarrow \idxcourse$, we have a cycle that includes courses $\idxcourse, \idxcourse''$ and $\idxcoursealt$ that are involved in two different \supernodes. Contradiction to the definition of the partition that there are no cycles including nodes from more than one \supernode in the final partition, completing the proof that each \supernode contains consecutive nodes. Hence, we order the \supernodes as $\setsupernode_1,\ldots \setsupernode_\numsupernodes$, such that the indexing of the nodes increases with respect to the indexing of the \supernodes.

\paragraph{Step 3: The ranks in each \supernode are consecutive}
We show that the ranks of the \elements within each \supernode are consecutive, and also in the increasing order of the indexing of the \supernodes. Assume for contradiction that there exists some \element of rank $\rankalt$ in $\setsupernode_\idxnodealt$, and some \element of rank $\rank$ in $\setsupernode_\idxnode$ with $\idxnode < \idxnodealt$ and $\rank > \rankalt$. Denote the corresponding courses as $\idxcourse\in \setsupernode_\idxnode$ and $\idxcoursealt\in \setsupernode_\idxnodealt$. On the one hand, by Step 2 we have $\idxcourse<\idxcoursealt$ due to $\idxnode < \idxnodealt$. Then by Step 1, we have a path $\idxcourse\rightarrow \idxcoursealt$. On the other hand, we consider the \elements of ranks $\{\rankalt, \ldots, \rank\}$ and construct a path $\idxcoursealt\rightarrow\idxcourse$  similar to the construction of the path in Step 1. Concatenating the paths $\idxcourse\rightarrow \idxcoursealt$ and $\idxcoursealt\rightarrow\idxcourse$ gives a cycle that include courses $\idxcourse\in \setsupernode_\idxnode$ and $\idxcoursealt\in\setsupernode_\idxnodealt$ that from two different \supernodes. Contradiction to the definition of the partition that there does not exist cycles including more than one \supernode.

\paragraph{Step 4: The only edges on the \supernodes are $(\setsupernode_\idxnode, \setsupernode_{\idxnode+1})$ for all $\idxnode\in [\numsupernodes-1]$}
For total orderings, the edges exist between \elements of adjacent ranks. That is, consider the \elements of ranks $\rank$ and $\rank+1$ for any $\rank\in [\numcourse\numstudent-1]$. If their corresponding courses $\idxcourse_\rank$ and $\idxcourse_{\rank+1}$ are different, then there exists an edge $\idxcourse_\rank \rightarrow \idxcourse_{\rank+1}$. Then Step 4 is a direct consequence of Step 3.

\subsection{Proof of auxiliary results for Theorem~\ref{thm:cv_bias_only}}

In this section, we present the proofs of the auxiliary results for Theorem~\ref{thm:cv_bias_only}.

\subsubsection{Proof of Theorem~\ref{thm:consistency_train}}\label{app:proof_thm_consistency_train}

The proof closely follows part~\ref{part:thm_consistency_constant_fraction} and part~\ref{part:thm_consistency_total_order} of Theorem~\ref{thm:consistency} (see Appendix~\ref{app:proof_thm_consistency}). Therefore, we outline the modifications to the proof of Theorem~\ref{thm:consistency}, in order to extend to any $\settrain\subseteq [\numcourse]\times[\numstudent]$ obtained by Algorithm~\ref{alg:cv}.

\paragraph{Proof Theorem~\ref{thm:consistency_train}\ref{part:thm_consistency_train_const_frac}}

    The proof closely follows the proof of Theorem~\ref{thm:consistency}\ref{part:thm_consistency_constant_fraction} (see Appendix~\ref{app:proof_consistency_const_frac}) with the modifications discussed in what follows.

\paragraph{Extending $\setpairconstraint_\const$ to $\setpairconstrainttrain_\const$} Recall from~\eqref{eq:def_blocklength_course_group_set} that $\blocklengthcoursegrouptrain{\idxcourse}{\idxgroup}$ denotes the number of students in course $\idxcoursescope$ of group $\idxgroupscope$ restricted to the training set $\settrain$, and $\blocklengthgrouptrain{\idxgroup}$ denotes the number of students in group $\idxgroup$ restricted to the training set $\settrain$. We extend the definition~\eqref{eq:consistency_def_set_pair_constraint} of $\setpairconstraint_\const$ and define
\begin{align*}
    \setpairconstrainttrain_\const \defn \left\{(\idxcourse, \idxcoursealt)\in [\numcourse]^2: \exists \idxgroup\in [\numgroup] \textrm{ such that } \frac{\blocklengthcoursegrouptrain{\idxcourse}{\idxgroup}}{\blocklengthgrouptrain{\idxgroup}}, \frac{\blocklengthcoursegrouptrain{\idxcoursealt}{\idxgroup+1}}{\blocklengthgrouptrain{\idxgroup+1}} \ge \const\right\}.
\end{align*}

\paragraph{Extending Lemma~\ref{lem:constraint_adjacent_group} to  $\setpairconstrainttrain_\const$ restricted to the training set $\settrain$}

We show that Lemma~\ref{lem:constraint_adjacent_group} holds for any $(\idxcourse, \idxcoursealt)\in \setpairconstrainttrain_\const$, and the estimator~\eqref{eq:optimization_restricted_general} $\estmeanat{0}$ restricted to $\settrain$.

Denote $\biascoursegroupmaxtrain{\idxcourse}{\idxgroup}$ as the largest bias in course $\idxcourse$ of group $\idxgroup$ restricted to the training set $\settrain$, and denote $\biasgroupmaxtrain{\idxgroup}$ as the largest bias of group $\idxgroup$ restricted to the training set $\settrain$. We extend~\eqref{eq:consistency_rank_bound_one} to show that the difference between the ranks of $\biascoursegroupmaxtrain{\idxcourse}{\idxgroup}$ and $\biasgroupmaxtrain{\idxgroup}$ is bounded by some constant with high probability. 

Moreover, it can be verified that the difference between the ranks of $\biasgroupmaxtrain{\idxgroup}$ and $\biasgroupmax{\idxgroup}$ is bounded by a constant with high probability. Combining these two bounds, the difference between the ranks of $\biascoursegroupmaxtrain{\idxcourse}{\idxgroup}$ and $\biasgroupmax{\idxgroup}$ is bounded by a constant with high probability. We define $\biascoursegroupmintrain{\idxcoursealt}{\idxgroup+1}$ and $\biasgroupmin{\idxgroup+1}$ likewise, and extend~\eqref{eq:consistency_rank_bound_two} to show that the difference between the ranks of $\biascoursegroupmintrain{\idxcoursealt}{\idxgroup+1}$ and $\biasgroupmin{\idxgroup+1}$ is bounded by a constant with high probability. Therefore, we extend~\ref{eq:consistency_bias_bound} to:
\begin{align*}
    \biascoursegroupmintrain{\idxcoursealt}{\idxgroup+1} - \biascoursegroupmaxtrain{\idxcourse}{\idxgroup} < \errbound, \quad\text{with probability at least } 1-\probbound.
\end{align*}
Following the rest of the original arguments for Lemma~\ref{lem:constraint_adjacent_group} (see Appendix~\ref{app:proof_consistency_auxiliary}) completes the extension of Lemma~\ref{lem:constraint_adjacent_group} to being restricted to $\settrain$.

\paragraph{Extending Lemma~\ref{lem:consistency_cycle_to_diff} to $\setpairconstrainttrain_\const$ restricted to $\settrain$}

We replace the set $\setpairconstraint_\const$ in Lemma~\ref{lem:consistency_cycle_to_diff} by the set $\setpairconstrainttrain_\const$. It can be verified that Lemma~\ref{lem:consistency_cycle_to_diff} holds under this extension following its original proof (see Appendix~\ref{app:proof_consistency_auxiliary}).

\paragraph{Extending the rest of the arguments}
For any $\idxcoursescope, \idxgroupscope$, by~\eqref{eq:bound_train_size_per_course} and~\eqref{eq:bound_train_size} from Lemma~\ref{lem:bound_size} we have
\begin{align*}
    \frac{\blocklengthcoursegrouptrain{\idxcourse}{\idxgroup}}{\blocklengthgrouptrain{\idxgroup}}\ge \frac{
        \frac{\blocklength_{\idxcourse\idxgroup}}{4}
    }{
        \frac{3\blocklengthgroup{\idxgroup}}{4}}
    = \frac{\blocklengthcoursegroup{\idxcourse}{\idxgroup}}{3\blocklengthgroup{\idxgroup}}.
\end{align*}
Hence, any $(\idxcourse, \idxcoursealt)\in \setpairconstraint_{\frac{\constfrac}{\numcourse}}$, we have $(\idxcourse, \idxcoursealt) \in \setpairconstrainttrain_{\frac{\constfrac}{3\numcourse}}$. The rest of the arguments follow from the original proof of Theorem~\ref{thm:consistency}\ref{part:thm_consistency_constant_fraction} (see Appendix~\ref{app:proof_consistency_const_frac}).
    
\paragraph{Proof of Theorem~\ref{thm:consistency_train}\ref{part:thm_consistency_train_total_order}}

The proof closely follows the proof of Theorem~\ref{thm:consistency}\ref{part:thm_consistency_total_order} (see Appendix~\ref{app:proof_consistency_total_order}) with the modifications discussed in what follows. 

\paragraph{Extending $\setpairconstraint_\const$ to $\setpairconstraint^{\texttrain'}_\const$} 

Recall that for total orderings, we have $(\idxcourse, \idxcoursealt)\in \setpairconstraint_1$ if and only if there exists some $\idxgroup \in [\numcourse\numstudent-1]$ such that course $\idxcourse$ contains the \element of rank $\idxgroup$, and course $\idxcoursealt$ contains the \element of rank $(\idxgroup + 1)$. We define the following set $\setpairconstraint^{\texttrain'}$, where we consider the rank with respect to the total ordering restricted to the \elements in $\settrain$. That is, we extend the definition~\eqref{eq:consistency_def_set_pair_constraint} of $\setpairconstraint_\const$ and define
\begin{align}\label{eq:cv_only_bias_total_order_def_set_restricted}
    \setpairconstraint^{\texttrain'}\defn\; 
    \left\{
        \begin{array}{l}
        (\idxcourse, \idxcoursealt)\in [\numcourse]^2:  \exists 1\le \idxgroup < \idxgroupalt\le \sizesettrain \\
        \qquad\qquad\text{ such that } 
        \begin{array}[t]{l}
            \text{the \element of rank $\idxgroup$ is in $\settraincourse{\idxcourse}$},\\
            \text{the \element of rank $\idxgroupalt$ is in $\settraincourse{\idxcourse+1}$},\\
            \text{the \elements of ranks $(\idxgroup+1)$ through $(\idxgroupalt-1)$ are in $\setval$}
        \end{array}
        \end{array}
    \right\}.
\end{align}

\paragraph{Extending Lemma~\ref{lem:constraint_adjacent_group}}

By Lemma~\ref{lem:rank_difference_adjacent_elements}\ref{part:rank_difference_adjacent_within_train} we have that for any $(\idxcourse,\idxcoursealt)\in \setpairconstraint^{\texttrain'}$, the corresponding values of $\idxgroup$ and $\idxgroupalt$ in~\eqref{eq:cv_only_bias_total_order_def_set_restricted} satisfy $\idxgroupalt-\idxgroup\le 2\numcourse+1$. We define $\maxspacing'$ as the maximal difference between \elements that are adjacent within $\settrain$. Then by Lemma~\ref{lem:maximal_spacing_order_stats} we extend the bound of $\maxspacing$ in~\eqref{eq:consistency_max_spacing} to $\maxspacing'$ as
\begin{align*}
    \Prob\left(\maxspacing' < \epsilon\right) > 1-\frac{\probbound}{2}.
\end{align*}
Following the rest of the arguments in Appendix~\ref{app:proof_lem_constraint_adjacent_group}, we have that Lemma~\ref{lem:constraint_adjacent_group} holds restricted to the training set $\settrain$.

\paragraph{Extending Lemma~\ref{lem:consistency_cycle_to_diff} to $\setpairconstrainttrain_\const$ restricted to $\settrain$}

We replace the set $\setpairconstraint_\const$ in Lemma~\ref{lem:consistency_cycle_to_diff} by the set $\setpairconstraint^{\texttrain'}$.
It can be verified that Lemma~\ref{lem:consistency_cycle_to_diff} holds under this extension following its original proof (see Appendix~\ref{app:proof_consistency_auxiliary}).

\paragraph{Extending the rest of the arguments} 

The rest of the arguments follow from the original proof of Theorem~\ref{thm:consistency}\ref{part:thm_consistency_total_order} (see Appendix~\ref{app:proof_consistency_total_order}).
Specifically, we replace the set $\setpairconstraint_1$ by $\setpairconstraint^{\texttrain'}$. We consider the total ordering restricted to the training set $\settrain$.
We extend the definition~\eqref{eq:def_mean_b_low_high} of $(\estmeanbiasleft,\estmeanbiasright)$ to $(\estmeanbiasleft', \estmeanbiasright')$ defined as:
\begin{align*}
    & \estmeanbiasleft' \defn \frac{1}{\sum_{\idxcourse\in \setsupernodeleft}\abs*{\settraincourse{\idxcourse}}} 
    \sum_{\idxcourse\in \setsupernodeleft}\sum_{\idxstudent\in \settraincourse{\idxcourse}} \estbias_\idxpair\\
    & \estmeanbiasright' \defn \frac{1}{\sum_{\idxcourse\in \setsupernoderight}\abs*{\settraincourse{\idxcourse}}}\sum_{\idxcourse\in \setsupernoderight}\sum_{\idxstudent\in \settraincourse{\idxcourse}} \estbias_\idxpair.
\end{align*}

\subsubsection{Proof of Lemma~\ref{lem:cv_only_bias_bound_interp_mean}}\label{app:proof_lem_cv_only_bias_bound_interp_mean}

We fix any partial ordering $\setpartialorder$ that satisfies the all $\constfrac$-fraction assumption, and fix any training-validation split $(\settrain, \setval)$ obtained by Algorithm~\ref{alg:cv}.  Recall that $\settotalorder$ denotes the set of all total orderings that are consistent with the partial ordering $\setpartialorder$. Recall from Line~\ref{line:interpolation} of Algorithm~\ref{alg:cv} that the interpolated bias is computed as:
\begin{align}\label{eq:interpolation}
    \estbiasmtxvalat{\reg} = \frac{1}{\abs*{\settotalorder}} \sum_{\totalorder\in \settotalorder} \estbiasmtxvalatoforder{\reg}{\totalorder},
\end{align}
where recall from Line~\ref{line:interpolation_per_total_order} of Algorithm~\ref{alg:cv} that $[\estbiasmtxvalatoforder{\reg}{\totalorder}]_\idxpair$ for any $\idxpairparen\in \setval$ is computed as the mean value of $\estbiasmtx$ on the nearest-neighbor(s) of $\idxpairparen$ with respect to the total ordering $\totalorder$. Recall that $\setnn(\idxcourse, \idxstudent;\totalorder)$ denotes the set (of size $1$ or $2$) of the nearest neighbor(s) of $\idxpairparen$. We have
\begin{align}\label{eq:interpolation_per_order}
    [\estbiasmtxvalatoforder{\reg}{\totalorder}]_\idxpair = \frac{1}{\abs*{\setnn(\idxcourse, \idxstudent;\totalorder)}} \sum_{(\idxcoursenn,\idxstudentnn)\in \setnn} \estbiasmtxat{\reg}_{\idxcoursenn\idxstudentnn}.
\end{align}
Plugging~\eqref{eq:interpolation_per_order} to~\eqref{eq:interpolation}, we have
\begin{align*}
    \estbiasmtxvalat{\reg}_{\idxcourse\idxstudent} = \frac{1}{\abs*{\settotalorder}} \sum_{\totalorder\in \settotalorder}\frac{1}{\abs*{\setnn(\idxcourse, \idxstudent;\totalorder)}} \sum_{(\idxcoursenn,\idxstudentnn)\in \setnn} \estbiasmtxat{\reg}_{\idxcoursenn\idxstudentnn}.
\end{align*}
The remaining of the proof is outlined as follows. We decompose the summation over $\totalorder\in \settotalorder$ on the RHS of~\eqref{eq:interpolation} into two parts: total orderings $\totalorder\in \settotalorder$ where the set of nearest-neighbors $\setnn(\idxcourse, \idxstudent; \totalorder)$ is within group $\idxgroup$, and total orderings $\totalorder\in \settotalorder$ where at least one nearest-neighbor in $\setnn$ is outside group $\idxgroup$. We show $\estbiasinterpgroup{\idxgroup} =\estbiasmeangrouptrain{\idxgroup}$ in the first case, and then show that the second case happens with low probability.

We consider any group $\idxgroupscope$, and any \element in the validation set of group $\idxgroup$, that is, $\idxpairparen\in \setvalgroup{\idxgroup}$. Let $\settotalorderin\subseteq \settotalorder$ denote the subset of total orderings where the nearest-neighbor set $\setnn(\idxcourse, \idxstudent;\totalorder)$ is contained within group $\idxgroup$:
\begin{align*}
    \settotalorderin \defn \{\totalorder\in\settotalorder: \setnn(\idxcourse, \idxstudent;\totalorder) \subseteq \settraingroup{\idxgroup}\}.
\end{align*}
Let $\settotalorderout \defn \settotalorder\setminus \settotalorderin$ denote the subset of total orderings where at least one nearest-neighbor from $\setnn(\idxcourse, \idxstudent;\totalorder)$ is from outside group $\idxgroup$. It can be verified by symmetry that the value of $\estbiasmtxvalat{\reg}_{\idxpair}$ is identical for all $\idxpairparen\in \setvalgroup{\idxgroup}$. Recall that we denote this value as $\estbiasinterpgroup{\idxgroup}\defn \estbiasmtxvalat{\reg}_{\idxpair}$ for $\idxpairparen\in \setvalgroup{\idxgroup}$.

\paragraph{Case of $\totalorder\in \settotalorderin$:}

By the definition of $\settotalorderin$, we have $\setnn(\idxcourse, \idxstudent;\totalorder) \subseteq \settraingroup{\idxgroup}$. By symmetry, it can be verified that the mean of the \nn set of the \element $\idxpairparen$ over $\settotalorderin$ is simply the mean of all training \elements in $\settraingroup{\idxgroup}$. That is,
\begin{align}\label{eq:cv_only_bias_interp_bias_case_one}
    \frac{1}{\abs*{\settotalorderin}} \sum_{\totalorder\in \settotalorderin} [\estbiasmtxvalatoforder{\reg}{\totalorder}]_{\idxpair} = \frac{1}{\abs*{\settraingroup{\idxgroup}}} \sum_{(\idxcoursealt, \idxstudentalt)\in \settraingroup{\idxgroup}} \estbiasat{\reg}_{\idxcoursealt\idxstudentalt} \stackrel{\stepone}{=} \estbiasmeangrouptrain{\idxgroup},
\end{align}
where step~\stepone is true by the definition of $\estbiasmeangrouptrain{\idxgroup}$.

\paragraph{Case of $\totalorder\in \settotalorderout$:}

We bound the size of $\settotalorderout$. If a nearest-neighbor of the \element $\idxpairparen$ is outside group $\idxgroup$, then this \nn can only come from group $(\idxgroup-1)$ or $(\idxgroup+1)$. First consider the case where a \nn is from group $(\idxgroup-1)$. Assume that the \element $\idxpairparen$ is ranked $\rank\in [\blocklengthgroup{\idxgroup}]$ within the set $\setgroupof{\idxgroup}$ of all \elements from group $\idxgroup$ with respect to $\totalorder$. A \nn is from group $(\idxgroup-1)$, only if all \elements ranked $1$ through $\rank-1$ are all in the validation set (otherwise there is some training \element whose rank is between $1$ and $(\rank-1)$ within group $\idxgroup$, and this \element is closer to $\idxpairparen$ than any \element from group $(\idxgroup-1)$, giving a contradiction). Out of the total orderings in $\settotalorder$ where $\idxpairparen$ is ranked $\rank$ within group $\idxgroup$, the fraction of total orderings that the \elements ranked $1$ through $(\rank-1)$ within group $\idxgroup$ are all in the validation set $\setval$ is:
\begin{align*}
    \prod_{i=1}^{\rank-1} \frac{\blocklengthgroupval{\idxgroup}-i}{\blocklengthgroup{\idxgroup}-i} \le \left(\frac{\blocklengthgroupval{\idxgroup}}{\blocklengthgroup{\idxgroup}}\right)^{\rank-1} \stackrel{\stepone}{<} \left(\frac{3}{4}\right)^\rank,
\end{align*}
where~\stepone is true due to~\eqref{eq:bound_val_size} from Lemma~\ref{lem:bound_size}. By symmetry, the fraction of $\totalorder\in \settotalorder$ such that $\idxpairparen$ is placed in each position $\rank\in [\blocklengthgroup{\idxgroup}]$ is $\frac{1}{\blocklengthgroup{\idxgroup}}$. Therefore, the fraction of total orderings that a \nn is from group $(\idxgroup-1)$ is upper-bounded by:
\begin{align*}
    \frac{1}{\blocklengthgroup{\idxgroup}}\sum_{\rank=1}^{\blocklengthgroup{\idxgroup}} \left(\frac{3}{4}\right)^\rank \le \frac{3}{\blocklengthgroup{\idxgroup}}\stackrel{\stepone}{<} \frac{3}{\numcourse\constfrac\numstudent},
\end{align*}
where inequality~\stepone holds because $\blocklengthgroup{\idxgroup}= \sum_{\idxcoursescope} \blocklengthcoursegroup{\idxcourse}{\idxgroup}> \numcourse\constfrac\numstudent$ due to the all $\constfrac$-fraction assumption. By the same argument, the fraction of total orderings that at least one \nn is from group $(\idxgroup+1)$ is also upper-bounded by $\frac{3}{\numcourse\constfrac\numstudent}$. Hence, we have
\begin{align}\label{eq:cv_only_bias_const_frac_set_total_out_bound}
    \frac{\abs*{\settotalorderout}}{\abs*{\settotalorder}}< \frac{6}{\numcourse\constfrac\numstudent}.
\end{align}
For any $\idxpairparen\in \setvalgroup{\idxgroup}$, we have
\begin{align*}
    \estbiasinterpgroup{\idxgroup} = \frac{1}{\abs*{\settotalorder}} \left( \sum_{\totalorder\in \settotalorderin} [\estbiasmtxvalatoforder{\reg}{\totalorder}]_\idxpair + \sum_{\totalorder\in \settotalorderout} [\estbiasmtxvalatoforder{\reg}{\totalorder}]_\idxpair \right)  \stackrel{\stepone}{=} \frac{1}{\sizesettotalorder}\left( \abs*{\settotalorderin}\cdot \estbiasmeangrouptrain{\idxgroup} + \sum_{\totalorder\in \settotalorderout} [\estbiasmtxvalatoforder{\reg}{\totalorder}]_\idxpair\right),
\end{align*}
where equality~\stepone is true by plugging in~\eqref{eq:cv_only_bias_interp_bias_case_one}. Hence, we have
\begin{align*}
    \abs*{\estbiasinterpgroup{\idxgroup} - \estbiasmeangrouptrain{\idxgroup}} & = \frac{1}{\abs*{\settotalorder}}\abs*{\sum_{
    \totalorder\in \settotalorderout} [\estbiasmtxvalatoforder{\reg}{\totalorder}]_\idxpair - \estbiasmeangrouptrain{\idxgroup} }\\
    & \le \frac{1}{\abs*{\settotalorder}}\sum_{
    \totalorder\in \settotalorderout}\left(\abs*{ [\estbiasmtxvalatoforder{\reg}{\totalorder}]_\idxpair} + \abs*{\estbiasmeangrouptrain{\idxgroup}} \right)\\
    & \stackrel{\stepone}{\le} \frac{2\abs*{\settotalorderout}}{\abs*{\settotalorder}} \max_{\idxcoursescope, \idxstudentscope} \abs*{\estbias_{\idxcourse\idxstudent}} 
    \stackrel{\steptwo}{\le} \frac{12}{\constfrac\numcourse\numstudent} \cdot 
    \max_{\idxcoursescope,\idxstudentscope}\abs*{\estbias_\idxpair},
\end{align*}
where inequality~\stepone is true because $[\estbiasmtxvalatoforder{\reg}{\totalorder}]_\idxpair$ and $\estbiasmeangrouptrain{\idxgroup}$ are both the mean of $\estbiasmtx$ on a subset of its \elements, so we have $\abs*{[\estbiasmtxvalatoforder{\reg}{\totalorder}]_\idxpair}\le \max_{\idxcoursescope, \idxstudentscope} \abs*{\estbias_\idxpair}$ and $\abs*{\estbiasmeangrouptrain{\idxgroup} }\le \max_{\idxcoursescope, \idxstudentscope} \abs*{\estbias_\idxpair}$. Then step~\steptwo is true by plugging in~\eqref{eq:cv_only_bias_const_frac_set_total_out_bound}. This completes the proof.

\subsubsection{Proof of Corollary~\ref{cor:cv_only_bias_bound_interp_mean}}\label{app:proof_cor_cv_only_bias_bound_interp_mean}

Fix any $\errbound > 0$. By the consistency of $\estbiasmtxat{0}$ from ~\eqref{eq:cv_only_bias_group_order_bias_converge_in_prob}, we have
\begin{align}\label{eq:cv_only_bias_group_order_bias_converge_in_prob_repeat}
    \limn \Probbig{\abs*{\estbiasmtxat{0}_\idxpair - \biasmtx_\idxpair} < \frac{\errbound}{2}, \quad\forall \idxpairparen\in \settrain} = 1.
\end{align}
Since $\estbiasmeangrouptrain{\idxgroup}$ and $\biasmeangrouptrain{\idxgroup}$ are simply the mean of $\estbiasmtx$ and $\biasmtx$ over $\settraingroup{\idxgroup}\subseteq \settrain$. We have
\begin{align}\label{eq:cv_only_bias_const_frac_bias_mean_group_converge}
    \limn \Probbig{\abs*{\estbiasmeangrouptrain{\idxgroup} - \biasmeangrouptrain{\idxgroup}} < \frac{\errbound}{2}, \quad \forall \idxgroupscope } = 1.
\end{align}
For each $\idxgroupscope$, we have
\begin{align}
    \abs*{\estbiasinterpgroup{\idxgroup} - \biasmeangrouptrain{\idxgroup}} & \le \abs*{\estbiasinterpgroup{\idxgroup} - \estbiasmeangrouptrain{\idxgroup}} +  \abs*{\estbiasmeangrouptrain{\idxgroup} - \biasmeangrouptrain{\idxgroup}}\nonumber\\
    & \stackrel{\stepone}{\le} \frac{12}{\constfrac\numcourse\numstudent}\cdot \max_{\idxcoursescope,\idxstudentscope} \abs*{\estbias_\idxpair} + \abs*{\estbiasmeangrouptrain{\idxgroup} - \biasmeangrouptrain{\idxgroup}}\nonumber\\
    & \le \frac{12}{\constfrac\numcourse\numstudent}\left(\max_{\idxcoursescope, \idxstudentscope} \abs*{\bias_\idxpair } + \max_{\idxcoursescope, \idxstudentscope} \abs*{\bias_\idxpair - \estbias_\idxpair} \right)+ \abs*{\estbiasmeangrouptrain{\idxgroup} - \biasmeangrouptrain{\idxgroup}},\label{eq:cv_only_bias_const_frac_interp_mean_bound}
\end{align}
where~\stepone is true by combining Lemma~\ref{lem:cv_only_bias_bound_interp_mean}. In~\eqref{eq:cv_only_bias_const_frac_interp_mean_bound}, we bound the term $\max_{\idxcoursescope, \idxstudentscope}\abs*{\bias_\idxpair}$ by Lemma~\ref{lem:maximal_gaussian} as
\begin{align}\label{eq:cv_bias_bias_maximal_converge}
    \limn \prob\left(\max_{\idxcoursescope, \idxstudentscope}\abs*{\bias_\idxpair} < 2\sqrt{\log\numcourse\numstudent}\right) = 1.
\end{align}
We bound the term $\max_{\idxcoursescope, \idxstudentscope} \abs*{\bias_\idxpair - \estbias_\idxpair}$ by~\eqref{eq:cv_only_bias_group_order_bias_converge_in_prob_repeat}, and the term $\abs*{\estbiasmeangrouptrain{\idxgroup} - \biasmeangrouptrain{\idxgroup}}$ by~\eqref{eq:cv_only_bias_const_frac_bias_mean_group_converge}. Hence, plugging~\eqref{eq:cv_bias_bias_maximal_converge},~\eqref{eq:cv_only_bias_group_order_bias_converge_in_prob_repeat} and~\eqref{eq:cv_only_bias_const_frac_bias_mean_group_converge} into~\eqref{eq:cv_only_bias_const_frac_interp_mean_bound}, we have 
\begin{align*}
    \limn \Prob\left( \abs*{\estbiasinterpgroup{\idxgroup} - \biasmeangrouptrain{\idxgroup}} \le \frac{12}{\constfrac\numcourse\numstudent}\left(2\sqrt{\log\numcourse\numstudent} + \frac{\errbound}{2}\right) + \frac{\errbound}{2}, \quad \forall \idxgroupscope\right) = 1.
\end{align*}
Equivalently,
\begin{align*}
     \limn \Prob\left(\abs*{\estbiasinterpgroup{\idxgroup} - \biasmeangrouptrain{\idxgroup}} \le \errbound, \quad \forall \idxgroupscope\right) = 1,
\end{align*}
completing the proof.

\subsubsection{Proof of Lemma~\ref{lem:cv_only_bias_const_frac_hoeffding}}\label{app:proof_lem_cv_only_bias_const_frac_hoeffding}

We fix any training-validation split $(\settrain, \setval)$ and fix any $\errbound > 0$ and $\probbound>0$. We first condition on any value of the bias as $\biasmtx =\truebiasmtx$. Then the bias terms in $\setvalcoursegroup{\idxcourse}{\idxgroup}$ (whose mean is $\biasmeancoursegroupval{\idxcourse}{\idxgroup}$) can be considered as randomly sampling $\blocklengthcoursegroupval{\idxcourse}{\idxgroup}$ values from the $\blocklengthgroup{\idxgroup}$ terms in $\setgroupof{\idxgroup}$ (whose mean is $\bias_{\idxgroup})$. Denote $\truebiasmaxminusmin\defn \max_{\idxcoursescope, \idxstudentscope} \truebias_\idxpair - \min_{\idxcoursescope, \idxstudentscope} \truebias_\idxpair$, and denote $\biasmaxminusmin\defn \max_{\idxcoursescope, \idxstudentscope} \bias_\idxpair - \min_{\idxcoursescope, \idxstudentscope} \bias_\idxpair$. By Hoeffding's inequality without replacement~\cite[Section~6]{hoeffding1963bound}, we have
    \begin{align}\label{eq:cv_only_bias_const_frac_hoeffding_blocklength}
        \Prob\left( \abs*{\biasmeancoursegroupval{\idxcourse}{\idxgroup} - \truebias_{\idxgroup}} > \truebiasmaxminusmin \sqrt{\frac{\log\left(\frac{1}{\probbound}\right)}{\blocklengthcoursegroupval{\idxcourse}{\idxgroup}}} \;\middle|\; \biasmtx = \truebiasmtx \right) \le 2\exp\left(-\frac{2\blocklengthcoursegroupval{\idxcourse}{\idxgroup} \truebiasmaxminusmin^2 \log\left(\frac{1}{\probbound} \right)}{\blocklengthcoursegroupval{\idxcourse}{\idxgroup}\truebiasmaxminusmin^2}\right) = 2\probbound^2 \stackrel{\stepone}{<} \frac{\probbound}{2},
    \end{align}
    where inequality~\stepone is true for any $\probbound\in (0, \frac{1}{4})$.
    Invoking~\eqref{eq:bound_val_size_per_course} from Lemma~\ref{lem:bound_size} and using the all $\constfrac$-fraction assumption, we have
    \begin{align}
        \blocklengthcoursegroupval{\idxcourse}{\idxgroup} \ge \frac{\blocklengthcoursegroup{\idxcourse}{\idxgroup}}{4} > \frac{\constfrac\numstudent}{4}.\label{eq:cv_only_bias_const_frac_size_val_bound}
    \end{align}
    Combining~\eqref{eq:cv_only_bias_const_frac_hoeffding_blocklength} with~\eqref{eq:cv_only_bias_const_frac_size_val_bound}, we have that for any $\probbound\in (0, \frac{1}{4})$,
    \begin{align}\label{eq:cv_bias_const_frac_hoeffding_n}
        \Prob\left( \abs*{\biasmeancoursegroupval{\idxcourse}{\idxgroup} - \truebias_{\idxgroup}} > 2\truebiasmaxminusmin\sqrt{\frac{\log\left(\frac{1}{\probbound}\right)}{\constfrac\numstudent}} \;\middle|\; \biasmtx = \truebiasmtx \right) <\frac{\probbound}{2}.
    \end{align}
    Now we analyze the term $\biasmaxminusmin$ in~\eqref{eq:cv_bias_const_frac_hoeffding_n}. By Lemma~\ref{lem:maximal_gaussian}, there exists integer $\numstudent_0$ such that for any $\numstudent\ge \numstudent_0$,
    \begin{align}\label{eq:cv_bias_const_frac_maximal_gaussian}
        \Prob\left(
            \biasmaxminusmin \le 4\sqrt{\log\numcourse\numstudent}
        \right)  \ge 1- \frac{\probbound}{2}.
    \end{align}
    Let $\numstudent_1$ be a sufficiently large constant such that $\numstudent_1 \ge \numstudent_0$ and $8\sqrt{\log\numcourse\numstudent}\cdot \sqrt{\frac{\log\left(\frac{1}{\probbound}\right)}{\constfrac\numstudent}} < \errbound$. Then combining~\eqref{eq:cv_bias_const_frac_maximal_gaussian} with~\eqref{eq:cv_bias_const_frac_hoeffding_n}, for any $\numstudent\ge \numstudent_1$,
    \begin{align*}
        \Probbig{ \abs*{\biasmeancoursegroupval{\idxcourse}{\idxgroup} - \bias_{\idxgroup}}  <\errbound}
        & = \int_{\truebiasmtx\in \reals^{\numcourse\times \numstudent} } \left.\Probbig{ \abs*{\biasmeancoursegroupval{\idxcourse}{\idxgroup} - \bias_{\idxgroup}}  < \errbound \;\middle|\; \biasmtx=\truebiasmtx}\right. \cdot \Prob(\truebiasmtx)\dd \truebiasmtx \nonumber\\
        & \ge \int_{\substack{
            \truebiasmtx\in \reals^{\numcourse\times \numstudent} \nonumber\\
            \truebiasmaxminusmin \le 4\sqrt{\log\numcourse\numstudent}}
        }
        \left.\Probbig{ \abs*{\biasmeancoursegroupval{\idxcourse}{\idxgroup} - \bias_{\idxgroup}}  < \errbound\;\middle|\; \biasmtx }\right. \cdot \Prob(\biasmtx)\dd\truebiasmtx \nonumber\\
        & \stackrel{\stepone}{\ge} \left(1-\frac{\probbound}{2}\right) \cdot \Probbig{\biasmaxminusmin\le  \sqrt{4\log\numcourse\numstudent}}\nonumber\\
        & \stackrel{\steptwo}{\ge} \left(1-\frac{\probbound}{2}\right)^2\ge 1- \probbound,
    \end{align*}
    where inequality~\stepone is true by~\eqref{eq:cv_bias_const_frac_hoeffding_n} and inequality~\steptwo is true by~\eqref{eq:cv_bias_const_frac_maximal_gaussian}. Equivalently, we have
    \begin{align}\label{eq:cv_bias_const_frac_hoeffding_lim}
        \limn \Probbig{
            \abs*{\biasmeancoursegroupval{\idxcourse}{\idxgroup} - \bias_{\idxgroup}}  <\errbound
        } = 1.
    \end{align}
    Due to the all $\const$-fraction assumption, the number of groups is upper-bounded as $\numgroup\le \frac{1}{\constfrac}$. Taking a union bound of~\eqref{eq:cv_bias_const_frac_hoeffding_lim} over $\idxcoursescope, \idxgroupscope$, we have 
 \begin{align*}
    & \limn \Probbig{ \abs*{\biasmeancoursegroupval{\idxcourse}{\idxgroup} - \bias_{\idxgroup}} < \errbound, \quad \forall \idxcoursescope, \idxgroupscope} =1,
    \end{align*}
    completing the proof of~\eqref{eq:cv_only_bias_const_frac_bound_diff_block_and_interp}. A similar argument yields~\eqref{eq:cv_only_bias_const_frac_bound_dif_group_train}, where in~\eqref{eq:cv_only_bias_const_frac_size_val_bound} we invoke~\eqref{eq:bound_train_size} from Lemma~\ref{lem:bound_size} instead of~\eqref{eq:bound_val_size_per_course}.

\subsubsection{Proof of Lemma~\ref{lem:iso_regularized_obj_close_form}}\label{app:proof_lem_iso_regularized_obj_close_form}

In the proof, we use the following lemma.
\begin{lemma}\label{lem:isotonic_equivalence}
Let $d\ge 1$ be an integer. For any $y\in \reals^d$, we have
\begin{align}\label{eq:isotonic_equivalent}
    \argmin_{\soliso\in \conemonotone} \normtwo{y - \soliso}^2 + \reg \normtwo{\soliso}^2 = \argmin_{u\in \conemonotone} \normtwo{\proj_\conemonotone (y) - \soliso}^2 + \reg \normtwo{\soliso}^2
\end{align}
\end{lemma}
The proof of Lemma~\ref{lem:isotonic_equivalence} is presented at the end of this section.
We now derive a the closed-form solution to~\eqref{eq:isotonic_equivalent}. Consider the optimization problem on the RHS of~\eqref{eq:isotonic_equivalent}. We take the derivative of the objective with respect to $\soliso$, and solve for $\soliso$ by setting the derivative to $0$. It can be verified that the unconstrained solution $\solopt_\text{un}$ to the RHS of~\eqref{eq:isotonic_equivalent} is:
\begin{align}\label{eq:isotonic_sol}
    \solopt_\text{un} = \frac{1}{1 + \reg}\proj_\conemonotone(y).
\end{align}
Note that this unconstrained solution $\solopt_\text{un}$ satisfies $\solopt_\text{un}\in \conemonotone$, so $\solopt_\text{un}$ is also the (constrained) solution to~\eqref{eq:isotonic_equivalent}.
Plugging~\eqref{eq:isotonic_sol} to the objective on the LHS of~\eqref{eq:isotonic_equivalent} and rearranging the terms complete the proof.

\paragraph{Proof of Lemma~\ref{lem:isotonic_equivalence}}
We apply induction on the Pool-Adjacent-Violators algorithm (PAVA)~\cite[Section 1.2]{barlow1972isotonic}. For completeness, the Pool-Adjacent-Violators algorithm is shown in Algorithm~\ref{alg:pava}. For any integer $d\ge 1$ and any input $y\in \reals^d$, PAVA returns $\argmin_{\soliso\in \conemonotone} \normtwo{y - \soliso}^2$.

\begin{algorithm}
    \DontPrintSemicolon
Initialize $\soliso = y$\;
Initialize the partition $P = \{S_1, \ldots, S_d\}$, where $S_i = \{i\}$ for every $i \in [d]$.\;
\While{$\soliso \not\in \conemonotone$}{
    Find any $i\in [d]$ such that $\soliso_{i} > \soliso_{i + 1}$.\label{line:pava_find_violator}\;
    Find $S, S'\in P$ such that $i\in S$ and $i+1 \in S'$.\label{line:pava_find_index_violator}\;
    Update $\soliso_r \leftarrow \frac{1}{\abs*{S} + \abs*{S'}} (\sum_{i\in S} \soliso_i + \sum_{i\in S'} \soliso_i)$ for each $r\in S \union S'$.\;\label{line:pava_update_u}
    Update the partition as $P \leftarrow P\setminus \{S, S'\}+ \{S \union S'\}$.
}
\Return{$\soliso$}
\caption{The Pool-Adjacent-Violators algorithm (PAVA). Input: $y\in \reals^d$.}\label{alg:pava}
\end{algorithm}

Assume that the while loop in Algorithm~\ref{alg:pava} is executed $\numpava$ times. Let $\yseq{0}\rightarrow \yseq{1}\rightarrow\ldots\rightarrow \yseq{\numpava}$ be any sequence of the value of $x$ obtained in Algorithm~\ref{alg:pava}. We have $ \yseq{0} = y$ and $\yseq{\numpava} = \proj_\conemonotone y$. 
In what follows, we show that for any $0\le \idxpava\le \numpava-1$,
\begin{align}\label{eq:pava_induction}
    \argmin_{\soliso\in \conemonotone} \normtwo{\yseq{\idxpava} - \soliso}^2 + \reg \normtwo{\soliso}^2 = \argmin_{\soliso\in \conemonotone} \normtwo{\yseq{\idxpava + 1} - \soliso}^2 + \reg \normtwo{\soliso}^2.
\end{align}
By induction on~\eqref{eq:pava_induction}, we have
\begin{align}\label{eq:pava_final}
     \argmin_{\soliso\in \conemonotone} \normtwo{\yseq{0} - \soliso}^2 + \reg \normtwo{\soliso}^2 = \argmin_{\soliso\in \conemonotone} \normtwo{\yseq{\numpava} - \soliso}^2 + \reg \normtwo{\soliso}^2.
\end{align}
Combining~\eqref{eq:pava_final} with the fact that $ \yseq{0} = y$ and $\yseq{\numpava} = \proj_\conemonotone y$  completes the proof.

\paragraph{Proof of~\eqref{eq:pava_induction}:} Consider any $\idxpava$ such that $0\le \idxpava\le \numpava-1$. We consider Line 4-6 of PAVA in Algorithm~\ref{alg:pava}. For clarity of notation, we denote the partition corresponding to $\yseq{\idxpava}$ as $\setpartition^{(\idxpava)}$ and the partition corresponding to $\yseq{\idxpava+1}$ as $\setpartition^{(\idxpava+1)}$. Then we have $S, S'\in \setpartition^{(\idxpava)}$ and $S\union S'\in \setpartition^{(\idxpava+1)}$.

First, by PAVA it is straightforward to verify that $S$ and $S'$ both contain consecutive indices. That is, there exists integers $m_1, m_2$ such that $1 \le m_1 \le i< m_2 \le d$, such that
\begin{align*}
    & S = \{m_1, \ldots, i\}\\
    & S' = \{i+1, \ldots, m_2\}.
\end{align*}
Furthermore, by PAVA it can be verified that
\begin{subequations}\label{eq:pava_u_equal_in_each_block}
\begin{align}
    a\defn & \yseq{\idxpava}_i = \yseq{\idxpava}_{i'} \qquad \forall i, i'\in S\\
    b\defn & \yseq{\idxpava}_i = \yseq{\idxpava}_{i'} \qquad \forall i, i'\in S'\\
    z\defn & \yseq{\idxpava+1}_i = \yseq{\idxpava+1}_{i'} \qquad \forall i, i'\in S\union S'.
\end{align}
\end{subequations}
Denote these values in~\eqref{eq:pava_u_equal_in_each_block} as $a, b$ and $z$, respectively. By the update of $\soliso$ in Line~\ref{line:pava_update_u} of Algorithm~\ref{alg:pava}, we have the relation
\begin{align}\label{eq:pava_relation_a_b_z}
    z = \frac{1}{\abs*{S} + \abs*{S'}}\left(\abs*{S}\cdot a + \abs*{S'}\cdot b\right).
\end{align}
Denote $\soliso^{*(\idxpava)}$ and $\soliso^{*(\idxpava+1)}$ as the minimizer to the LHS and RHS of~\eqref{eq:pava_induction}, respectively. Using~\eqref{eq:pava_u_equal_in_each_block}, it can be verified that
\begin{subequations}\label{eq:pava_opt_equal_in_each_block}
\begin{align}
    a^*\defn & \soliso^{*(\idxpava)}_i  = \soliso^{*(\idxpava)}_{i'} \qquad \forall i, i' \in S\label{eq:pava_equal_in_each_block_one}\\
    b^*\defn & \soliso^{*(\idxpava)}_i  = \soliso^{*(\idxpava)}_{i'} \qquad \forall i, i' \in S'\label{eq:pava_equal_in_each_block_two}\\
    & \soliso^{*(\idxpava+1)}_i  = \soliso^{*(\idxpava+1)}_{i'} \qquad \forall i, i' \in S\union S'.\label{eq:pava_equal_in_each_block_three}
\end{align}
\end{subequations}
Denote the values in~\eqref{eq:pava_equal_in_each_block_one} and~\eqref{eq:pava_equal_in_each_block_two} as $a^*$ and $b^*$, respectively.

We now show that $a^* = b^*$. Assume for contradiction that $a^* \ne b^*$. Since the solution $\soliso^{*(\idxpava)}\in \conemonotone$, we have $a^* \le b^*$. Hence, we have $a^* < b^*$. By Line~\ref{line:pava_find_violator} of Algorithm~\ref{alg:pava}, we have $a > b$. We construct the alternative solution
\begin{align*}
    v^{*(t)}_i = \begin{cases}
        \soliso^{*(\idxpava)}_i & i\not \in S\union S\\
        \frac{1}{\abs*{S} + \abs*{S'}} (\abs*{S}\cdot a^* + \abs*{S} \cdot b^*) & i \in S\union S'.
    \end{cases}
\end{align*}
It can be verified that $v^{*(t)}$ attains a strict strictly smaller objective than $\soliso^{*(\idxpava)}$ for the objective on the LHS of~\eqref{eq:pava_induction}. Contradiction to the assumption that $\soliso^{*(\idxpava)}$ is the minimizer to the LHS of~\eqref{eq:pava_induction}. Hence, we have $a^* = b^*$, implying
\begin{align*}
    \soliso^{*(\idxpava)}_i  = \soliso^{*(\idxpava)}_{i'} \qquad \forall i, i'\in S \union S'.
\end{align*}
The LHS of~\eqref{eq:pava_induction} is equivalent to
\begin{align}
    & \argmin_{\substack{\soliso\in \conemonotone, t\in \reals\\ t = u_i,\;\forall i, i'\in S\union S'}}
    \sum_{i\not\in S\union S'} (\yseq{\idxpava}_i - x_i)^2+  \sum_{i\in S\union S'} (\yseq{\idxpava}_i - x_i)^2 + \reg \normtwo{u}^2\nonumber\\
    & \argmin_{\substack{\soliso\in \conemonotone\\ t = u_i,\;\forall i, i'\in S\union S'}} \sum_{i\not\in S\union S'} (\yseq{\idxpava}_i - x_i)^2 +  \underbrace{\abs*{S} \cdot (a - t)^2 + \abs*{S'} \cdot (b - t)^2}_{\term} + \reg \normtwo{u}^2.\label{eq:pava_obj_lhs}
\end{align}
We write the term $\term$ as
\begin{align}
    \term & = \abs*{S}\cdot a^2 + \abs*{S'}\cdot b^2- 2 \left(\abs*{S}\cdot a + \abs*{S'} \cdot b\right) \cdot t +(\abs*{S} + \abs*{S'})\cdot t^2  \nonumber\\
    & = (\abs*{S} + \abs*{S'}) \cdot \left( \frac{\abs*{S}\cdot a + \abs*{S'} b}{\abs*{S} + \abs*{S'}}- t\right)^2 + \text{term}(a, b, S, S') \nonumber\\
    & \stackrel{\stepone}{=} (\abs*{S} + \abs*{S'}) \cdot \left(z- t\right)^2 + \text{term}(a, b, S, S'),\label{eq:pava_term}
\end{align}
where equality~\stepone is true by~\eqref{eq:pava_relation_a_b_z}.

Using the relation $\yseq{\idxpava}_i = \yseq{\idxpava+1}_i$ for every $i\not\in S\union S'$, the RHS of~\eqref{eq:pava_induction} is equivalent to
\begin{align}
    & \argmin_{\substack{\soliso\in \conemonotone, t\in \reals\\ t = u_i,\;\forall i\in S\union S'}}
    \sum_{i\not\in S\union S'} (\yseq{\idxpava+1}_i - x_i)^2+  \sum_{i\in S\union S'} (\yseq{\idxpava+1}_i - x_i)^2 + \reg \normtwo{u}^2\nonumber\\
    & \argmin_{\substack{\soliso\in \conemonotone, t\in \reals\\ t = u_i,\;\forall i\in S\union S'}} \sum_{i\not\in S\union S'} (\yseq{\idxpava}_i - x_i)^2 +  (\abs*{S} + \abs*{S'})\cdot (z - t)^2 + \reg\normtwo{u}^2.\label{eq:pava_obj_rhs}
\end{align}
The equivalence of the LHS and RHS of~\eqref{eq:pava_induction} can be verified by combining~\eqref{eq:pava_obj_lhs},~\eqref{eq:pava_term}, and~\eqref{eq:pava_obj_rhs}.

\subsubsection{Proof of Lemma~\ref{lem:index_pair_proportional_size_val}}\label{app:proof_lem_index_pair_proportional_size_val}

    Let $\const'>0$ be a constant. Denote $\eventnonoverlaplinear{\const'}{\const}$ as the event that the number of non-overlapping pairs in $\setpairindices_\const$ (instead of $\setpairindices_\const\intersect \setval$ defined for the event $\eventnonoverlaplinearval{\const'}{\const}$) is at least $\const'\numstudent$. We delegate the main part of this proof to the following lemma.
    
    \begin{lemma}\label{lem:index_pair_proportional_size}
        Suppose $\numcourse=2$. Assume the bias is distributed according to assumption~\ref{assumption:bias} with $\gaussianwidthbias=1$. For any $\const > 0$, there exists a constant $\const' > 0$ such that
        \begin{align*}
            \limn \prob\left( \eventnonoverlaplinear{\const'}{\const}\intersect \event_2\right) = \limn \prob(\event_2).
        \end{align*}
    \end{lemma}
    The proof this result is provided at the end of this section.
    We first explain how to complete the proof of Lemma~\ref{lem:index_pair_proportional_size_val} given Lemma~\ref{lem:index_pair_proportional_size}. The proof of Lemma~\ref{lem:index_pair_proportional_size} is presented at the end of this section.

    Conditional on $\eventnonoverlaplinear{\const'}{\const}$, consider the $\const'\numstudent$ non-overlapping pairs in $\setpairindices_\const$. We denote this subset of non-overlapping pairs as  $\setpairindicesnonoverlap'$. For each $\rank \in [\frac{\numstudent}{2}]$ in Lines~\ref{line:assign_start}-\ref{line:assign_end} in Algorithm~\ref{alg:cv}, consider the \elements  $(1, \idxstudent^{(2\rank-1)})$ and $(1, \idxstudent^{(2\rank)})$ in Line~\ref{line:assign} of Algorithm~\ref{alg:cv}. If both $(1, \idxstudent^{(2\rank-1)})$ and $(1, \idxstudent^{(2\rank)})$ are involved in some pairs in $\setpairindicesnonoverlap'$, then we arbitrarily remove one of the pairs involving either $(1, \idxstudent^{(2\rank-1)})$ or $(1, \idxstudent^{(2\rank)})$ from $\setpairindicesnonoverlap'$. After the removal, the size of the remaining $\setpairindices''$ is at least $\frac{\const'\numstudent}{2}$. We repeat the same procedure to consider the \elements $(2, \idxstudent^{(2\rank-1)})$ and $(2, \idxstudent^{(2\rank)})$ and remove \elements. After this second removal, the size of the remaining $\setpairindicesnonoverlap'$ is at least $\frac{\const'\numstudent}{4}$. We now denote this set of non-overlapping pairs after the two removals as $\setpairindicesnonoverlap'$. Now consider any remaining pair $(\idxstudent, \idxstudent')\in \setpairindicesnonoverlap'$. The probability of $(1, \idxstudent)\in \setval$ is $\frac{1}{2}$ and the probability of $(2, \idxstudentalt)\in \setval$ is $\frac{1}{2}$. Hence, the probability of  $(\idxstudent, \idxstudentalt) \in \setpairindicesnonoverlap'\intersect \setval$  is $\frac{1}{4}$. Due to the removal, all of the \elements involved in $\setpairindicesnonoverlap'$ appear in different pairs during the training-validation split in Lines~\ref{line:assign_start}-\ref{line:assign_end} in Algorithm~\ref{alg:cv}. Hence, the probability of $(\idxstudent, \idxstudentalt)\in \setval$ is independent for each pair $(\idxstudent, \idxstudentalt)\in \setpairindicesnonoverlap'$. By Hoeffding's inequality, we have
    \begin{align*}
        \limn \prob\left(\abs*{\setpairindicesnonoverlap'\intersect \setval}\ge \frac{\const'\numstudent}{32} \;\middle|\; \eventnonoverlaplinear{\const'}{\const}\right) = 1.
    \end{align*}
    That is,
    \begin{align}\label{eq:cv_bias_total_order_pair_size_val_condition}
        \limn \prob\left(\eventnonoverlaplinearval{\frac{\const'}{32}}{\const} \;\middle|\;\eventnonoverlaplinear{\const'}{\const}\right) = 1.
    \end{align}
    Hence, we have
    \begin{align}
        \Prob(\eventnonoverlaplinearval{\frac{\const'}{32}}{\const}\intersect \event_2) & \ge \Prob(\eventnonoverlaplinearval{\frac{\const'}{32}}{\const}\intersect \eventnonoverlaplinear{\const'}{\const}\intersect \event_2) \nonumber\\
        & =
        \prob(\eventnonoverlaplinear{\const'}{\const}\intersect \event_2) - \Prob(\setcomplement{\eventnonoverlaplinearval{\frac{\const'}{32}}{\const}}\intersect \eventnonoverlaplinear{\const'}{\const}\intersect \event_2) \nonumber\\
        & \ge \prob(\eventnonoverlaplinear{\const'}{\const}\intersect \event_2) - \Prob(\setcomplement{\eventnonoverlaplinearval{\frac{\const'}{32}}{\const}}\intersect \eventnonoverlaplinear{\const'}{\const}).\label{eq:cv_bias_total_order_size_val_decompose}
    \end{align}
    Taking the limit of $\numstudent\rightarrow \infty$ in~\eqref{eq:cv_bias_total_order_size_val_decompose}, we have
    \begin{align*}
        \limn \Prob(\eventnonoverlaplinearval{\frac{\const'}{32}}{\const}\intersect \event_2) \stackrel{\stepone}{\ge} \limn \prob(\event_2),
    \end{align*}
    where inequality~\stepone is true by combining Lemma~\ref{lem:index_pair_proportional_size} and~\eqref{eq:cv_bias_total_order_pair_size_val_condition}, completing the proof of Lemma~\ref{lem:index_pair_proportional_size_val}. It remains to prove Lemma~\ref{lem:index_pair_proportional_size}.

    \paragraph{Proof of Lemma~\ref{lem:index_pair_proportional_size}}
    
    Recall the definition~\eqref{eq:cv_only_bias_total_order_def_set_pair} of $\setpairindices_\const = \{(\idxstudent, \idxstudent')\in [\numstudent]^2: 0 < \bias_{2\idxstudentalt} - \bias_{1\idxstudent} < \const\}$.
    We first convert the constraint $0< \bias_{2\idxstudentalt} - \bias_{1\idxstudent} < \const$ to a constraint on the ranks of the \elements $(1, \idxstudent)$ and $(2, \idxstudentalt)$. 
    
    Recall that $\pdf$ denotes the p.d.f. of $\normal(0, 1)$. Recall that $\rank(\idxcourse\idxstudent)$ is the rank of the \element $\idxpairparen$ (in the total ordering of all $2\numstudent$ \elements since we assume $\numcourse=2$). 
    For any constant $\marginpair \in (0, 1/2)$, we define the following set of pairs:
    \begin{align*}
        \setpairrank_{\marginpair, \const} = \left\{
        \begin{tabular}{ll}
            $(\idxstudent, \idxstudent')\in [\numstudent]^2:$  & $\marginpair \numstudent < \rank_{1\idxstudent} < \rank_{2\idxstudentalt} < (2-\marginpair) \numstudent$,\\
            & $\rank_{2\idxstudentalt} - \rank_{1\idxstudent} \le \const\pdf(\frac{\marginpair}{2})\numstudent$
        \end{tabular}
        \right\}.
    \end{align*}
    The following lemma shows that $\setpairrank_{\marginpair, \const}$ is a subset of $\setpairindices_\const$ for each $\marginpair > 0$ with high probability, and therefore we only need to lower-bound the number of non-overlapping pairs in $\setpairrank_{\marginpair, \const}$.

    \begin{lemma}\label{lem:cv_only_bias_total_order_set_pair_set_rank}
        For each $\const > 0$, for any $\marginpair\in \left(0, \frac{1}{2}\right)$, we have
        \begin{align*}
            \limn \Probbig{\setpairrank_{\marginpair,\const} \subseteq \setpairindices_{2\const}} = 1.
        \end{align*}
    \end{lemma}
    The proof of this result is provided in Appendix~\ref{app:proof_lem_cv_only_bias_total_order_set_pair_set_rank}.
    Denote $\eventnonoverlaplinearrank{\marginpair}{\const'}{\const}$ as the event that the set $\setpairrank_{\marginpair,\const}$ contains at least $\const'\numstudent$ non-overlapping pairs. We have that $\eventnonoverlaplinearrank{\marginpair}{\const'}{\const}$  is deterministic (depending on $\marginpair, \const', \const$ and the total ordering $\totalorder$). Then Lemma~\ref{lem:cv_only_bias_total_order_set_pair_set_rank} implies that for any $\marginpair\in \left(0, \frac{1}{2}\right)$ and any $\const'\in (0, 1)$,
    \begin{align}\label{eq:cv_bias_total_order_event_property_one}
        \limn \Probbig{\eventnonoverlaplinearrank{\marginpair}{\const'}{\const}\intersect \setcomplement{\eventnonoverlaplinear{\const'}{2\const}}} = 0.
    \end{align}
    In what follows, we establish that there exists $\marginpair>0$ and $\const' > 0$ such that
    \begin{align}\label{eq:cv_bias_total_order_event_property_two_goal}
        \limn \Probbig{\setcomplement{\event_{\marginpair, \const', \const}}\intersect \event_2} = 0,
    \end{align}
    where the choices of $\marginpair$ and $\const'$ are specified later.

    \paragraph{Proof of~\eqref{eq:cv_bias_total_order_event_property_two_goal}:} Assume there exists maximally $\numnonoverlappairs$ such non-overlapping pairs in $\setpairrank_{\marginpair, \const}$ (that is, $\setpairrank_{\marginpair, \const}$ does not have any subset of non-overlapping pairs of size greater than $\numnonoverlappairs$). Assume for contradiction that 
    \begin{align}\label{eq:cv_bias_total_order_num_nonoverlapping_assumption_contradiction}
        \numnonoverlappairs <\min\left\{\frac{\const\pdf(\frac{\marginpair}{2})}{2}, \marginpair\right\}\cdot\numstudent.
    \end{align}
    We ``remove'' these $\numnonoverlappairs$ pairs from the total ordering of $2\numstudent$ \elements, and then there are  $2(\numstudent - \numnonoverlappairs)$ remaining \elements after the removal. In what follows, we derive a contradiction by using the fact that theses \elements are not in $\setpairrank_{\marginpair, \const}$.
    
\begin{figure}[ht]
        \centering
            \includegraphics[width=0.5\linewidth]{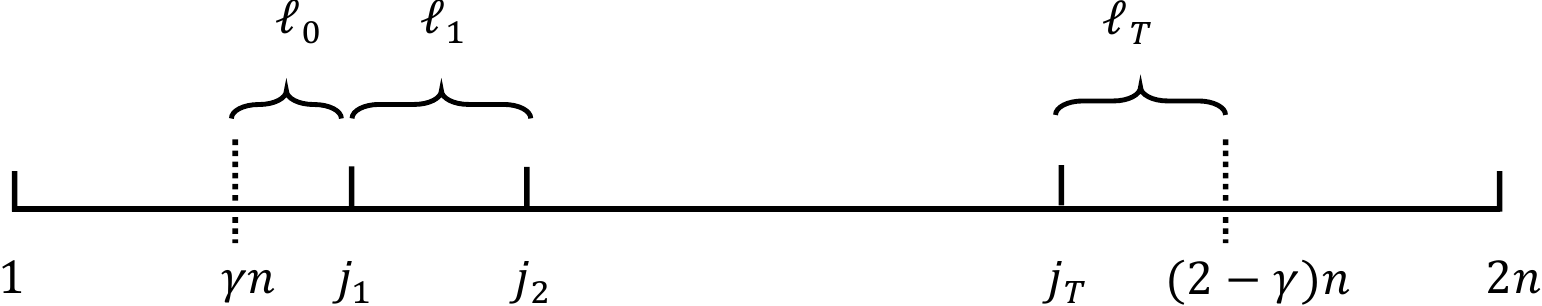}        \caption{The definition~\eqref{eq:cv_only_bias_total_order_def_l} of $\blocklength$.}
        \label{fig:pair_blocklength}
    \end{figure}

    Denote the ranks corresponding to the remaining \elements from course $2$ with rank between $(\marginpair\numstudent, (2-\marginpair)\numstudent]$ as $\idxstudent_1<  \ldots< \idxstudent_{\numremaining}$. Since $\numnonoverlappairs$ \elements are removed from each course, we have 
    \begin{align}\label{eq:cv_bias_total_order_num_remaining}
        \numremaining\le \numstudent- \numnonoverlappairs.
    \end{align}
    Since there are $(\numstudent-\numnonoverlappairs)$ remaining \elements in course $2$, and the number of \elements whose rank is outside the range $(\marginpair\numstudent, (2-\marginpair)\numstudent]$ is $2\marginpair\numstudent$, we also have $\numremaining \ge \numstudent -\numnonoverlappairs - 2\marginpair\numstudent > 0$. Denote the difference of the ranks between adjacent remaining \elements in course $2$ as \begin{align}\label{eq:cv_only_bias_total_order_def_l}
        \blocklength_i = \begin{cases}
            \idxstudent_1 -\marginpair\numstudent-1 & \text{if } i = 0\\
            \idxstudent_{i+1} - \idxstudent_i -1 & \text{if } 1 \le i \le \numremaining-1\\
            (2-\marginpair)\numstudent-\idxstudent_{i} & \text{if } i = \numremaining.
        \end{cases}
    \end{align}
    The definition~\eqref{eq:cv_only_bias_total_order_def_l} of $\blocklength$ is also visualized in Fig.~\ref{fig:pair_blocklength}.
    
    By in the definition of~\eqref{eq:cv_only_bias_total_order_def_l}, we have
    \begin{align*}
        \sum_{i=0}^{\numremaining} \blocklength_i = (2-2\marginpair)\numstudent - \numremaining \stackrel{\stepone}{\ge} (1-2\marginpair)\numstudent + \numnonoverlappairs,
    \end{align*}
    where inequality~\stepone is true by~\eqref{eq:cv_bias_total_order_num_remaining}.
    
    There are also $(\numstudent - \numnonoverlappairs)$ remaining \elements in course $1$. We consider the ranks where these \elements can be placed. Again, the number of positions outside the range $(\marginpair\numstudent, (2-\marginpair)\numstudent]$ is $2\marginpair\numstudent$. Therefore, at least $(1-2\marginpair)\numstudent - \numnonoverlappairs$ \elements form course $1$ need to placed within the range of $(\marginpair\numstudent, (2-\marginpair)\numstudent]$. Inside this range, the $\const\pdf\left(\frac{\marginpair}{2}\right)\numstudent$ ranks before each \element in course $2$ cannot be placed, because otherwise this \element from course $1$ and the corresponding \element from course $2$ form a pair in $\setpairrank_{\marginpair, \const}$. Contradiction to the assumption that a maximal subset of non-overlapping pairs has been removed. Hence, inside the range, the number of ranks where \elements from course $1$ can be placed is
    \begin{align*}
        \sum_{i=0}^{\numremaining-1} \max\left\{\blocklength_i - \const\pdf\left(\frac{\marginpair}{2}\right) \numstudent, 0\right\} + \blocklength_{\numremaining}.
    \end{align*}
    Since we need to place at least $(1-2\marginpair)\numstudent - \numnonoverlappairs$ \elements from course $1$ to these ranks, we have
    \begin{align}\label{eq:cv_only_bias_total_order_positions_to_place}
        \sum_{i=0}^{\numremaining - 1} \max\left\{\blocklength_i - \const\pdf\left(\frac{\marginpair}{2}\right) \numstudent, 0\right\} + \blocklength_{\numremaining} \ge (1-2\marginpair)\numstudent - \numnonoverlappairs.
    \end{align}
    Now we separately discuss the following two cases.
    
    \noindent\textbf{Case 1:} $\blocklength_i \ge \const\pdf\left(\frac{\marginpair}{2}\right)\numstudent$ for some $0\le i \le \numremaining-1$. Then consider the interval $[\idxstudent_i - \const\pdf(\frac{\marginpair}{2})\numstudent, \idxstudent_i)$. On the one hand, there cannot be \elements from course 2 in this interval, because we define $\blocklength_i$ as the difference of ranks between \elements $\idxstudent_{i+1}$ and $\idxstudent_{i}$ that are already adjacent among \elements in course $2$. On the other hand, there cannot be \elements $\idxstudent$ from course $1$ in this interval, because otherwise we have $(\idxstudent, \idxcourse_i)\in \setpairrank_{\marginpair, \const}$. Contradiction to the assumption that the removed subset of non-overlapping pairs is maximal. Hence, all of the $\const\pdf\left(\frac{\marginpair}{2}\right)\numstudent$ \elements from this interval $[\idxstudent_i - \const\pdf(\frac{\marginpair}{2})\numstudent, \idxstudent_i)$ have been removed, and we have $\numnonoverlappairs\ge\frac{\const\pdf\left(\frac{\marginpair}{2}\right)\numstudent}{2}$. Contradiction to the assumption~\eqref{eq:cv_bias_total_order_num_nonoverlapping_assumption_contradiction}.
    
    \noindent\textbf{Case 2: } $\blocklength_i < \const\pdf\left(\frac{\marginpair}{2}\right)\numstudent$ for all $0\le i \le \numremaining-1$. Then inequality~\eqref{eq:cv_only_bias_total_order_positions_to_place} reduces to
    \begin{align}\label{eq:cv_bias_total_order_position_last}
        \blocklength_{\numremaining}\ge (1-2\marginpair)\numstudent-\numnonoverlappairs\stackrel{\stepone}{\ge} (1-3\marginpair)\numstudent,
    \end{align}
    where inequality~\stepone is true by the assumption~\eqref{eq:cv_bias_total_order_num_nonoverlapping_assumption_contradiction} that $\numnonoverlappairs < \marginpair\numstudent$.
    
In what follows, we consider the construction of ranks of all \elements (either removed or not) that maximizes $\sum_{\idxstudentscope} (\bias_{2\idxstudent} - \bias_{1\idxstudent})$. Then we show that under the assumption~\eqref{eq:cv_bias_total_order_num_nonoverlapping_assumption_contradiction}, we have
\begin{align*}
    \limn \Prob\left(\sum_{\idxstudentscope} (\bias_{2\idxstudent} - \bias_{1\idxstudent}) < 0\right) =1.
\end{align*}

\begin{figure}[ht]
    \centering
    \includegraphics[width=0.65\linewidth]{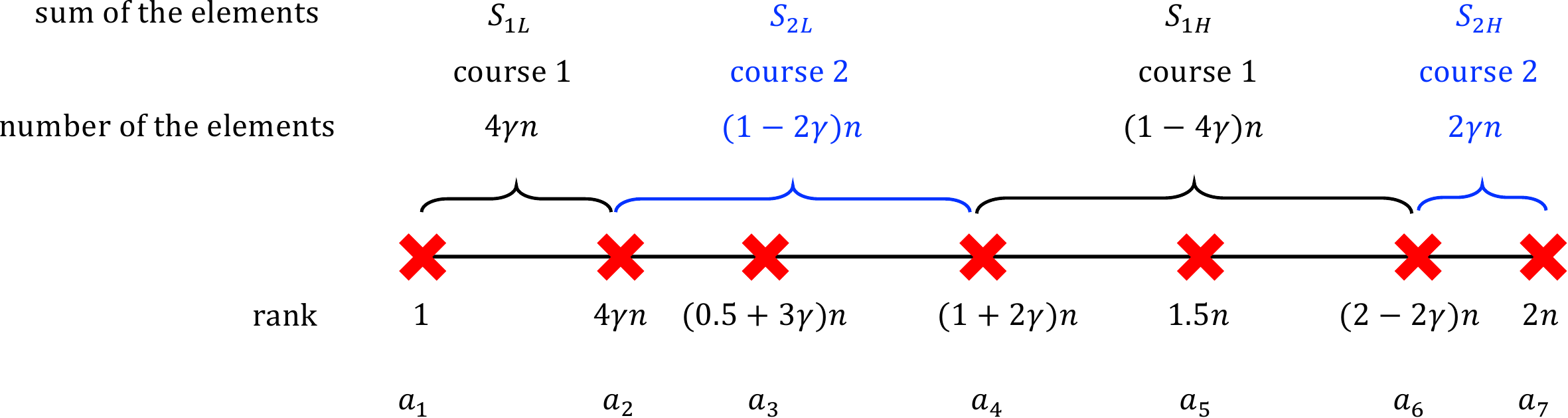}\hspace{2cm}
    \caption{Assignment of biases to the $2$ courses.}
    \label{fig:pair_assign}
\end{figure}

\paragraph{Construction of the ranks:} To maximize $\sum_{\idxstudent} (\bias_{2\idxstudent} - \bias_{1\idxstudent})$, we want to assign \elements in course $2$ to higher ranks, and \elements in course $1$ to lower ranks. We consider the course assigned to the following ranges of the rank.
\begin{itemize}
    \item \textbf{Ranks $((2-\marginpair)\numstudent, 2\numstudent]$} : The size of this range is $2\marginpair\numstudent$. We assign \elements from the course 2 to these ranks, since these are the highest possible ranks. 
    
    \item \textbf{Ranks $((1+2\marginpair)\numstudent, (2-\marginpair)\numstudent]$: } The size of this range is $(1 - 3\marginpair)\numstudent$. Note that the rank $\idxstudent_\numremaining$ is 
    \begin{align*}
        \idxstudent_\numremaining & \stackrel{\stepone}{=} (2-\marginpair)\numstudent - \blocklength_\numremaining \\
        & \stackrel{\steptwo}{\le} (2-\marginpair)\numstudent - (1-3\marginpair)\numstudent = (1+2\marginpair)\numstudent,
    \end{align*}
    where equality~\stepone is true by the definition~\eqref{eq:cv_only_bias_total_order_def_l}, and inequality~\steptwo is true by~\eqref{eq:cv_bias_total_order_position_last}. We consider the number of \elements from course $2$ in this range, remaining or removed.
    By the definition of $\idxstudent_\numremaining$ from~\eqref{eq:cv_only_bias_total_order_def_l} there cannot exist remaining \elements from course $2$ in this range. The number of removed \elements from course $2$ is $\numnonoverlappairs\le \marginpair\numstudent$ by assumption~\eqref{eq:cv_bias_total_order_num_nonoverlapping_assumption_contradiction}. Hence, the number of \elements from course $2$ in this range is at most $\marginpair\numstudent$. The other \elements in this range are from course $1$. Hence, the number of \elements from course $1$ in this range is at least $(1-4\marginpair)\numstudent$. We assign the \elements in course $2$ to higher ranks than the \elements in course $1$.
    
    \item \textbf{Ranks $[1, (1-2\marginpair)\numstudent]$} There are $4\marginpair\numstudent$ \elements from course $1$, and $(1-2\marginpair)\numstudent$ \elements from course $2$ that have not been assigned to ranks. We simply assign the $(1-2\marginpair)\numstudent$ \elements from course $2$ to be higher ranks than the $4\marginpair\numstudent$ \elements from course $1$.
\end{itemize} 
This construction of ranks is also shown in Fig.~\ref{fig:pair_assign}. We denote $S_{1L}, S_{2L}, S_{1H}, S_{2H}$ respectively as the sums of the subset of \elements as shown in Fig.~\ref{fig:pair_assign}.

The following lemma now bounds the difference between the sums of the bias in the two courses, under this construction.
\begin{lemma}\label{lem:assign_sum_diff}
    Consider $2\numstudent$ i.i.d. samples from $\normal(0, 1)$, ordered as $\vargaussian^{(1)} \le \ldots \le \vargaussian^{(2\numstudent)}$. Let 
    \begin{align*}
        & I_{1L}  \defn \{1, \ldots, 4\margin \numstudent\}\\
        & I_{2L} \defn \{4\margin\numstudent+1, \ldots, (1+2\margin)\numstudent\}\\
        & I_{1H} \defn \{(2-2\margin) \numstudent, \ldots, 2\numstudent\}\\
        & I_{2H} \defn \{(2-2\margin)\numstudent, \ldots, 2\numstudent\},
    \end{align*}
    and let
    \begin{align*}
        I_1 & \defn I_{1L}\union I_{1H},\\
        I_2 & \defn I_{2L} \union I_{2H}.
    \end{align*}
    Then there exists some constant $\marginpair > 0$, such that \begin{align*}
        \limn  \left(\sum_{i\in I_2} \vargaussian^{(i)} - \sum_{i\in I_1} \vargaussian^{(i)} < 0\right) = 1.
    \end{align*}
\end{lemma}
The proof of this result is provided in Appendix~\ref{app:proof_lem_assign_sum_diff}.
Denote the constant $\marginpair$ in Lemma~\ref{lem:assign_sum_diff} as $\marginpair_0$. By Lemma~\ref{lem:assign_sum_diff}, we have that under the assumption~\eqref{eq:cv_bias_total_order_num_nonoverlapping_assumption_contradiction} of $\numnonoverlappairs < \min\left\{\frac{\const\pdf\left(\frac{\marginpair_0}{2}\right)}{2}, \marginpair_0\right\}\numstudent$, then
\begin{align*}
    \limn \Prob\left(\sum_{\idxstudentscope} (\bias_{2\idxstudent} - \bias_{1\idxstudent}) < 0\right) = 1.
\end{align*}
Equivalently, let $\const'_0 =\min\left\{\frac{\const\pdf\left(\frac{\marginpair_0}{2}\right)}{\marginpair_0}\right\}$, we have
\begin{align*}
    \limn \Probbig{ \setcomplement{\eventnonoverlaplinearrank{\margin_0}{\const'_0}{\const}}\intersect \event_2}=0,
\end{align*}
completing the proof of~\eqref{eq:cv_bias_total_order_event_property_two_goal}.

\paragraph{Combining~\eqref{eq:cv_bias_total_order_event_property_one} and~\eqref{eq:cv_bias_total_order_event_property_two_goal}:}
We have
\begin{align}
    \limn \Prob\left(\eventnonoverlaplinear{\const'_0}{\const}\intersect \event_2\right) & = \Prob(\event_2) - \Prob(\event_2\intersect\setcomplement{\event_{\const', \const}})\nonumber\\
    & = \Prob(\event_2) - \Prob(\event_2\intersect\setcomplement{\eventnonoverlaplinear{\const'}{\const}})\nonumber\\
    & = \Prob(\event_2) - \Prob(\event_2\intersect\setcomplement{\eventnonoverlaplinear{\const'}{\const}}\intersect \eventnonoverlaplinearrank{\marginpair_0}{\const'_0}{\const}) - \Prob(\event_2\intersect\setcomplement{\eventnonoverlaplinear{\const'_0}{\const}}\intersect \setcomplement{\eventnonoverlaplinearrank{\marginpair_0}{\const'_0}{\const}}).\label{eq:cv_bias_total_order_event_nonoverlap_decompose}
\end{align}
Taking the limit of $\numstudent\rightarrow \infty$ in~\eqref{eq:cv_bias_total_order_event_nonoverlap_decompose}, we have
\begin{align*}
    \Prob\left(\eventnonoverlaplinear{\const'_0}{\const}\intersect \event_2\right) \stackrel{\stepone}{=} \limn \prob(\event_2),
\end{align*}
where equality~\stepone is true by combining~\eqref{eq:cv_bias_total_order_event_property_one} and~\eqref{eq:cv_bias_total_order_event_property_two_goal}. This completes the proof of Lemma~\ref{lem:index_pair_proportional_size}.

\subsubsection{Proof of Lemma~\ref{lem:cv_only_bias_total_order_set_pair_set_rank}}\label{app:proof_lem_cv_only_bias_total_order_set_pair_set_rank}

    We show that for any $(\idxstudent, \idxstudentalt)\in \setpairrank_{\marginpair, \const}$ we have $(\idxstudent, \idxstudentalt)\in \setpairindices_{2\const}$ due to the assumption~\eqref{assumption:bias}. First, by the definition of $\setpairrank_{\marginpair, \const}$ we have $\rank_{1\idxstudent} < \rank_{2\idxstudentalt}$, and hence $\bias_{2\idxstudentalt} > \bias_{1\idxstudent}$. It remains to show that $\bias_{2\idxstudentalt} - \bias_{1\idxstudent} < \const$. We denote $(t_0, \ldots, t_{T}) \defn (\marginpair, \marginpair +\const \pdf(\frac{\marginpair}{2}), \ldots, (2-\marginpair))$, where $T = \frac{2-2\marginpair}{\const \pdf(\frac{\marginpair}{2})}$ which is a constant. Recall that $\bias^{(k \,: \,2\numstudent)}$ denotes the $k^\thcount$ order statistics among the $2\numstudent$ random variables. Recall that $\invcdf$ denotes the inverse c.d.f. of $\normal(0, 1)$. By Lemma~\ref{lem:order_stats_consistent} we have \begin{align}\label{eq:cv_bias_total_order_converge_inv_cdf}
        \bias^{(t_i\numstudent \,: \,2\numstudent)}\convprob \cdfinv\left(\frac{t_i}{2}\right)\qquad \forall 0\le i \le T.
    \end{align}
    Taking a union bound of~\eqref{eq:cv_bias_total_order_converge_inv_cdf} over $0\le i \le T$, we have
    \begin{align}\label{eq:cv_only_bias_total_order_anchor_points}
        \limn \left(\vphantom{\frac{t_i}{2}}\right.
            \underbrace{
            \abs*{\bias^{(t_i\numstudent \,:\, 2\numstudent)} - \cdfinv\left(\frac{t_i}{2}\right)} < \frac{\const}{2} \quad\forall 0\le i \le T
        }_{\event}
        \left.\vphantom{\frac{t_i}{2}}\right)= 1.
    \end{align}
    Denote this event in~\eqref{eq:cv_only_bias_total_order_anchor_points} as $\event$.
    By the definition of $\setpairrank_{\marginpair, \const}$, for any $(\idxstudent, \idxstudentalt)\in \setpairrank_{\marginpair, \const}$ we have $\marginpair \numstudent <\rank_{1\idxstudent} < \rank_{2\idxstudent'} < (2-\marginpair)\numstudent$ and $\rank_{2\idxstudent'} -\rank_{1\idxstudent} < \const\pdf(\frac{\marginpair}{2})\numstudent$. Hence, there exists some integer $0\le i \le T-2$ such that $t_i\numstudent \le \rank_{1\idxstudent} < \rank_{2\idxstudentalt} \le t_{i+2}\numstudent$. Conditional on the event $\event$ from~\eqref{eq:cv_only_bias_total_order_anchor_points}, for any $(\idxstudent, \idxstudentalt)\in \setpairrank_{\marginpair, \const}$,
    \begin{align*}
        \bias_{2\idxstudent'} - \bias_{1\idxstudent} \le \bias^{(t_{i+2}\numstudent \,:\, 2\numstudent)} - \bias_{(t_{i}\numstudent \,:\, 2\numstudent)} & < \cdfinv\left(\frac{t_{i+2}}{2}\right) - \cdfinv\left(\frac{t_i}{2}\right) + \const\\
        & < \frac{(t_{i+2} - t_i)}{2}  \cdot\max_{x\in (\frac{\marginpair}{2} , 1-\frac{\marginpair}{2})}(\cdfinv)'(x) + \const\\
        & \stackrel{\stepone}{=}\const\pdf\left(\frac{\marginpair}{2}\right)\cdot \max_{x\in \left(\frac{\marginpair}{2}, 1-\frac{\marginpair}{2}\right)} \frac{1}{\pdf(x)} + \const\\
        & \left.=\const\pdf\left(\frac{\marginpair}{2}\right) \cdot \frac{1}{\pdf\left(\frac{\marginpair}{2}\right)} + \const = 2\const \;\middle|\; \event.\right.
    \end{align*}
    where~\stepone holds due to the equality $(\cdfinv)'(x) = \frac{1}{\cdf'(x)}=\frac{1}{\pdf(x)}$ for all $x\in (0, 1)$. Hence, 
   $\setpairrank_{\marginpair, \const}\subseteq \setpairindices_{2\const}$ conditional on $\event$,
    and we have
    \begin{align*}
        \limn \prob(\setpairrank_{\marginpair, \const}\subseteq \setpairindices_{2\const})\ge \limn \prob(\event) \stackrel{\stepone}{=} 1,
    \end{align*}
    where equality~\stepone is true by~\eqref{eq:cv_only_bias_total_order_anchor_points}, completing the proof.

\subsubsection{Proof of Lemma~\ref{lem:assign_sum_diff}}\label{app:proof_lem_assign_sum_diff}

We denote the random variables $\sumconstruct_{1L}, \sumconstruct_{2L}, \sumconstruct_{1H}$ and $\sumconstruct_{2H}$ as the sums over $I_{1L}, I_{2L}, I_{1H}$ and $I_{2H}$, respectively. To bound these sums, we consider the values of $\vargaussianorder{i}$ at the following $7$ ranks:
\begin{align*}
    i\in \{1,
    4\marginpair\numstudent,
    (0.5+3\marginpair)\numstudent, 
    (1+2\marginpair)\numstudent,
    1.5\numstudent, 
    (2-2\marginpair)\numstudent, 
    2\numstudent\},
\end{align*}
as shown by the cross marks in Fig.~\ref{fig:pair_assign}. Let $\valuecondition\in \reals^7$. In what follows we condition on the event that
\begin{align*}
    \begin{bmatrix}
        \vargaussianorder{1}, 
        \vargaussianorder{4\marginpair\numstudent},
        \vargaussianorder{(0.5+3\marginpair)\numstudent},
        \vargaussianorder{(1+2\marginpair)\numstudent},
        \vargaussianorder{1.5\numstudent},
        \vargaussianorder{(2-2\marginpair)\numstudent},
        \vargaussianorder{2\numstudent}
    \end{bmatrix}^T
     = \valuecondition.
\end{align*}
Denote the expected means of $\sumconstruct_{1L}, \sumconstruct_{2L}, \sumconstruct_{1H}$ and $\sumconstruct_{2H}$ conditional on $\valuecondition$ as $\meanconstruct_{1L\given \valuecondition}, \meanconstruct_{2L\given \valuecondition}, \meanconstruct_{1H\given \valuecondition}$ and $\meanconstruct_{2H\given \valuecondition}$, respectively. 

\paragraph{Bounding the sums $\sumconstruct_{1L}, \sumconstruct_{2L}, \sumconstruct_{1H}$ and $\sumconstruct_{2H}$ conditional on $\valuecondition$:} We first consider the sum $\sumconstruct_{2H}$. By Hoeffding's inequality, we have
\begin{subequations}\label{eq:cv_only_bias_total_order_sum_construct_hoeffding}
\begin{align}
    & \limn \Prob\left(\abs*{\sumconstruct_{1L} - 4\marginpair\numstudent\meanconstruct_{1L\given \valuecondition}} < (\valueconditionmax-\valueconditionmin)\sqrt{\numstudent\log\numstudent} \;\middle|\; \valuecondition\right) = 1\\
    & \limn \Prob\left(\abs*{S_{2L} - (1-2\marginpair) \numstudent \meanconstruct_{2L\given \valuecondition}} < (\valueconditionmax-\valueconditionmin)\sqrt{\numstudent\log\numstudent}\;\middle|\; \valuecondition\right) =1\\
    & \limn \Prob\left(\abs*{S_{1H} - (1-4\marginpair) \numstudent \meanconstruct_{1H\given \valuecondition}} < (\valueconditionmax-\valueconditionmin)\sqrt{\numstudent\log\numstudent}\;\middle|\; \valuecondition\right)= 1\\
    & \limn \Prob\left(\abs*{\sumconstruct_{2H} - 2\marginpair\numstudent\meanconstruct_{2H\given \valuecondition}} < (\valueconditionmax - \valueconditionmin)\sqrt{\numstudent\log\numstudent}\;\middle|\; \valuecondition\right) = 1.
\end{align}
\end{subequations}
Taking a union bound of~\eqref{eq:cv_only_bias_total_order_sum_construct_hoeffding} and using the equality $\sum_{i\in I_2} \vargaussianorder{i} - \sum_{i\in I_1} \vargaussianorder{i} = S_{2L} + S_{2H} - S_{1L} - S_{1H}$, we have 
\begin{align*}
    \limn \prob& 
    \left(\vphantom{\sum_{i\in I_2}}\right.
    \sum_{i\in I_2} \vargaussianorder{i} - \sum_{i\in I_1} \vargaussianorder{i}\\
    & \le \numstudent
    \left[\vphantom{\sum_{i\in I_2}}\right.
    \underbrace{
        (1-2\marginpair) \meanconstruct_{2L\given \valuecondition} - (1-4\marginpair)\meanconstruct_{1H\given \valuecondition} + 2\marginpair \meanconstruct_{2H\given \valuecondition}- 4\marginpair \meanconstruct_{1L\given \valuecondition}+ 4(\valueconditionmax-\valueconditionmin)\sqrt{\frac{\log\numstudent}{\numstudent}}
    }_{\term}
    \left.\vphantom{\sum_{i\in I_2}}\;\middle|\; \valuecondition\right]
    \left.\vphantom{\sum_{i\in I_2}}\right)
    = 1.
\end{align*}
We rearrange the terms in $\term$ as
\begin{align}\label{eq:cv_bias_total_order_bias_sum_diff_rearrange}
    \term & =
        (1-4\marginpair) (\meanconstruct_{2L\given \valuecondition} -\meanconstruct_{1H\given \valuecondition}) + 4\marginpair( \meanconstruct_{2H\given \valuecondition}-\mu_{1L\given \valuecondition})  +2\marginpair(\meanconstruct_{2L\given \valuecondition} - \mu_{2H\given \valuecondition}) +  4(\valueconditionmax-\valueconditionmin)\sqrt{\frac{\log\numstudent}{\numstudent}}.
\end{align}
In what follows, we define a range $\setcondition$ on the values of $\valuecondition$, show that $\limn \prob(\valuecondition\in \setcondition) = 1$ and show that $\term < 0$ conditional on any $\valuecondition\in \setcondition$.

\paragraph{Defining the range $\setcondition$ and showing $\limn \prob(\valuecondition\in \setcondition) = 1$:} 

We define the range $\setcondition\subseteq \reals^7$ as
\begin{align}\label{eq:cv_bias_total_order_event_a}
    \setcondition \defn \; \left\{ 
            \begin{array}{l}
            \valueconditionmin < \invcdf(1.5\margin)\\
            \valuecondition_2 > \invcdf(1.99\margin)\\
            \valuecondition_3 < \invcdf(0.25 + 1.5\margin) + 0.01\\
            \valuecondition_5 > \invcdf(0.75) -0.01 \\
            \valuecondition_6 < \invcdf(1-0.99\margin)\\
            \valueconditionmax > \invcdf(1-0.5\margin)
            \end{array}
        \right\} \intersect \left\{
            \begin{array}{l}
                \valueconditionmin > -2\sqrt{\log 2\numstudent}\\
                \valueconditionmax < 2\sqrt{\log2\numstudent}
            \end{array}
    \right\}.
\end{align}
By Lemma~\ref{lem:order_stats_consistent}, we have
\begin{subequations}\label{eq:cv_bias_total_order_conv_invcdf}
\begin{align}
    \valuecondition_2 & \convprob\invcdf (2\margin)\\
    \valuecondition_3 & \convprob \invcdf(0.25 + 1.5\margin)\\
    \valuecondition_5 & \convprob \invcdf(0.75)\\
    \valuecondition_6 & \convprob \invcdf(1-\margin).
\end{align}
\end{subequations}
Moreover, for the extremal values $\valueconditionmin$ and $\valueconditionmax$, we have that for any $\const \in \reals$,
\begin{subequations}\label{eq:cv_bias_total_order_conv_extremal}
\begin{align}
    & \limn \prob(\valueconditionmin < \const) = 1\\
    & \limn \prob(\valueconditionmax > \const) = 1.
\end{align}
\end{subequations}
Combining~\eqref{eq:cv_bias_total_order_conv_invcdf},~\eqref{eq:cv_bias_total_order_conv_extremal} and Lemma~\ref{lem:maximal_gaussian}, we have that for any $\margin > 0$,
\begin{align*}
    \limn \prob(\event) = 1.
\end{align*}

\paragraph{Analyzing the expected means $\meanconstruct_{1L\given \valuecondition}, \meanconstruct_{2L\given \valuecondition}, \meanconstruct_{1H\given \valuecondition}, \meanconstruct_{2H\given \valuecondition}$:} We analyze the terms on the RHS of~\eqref{eq:cv_bias_total_order_bias_sum_diff_rearrange}.

\paragraph{Term $(\meanconstruct_{2L\given \valuecondition} - \meanconstruct_{1H\given \valuecondition})$: } We have $\meanblock_{2L} \le \frac{\valuecondition_3 + \valuecondition_4}{2}$ and $\meanblock_{1H} \ge \frac{\valuecondition_4 + \valuecondition_5}{2}$. Therefore, conditional on any $\valuecondition\in \setcondition$, for any $\margin < 0.1$,
\begin{align}\label{eq:cv_bias_total_order_mean_term_one}
    \meanconstruct_{2L\given \valuecondition} - \meanconstruct_{1H\given \valuecondition} \le \frac{\valuecondition_3 - \valuecondition_5}{2} \stackrel{\stepone}{\le}  -0.5,
\end{align}
where inequality~\stepone is true by the definition~\eqref{eq:cv_bias_total_order_event_a} of $\setcondition$.

\paragraph{Term $(\meanconstruct_{2H} - \meanconstruct_{1L})$:} Let $\vargaussian$ denote a random variable of $\normal(0, 1)$. Conditional on any $\valuecondition\in \setcondition$,
\begin{subequations}\label{eq:cv_bias_total_order_mean_term_two}
\begin{align}
    \meanconstruct_{2H\given \valuecondition} & = \frac{1}{\sqrt{2\pi}}\frac{1}{\Prob\left(\valuecondition_6 < X < \valueconditionmax\right)} \int_{\valuecondition_6}^{\valueconditionmax} x e^{-\frac{x^2}{2}} \dd x\nonumber\\
    & = \frac{1}{\sqrt{2\pi}}\frac{1}{\Prob\left(\valuecondition_6 < X < \valueconditionmax\right)} \left[-e^{-\frac{x^2}{2}}\right]_{x=\valuecondition_6}^{\valueconditionmax} \nonumber\\
    & \le \frac{1}{\sqrt{2\pi}}\frac{1}{\Prob\left(\valuecondition_6 < X < \valueconditionmax\right)} e^{-\frac{\valuecondition_6^2}{2}} \nonumber\\
    & \stackrel{\stepone}{\le} \frac{1}{\sqrt{2\pi}} \frac{1}{0.49\margin} e^{-\frac{\left[\invcdf(1-0.99\margin)\right]^2}{2}},
\end{align}
where~\stepone is true by the definition~\eqref{eq:cv_bias_total_order_event_a} of $\setcondition$.
Similarly, conditional on the event $\event$ and on any $\valuecondition$,
\begin{align}
    \mu_{1L\given \valuecondition} > -\frac{1}{\sqrt{2\pi}}\frac{1}{0.49\margin} e^{-\frac{\left[\invcdf(1.99\marginpair)\right]^2}{2}}.
\end{align}
\end{subequations}

\paragraph{Term: $(\meanconstruct_{2L\given \valuecondition} - \mu_{2H\given \valuecondition})$:} For any $\valuecondition\in \reals^7$, we have 
\begin{align}\label{eq:cv_bias_total_order_mean_term_three}
    (\meanconstruct_{2L\given \valuecondition} - \mu_{2H\given \valuecondition}) < 0.
\end{align}

\paragraph{Showing $\term < 0$: } Plugging the three terms from~\eqref{eq:cv_bias_total_order_mean_term_one},~\eqref{eq:cv_bias_total_order_mean_term_two} and~\eqref{eq:cv_bias_total_order_mean_term_three} back to~\eqref{eq:cv_bias_total_order_bias_sum_diff_rearrange}, conditional on any $\valuecondition\in \setcondition$,
\begin{align*}
    \term < -0.5(1-4\margin) + 4 \cdot \frac{1}{\sqrt{2\pi}}\frac{1}{0.49}\left(
        e^{-\frac{[\invcdf(1-0.99\margin)]^2}{2}} + e^{-\frac{[\invcdf(1.99\marginpair)]^2}{2}}
    \right)  + 8\sqrt{\log\numstudent}\sqrt{\frac{\log2\numstudent}{\numstudent}}.
\end{align*}
As $\margin \rightarrow 0$, we have $\invcdf(1.99\marginpair) \rightarrow -\infty$ and $\invcdf(1-0.99\marginpair) \rightarrow \infty$. It can be verified that there exists some sufficiently small $\margin_0 >0$, such that 
\begin{align*}
   \left. \limn \term < 0 \;\middle|\;\valuecondition\in \setcondition.\right.
\end{align*}
Hence, we have
\begin{align*}
    \limn \prob\left(\sum_{i\in I_2} \vargaussianorder{i} - \sum_{i\in I_1} \vargaussianorder{i} \le 0\right) & \ge \limn \int_{\valuecondition\in \reals^7}\prob\left(\term < 0 \;|\; \valuecondition\right) \prob(\valuecondition) \\
    & \ge \limn \prob(\valuecondition\in\setcondition) = 1,
\end{align*}
completing the proof.

\subsection{Proof of auxiliary results for Theorem~\ref{thm:cv_noise_only}}

In this section, we present the proofs of the auxiliary results for Theorem~\ref{thm:cv_noise_only}.

\subsubsection{Proof of Lemma~\ref{lem:bound_b_whp_given_x}}\label{app:proof_lem_bound_b_whp_given_x}
First, at $\reg=\infty$ we have $\estbiasmtxat{\infty}=0$ by Proposition~\ref{prop:property_existence_at_infty}, and hence the claimed result is trivially true.

Now consider any $\reg\in [0, \infty)$. We fix any value of $\obsmtx\in \reals^{\numcourse\times \numstudent}$ and any value of $\meancourse\in \reals^\numcourse$. Denote $\mtxdiff \defn \obsmtx - \meancourse\vecone^T$. By triangle's inequality, we have $\max_{\idxpairparen\in \set} \abs*{\valdiff_\idxpair} \le \max_{\idxpairparen \in \set} \abs*{\obs_\idxpair} + \norminf{\meancourse}$.
It then suffices to establish the inequality\begin{align*}
    \max_{\idxpairparen\in \set} \abs{\biasat{\reg}_\idxpair} \le \max_{\idxpairparen\in \set}\abs*{\valdiff_\idxpair},
\end{align*}
where $\biasmtxat{\reg}$ is the solution to the optimization
\begin{align}\label{eq:cv_noise_b_optimization}
    \argmin_{\biasmtx \text{ satisfies } \setpartialorder} \norm*{\mtxdiff - \biasmtx}_\set^2 + \reg\norm*{\biasmtx}_\set^2,
\end{align}
with ties broken by minimizing $\norm{\biasmtx}_\set^2$.
Assume for contradiction that we have 
\begin{align}\label{eq:bound_b_whp_given_x_contradiction}
    \max_{\idxpairparen\in \set} \abs{\biasat{\reg}_\idxpair}> \max_{\idxpairparen\in \set}\abs*{\valdiff_\idxpair}.
\end{align}
Denote $\diffmax \defn \max_{\idxpairparen\in \set} \valdiff_\idxpair$ and $\diffmin \defn \min_{\idxpairparen\in \set} \valdiff_\idxpair$. Then we consider an alternative solution $\biasmtx'$ constructed from $\biasmtxat{\reg}$ as:
\begin{align*}
    \bias'_\idxpair = 
    \begin{cases}
        \max_{\idxpairparen\in \set} \valdiff_\idxpair &\text{if }\biasat{\reg}_\idxpair \in (\diffmax, \infty) \\
        \biasat{\reg}_\idxpair & \biasat{\reg}_\idxpair\in [\diffmin, \diffmax]\\
        \min_{\idxpairparen\in \set} \valdiff_\idxpair &\text{if } \biasat{\reg}_\idxpair \in (-\infty, \diffmin).
    \end{cases}
\end{align*}
By the assumption~\eqref{eq:bound_b_whp_given_x_contradiction}, there exists some $\idxpairparen\in \set$ such that $\biasat{\reg}_\idxpair\not\in [\diffmin, \diffmax]$. Hence, we have $\biasmtx' \ne \biasmtxat{\reg}$. It can be verified that $\biasmtx'$ satisfies the partial ordering $\setpartialorder$ because $\biasmtxat{\reg}$ satisfies $\setpartialorder$. Furthermore, it can be verified that
\begin{align*}
    & \norm{\mtxdiff-\biasmtx'}_\set^2 < \norm{\mtxdiff-\biasmtxat{\reg}}_\set^2
\end{align*}
and also
\begin{align*}
     \norm{\biasmtx'}_\set^2 < \norm{\biasmtxat{\reg}}_\set^2
\end{align*}
Hence, $\biasmtx'$ attains a strictly smaller objective of~\eqref{eq:cv_noise_b_optimization} than $\biasmtxat{\reg}$.
Contradiction to the assumption that $\estbiasmtxat{\reg}$ is the optimal solution of~\eqref{eq:cv_noise_b_optimization}.

\subsubsection{Proof of Lemma~\ref{lem:gaussian_width}}\label{app:proof_gaussian_width}

    Recall that the monotone cone is denoted as $M\defn \{\theta\in \reals^d: \theta_1 \le \ldots \le \theta_d\}$, and $\proj_M$ denotes the projection~\eqref{eq:isotonic_proj} onto $M$.
    
    From known results on the monotone cone (see~\cite[Section 3.5]{amelunxen2014edge}), we have $\Expect[\proj_M Z] \le \const \sqrt{\log d}$ for some fixed constant $\const> 0$. Using the Moreau decomposition, we have (see~\cite[Eq. 20]{wei2019geometry}):
    \begin{align*}
        \Expect
        \left[\vphantom{\sup_{\normtwo{\theta} = 1}}\right.
        \sup_{\substack{\normtwo{\theta} = 1\\ \theta\in M}} \theta^T Z
        \left.\vphantom{\sup_{\normtwo{\theta} = 1}}\right] 
        = \Expect\norm*{\proj_M Z}_2 \le \const\sqrt{\log d}.
    \end{align*}
    Note that we have the deterministic equality $\sup_{\theta\in M, \normtwo{\theta} = 1} \theta^T Z \ge 0$ by taking $\theta = 0$. By Markov's inequality, we have
    \begin{align*}
        \Prob\left(\vphantom{\sup_{\normtwo{\theta} = 1}}\right.
        \sup_{\substack{
                \normtwo{\theta} = 1\\
                \theta\in M}
            } \theta^T Z > d^{\frac{1}{4}}
            \left.\vphantom{\sup_{\normtwo{\theta} = 1}}\right)
            \le \frac{\Expect\left[ \sup_{\theta\in M, \normtwo{\theta} = 1} \theta^T Z \right]}{d^{\frac{1}{4}} } \le \frac{\const\sqrt{\log d}}{d^{\frac{1}{4}} },
    \end{align*}
    completing the proof.

\subsubsection{Proof of Lemma~\ref{lem:cv_only_noise_total_order_size_set_interleaving}}\label{app:proof_lem_cv_only_noise_total_order_size_set_interleaving}

In the proof, we first bound the event $\event_\frac{1}{36}$, and then combine the events $\event_\frac{1}{36}$ and $\event'_\frac{1}{36}$.

\paragraph{Bounding $\event_\frac{1}{36}$}
We denote the interleaving points in $\setinterleavingpairs$ as $\rank^{(1)} < \ldots < \rank^{(\abs*{\setinterleavingpairs})}$. It can be verified that for any $\idxrank\in [\abs*{\setinterleavingpairs} - 1]$, if $\rank^{(\idxrank)}\in \setinterleaving_1$ then then we have  $\rank^{(\idxrank+1)}\in\setinterleaving_2$, and vice versa. Hence, we have
\begin{align}\label{eq:cv_only_noise_total_order_interleaving_set_differ_by_one}
    -1 \le \abs*{\setinterleaving_1} - \abs*{\setinterleaving_2} \le 1.
\end{align}
By Definition~\ref{def:interleaving} of the $\constfrac$-fraction interleaving assumption, we have
\begin{align}\label{eq:cv_only_noise_total_order_const_interleaving}
    \abs*{\setinterleaving_1} + \abs*{\setinterleaving_2} = \abs*{\setinterleaving} \ge \constfrac\numstudent.
\end{align}
Combining~\eqref{eq:cv_only_noise_total_order_interleaving_set_differ_by_one} and~\eqref{eq:cv_only_noise_total_order_const_interleaving}, we have
\begin{align*}
    \abs*{\setinterleaving_1}, \abs*{\setinterleaving_2} >  \frac{\constfrac\numstudent}{3}.
\end{align*}
Suppose the smallest interleaving point in $\setinterleaving_1$ is $\rank_1\defn \min \setinterleaving_1$. We now denote the interleaving points in the increasing order of their rank as:
\begin{align*}
    \ldots <\rank_1 < \rank_1' < \ldots < \rank_{\frac{\constfrac\numstudent}{3}} < \rank'_{\frac{\constfrac\numstudent}{3}} < \ldots\;.
\end{align*}
Then we have $\rank_\idxrank\in \setinterleaving_1$ and $\rank_\idxrank' \in \setinterleaving_2$ for all $\idxrank \in \left[\frac{\constfrac\numstudent}{3}\right]$.

we construct the set of distinct pairs as:
\begin{align*}
\setinterleavingval \defn \left\{(\rank_{2\idxrank - 1}, \rank'_{2\idxrank}): \idxrank\in \left[\frac{\constfrac\numstudent}{6}\right]\right\}\intersect ( \setval\times \setval).
\end{align*}
Now we lower-bound the size of $\setinterleavingval$. For each $\idxrank\in \left[\frac{\constfrac\numstudent}{6}\right]$, consider the probability that the pair $(\rank_{2\idxrank - 1}, \rank'_{2\idxrank})$ is in $\setval$.
It can be verified that the \elements of ranks $\{\rank_{2\idxrank-1}\}_{\idxrank\in \left[\frac{\constfrac\numstudent}{6}\right]}$ are not adjacent in the sub-ordering of $\totalorder$ restricted to course $1$, and hence appear in distinct pairs in Line~\ref{line:assign_start}-\ref{line:assign_end} of Algorithm~\ref{alg:cv} when generating the training-validation split of $(\settrain, \setval)$. Hence, the probability that each \element $\{\rank_{2\idxrank-1}\}_{\idxrank\in \left[\frac{\constfrac\numstudent}{6}\right]}$ is assigned to $\setval$ is independently $\frac{1}{2}$. Similarly, the probability that each \element $\{\rank'_{2\idxrank}\}_{\idxrank\in \left[\frac{\constfrac\numstudent}{6}\right]}$ is assigned to $\setval$ is $\frac{1}{2}$. Hence, the probability of each pair $(\rank_{2\idxrank-1}, \rank'_{2\idxrank})$ is assigned to $\setval$ is $\frac{1}{4}$.
By Hoeffding's inequality, we have
\begin{align*}
     \lim_{\numstudent\rightarrow \infty} \prob\left(\abs*{\setinterleavingval} > \frac{\constfrac\numstudent}{36}\right) = 1.
\end{align*}
That is, $\limn \prob\left(\event_\frac{1}{36}\right) = 1$.

\paragraph{Combining $\event_\frac{1}{36}$ and $\event'_\frac{1}{36}$}
By a similar argument, we have $\limn \prob\left(\event'_\frac{1}{36}\right) = 1$. Taking a union bound of $\event_\frac{1}{36}$ and $\event'_\frac{1}{36}$ completes the proof.

\subsubsection{Proof of Lemma~\ref{lem:cv_only_noise_total_order_size_half_set}}\label{app:proof_lem_cv_only_noise_total_order_size_half_set}

Consider any $T'\in \{\setinterleaving^+ \intersect \setinterleaving_1, \setinterleaving^- \intersect \setinterleaving_1, \setinterleaving^+ \intersect \setinterleaving_2, \setinterleaving^- \intersect \setinterleaving_2\}$. Similar to the proof of Lemma~\ref{lem:cv_only_noise_total_order_size_set_interleaving}, using the fact that the interleaving points alternate between $\setinterleaving_1$ and $\setinterleaving_2$, we have
\begin{align*}
    \abs*{T'} > \frac{\constfrac\numstudent}{6}.
\end{align*}
We write the \elements in $T'$ in the increasing order as $\idxrank_1 < \ldots < \idxrank_{\frac{\constfrac\numstudent}{6}} < \ldots < \idxrank_{\abs*{T'}}$. It can be verified that the \elements in $\{\rank_{2\idxrank}\}_{\idxrank\in \left[\frac{\constfrac\numstudent}{12}\right]}$ appear in different pairs when generating the training-validation split $(\settrain, \setval)$ in Line~\ref{line:assign_start}-\ref{line:assign_end} of Algorithm~\ref{alg:cv}. Hence, each \element in $\{\rank_{2\idxrank}\}_{\idxrank\in \left[\frac{\constfrac\numstudent}{12}\right]}$ is assigned to $\setval$ independently with probability $\frac{1}{2}$. Using Hoeffding's inequality, we lower-bound the size of $T'\intersect\setval$ as:
\begin{align}\label{eq:cv_noise_size_half_set_hoeffding}
    \lim_{\numstudent\rightarrow \infty} \Prob\left(\abs*{T'\intersect\setval} > \frac{\constfrac\numstudent}{36}\right) = 1.
\end{align}
Taking a union bound of~\eqref{eq:cv_noise_size_half_set_hoeffding} over $T'\in \{\setinterleaving^+ \intersect \setinterleaving_1, \setinterleaving^- \intersect \setinterleaving_1, \setinterleaving^+ \intersect \setinterleaving_2, \setinterleaving^- \intersect \setinterleaving_2\}$ completes the proof.